\documentclass[twoside]{article}

\usepackage[accepted]{aistats2026}
\usepackage[round]{natbib}

\usepackage[utf8]{inputenc} 
\usepackage[T1]{fontenc}    
\usepackage{hyperref}       
\usepackage{url}            
\usepackage{booktabs}       
\usepackage{amsfonts}       
\usepackage{nicefrac}       
\usepackage{microtype}      
\usepackage{xcolor}         
\usepackage{algorithmic,algorithm} 

\usepackage{amsmath}
\usepackage{amssymb}
\usepackage{mathtools}
\usepackage{amsthm}

\theoremstyle{plain}
\newtheorem{theorem}{Theorem}[section]
\newtheorem{proposition}[theorem]{Proposition}
\newtheorem{lemma}[theorem]{Lemma}

\theoremstyle{definition}
\newtheorem{definition}[theorem]{Definition}
\newtheorem{assumption}[theorem]{Assumption}
\theoremstyle{remark}
\newtheorem{remark}[theorem]{Remark}
\newtheorem{example}[theorem]{Example}
\newtheorem{baseline}[theorem]{Baseline}

\usepackage[textsize=tiny]{todonotes}

\usepackage[normalem]{ulem} 

\newcommand{\bm}[1]{\boldsymbol{#1}}

\newcommand{\vf}{{\mathbf{f}}}
\newcommand{\vg}{{\mathbf{g}}}

\newcommand{\vn}{{\mathbf{n}}}

\newcommand{\vp}{{\mathbf{p}}}

\newcommand{\vs}{{\mathbf{s}}}

\newcommand{\vv}{{\mathbf{v}}}
\newcommand{\vw}{{\mathbf{w}}}

\newcommand{\vlam}{{\bm{\lambda}}}

\newcommand{\vxi}{{\bm{\xi}}}

\newcommand{\cG}{{\mathcal{G}}}

\newcommand{\cN}{{\mathcal{N}}}

\newcommand{\cT}{{\mathcal{T}}}

\newcommand{\cW}{{\mathcal{W}}}

\DeclareMathOperator*{\argmin}{arg\,min} 

\newcommand{\bc}{\begin{center}}
\newcommand{\ec}{\end{center}}

\begin{document}

%

%

\twocolumn[

\aistatstitle{Enforcing Fair Predicted Scores on Intervals of Percentiles by Difference-of-Convex Constraints}

\aistatsauthor{ Yutian He \And Yankun Huang \And Yao Yao \And Qihang Lin }

\aistatsaddress{
Department of\\Mathematics\\
University of Iowa\\
\texttt{yutian-he@uiowa.edu}
\And
Department of\\Information Systems\\
Arizona State University\\
\texttt{yankun.huang@asu.edu}
\And
School of Mathematics\\
University of Minnesota\\Twin Cities\\
\texttt{yaoxx323@umn.edu}
\And
Department of\\Business Analytics\\
University of Iowa\\
\texttt{qihang-lin@uiowa.edu}
} ]

\begin{abstract}
 Fairness in machine learning has become a critical concern. Existing approaches often focus on achieving full fairness across all score ranges generated by predictive models, ensuring fairness in both high- and low-percentile populations. However, this stringent requirement can compromise predictive performance and may not align with the practical fairness concerns of stakeholders. In this work, we propose a novel framework for building partially fair machine learning models that enforce fairness only within a specific percentile interval of interest while maintaining flexibility in other regions. We introduce statistical metrics to evaluate partial fairness within a given percentile interval. To achieve partial fairness, we propose an in-processing method by formulating the model training problem as constrained optimization with difference-of-convex constraints, which can be solved by an inexact difference-of-convex algorithm (IDCA). We provide the complexity analysis of IDCA for finding a nearly KKT point. Through numerical experiments on real-world datasets, we demonstrate that our framework achieves high predictive performance while enforcing partial fairness where it matters most.
\end{abstract}

\section{Introduction}
\label{sec:intro}
\vspace{-0.1cm}
Fairness in machine learning has emerged as a critical issue, especially when AI systems are employed to assist in high-stakes decision-making~\citep{barocas2023fairness}. Existing research on fairness has introduced multiple criteria to evaluate and enforce fairness of machine learning models among different groups~\citep{feldman2015certifying,dwork2012fairness,hardt2016equality,chzhen2020fair,vogel2021learning}. However, most of these metrics assume that fairness is equally important at all levels of the output predicted scores-a requirement that may impose unnecessary constraints and compromise models' predictive performance. In some applications, fairness considerations are instead localized within specific percentile intervals of the score distribution (e.g., the top/bottom 0--20\% or regions near decision boundaries), where outcomes are more contested and socially consequential. 

For example, in school admissions, machine learning models may be used to rank applicants by predicted admissibility/qualification scores. Fairness is especially critical for those near the admission threshold, where small disparities in prediction can determine acceptance or rejection and are most likely to be scrutinized. By contrast, applicants with very high scores are automatically admitted, making fairness concerns at the top percentiles less contentious. Ensuring fairness around the threshold both reduces complaints and promotes equal educational opportunities. 

The uneven importance of fairness across score percentiles is supported by findings in social psychology and human factors. According to the \emph{uncertainty management theory}~\citep{van2001uncertainty}, individuals rely more on fairness judgments when facing uncertainty. Those with scores in the medium percentiles, e.g., near decision boundaries, experience more uncertainty regarding the decisions made about them than those in high or low percentiles. \citet{van2001uncertainty} shows through experiments that participants exposed to uncertainty manipulations reacted more strongly to unfair procedures, highlighting that uncertainty amplifies sensitivity to fairness. Additionally, \citet{wang2020factors} found that, in algorithmic decision-making, individuals with favorable outcomes perceived algorithms as fairer than those with unfavorable ones. These findings suggest that when outcomes are near borderline or unfavorable, people increasingly rely on fairness perceptions for evaluating the legitimacy of the decision. Consequently, fairness concerns are more pronounced within certain percentile ranges of the score distribution. This implies that enforcing fairness uniformly across all percentiles, as many existing techniques do, may impose unnecessary restrictions.

To address this gap, we propose the concept of \emph{partial fairness}, which ensures fairness within specific percentile intervals of predicted scores, providing a practical alternative to full fairness by focusing on contested cases while allowing flexibility elsewhere. We propose multiple metrics to quantify partial fairness and adopt an \emph{in-processing} approach, formulating training as a constrained optimization problem that maximizes predictive performance subject to partial fairness constraints based on the proposed metrics. Because partial fairness targets only critical score percentiles, it reduces unnecessary restrictions, enabling models to maintain stronger predictive performance.


Following standard modeling practices in machine learning, we formulate the constrained optimization problems using empirical approximations and surrogate functions. We show that each proposed fairness constraint can be expressed as the difference of two convex (possibly non-smooth) functions. Embedding these constraints into the training process leads to a \emph{difference-of-convex} (DC) constrained optimization problem.

DC optimization has been studied extensively for several decades~\citep{tao1997convex,hiriart1985generalized,le2018dc,pham2014recent,le2024open}.  The best-known algorithm for DC optimization is the \emph{difference-of-convex algorithm} (DCA), which iteratively updates the solution by solving a sequence of convex subproblems obtained by linearizing the concave component of each DC function~\citep{tao1986algorithms,pham2014recent}. When these subproblems are computationally challenging, they are often solved approximately, leading to the \emph{inexact DCA} (IDCA), which is more practical and widely applicable than the standard DCA~\citep{cruz2020generalized,ferreira2026inexact}. Although the asymptotic convergence theory of DCA for DC-constrained optimization has been well established~\citep{pang2017computing,sriperumbudur2012proof,pham2014recent}, the convergence rate and the complexity of IDCA for DC-constrained optimization has not been analyzed.


To bridge this gap, we study the IDCA for general non-smooth DC-constrained optimization and provide a theoretical analysis of its oracle complexity for computing a nearly $\epsilon$-Karush--Kuhn--Tucker (KKT) point. To the best of our knowledge, this is the first complexity analysis of IDCA for DC-constrained problems, making it an important contribution in its own right. 

\vspace{-0.3cm}
\section{Related Works}
\vspace{-0.1cm}
Various fairness metrics have been studied in the literature, including demographic parity~\citep{calders2009building,zafar2017fairness, feldman2015certifying, dwork2012fairness}, equality of opportunity~\citep{hardt2016equality}, equality of odds~\citep{hardt2016equality}, predictive quality parity~\citep{chouldechova2017fair},  counterfactual fairness~\citep{kusner2017counterfactual} and predicted parity (calibration)~\citep{chouldechova2017fair}. 

Most of the above metrics are tailored to classification tasks, as they evaluate fairness based on predicted class labels. In contrast, fairness metrics have also been proposed directly on predicted scores/probabilities, so they are applicable to both classification and regression tasks. For examples, fairness metrics based on pairwise comparison (AUC-based metrics) have been studied in~\citet{borkan2019nuanced, kallus2019fairness, dixon2018measuring, vogel2021learning, narasimhan2020pairwise,beutel2019fairness,yao2022large,yang2023minimax}. 
Other score-based fairness metrics include the ones based on Wasserstein distance~\citep{miroshnikov2022wasserstein, miroshnikov2021model, chzhen2020fair, chzhen2020fairplugin, chzhen2022minimax}, ROC~\citep{vogel2021learning}, Gini coefficient~\citep{chen2024measuring}, and the Kolmogorov-Smirnov test~\citep{chen2024measuring}.

The fairness metrics mentioned above aim to achieve fairness over the entire distribution of scores generated by predictive models. In contrast, the partial fairness metrics proposed in this work focus on measuring fairness over a specific percentile interval, where prediction outcomes are more contested and fairness concerns are more frequently raised. 

The ROC-based fairness notion proposed by \citet{vogel2021learning} requires $\text{ROC}(\alpha)=\alpha$ for all $\alpha \in [0,1]$, where $\text{ROC}(\alpha)$ denotes the ROC curve comparing the predicted scores of two groups. Although the authors do not explicitly introduce the concept of partial fairness, one can restrict $\alpha$ to a subinterval of $[0,1]$ to recover a form of partial statistical parity (Definition~\ref{def:pdp}) considered in our work; see Section 3.3 of \citet{vogel2021learning} for details. That said, our notion of partial fairness is more general, encompassing not only partial statistical parity but also partial demographic parity (Definition~\ref{eq:gauc}) and partial group AUC fairness (Definition~\ref{eq:gauc}). Moreover, the numerical approach used to enforce ROC-based fairness in \citet{vogel2021learning} (Algorithm 2) is heuristic and lacks theoretical guarantees, whereas our method is grounded in DC optimization and admits rigorous convergence analysis.



Fairness metrics have also been studied for selection tasks~\citep{khalili2021improving}. These tasks typically require a model that predicts qualification scores, which are then used to make selection decisions. The techniques proposed in this paper can further promote fairness in selection by producing fair qualification scores.

The three main approaches for building a fairness-aware machine learning model include the pre-processing~\citep{dwork2012fairness}, post-processing~\citep{hardt2016equality,chzhen2020fair,chzhen2022minimax,xian2023fair}, and in-processing methods~\citep{agarwal2018reductions,goh2016satisfying,wan2023processing}. The methods in this paper are the in-processing methods, where we optimize a model's prediction performance subject to partial fairness constraints, which can be formulated as a DC-constrained optimization problem. 

DCA is the most studied numerical algorithm for DC optimization~\citep{tuy1984global,tao1986algorithms,sriperumbudur2012proof,tao1997convex,le2018convergence,pham2014recent,abbaszadehpeivasti2024rate}. Each iteration of DCA requires solving a convex subproblem exactly, which can be challenging. To address this, we employ IDCA, a variant of DCA that allows for inexact solutions to the subproblems~\citep{xu2019stochastic,cruz2020generalized,ferreira2026inexact,kanzow2024bundle,moudafi2006convergence,liu2022inexact}. In addition to traditional DCA and IDCA, there exist many other numerical algorithms for DC optimization, including proximal DCA~\citep{cruz2020generalized,souza2016global,sun2003proximal,moudafi2006convergence,pang2017computing,moudafi2024dc},  boosted DCA~\citep{aragon2018accelerating,aragon2020boosted,aragon2022boosted,zhang2024boosted,ferreira2024boosted,ferreira2026inexact}, inertial DCA~\citep{mainge2008convergence,de2019inertial}, DC composite algorithm~\citep{le2024minimizing,hu2024single}, gradient descent method~\citep{khamaru2019convergence}, proximal gradient method~\citep{khamaru2019convergence}, Frank-Wolfe method~\citep{khamaru2019convergence}, smoothing method~\citep{sun2023algorithms,moudafi2021complete,yao2022large,hu2024single}, augmented Lagrangian method~\citep{pang2018decomposition,sun2023algorithms}, and codifferential method~\citep{bagirov2011codifferential}.

However, most of the aforementioned works focus exclusively on unconstrained DC optimization, making their results inapplicable to our problems. The convergence of DCA and its variants in the constrained setting has been established by~\citet{sriperumbudur2012proof,pham2014recent,pang2017computing,pang2018decomposition,le2024minimizing}. Notably, \citet{pang2017computing,pang2018decomposition} proved convergence to a directional stationary point, which is a stronger notion of stationarity than the nearly KKT point considered in this paper. However, for our problems, those results of DCA are not applicable due to the difficulty of solving the subproblem exactly. To the best of our knowledge, no existing work has established convergence results for IDCA in constrained settings. Our work, therefore, contributes to the convergence and complexity analysis of IDCA.

The theoretical analysis in this work is motivated by recent developments in the complexity analysis of first-order methods for weakly convex non-smooth constrained optimization~\citep{ma2020quadratically, boob2023stochastic, huang2023oracle, jia2025first}. 
Assuming different constraint qualifications on the subproblems, those works establish a complexity of $\mathcal{O}(\epsilon^{-4})$ for finding a nearly $\epsilon$-KKT point, which matches the complexity bound achieved in this work. While a weakly convex function is a DC function, the converse does not hold. Hence, the results from these prior works cannot be directly applied to IDCA. 

\vspace{-0.1cm}
\section{Preliminary and Notations}
\label{sec:preliminary}
We consider a prediction task in which the goal is to develop a model that predicts a continuous or discrete target variable $\zeta \in \mathbb{R}$ using a feature vector $\vxi \in \mathbb{R}^d$. Each data point has a categorical sensitive attribute (e.g., gender and race), denoted by $\gamma \in \mathcal{G}:=\{1,2,\dots,G\}$, which may or may not be a component of $\vxi$. 
A data point is represented as a random triplet $ (\vxi, \zeta, \gamma) \in \mathbb{R}^{d+2}$ following a ground truth distribution. 

Let $h_{\vw} : \mathbb{R}^d \to \mathbb{R}$ be a predictive model that assigns a predicted score $h_{\vw}(\vxi)$ to a feature vector $\vxi$. When $\zeta$ is continuous, this score directly predicts $\zeta$. When $\zeta$ is binary,   the score can be converted into a prediction of $\zeta$ after thresholding. We are interested in the fairness of $h_{\vw}(\vxi)$ among the groups of data defined by $\gamma$. Here, $h_{\vw}$ depends on a parameter vector $\vw$ that belongs to a convex and closed set $\cW \subset \mathbb{R}^n$. We denote the gradient of $h_{\vw}$ with respect to $\vw$ by $\nabla_{\vw} h_{\vw}(\vxi)$ and make the following assumption throughout the paper.
\begin{assumption}
\label{assume:hsmooth}
For any $\vxi$, $h_{\vw}(\vxi)$ is differentiable and $\nabla_{\vw} h_{\vw}(\vxi)$ is $L_h$-Lipschitz continuous in $\vw$. 
\end{assumption}

Let $\text{Pr}(\mathcal{A})$ and $\text{Pr}(\mathcal{A}|\mathcal{B})$ denote, respectively, the probability of $\mathcal{A}$ and the conditional probability of $\mathcal{A}$ conditioning on $\mathcal{B}$. Let $\mathbf{1}_{\mathcal{A}}$ be the $0$-$1$ indicator function of event $\mathcal{A}$. Let 
\begin{align}
\label{eq:cdf}
F_{\vw,k}(\theta):=\text{Pr}(h_{\vw}(\vxi) \leq\theta | \gamma=k),~\forall k\in\mathcal{G}, 
\end{align}
i.e., the cumulative distribution function (CDF) of the predicted scores in group $k$. Let $[m]:=\{1, \dots, m\}$. For a convex function $f(\cdot)$, let  $\partial f(\cdot)$ be its subdifferential. Let $\|\cdot\|$ be the Euclidean norm and $\mathcal{N}_{\cW}(\vw)$ be the normal cone of $\cW$ at $\vw\in\cW$.


We say a model satisfies \emph{statistical parity} (also called \emph{distributional parity}) when the distribution of $h_{\vw}(\vxi)$ is independent of $\gamma$, which means 
\begin{align}
\nonumber
&~\text{SP}_{k,k'}(\vw)\\\nonumber
:=&\max_{\theta}\left|\text{Pr}(h_{\vw}(\vxi) >\theta | \gamma=k )-\text{Pr}(h_{\vw}(\vxi) > \theta | \gamma=k' )\right|\\\label{eq:dp}
=&~0,~\forall k, k'\in\mathcal{G}.
\end{align}
For a binary classification task where $\zeta\in\{1,-1\}$, a prediction $\widehat\zeta$ can be generated with a predetermined threshold $\widehat\theta$ (typically $\widehat\theta=0$) with $\widehat\zeta=1$ if $h_{\vw}(\vxi)>\widehat\theta$ and $\widehat\zeta=-1$, otherwise. In this case,  \eqref{eq:dp} is often relaxed to the \emph{demographic parity}, i.e.,
\begin{align}
\nonumber
&~\text{DP}_{k,k'}(\vw)\\\nonumber
:=&\left|\text{Pr}(h_{\vw}(\vxi) >\widehat\theta | \gamma=k )-\text{Pr}(h_{\vw}(\vxi) > \widehat\theta | \gamma=k' )\right|\\\label{eq:ndp}
=&~0,~\forall k, k'\in\mathcal{G}.
\end{align}

\vspace{-0.1cm}
\section{Partial Fairness}\label{sec:partialfairness}
We propose new criteria of fairness where the fairness is only enforced among the instances within a certain percentile interval in different groups. For simplicity, in this and the next sections, we assume that $h_{\vw}(\vxi)$ has an absolutely continuous distribution conditioning on $\gamma=k$ for each $k\in\mathcal{G}$. 
\begin{definition}
\label{def:rank}
For $k\in\mathcal{G}$ and $\theta\in\mathbb{R}$, we call $\bar{F}_{\vw,k}(\theta)$ the \emph{rank} of score $\theta$ in group $k$,  where
$$
\bar{F}_{\vw,k}(\theta):=\text{Pr}(h_{\vw}(\vxi) >\theta | \gamma=k)=1-F_{\vw,k}(\theta).
$$
\end{definition}
Note that $\bar{F}_{\vw,k}(\theta)$ is also called \emph{the complementary cumulative distribution function} (CCDF) of $h_{\vw}(\vxi)$ over group $k$. 
By definition, $\bar{F}_{\vw,k}(\theta)=p$ means $\theta$ is the $100(1-p)$th percentile of the scores in group $k$.
Then, we propose a generalization of \eqref{eq:dp} as follows.
\begin{definition}
\label{def:pdp}
We say $h_{\vw}(\vxi)$ satisfies the \emph{partial statistical parity (pSP) with respect to percentile interval} $\mathcal{I}=[\alpha, \beta)\subset[0,1]$ if 
\begin{align}
\nonumber
&~\text{SP}_{k,k'}^{\mathcal{I}}(\vw)\\\nonumber
:=&\max_{\theta}\left|
\begin{array}{l}
\text{Pr}\big(h_{\vw}(\vxi) >\theta \big| \gamma=k, \bar{F}_{\vw,k}(h_{\vw}(\vxi))\in\mathcal{I}\big)\\
-\text{Pr}\big(h_{\vw}(\vxi) > \theta \big| \gamma=k', \bar{F}_{\vw,k'}(h_{\vw}(\vxi))\in\mathcal{I}\big)
\end{array}
\right|\\\label{eq:pdp}
=&~0,~\forall k, k'\in\mathcal{G}.
\end{align}
\end{definition}
For example, when $\mathcal{I}=[0\%,20\%)$, the condition $\bar{F}_{\vw,k}(h_{\vw}(\vxi))\in\mathcal{I}$ means score $h_{\vw}(\vxi)$ is among the top 20\% scores in group $k$, and partial statistical parity requires the top 20\% scores in each group to have the same distribution but puts no restriction on the bottom  80\% scores in any group. When $\mathcal{I}=[0,1]$, \eqref{eq:pdp} is reduced to \eqref{eq:dp}. Similarly, \eqref{eq:ndp} can also be generalized for partial fairness.

\begin{definition}
\label{def:wpdp}
We say $h_{\vw}(\vxi)$ satisfies the \emph{partial demographic parity  (pDP) with respect to percentile interval} $\mathcal{I}=[\alpha, \beta)\subset[0,1]$ and threshold $\widehat\theta$ if 
\begin{align}
\nonumber
&~\text{DP}_{k,k'}^{\mathcal{I}}(\vw)\\\nonumber
:=&\left|
\begin{array}{l}
\text{Pr}\big(h_{\vw}(\vxi) >\widehat\theta \big| \gamma=k, \bar{F}_{\vw,k}(h_{\vw}(\vxi))\in\mathcal{I}\big)\\
-\text{Pr}\big(h_{\vw}(\vxi) > \widehat\theta  \big| \gamma=k', \bar{F}_{\vw,k'}(h_{\vw}(\vxi))\in\mathcal{I}\big)
\end{array}
\right|\\\label{eq:wpdp}
=&~0,~\forall k, k'\in\mathcal{G}.
\end{align}
\end{definition}


We target on an in-processing method to produce a $\vw\in\cW$ that satisfies condition \eqref{eq:pdp} or \eqref{eq:wpdp} by solving the following optimization problems subject to fairness constraints:
\begin{align}
\label{eq:inprocess_pdp}
\min_{\vw\in\cW} f_0(\vw) &~\text{  s.t. }~\text{SP}_{k,k'}^{\mathcal{I}}(\vw)\leq \kappa,~\forall k, k'\in\mathcal{G},\\\label{eq:inprocess_wpdp}
\min_{\vw\in\cW} f_0(\vw) &~\text{  s.t. }~\text{DP}_{k,k'}^{\mathcal{I}}(\vw)\leq \kappa,~\forall k, k'\in\mathcal{G},
\end{align}
where $f_0(\vw)$ represents the loss function defined on training data, which is designed such that minimizing $f_0(\vw)$ optimizes the predictive performance of $h_{\vw}(\vxi)$. The right-hand side $\kappa\in[0,1]$ is the allowed violation of the fairness constraints, which is specified by the users of our techniques based on their tolerance to unfairness according to the applications. Our model can be extended to the partial variants of other fairness metrics, e.g., the \textbf{group AUC fairness}~\citep{vogel2021learning,yao2022large}, \textbf{equality of opportunity}~\citep{hardt2016equality}, and \textbf{equality of odds}~\citep{hardt2016equality}. See Section \ref{sec: gAUC}. 

Parameter $\kappa$ and interval $\mathcal{I}$ are chosen by users of our technique based on their specific fairness concerns while the algorithm and theoretical results we develop are applicable to any choice. When users do not have a targeted interval, we provide some practical guidelines for classification problem in Section \ref{sec:chooseI}. 

\section{Optimization Models}\vspace{-0.1cm}
\label{sec:optimization_models}
There are two challenges for solving \eqref{eq:inprocess_pdp} and \eqref{eq:inprocess_wpdp}. First, the probabilities in \eqref{eq:pdp} and \eqref{eq:wpdp} are defined on the unknown population distribution. In practice, we must approximate them using a training dataset, denoted by $\mathcal{D}=\{(\vxi_i,\zeta_i,\gamma_i)\}_{i=1}^n$. Second, approximating the probabilities using the empirical distribution over $\mathcal{D}$ will introduce discontinuous $0$-$1$ indicator functions in \eqref{eq:inprocess_pdp} and \eqref{eq:inprocess_wpdp}. To obtain a continuous optimization problem, we have to further approximate the indicator function by a continuous surrogate, which often leads to non-convexity and non-smoothness. In this section, we will follow these two steps to derive a data-driven continuous approximation of \eqref{eq:inprocess_pdp} and \eqref{eq:inprocess_wpdp}.

We first provide an equivalent formulation of the constraint in \eqref{eq:inprocess_pdp} whose proof is in Section~\ref{sec:equivreform}.
\begin{lemma}
\label{eq:pdp_equiv}
A solution $\vw\in\cW$ is feasible to \eqref{eq:inprocess_pdp} if and only if, for any $p\in[\alpha,\beta-\kappa(\beta-\alpha))$, there exists $\theta_p\in\mathbb{R}$ such that
\begin{equation}
\begin{split}
\label{eq:pdpnew}
&\textup{Pr}(h_{\vw}(\vxi) >\theta_p | \gamma=k )\geq p~~\text{  and  }\\
&\textup{Pr}(h_{\vw}(\vxi) > \theta_p | \gamma=k)\leq p+\kappa(\beta-\alpha),~~\forall k\in\cG.
\end{split}
\end{equation}
\end{lemma}
However, there are uncountably many constraints in \eqref{eq:pdpnew}, which are computationally intractable. To derive a tractable approximation of \eqref{eq:pdpnew}, we first partition the training data into subsets based on $\gamma$, namely, $\mathcal{D}=\cup_{k\in\cG}\mathcal{D}_k$, where
$$
\mathcal{D}_k=\{(\vxi,\zeta,\gamma)\in\mathcal{D}|\gamma=k\}=\{(\vxi_i^k,\zeta_i^k,k)\}_{i=1}^{n_k},
$$
and $n_k$ is the size of data from group $k$. Then, we approximate the probabilities in \eqref{eq:pdpnew} using the expectation of the indicator $\mathbf{1}_{h_{\vw}(\vxi)>\theta}$ over the empirical distribution over $\mathcal{D}_k$. Additionally, we create a finite subset $\widehat{\mathcal{I}}\subset[\alpha,\beta-\kappa(\beta-\alpha))$ and only impose the constraints in \eqref{eq:pdpnew} with $p\in\widehat{\mathcal{I}}$. This leads to an approximation of  \eqref{eq:pdpnew} that requires, for  any $p\in\widehat{\mathcal{I}}$, there exists $\theta_p\in\mathbb{R}$ such that
\small
\begin{align*}
&\textstyle{\frac{1}{n_k} \sum_{i=1}^{n_k} \mathbf{1}_{h_{\vw}(\vxi_i^k)>\theta_{p}}}\geq p~~\text{  and  }\\
&\textstyle{\frac{1}{n_k} \sum_{i=1}^{n_k} \mathbf{1}_{h_{\vw}(\vxi_i^k)>\theta_{p}}}\leq p+\kappa(\beta-\alpha),~~\forall k\in\mathcal{G}.
\end{align*}
\normalsize
Next, we approximate the indicator function above by a continuous surrogate $\sigma(x)\approx \mathbf{1}_{x>0}$ and obtain the final approximation of  \eqref{eq:inprocess_pdp}:
\begin{align}
\label{eq:inprocess_pdp_approx}
&~\min_{\vw\in\cW,~(\theta_p)_{p\in\widehat{\mathcal{I}}}} f_0(\vw)\quad\text{s.t.}\\\nonumber
&~\frac{1}{n_k} \sum_{i=1}^{n_k} \sigma(h_{\vw}(\vxi_i^k)-\theta_{p})\geq p,\forall p\in\widehat{\mathcal{I}},k\in\mathcal{G},\\\nonumber
&~\frac{1}{n_k} \sum_{i=1}^{n_k} \sigma(h_{\vw}(\vxi_i^k)-\theta_{p})\leq p+\kappa(\beta-\alpha),\forall p\in\widehat{\mathcal{I}},k\in\mathcal{G},
\end{align}
where $(\theta_p)_{p\in\widehat{\mathcal{I}}}\in\mathbb{R}^{|\widehat{\mathcal{I}}|}$ is the auxiliary decision variable. The examples of $\sigma$ include   $\sigma(x)=\max\{\min\{x+0.5,1\},0\}$ and $\sigma(x)=\exp(x)/(1+\exp(x))$.




We next provide an equivalent formulation of the constraint in \eqref{eq:inprocess_wpdp} whose proof is in Section~\ref{sec:equivreform} also.
\begin{lemma}
\label{eq:wpdp_equiv}
A solution $\vw\in\cW$ is feasible to \eqref{eq:inprocess_wpdp} if and only if, for any $k$ and $k'$ in $\mathcal{G}$,
\small
\begin{align}
\label{eq:wpdpnew}
\left|
\begin{array}{l}
\min\left\{\textup{Pr}\big(h_{\vw}(\vxi) >\widehat\theta \big| \gamma=k\big),\beta\right\}\\
-\min\left\{\textup{Pr}\big(h_{\vw}(\vxi) >\widehat\theta \big| \gamma=k\big),\alpha\right\}\\
-\min\left\{\textup{Pr}\big(h_{\vw}(\vxi) >\widehat\theta \big| \gamma=k'\big),\beta\right\}\\
+\min\left\{\textup{Pr}\big(h_{\vw}(\vxi) >\widehat\theta \big| \gamma=k'\big),\alpha\right\}
\end{array}
\right|\leq\kappa(\beta-\alpha),
\end{align}
\normalsize
\end{lemma}

Following the same steps, we approximate the probabilities in \eqref{eq:wpdpnew} using data $\mathcal{D}$ and approximate the indicator function using a surrogate $\sigma(\cdot)$. The final approximation of  \eqref{eq:inprocess_wpdp} is as follows.
\begin{align}
\label{eq:inprocess_wpdp_approx}
&\min_{\vw\in\cW} f_0(\vw)\\\nonumber
\text{s.t.}&
\begin{array}{l}
\min\left\{\frac{1}{n_k} \sum_{i=1}^{n_k} \sigma(h_{\vw}(\vxi_i^k)-\widehat\theta),\beta\right\}\\
-\min\left\{\frac{1}{n_k} \sum_{i=1}^{n_k} \sigma(h_{\vw}(\vxi_i^k)-\widehat\theta),\alpha\right\}\\
-\min\left\{\frac{1}{n_{k'}} \sum_{i=1}^{n_{k'}} \sigma(h_{\vw}(\vxi_i^{k'})-\widehat\theta),\beta\right\}\\
+\min\left\{\frac{1}{n_{k'}} \sum_{i=1}^{n_{k'}} \sigma(h_{\vw}(\vxi_i^{k'})-\widehat\theta),\alpha\right\}\leq \kappa(\beta-\alpha),
\end{array}\\\nonumber
&~~\forall k, k'\in\mathcal{G}.
\end{align}
Although \eqref{eq:inprocess_pdp_approx} and \eqref{eq:inprocess_wpdp_approx} are formulated very differently, we can show that, under mild assumptions on $f_0$, $\sigma$ and $h_{\vw}(\cdot)$, both problems can be formulated as the following optimization model. 
\begin{equation} \label{DC}
    \begin{split}
        \min_{\vw \in \cW} &~ f_0(\vw):=f_0^+(\vw) - f_0^-(\vw)\\
        ~\text{ s.t.}\;&~  f_i(\vw):=f_i^+(\vw) - f_i^-(\vw) \leq 0,~ i\in[m],
    \end{split}
\end{equation}
where functions in $\{ f_i^+ \}_{i=0}^m$ and $\{ f_i^- \}_{i=0}^m$ are real-valued, convex but potentially non-smooth. Since each $f_i$ is the difference of two convex functions, \eqref{DC} is a difference-of-convex (DC) constrained optimization problem. In Section~\ref{sec:reform}, we present some specific examples of \eqref{eq:inprocess_pdp_approx} and \eqref{eq:inprocess_wpdp_approx} that are instances of \eqref{DC}. 

Although the primary focus of this work is on optimization methods for enforcing partial fairness, we also provide statistical guarantees for models obtained from \eqref{eq:inprocess_pdp_approx}. In particular, we analyze the generalization error arising from three sources: the discretization of $\mathcal{I}$, the surrogate approximation $\sigma(x)\approx \mathbf{1}_{x>0}$, and sampling error. The results are presented in Section~\ref{sec:generalization}.

\begin{remark}
Definitions~\ref{def:pdp} and~\ref{def:wpdp} can be easily extended to settings where the percentile interval $\mathcal{I}$ varies across groups, as well as to the case where $\mathcal{I}$ is not a single interval but a union of multiple disjoint intervals. In both cases, the corresponding optimization models \eqref{eq:inprocess_pdp_approx} and \eqref{eq:inprocess_wpdp_approx} can be naturally extended, formulated as DC programs, and efficiently solved within the IDCA framework.
\end{remark}


\section{Inexact DCA}
\label{sec:IDCA}

Since  \eqref{DC} is non-convex, finding an optimal or $\epsilon$-optimal solution of \eqref{DC} is intractable in general. Therefore, the recent studies on non-convex optimization are mainly motivated to find an $\epsilon$-Karush--Kuhn--Tucker ($\epsilon$-KKT) solution. 
\begin{definition}
\label{def:eKKT}
A point $\vw\in\cW$ is an \emph{$\epsilon$-KKT point} of \eqref{DC} if there exist $\vn_{\cW}\in \cN_\cW(\vw)$, $\vf_i^+\in\partial f_i^+(\vw)$, $\vf_i^-\in\partial f_i^-(\vw)$, and $\lambda_i\geq 0$ for $i\in[m]$ satisfying 
\begin{align}
\label{eq:stationarity}
\left\|\vf_0^+-\vf_0^-+\textstyle\sum_{i=1}^m\lambda_i(\vf_i^+-\vf_i^-)+\vn_{\cW}\right\|\leq \epsilon,\\\label{eq:feasibility}
f_i^+(\vw)-f_i^-(\vw)\leq \epsilon,~ i\in[m],\\\label{eq:complementary}
|\lambda_i(f_i^+(\vw)-f_i^-(\vw))|\leq \epsilon,~i\in[m].
\end{align}
\end{definition}
However, when a problem is non-smooth, even finding an  $\epsilon$-KKT point is intractable in general as the set of  $\epsilon$-KKT points often has a zero measure (e.g., $\min_{\vw}\|\vw\|$). Following the recent literature on non-convex non-smooth optimization, we consider the following weaker class of solutions.
\begin{definition}
\label{def:nearlyeKKT}
A point $\vw\in\cW$ is a \emph{nearly $\epsilon$-KKT point} of \eqref{DC} if there exist $\widehat\vw\in\cW$, $\vn_{\cW}\in \cN_\cW(\widehat\vw)$, $\vf_i^+\in\partial f_i^+(\widehat\vw)$, $\vf_i^-\in\partial f_i^-(\vw)$, and $\lambda_i\geq 0$ for $i\in[m]$ satisfying $\|\widehat\vw-\vw\|\leq \epsilon$, \eqref{eq:stationarity}, \eqref{eq:feasibility} and \eqref{eq:complementary}.
\end{definition}
The only difference between an $\epsilon$-KKT point and a nearly $\epsilon$-KKT point is that the former satisfies the stationarity condition \eqref{eq:stationarity} only with its own subgradients while the latter satisfies \eqref{eq:stationarity} with its own subgradients and the subgradients of a nearby point (i.e., $\widehat\vw$). 

We use an inexact diffference-of-convex algorithm (IDCA) to find a nearly $\epsilon$-KKT point of \eqref{DC}. At the $k$-th iteration of IDCA, if the current solution is $\vw^{(k)}\in\cW$, we linearize the concave component of each of the objective and constraint functions in \eqref{DC} at $\vw^{(k)}$ to obtain the convex approximation of \eqref{DC} below
\begin{align} \label{DCA}
       \min_{\vv \in \cW} &~ g_0(\vv):=f_0^+(\vv) - \widehat{f}_0^-(\vv;\vw^{(k)})\\\nonumber
        ~\text{ s.t. }&~ g_i(\vv):=f_i^+(\vv) - \widehat{f}_i^-(\vv;\vw^{(k)}) \leq 0,~ i\in[m],
\end{align}
where
\begin{equation}
\label{eq:convexifyf}
    \widehat{f}_i^-(\vv;\vw):= f_i^-(\vw) + (\vf_i^-)^\top(\vv - \vw)
\end{equation}
and $\vf_i^{-}$ is an arbitrary subgradient in $\partial f_i^-(\vw)$. Note that, by the convexity of $f_i^-$, we have $f_i(\cdot)\leq g_i(\cdot)$ for $i=0,1,\dots,m$. Then IDCA proceeds to generate the next iterate $\vw^{(k+1)}$ as an $\epsilon_k$-optimal solution of \eqref{DCA}, meaning that 
\begin{align}
\hspace{-0.1in}
\label{eq:epsilonwk}
   \max\left\{g_0(\vw^{(k+1)})-g_0(\widehat\vw^{(k+1)}),g(\vw^{(k+1)})\right\}\leq \epsilon_k,
\end{align}
where $g(\cdot):=\max_{i\in[m]}g_i(\cdot)$ and $\widehat\vw^{(k+1)}$ is the optimal solution of \eqref{DCA}. It is formally presented in Algorithm~\ref{alg:dca}.

\begin{algorithm}[t]
   \caption{Inexact Difference-of-Convex Algorithm (IDCA)}
   \label{alg:dca}
\begin{algorithmic}[1]
   \STATE {\bfseries Input:} $\vw^{(0)}$ feasible to \eqref{DC}, the number of iterations $K$,  precision $\epsilon_k$, $k=0,1,\dots,K-1$.
   \FOR{$k=0$ {\bfseries to} $K-1$} 
   \STATE Find $\vw^{(k+1)}$ that satisfies \eqref{eq:epsilonwk}.
   \ENDFOR
\end{algorithmic}
\end{algorithm}



Since \eqref{DCA} is convex, we will directly apply the switching subgradient (SSG) method \citep[(3.2.24)]{nesterov2018lectures} to \eqref{DCA} to find  $\vw^{(k+1)}$. This method is presented in Algorithm~\ref{alg:swg}. One of the inputs of SSG is an infeasibility tolerance, denoted by $\epsilon$. At iteration $t$ of SSG, if the current solution $\vv^{(t)}$ is nearly feasible, i.e., $g_i(\vv^{(t)})\leq \epsilon$, $\vv^{(t)}$ will be updated along the subgradient of $g_0$. Otherwise, $\vv^{(t)}$ will be updated along the subgradient of $\max_{i\in[m]}g_i$. The final output is the nearly feasible solution that has the smallest objective value $g_0(\vv^{(t)})$. 

The efficiency of Algorithm~\ref{alg:dca} is measured by its \emph{oracle complexity}, which is the total number of subgradients computed before finding a nearly $\epsilon$-KKT point of \eqref{DC}. In order to establish the oracle complexity of Algorithm~\ref{alg:dca}, we make the following assumptions on \eqref{DC}.

\begin{assumption}
The following statements hold.
\label{assume:DC}
\begin{enumerate}
\item[A.] Functions in $\{ f_i^+ \}_{i=0}^m$ and $\{ f_i^- \}_{i=0}^m$ are real-valued and $\mu$-strongly convex on $\mathbb{R}^n$.
\item[B.] There exists constant $\nu>0$ 
such that, if $f_i(\vw)\leq 0$ for $i\in[m]$, there exists $\vv\in\cW$ such that
$
f_i^+(\vv) - \widehat{f}_i^-(\vv;\vw)\leq -\nu,~\forall i\in[m].
$
\item[C.] $f_{\text{lb}}:=\min_{\vw\in\cW}f_0(\vw)\geq-\infty$.
\item[D.] $f_i^+(\vw)$ and $f_i^-(\vw)$ are $M$-Lipchitz continuous on $\cW$ for $i=0,1,\dots,m$.
\item[E.] A solution $\vw_{\text{feas}}$ satisfying $f_i(\vw_{\text{feas}})\leq 0$ for $i\in[m]$ is accessible. 
\end{enumerate} 
\end{assumption}

\begin{algorithm}[t]
   \caption{Switching Subgradient (SSG) Method for \eqref{DCA}: SSG($\vv^{(0)}$,$\epsilon$, $T$)}
   \label{alg:swg}
\begin{algorithmic}[1]
   \STATE {\bfseries Input:} Infeasibility tolerance $\epsilon >0$, initial solution $\vv^{(0)} \in \cW$ with $g_i(\vv^{(0)})\leq \epsilon$ for $i\in[m]$ and the number of iterations $T$. 
   \STATE $\cT\leftarrow\emptyset$.
    \FOR {$t=0$ {\bfseries to} $T-1$} 
   \IF{$g(\vv^{(t)}) \leq \epsilon$}
   \STATE $\vv^{(t+1)}\leftarrow \text{Proj}_{\cW}(\vv^{(t)}-\frac{\epsilon}{\|\vg_0^{(t)}\|^2} \vg_0^{(t)})$ for any $\vg_0^{(t)}\in \partial g_0(\vv^{(t)})$. $\quad\cT\leftarrow\cT\cup\{t\}.$
   \ELSE
   \STATE $\vv^{(t+1)}\leftarrow  \text{Proj}_{\cW}(\vv^{(t)}-\frac{g(\vv^{(t)})}{\|\vg^{(t)}\|^2} \vg^{(t)})$ for any $\vg^{(t)}\in \partial g(\vv^{(t)})$.
   \ENDIF
   \ENDFOR
   \STATE {\bfseries Output: }$\vv^{(\tau)}$ with $\tau=\argmin_{t\in\cT} g_0(\vv^{(t)})$
\end{algorithmic}
\end{algorithm}

\begin{figure*}[tb]
     \begin{tabular}{@{}c|ccc@{}}
      & a9a & bank & law school \\
		\hline \vspace*{-0.1in}\\
		\raisebox{12ex}{\small{\rotatebox[origin=c]{90}{pSP fairness}}}
		& \hspace*{-0.06in}\includegraphics[width=0.30\textwidth]{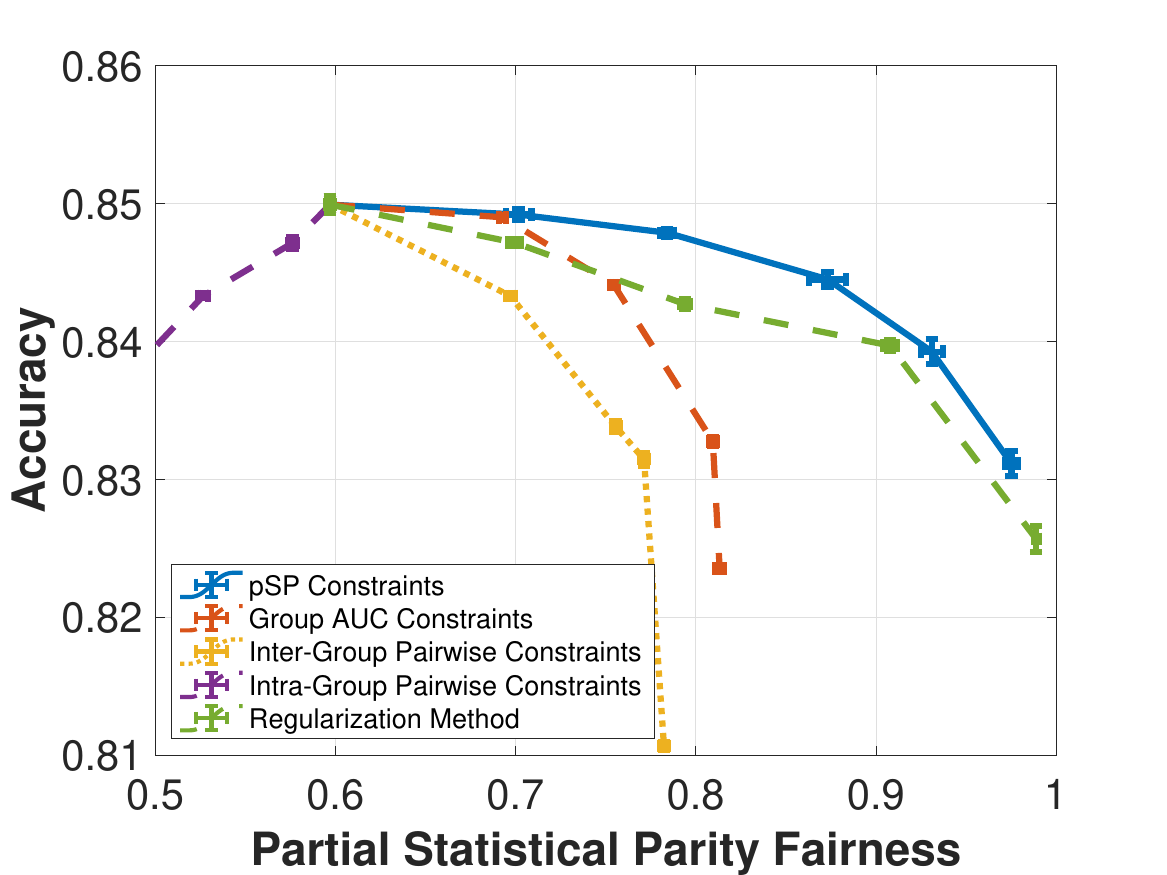}
		& \hspace*{-0.06in}\includegraphics[width=0.30\textwidth]{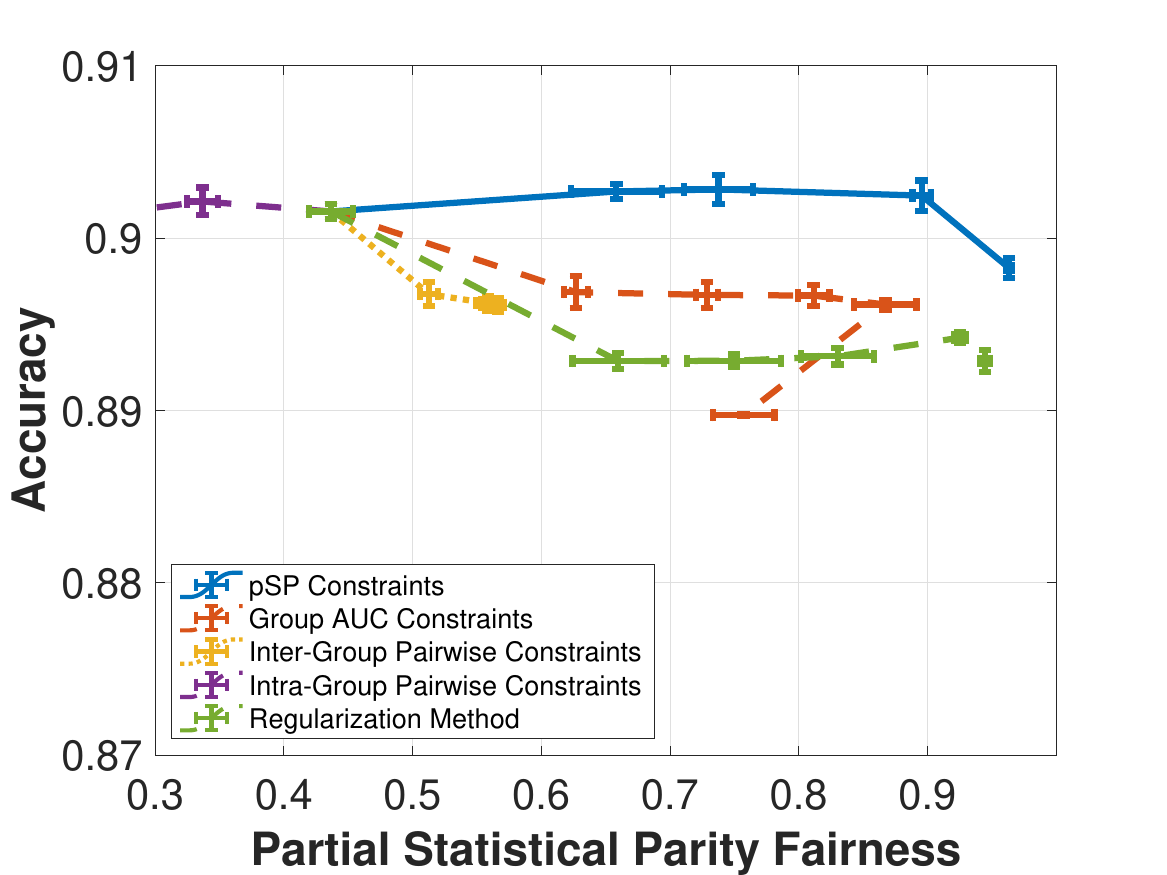}
		& \hspace*{-0.06in}\includegraphics[width=0.30\textwidth]{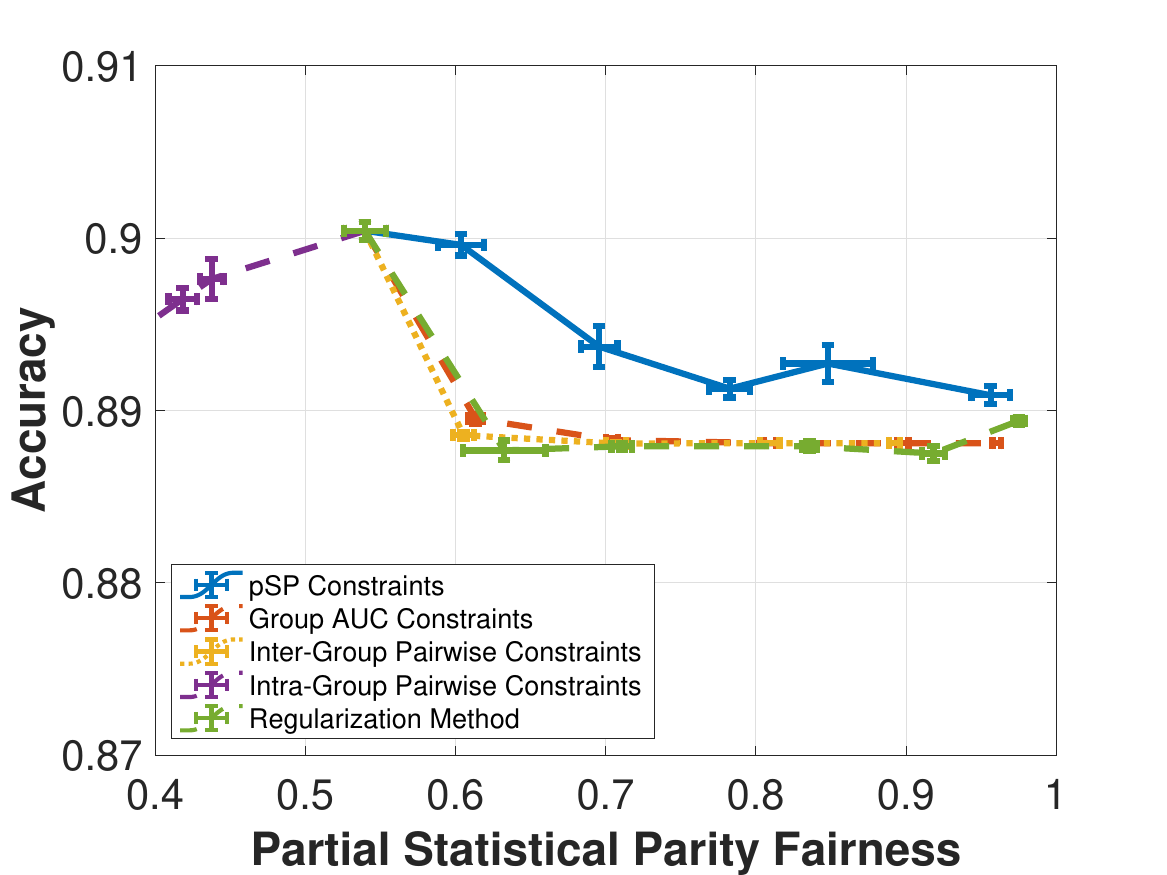}
         \\
	\raisebox{12ex}{\small{\rotatebox[origin=c]{90}{pDP fairness}}}
		& \hspace*{-0.06in}\includegraphics[width=0.30\textwidth]{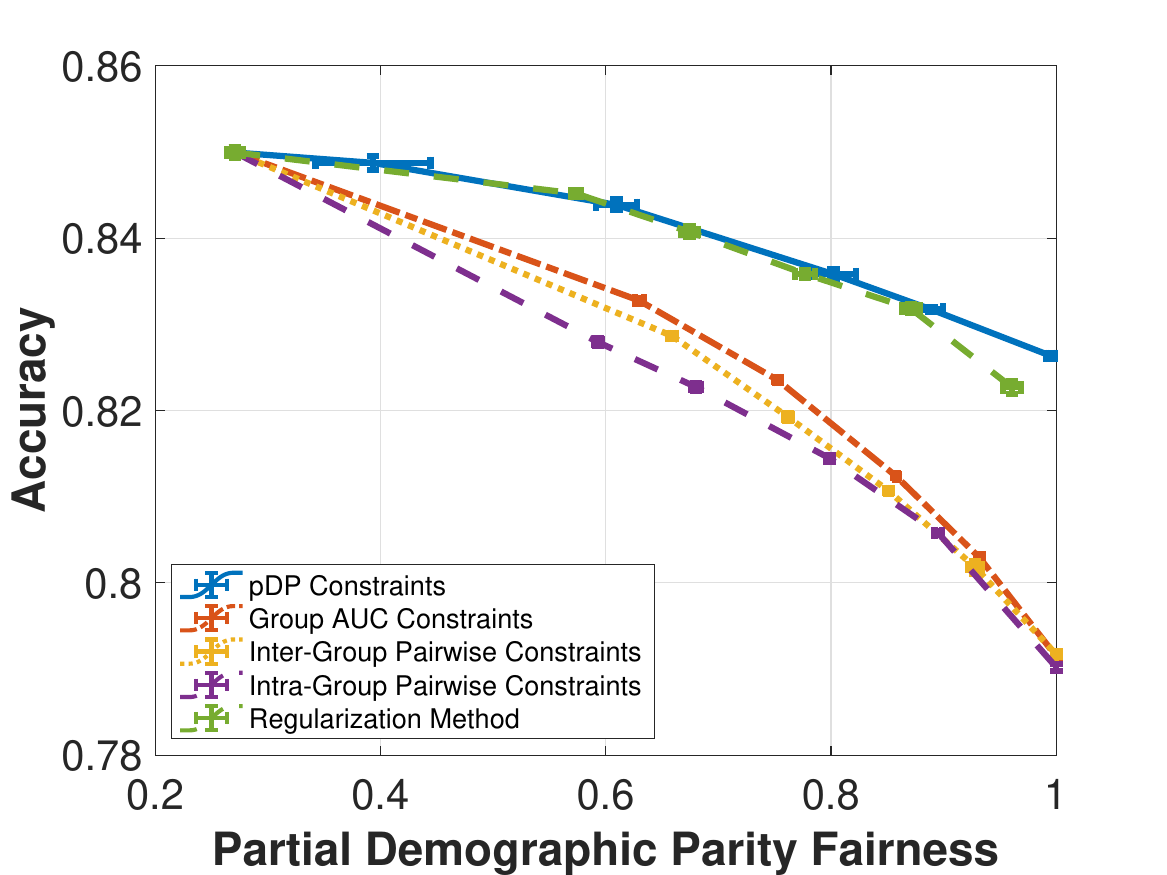}
		& \hspace*{-0.06in}\includegraphics[width=0.30\textwidth]{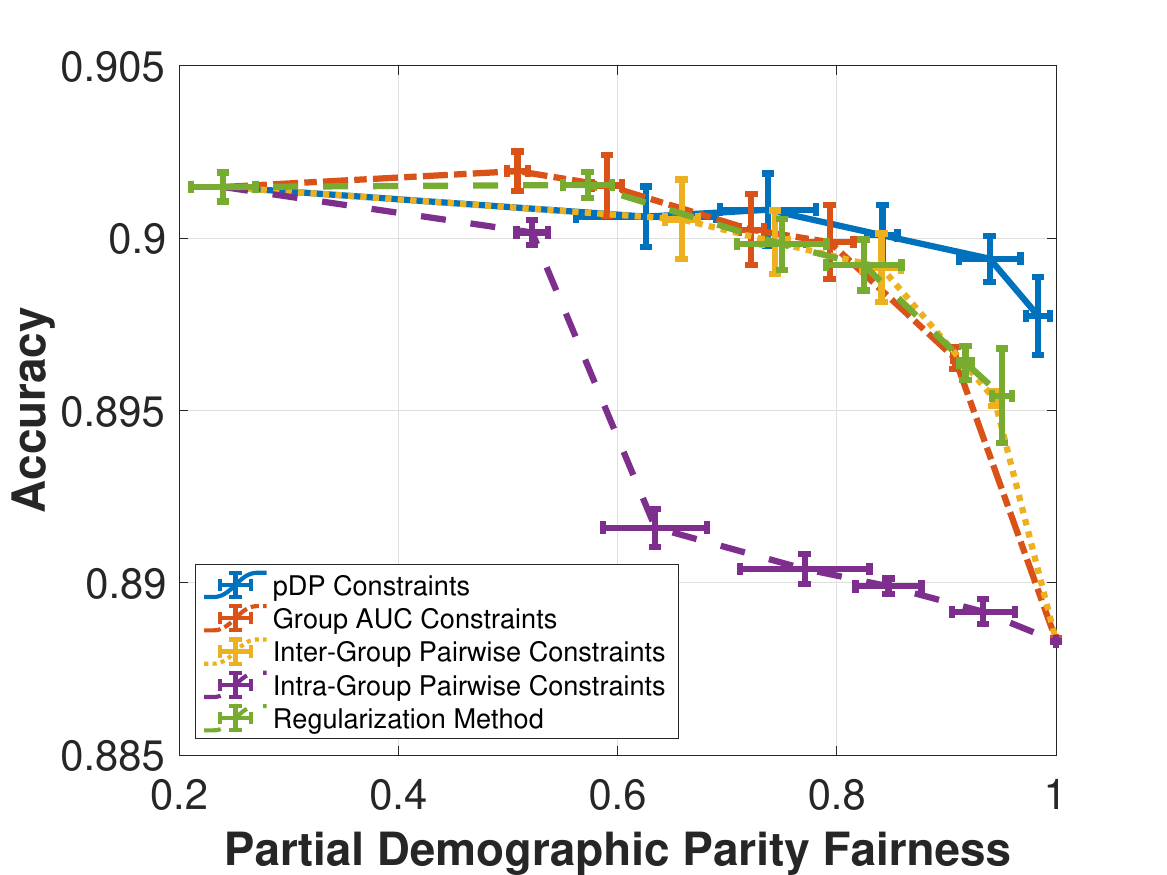}
		& \hspace*{-0.06in}\includegraphics[width=0.30\textwidth]{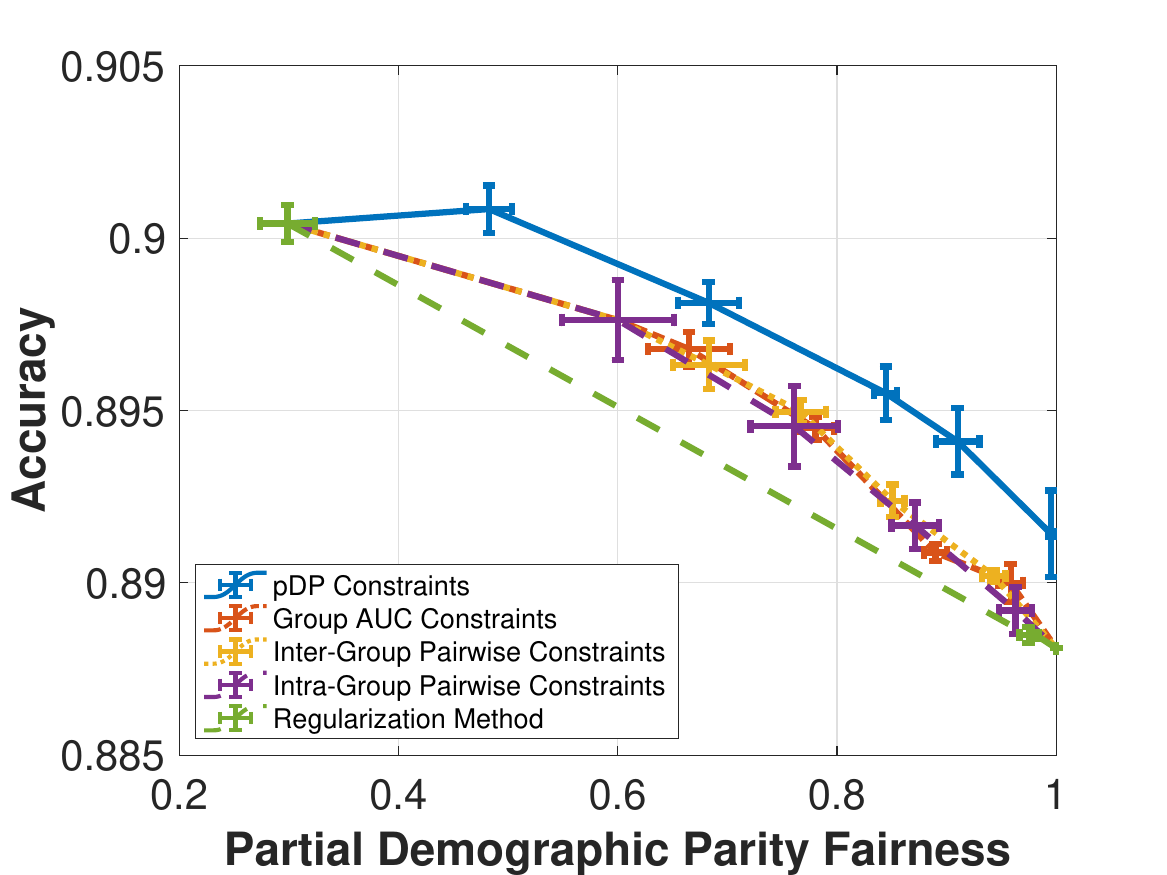}
        \end{tabular}
	\caption{Pareto frontiers showing the trade-off between classification accuracy and pSP fairness (the first row) and the trade-off between classification accuracy and pDP fairness (the second row) by each method. $\mathcal{I}$ is $[5\%, 30\%]$ for \textit{a9a}, $[0\%, 25\%]$ for \textit{bank}, and $[70\%, 100\%]$ for \textit{law school}. }  
    \label{fig:pDP_and_wpDP_efficiency_frontier}
	\vspace{-0.1in}
\end{figure*}

The strong convexity assumption in Assumption~\ref{assume:DC}A can be easily satisfied by replacing $f_i^+(\vw)$ and $f_i^-(\vw)$ with $f_i^+(\vw)+\frac{\mu}{2}\|\vw\|^2$ and $f_i^-(\vw)+\frac{\mu}{2}\|\vw\|^2$, respectively, if necessary. Assumption~\ref{assume:DC}B is the generalization of the uniform Slater's condition by~\citet{ma2020quadratically,huang2023oracle}. Two real-world examples satisfying this assumption, including a fair classification problem, are given~by \citet{huang2023oracle}, where $f_i^-(\vw)=\frac{\rho}{2}\|\vw\|^2$ for some $\rho>0$. Assumption~\ref{assume:DC}C and D are common in the literature. Assumption~\ref{assume:DC}E is not restricted because any $\vw_{\text{feas}}$ making $h_{\vw_{\text{feas}}}(\vxi)\equiv 0$ is a trivially fair model and thus satisfies most fairness constraints, including \eqref{eq:inprocess_pdp_approx} and \eqref{eq:inprocess_wpdp_approx}. Under Assumption~\ref{assume:DC}B, subproblem \eqref{DCA} is feasible as long as $\vw^{(k)}$ is feasible to \eqref{DC}. 

By a proof similar to~\citet[Theorem 3.2.3]{nesterov2018lectures}, the oracle complexity of Algorithm~\ref{alg:swg} for finding $\vw^{(k+1)}$ in \eqref{eq:epsilonwk} is known as follows. The proof is provided in Section~\ref{sec:ssg}.
\begin{proposition}
\label{thm:ssg}~For any $k$, if $f_i(\vw^{(k)})\leq 0$ for $i\in[m]$ and $\vw^{(k+1)}=\textup{SSG}(\vw^{(k)},\epsilon_k,T_k)$ with  $T_k\geq\frac{M^2}{\epsilon_k\mu}\ln\left(1+\frac{\mu \|\vw^{(k)}-\widehat\vw^{(k+1)}\|^2}{\epsilon_k}\right)$, $\vw^{(k+1)}$ satisfies \eqref{eq:epsilonwk}.
\end{proposition}
The oracle complexity of Algorithm~\ref{alg:dca} for finding a nearly $\epsilon$-KKT point of \eqref{DC} is as follows. The proof is in Section~\ref{sec:idca}.
\begin{theorem}\label{thm:idca}~Suppose 
\begin{align}
\label{eq:K}
K=&\left\lceil{2\max\left\{1,4M^2,\frac{8M^4}{\mu\nu}\right\}}\frac{\left( f_0(\vw^{(0)})-f_{\textup{lb}}\right)}{\mu\epsilon^2}\right\rceil,\\
\label{eq:epsilonk}
\epsilon_k=&~ \frac{\mu}{8}\min\left\{1,\frac{1}{4M^2},\frac{\mu\nu}{8M^4}\right\}\cdot\epsilon^2,
\end{align}
and set $\vw^{(k+1)}=\textup{SSG}(\vw^{(k)},\epsilon_k,T_k)$  along with $T_k=\frac{M^2}{\epsilon_k\mu} \ln\left(1+\frac{\mu \|\vw^{(k)}-\widehat\vw^{(k+1)}\|^2}{\epsilon_k}\right)$. Algorithm~\ref{alg:dca} finds a nearly $\epsilon$-KKT point for \eqref{DC} with oracle complexity $\tilde{O}(\epsilon^{-4})$.
\end{theorem}

\begin{figure*}[tb]
     \begin{tabular}{@{}c|cccc@{}}
     \toprule
      & unconstrained & $\kappa = 0.1$ & $\kappa = 0.05$ & $\kappa = 0.01$ \\
		\raisebox{7ex}{\small{\rotatebox[origin=c]{90}{a9a}}}
		& \hspace*{-0.07in}\includegraphics[width=0.22\textwidth]{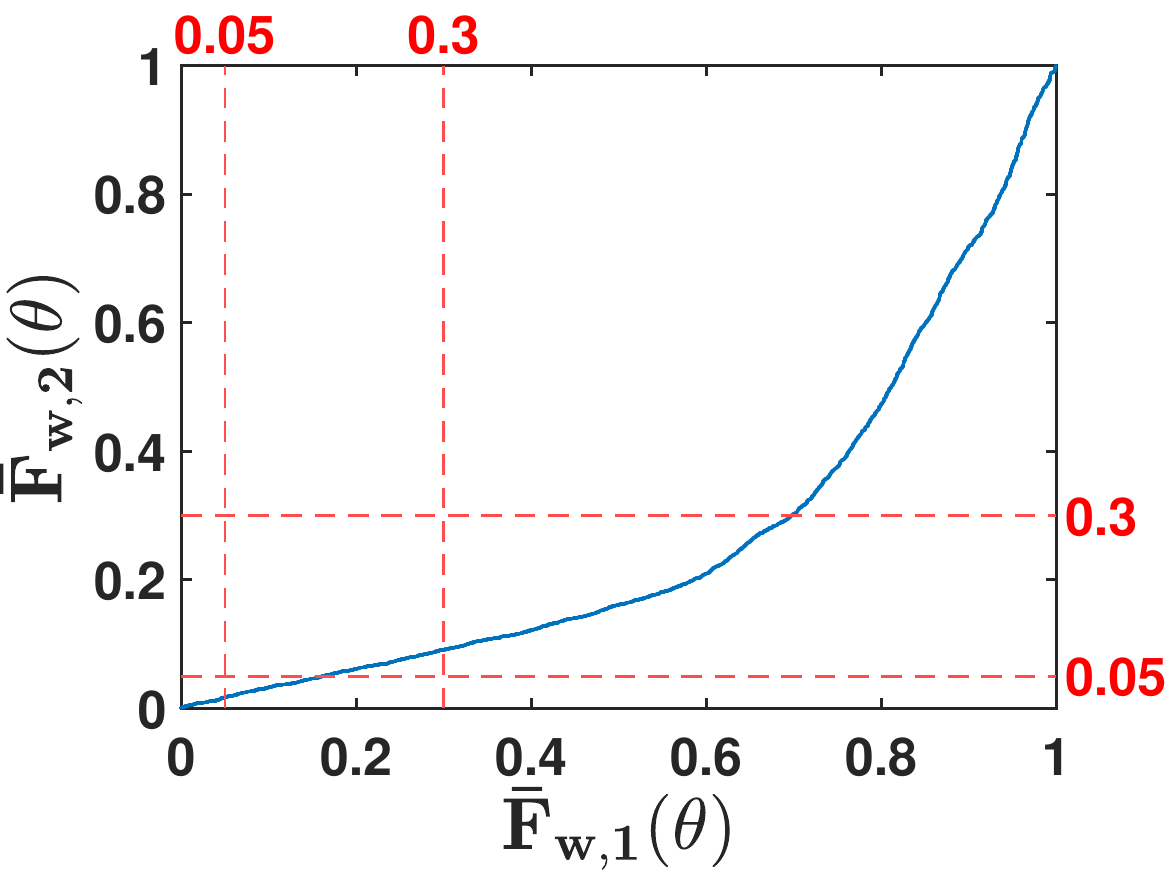}
		& \hspace*{-0.09in}\includegraphics[width=0.22\textwidth]{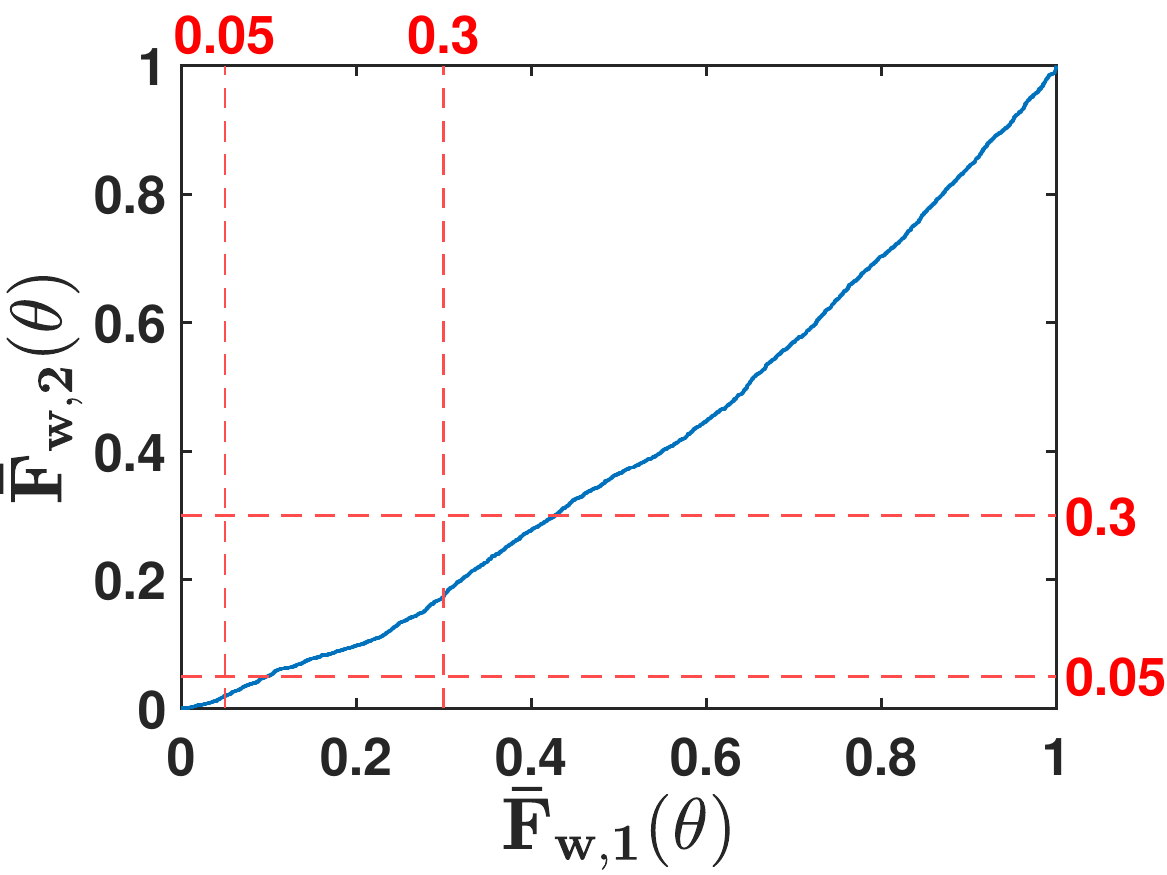}
		& \hspace*{-0.09in}\includegraphics[width=0.22\textwidth]{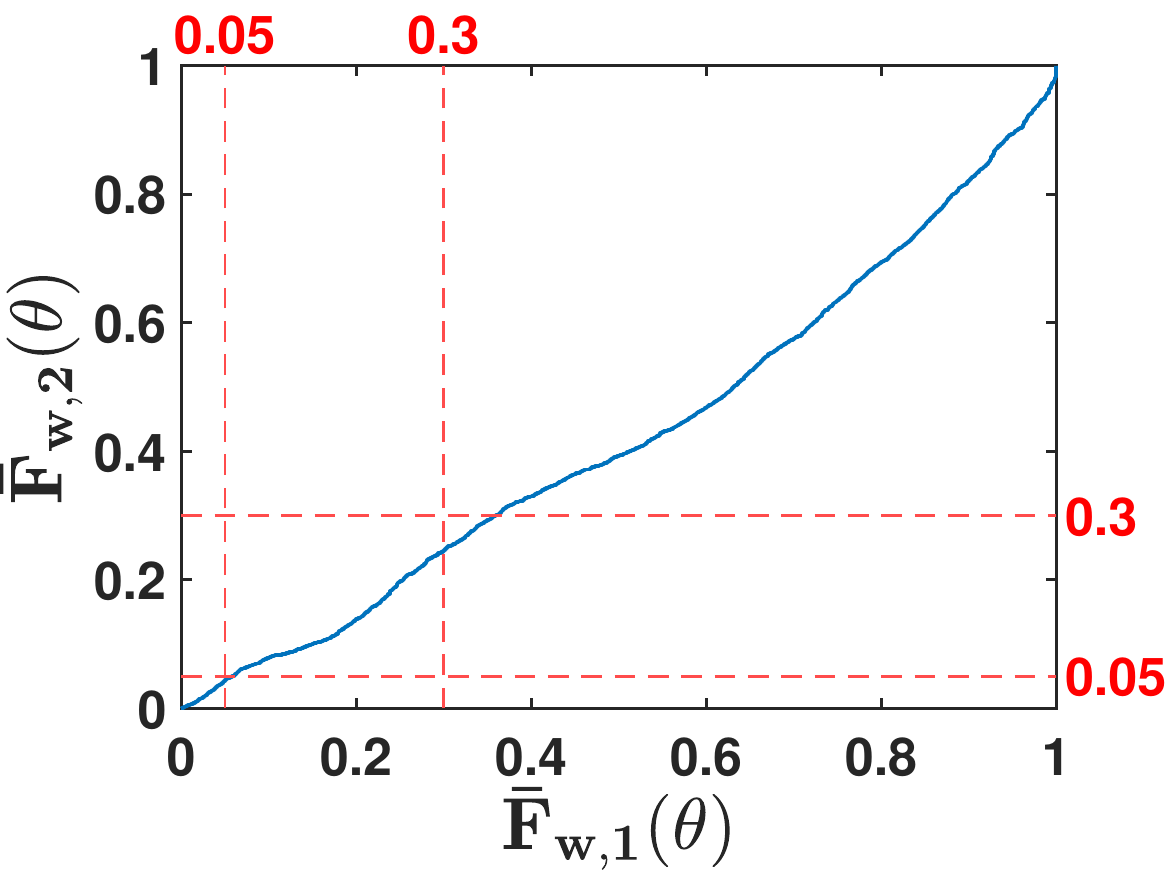}
		& \hspace*{-0.09in}\includegraphics[width=0.22\textwidth]{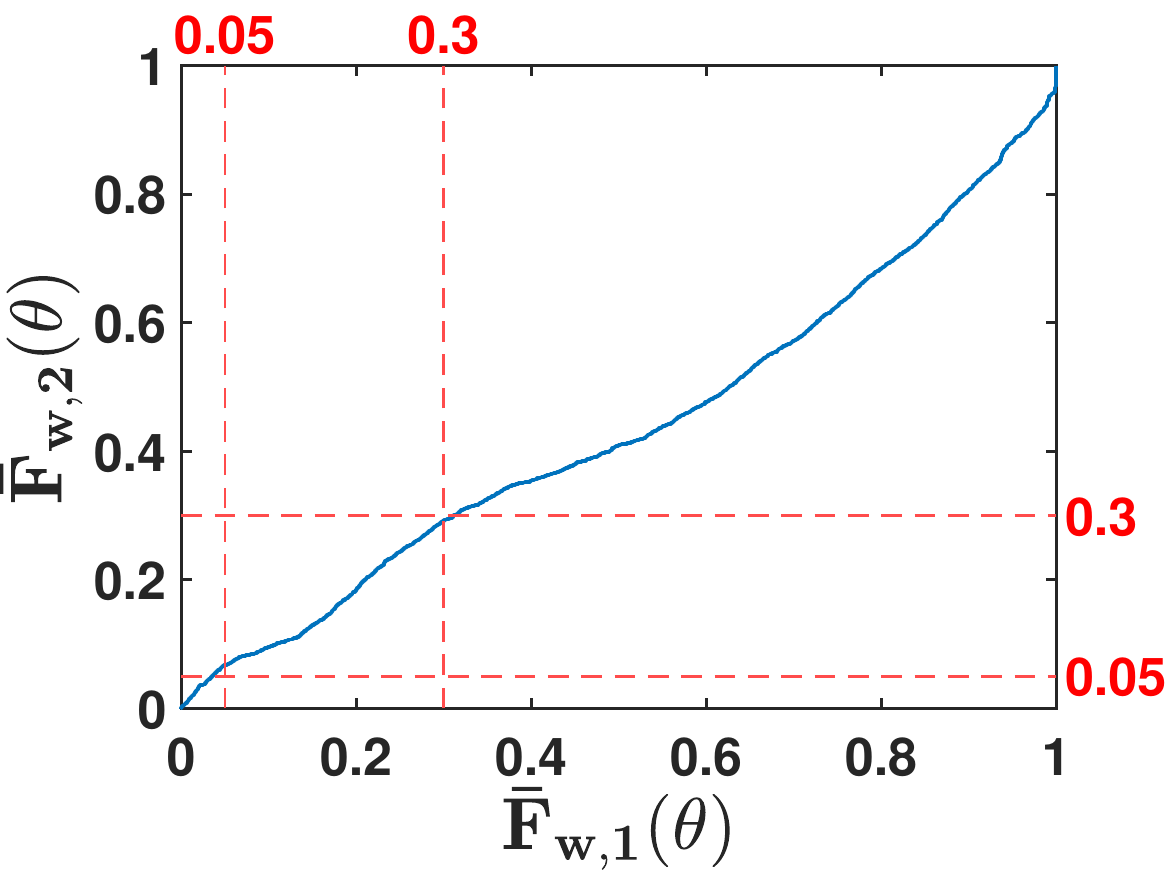}\\
        
        \midrule

        & unconstrained & $\kappa = 0.3$ & $\kappa = 0.08$ & $\kappa = 0.01$ \\
		\raisebox{7ex}{\small{\rotatebox[origin=c]{90}{bank}}}
		& \hspace*{-0.07in}\includegraphics[width=0.22\textwidth]{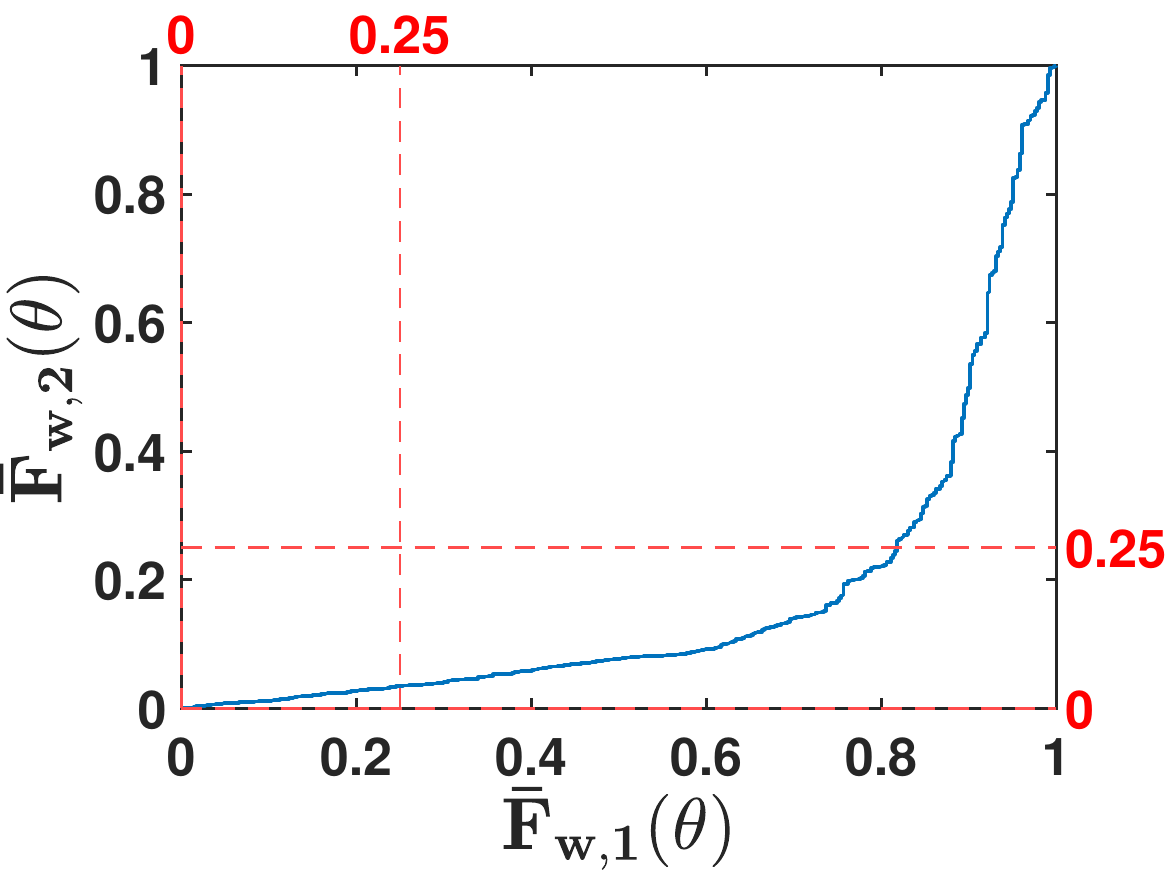}
		& \hspace*{-0.09in}\includegraphics[width=0.22\textwidth]{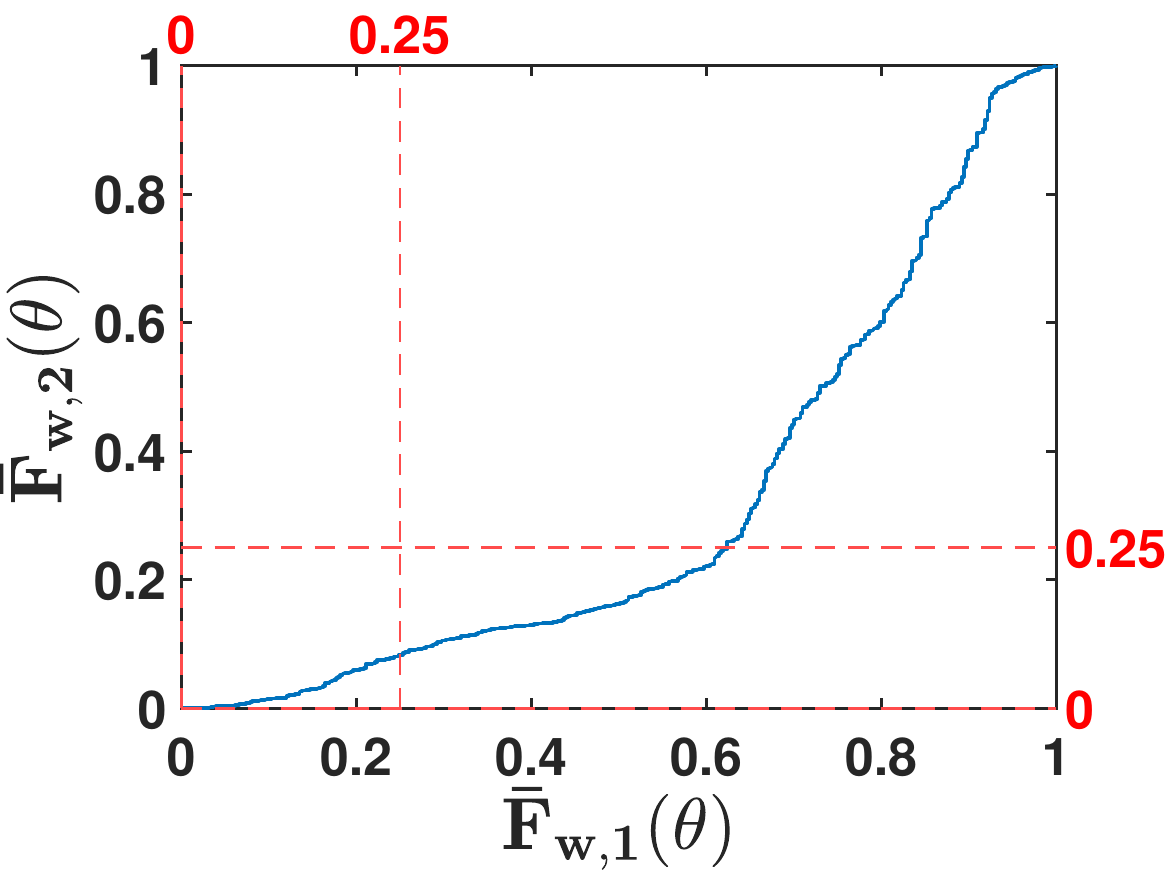}
		& \hspace*{-0.09in}\includegraphics[width=0.22\textwidth]{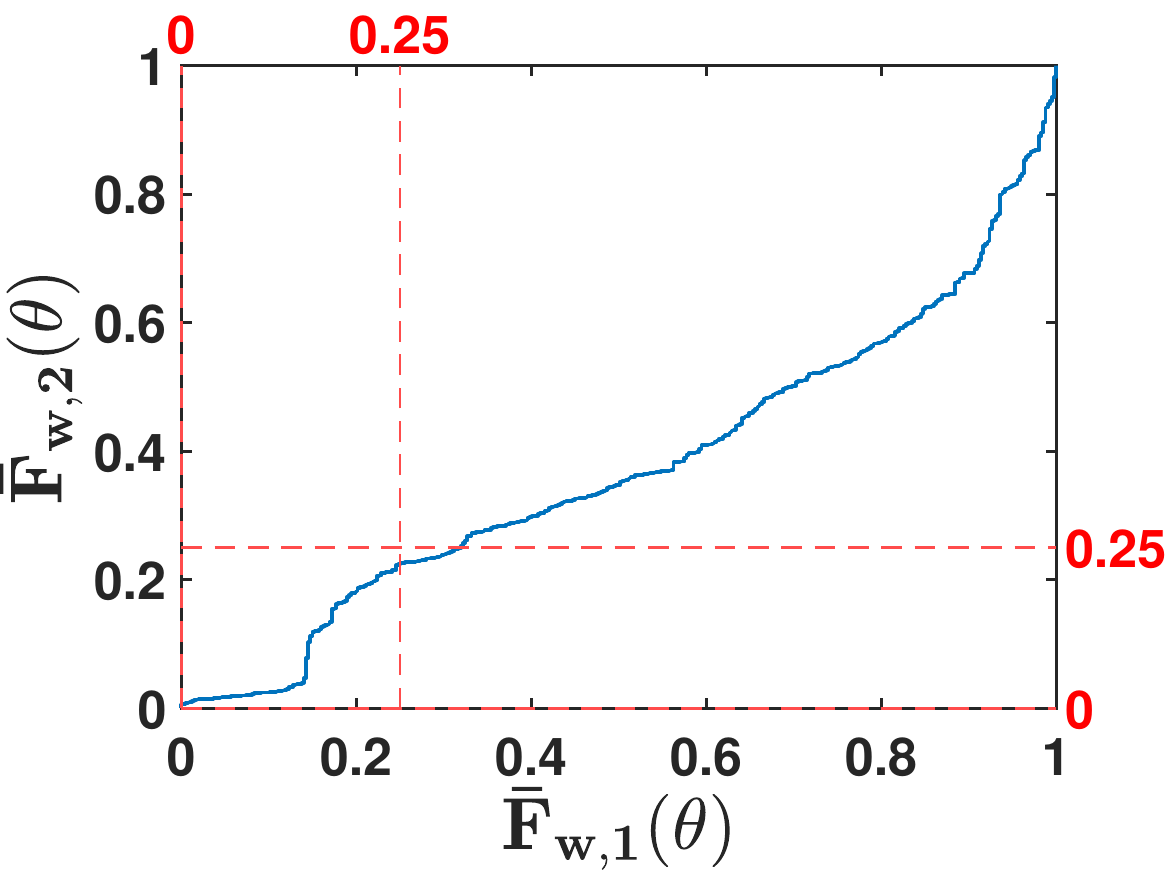}
		& \hspace*{-0.09in}\includegraphics[width=0.22\textwidth]{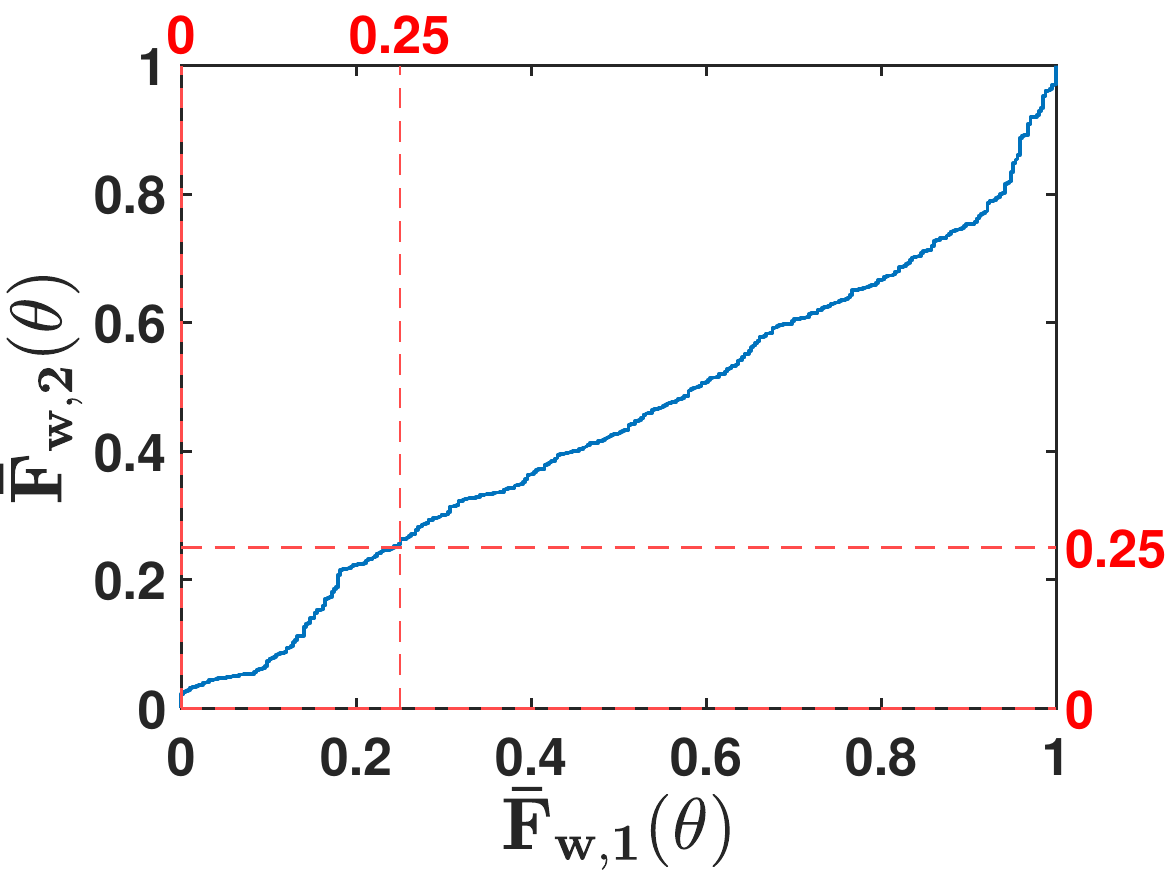}\\
        \midrule

        & unconstrained & $\kappa = 0.2$ & $\kappa = 0.1$ & $\kappa = 0.005$ \\
		\raisebox{7ex}{\small{\rotatebox[origin=c]{90}{law school}}}
		& \hspace*{-0.07in}\includegraphics[width=0.22\textwidth]{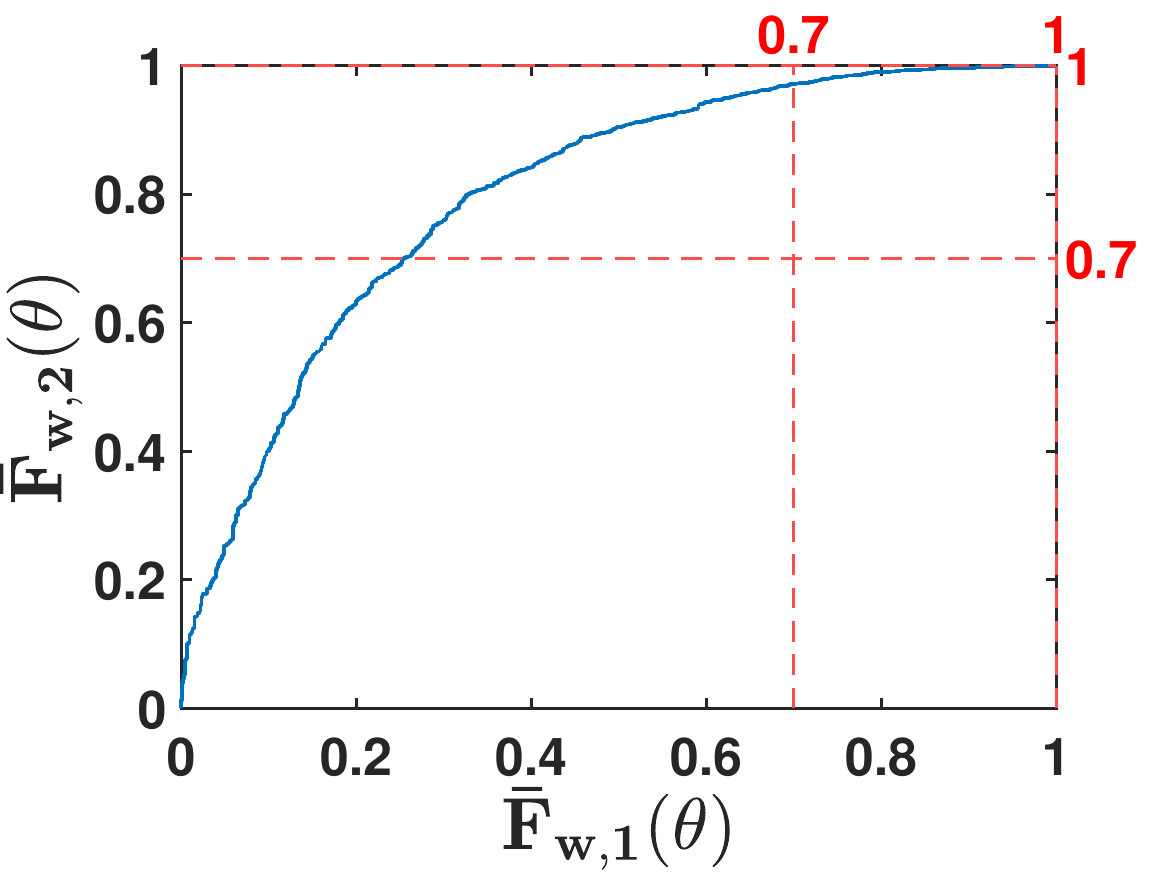}
		& \hspace*{-0.09in}\includegraphics[width=0.22\textwidth]{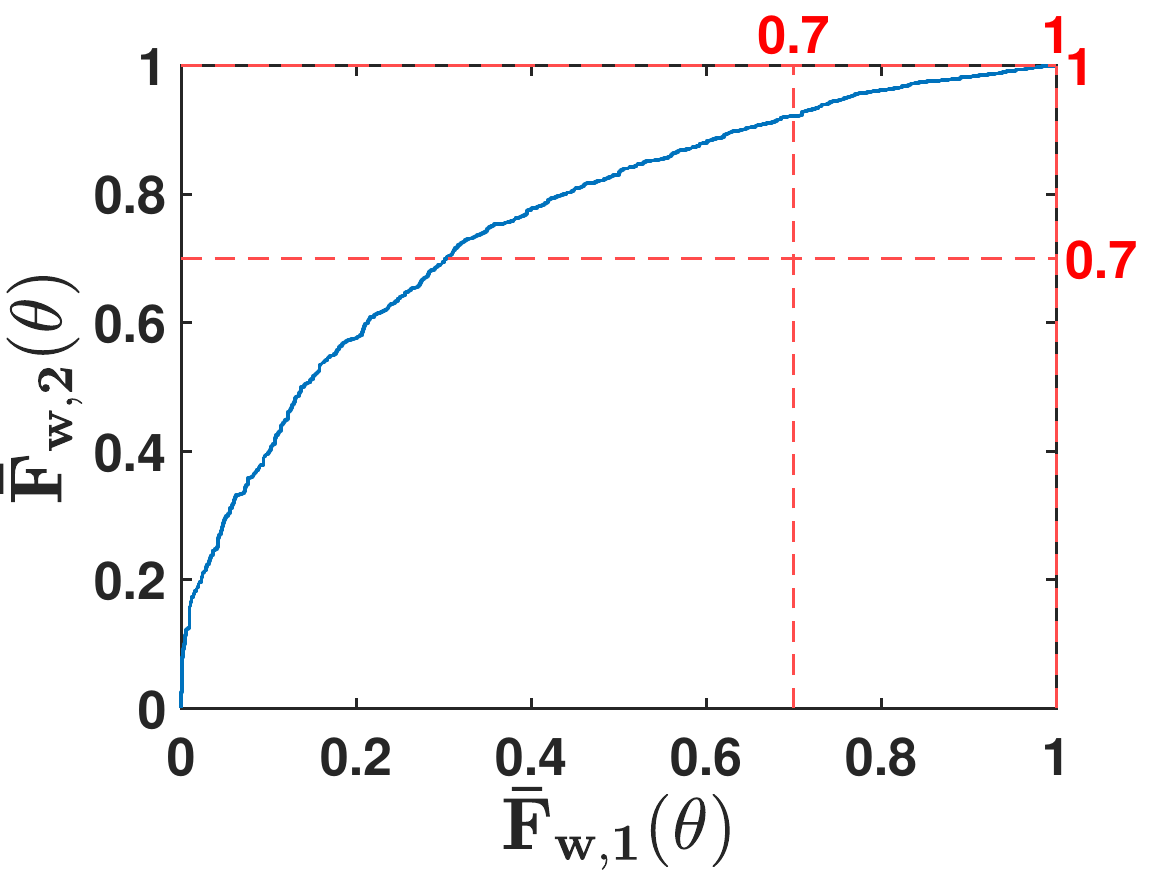}
		& \hspace*{-0.09in}\includegraphics[width=0.22\textwidth]{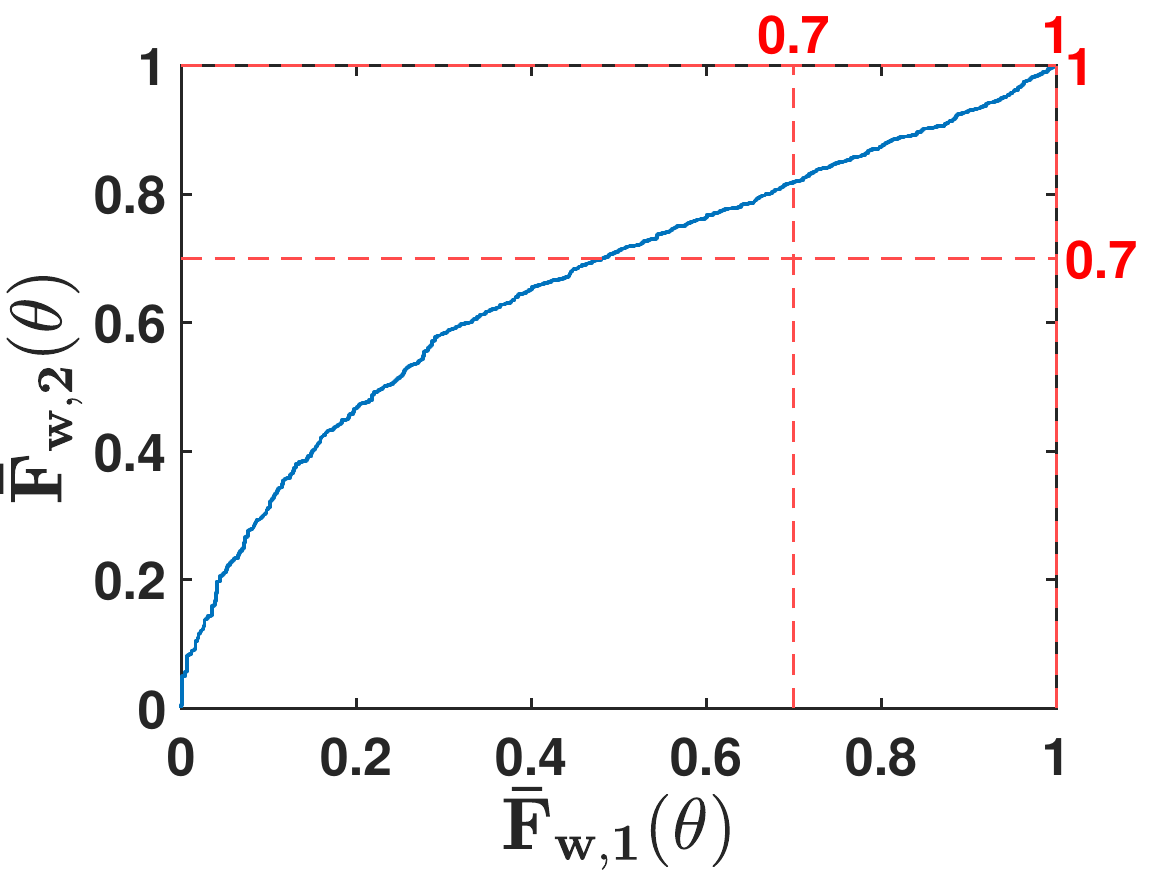}
		& \hspace*{-0.09in}\includegraphics[width=0.22\textwidth]{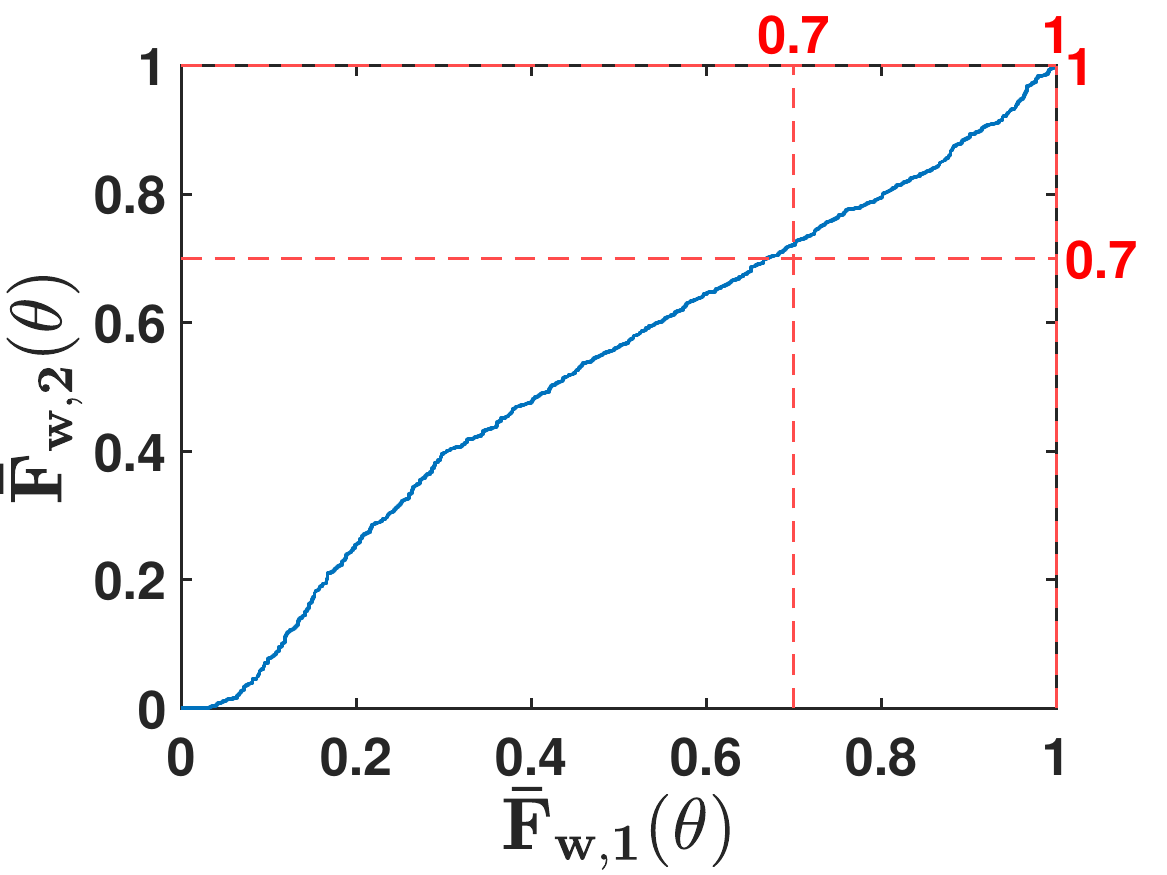}
        \end{tabular}
	\vspace{-0.05in}
	\caption{ How the predicted positive rates in both groups ($\bar{F}_{\vw,1}(\theta)$ and $\bar{F}_{\vw,2}(\theta)$) change with threshold $\theta$ when using different $\kappa$ in \eqref{eq:inprocess_pdp_approx} by the proposed method. $\mathcal{I}$ is $[5\%, 30\%]$ for \textit{a9a}, $[0\%, 25\%]$ for \textit{bank}, and $[70\%, 100\%]$ for \textit{law school} (between the two red numbers marked on both axises). }  
    \label{fig:pDP_accuracy_comparing_preficted}
	 \vspace{-0.1in}
\end{figure*}

\vspace{-0.2cm}
\section{Numerical Experiments}
\label{sec:exp}
In this section, we demonstrate the effectiveness of the proposed approach for training partially fair models for binary classification. The experiments are conducted on a MacBook Air (Apple M1 chip, 8-core CPU and 8GB of memory) in MATLAB using three publicly available datasets: \textit{a9a}~\citep{chang2011libsvm, kohavi1996scaling}, \textit{bank}~\citep{chang2011libsvm, moro2014data}, \textit{law school~}\citep{wightman1998lsac} and \textit{American Community Survey (ACS)} dataset \citep{ding2021retiring}. More details of the datasets can be found in Section~\ref{sec:data}.

We compare our methods with other in-processing methods that solve constrained optimization problems with the same objective function as our approach but different fairness constraints, including the constraints that enforce \textbf{Group AUC Fairness}~\citep{yao2023stochastic,vogel2021learning}, \textbf{Inter-Group  Pairwise Fairness}~\citep{kallus2019fairness,beutel2019fairness}, and \textbf{Intra-Group Pairwise Fairness}~\citep{beutel2019fairness}. We present the optimization models of these methods and their training procedures in Section~\ref{sec:benchmark}. Additionally, we compare our model with a regularization method whose details are given in Section~\ref{sec:regularization_method}. Using Pareto frontiers, we compare the efficiency of these methods in trading off prediction performance and partial fairness of the output models. Additional numerical results are presented in Sections \ref{sec:density}, \ref{sec:diffI} and \ref{sec:expopt}.


\subsection{Partial statistical parity}
\label{sec:pdpexp}
We consider a linear model with cross terms, i.e., $h_{\vw}(\vxi) =\left\langle \vw,(1,\vxi,\gamma,\gamma\cdot\vxi)\right\rangle$, and optimize its classification accuracy subject to \textbf{partial statistical parity (pSP) constraints} with different tolerances to violation. In particular, we apply IDCA (Algorithm~\ref{alg:dca}) to \eqref{eq:inprocess_pdp_approx} with $f_0(\vw)$ being the empirical loss in \eqref{eq:erm}, $\ell$ being the logistic loss in Example~\ref{ex:objerm} and $\cW=\mathbb{R}^{2d+2}$. We choose $\sigma(x) = \max\{ x+\frac{1}{2}, 0 \} - \max \{ x - \frac{1}{2}, 0 \}$ in \eqref{eq:inprocess_pdp_approx} just as Example~\ref{ex:hingesurrogate_pdp}. The interval $\mathcal{I}=[\alpha,\beta)$ is chosen for each dataset in the way described in Section~\ref{sec:partialfairness} with $p = \text{Pr}(h_{\vw^*}(\vxi) > 0)$, and is stated in the caption of each figure. We set $\widehat{\mathcal{I}}$ to be a set of ten equally spaced values in $\mathcal{I}$. To create the Pareto frontier, we choose different values for $\kappa$ in \eqref{eq:inprocess_pdp_approx} (see Section~\ref{sec:kappa}). The implementation details, such as choices of tuning parameters, for IDCA are presented in Section~\ref{sec:parameter_IDCA}.

For each model returned by each method, we evaluate its classification accuracy and fairness in terms of $1-\max_{k,k}\text{SP}_{k,k'}^{\mathcal{I}}(\vw)$ on the testing set. We randomly split the data into training, validation, and testing sets with five random seeds, and evaluate the means of both performance metrics along with their 95\% confidence intervals (error bars). The results are presented in Figure~\ref{fig:pDP_and_wpDP_efficiency_frontier} (pSP fairness), where the $y$-axis represents accuracy and the $x$-axis represents fairness. Note that the Pareto frontier of each method starts from the same unconstrained solution, i.e., $\vw^*=\argmin_{\vw\in\cW}f_0(\vw)$. On all three datasets, our method achieves a higher accuracy than all the baselines at all levels of fairness, showing the effectiveness of our methods in trading accuracy for partial fairness. The same experiment is also conducted on the ACS dataset and the results are presented in Section~\ref{sec:ACSresult}.

We also evaluate the effectiveness and robustness of our method on a deep learning task. Specifically, we consider an image classification problem formulated as \eqref{eq:inprocess_pdp_approx} on the CelebA dataset~\citep{liu2015deep}. In this setting, $h_{\vw}(\vxi)$ is set as a ResNet-18 model (a CNN pretrained on ImageNet) and fine-tuned to predict the “Attractive” attribute, while enforcing a partial statistical parity constraint with gender as the sensitive attribute. Experimental details and results are provided in Section~\ref{sec:CelebA}.




To demonstrate the effectiveness of our method in enforcing similar distributions of scores between groups, we compare the predicted positive rates between the groups at different thresholds $\theta$. Specifically, similar to an ROC curve, we plot $\bar{F}_{\vw,2}(\theta)$ against $\bar{F}_{\vw,1}(\theta)$ (see Definition~\ref{def:rank}) in Figure~\ref{fig:pDP_accuracy_comparing_preficted} with $\theta$ changing between $+\infty$ and $-\infty$. We show the curves of the unconstrained model $\vw^*$ and the models developed by our method with different $\kappa$ in \eqref{eq:inprocess_pdp_approx}. According to Figure~\ref{fig:pDP_accuracy_comparing_preficted}, the unconstrained model is highly unfair as $\bar{F}_{\vw,1}(\theta)$ and $\bar{F}_{\vw,2}(\theta)$ increase at very different speeds within the targeted interval $\mathcal{I}$ (between the two red numbers marked on both axises). In our method, as $\kappa$ decreases, $\bar{F}_{\vw,1}(\theta)$ and $\bar{F}_{\vw,2}(\theta)$ become closer and eventually increase at a similar speed, producing a curve of slope close to one (45-degree angle) within $\mathcal{I}$, which indicates the pSP fairness is mostly achieved.

\vspace{-0.2cm}
\subsection{Partial demographic parity}\label{sec:wpdpexp}
We consider the same linear model as in Section~\ref{sec:pdpexp}. This time, we want to optimize its classification accuracy subject to \textbf{partial demographic parity (pDP) constraints} with different tolerances to violation. In particular, we apply IDCA (Algorithm~\ref{alg:dca}) to \eqref{eq:inprocess_wpdp_approx} with $\widehat{\theta}=0$, where $f_0(\cdot)$, $\sigma(\cdot)$, $\mathcal{I}$, and $\cW$ are the same as in Section~\ref{sec:pdpexp}. The values of $\kappa$ used to create the Pareto frontiers can be found in Section~\ref{sec:kappa} and the implementation details of IDCA are given in Section~\ref{sec:parameter_IDCA}.



For each model returned by each method, we evaluate its classification accuracy and fairness in terms of $1-\max_{k,k}\text{DP}_{k,k'}^{\mathcal{I}}(\vw)$ on the testing set. The data is split in the same way as in Section~\ref{sec:pdpexp} to create the error bars. The results are presented in Figure~\ref{fig:pDP_and_wpDP_efficiency_frontier} (pDP fairness), where the $y$-axis represents accuracy and the $x$-axis represents fairness. We observe that, on the \textit{law school} dataset, our method achieves a higher accuracy than all the baselines for all levels of fairness. On the \textit{a9a} and \textit{bank} datasets, our method achieves the highest accuracy at the high levels of fairness (e.g., $\geq 0.85$) and achieves similar accuracy to the baselines at the medium and low levels of fairness.

\vspace{-0.2cm}
\section{Conclusion}
\label{sec:conclusion}
We introduced a novel in-processing framework for achieving partial fairness in machine learning only in critical regions, providing more flexibility than enforcing fairness across the entire score range. We formulate the problem as DC optimization with partial fairness constraints. The IDCA is introduced as a training algorithm, for which we prove novel theoretical guarantees and demonstrate strong empirical performance. 

\section*{Acknowledgment}
The authors would like to thank all the anonymous reviewers for their valuable comments and suggestions. This work was supported by the National Science Foundation under Grant No. 2147253 and No. 2246757.





%
%

\bibliography{ref}

\section*{Checklist}



\begin{enumerate}

  \item For all models and algorithms presented, check if you include:
  \begin{enumerate}
    \item A clear description of the mathematical setting, assumptions, algorithm, and/or model. [Yes]
    \item An analysis of the properties and complexity (time, space, sample size) of any algorithm. [Yes]
    \item (Optional) Anonymized source code, with specification of all dependencies, including external libraries. [Yes]
  \end{enumerate}

  \item For any theoretical claim, check if you include:
  \begin{enumerate}
    \item Statements of the full set of assumptions of all theoretical results. [Yes]
    \item Complete proofs of all theoretical results. [Yes]
    \item Clear explanations of any assumptions. [Yes]     
  \end{enumerate}

  \item For all figures and tables that present empirical results, check if you include:
  \begin{enumerate}
    \item The code, data, and instructions needed to reproduce the main experimental results (either in the supplemental material or as a URL). [Yes]
    \item All the training details (e.g., data splits, hyperparameters, how they were chosen). [Yes]
    \item A clear definition of the specific measure or statistics and error bars (e.g., with respect to the random seed after running experiments multiple times). [Yes]
    \item A description of the computing infrastructure used. (e.g., type of GPUs, internal cluster, or cloud provider). [Yes]
  \end{enumerate}

  \item If you are using existing assets (e.g., code, data, models) or curating/releasing new assets, check if you include:
  \begin{enumerate}
    \item Citations of the creator If your work uses existing assets. [Yes]
    \item The license information of the assets, if applicable. [Yes]
    \item New assets either in the supplemental material or as a URL, if applicable. [Not Applicable]
    \item Information about consent from data providers/curators. [Yes]
    \item Discussion of sensible content if applicable, e.g., personally identifiable information or offensive content. [Not Applicable]
  \end{enumerate}

  \item If you used crowdsourcing or conducted research with human subjects, check if you include:
  \begin{enumerate}
    \item The full text of instructions given to participants and screenshots. [Not Applicable]
    \item Descriptions of potential participant risks, with links to Institutional Review Board (IRB) approvals if applicable. [Not Applicable]
    \item The estimated hourly wage paid to participants and the total amount spent on participant compensation. [Not Applicable]
  \end{enumerate}

\end{enumerate}

\clearpage
\appendix
\thispagestyle{empty}

\onecolumn
\aistatstitle{Enforcing Fair Predicted Scores on Intervals of Percentiles by Difference-of-Convex Constraints: \\
Supplementary Materials}

\tableofcontents

\section{Extensions to group AUC fairness, equality of opportunity and equality of odds}\label{sec: gAUC}
Suppose the data points are ranked in descending order based on their scores $h_{\vw}(\vxi)$. \emph{Group AUC fairness} \citep{yao2023stochastic,vogel2021learning} requires that, with a probability of $50\%$, a random data point sampled from any group is not ranked lower than a random data point sampled from any other group,  that is, 
\begin{align}
\label{eq:gauc}
\text{GAUC}_{k,k'}(\vw):=\left|\text{Pr}(h_{\vw}(\vxi)\geq h_{\vw}(\vxi') | \gamma=k , \gamma'=k' )-0.5\right|=0,~\forall k, k'\in\mathcal{G},
\end{align}
where $ (\vxi,\zeta,\gamma)$ and $ (\vxi',\zeta',\gamma')$ are two i.i.d. random data points.  

Similar to Definitions~\ref{def:pdp} and \ref{def:wpdp}, we can also generalize group AUC fairness in \eqref{eq:gauc} by only considering the comparison between random data points sampled from some rank ranges of different groups. 
\begin{definition}
\label{def:pgauc}
We say $h_{\vw}(\vxi)$ satisfies the \emph{partial group AUC fairness with respect to percentile interval} $\mathcal{I}:=[\alpha, \beta)\subset[0,1]$ if 
\begin{align}
\label{eq:pgauc}
\text{GAUC}_{k,k'}^{\mathcal{I}}(\vw):=&\left|\text{Pr}\left(h_{\vw}(\vxi)\geq h_{\vw}(\vxi') \bigg| 
\gamma=k , \gamma'=k', 
\bar{F}_{\vw,k}(h_{\vw}(\vxi))\in\mathcal{I},
\bar{F}_{\vw,k'}(h_{\vw}(\vxi'))\in\mathcal{I} 
\right)- 0.5\right|=0,
\end{align}
for any $k$ and $k'$ in $\mathcal{G}$, where $ (\vxi,\zeta,\gamma)$ and $ (\vxi',\zeta',\gamma')$ are two i.i.d. random data points.
\end{definition}
When $\mathcal{I}=[0,1]$ for each $k\in\cG$, \eqref{eq:pgauc} is reduced to \eqref{eq:gauc}.

Consider a binary classification task where $\zeta\in\{1,-1\}$. Similar to demographic parity in \eqref{eq:ndp}, we say \textbf{equality of opportunity}~\citep{hardt2016equality} holds with respect to a decision threshold $\widehat\theta$ if
\begin{align}
\text{EOP}_{k,k'}(\vw)=&\left|\text{Pr}(h_{\vw}(\vxi) >\widehat\theta | \gamma=k,\zeta=1)-\text{Pr}(h_{\vw}(\vxi) > \widehat\theta | \gamma=k',\zeta=1 )\right|\label{eq:ndpeop}
=~0,~\forall k, k'\in\mathcal{G}.
\end{align}
Similarly, we say \textbf{equality of odds}~\citep{hardt2016equality} holds with respect to a decision threshold $\widehat\theta$ if
\begin{align}
\text{EOD}_{k,k'}(\vw)
:=\left|\text{Pr}(h_{\vw}(\vxi) >\widehat\theta | \gamma=k,\zeta=c )-\text{Pr}(h_{\vw}(\vxi) > \widehat\theta | \gamma=k',\zeta=c )\right|\label{eq:ndpeod}
=~0,~\forall k, k'\in\mathcal{G},~\forall c\in\{1,-1\}.
\end{align}
Once again, similar to Definitions~\ref{def:pdp} and \ref{def:wpdp}, we can also extend equality of opportunity and equality of odds above to partial fairness metrics as follows. 

\begin{definition}
\label{def:peop}
We say $h_{\vw}(\vxi)$ satisfies the \emph{partial equality of opportunity with respect to percentile interval} $\mathcal{I}=[\alpha, \beta)\subset[0,1]$ and threshold $\widehat\theta$ if 
\begin{align}
\text{EOP}_{k,k'}^{\mathcal{I}}(\vw)
:=\left|
\begin{array}{l}
\text{Pr}\big(h_{\vw}(\vxi) >\widehat\theta \big| \gamma=k, \zeta=1, \bar{F}_{\vw,k}(h_{\vw}(\vxi))\in\mathcal{I}\big)\\
-\text{Pr}\big(h_{\vw}(\vxi) > \widehat\theta  \big| \gamma=k', \zeta=1, \bar{F}_{\vw,k'}(h_{\vw}(\vxi))\in\mathcal{I}\big)
\end{array}
\right|\label{eq:peop}
=~0,~\forall k, k'\in\mathcal{G}.
\end{align}
\end{definition}

\begin{definition}
\label{def:peod}
We say $h_{\vw}(\vxi)$ satisfies the \emph{partial equality of odd  with respect to percentile interval} $\mathcal{I}=[\alpha, \beta)\subset[0,1]$ and threshold $\widehat\theta$ if 
\begin{align}
\text{EOD}_{k,k'}^{\mathcal{I}}(\vw)
:=\max_{c \in \{ -1, 1\}}\left|
\begin{array}{l}
\text{Pr}\big(h_{\vw}(\vxi) >\widehat\theta \big| \gamma=k, \zeta=c, \bar{F}_{\vw,k}(h_{\vw}(\vxi))\in\mathcal{I}\big)\\
-\text{Pr}\big(h_{\vw}(\vxi) > \widehat\theta  \big| \gamma=k', \zeta=c, \bar{F}_{\vw,k'}(h_{\vw}(\vxi))\in\mathcal{I}\big)
\end{array}
\right|\label{eq:peod}
=~0,~\forall k, k'\in\mathcal{G}.
\end{align}
\end{definition}

Just like \eqref{eq:inprocess_pdp} and \eqref{eq:inprocess_wpdp}, an in-processing method to produce a $\vw\in\cW$  satisfying \eqref{eq:pgauc}, \eqref{eq:peop} or \eqref{eq:peod} can be implemented by solving the following optimization problems subject to partial fairness constraints:
\begin{align}
\label{eq:inprocess_pgauc}
\min_{\vw\in\cW} f_0(\vw) &~\text{  s.t. }~\text{GAUC}_{k,k'}^{\mathcal{I}}(\vw)\leq \kappa,~\forall k, k'\in\mathcal{G},\\\label{eq:inprocess_peop}
\min_{\vw\in\cW} f_0(\vw) &~\text{  s.t. }~\text{EOP}_{k,k'}^{\mathcal{I}}(\vw)\leq \kappa,~\forall k, k'\in\mathcal{G},\\\label{eq:inprocess_peod}
\min_{\vw\in\cW} f_0(\vw) &~\text{  s.t. }~\text{EOD}_{k,k'}^{\mathcal{I}}(\vw)\leq \kappa,~\forall k, k'\in\mathcal{G},
\end{align}
where, just as in \eqref{eq:inprocess_pdp} and \eqref{eq:inprocess_wpdp} , $f_0(\vw)$ is the loss function measuring the predictive performance of $h_{\vw}(\vxi)$.

Following a process similar to the one that derives \eqref{eq:inprocess_pdp_approx} and \eqref{eq:inprocess_wpdp_approx}, we can construct the approximation of the constraints in \eqref{eq:inprocess_pgauc}. Let $n_k^\alpha=\lceil \alpha n_k\rceil$ and  $n_k^\beta=\lceil \beta n_k\rceil$ so $n_k^\beta-n_k^\alpha$ represents the number of data points in $\mathcal{D}_k$ whose ranks are in $\mathcal{I}$. Let $h_{\vw}(\vxi_{[i]}^k)$ be the $i$-th largest score in $\{h_{\vw}(\vxi_i^k)| i=1,\dots,n_k\}$ with ties broken in an arbitrary way. 
Then, for any $k$ and $k'$ in $\mathcal{G}$, $\text{GAUC}_{k,k'}^{\mathcal{I}}(\vw)$ defined in \eqref{eq:pgauc} can be approximated as 
\begin{equation*}
        \left|\frac{\sum_{i=n_k^\alpha+1}^{n_k^\beta}\sum_{j=n_{k'}^\alpha+1}^{n_{k'}^\beta} \mathbf{1}_{h_{\vw}(\vxi_{[i]}^k) \geq h_{\vw}(\vxi_{[j]}^{k'})}}{(n_k^\beta-n_k^\alpha)(n_{k'}^\beta-n_{k'}^\alpha)}-0.5\right|. 
\end{equation*}
Using a surrogate $\sigma$ to approximate the indicator functions above, we obtain the final approximation of  \eqref{eq:inprocess_pgauc}:
\small
\begin{align}
\label{eq:inprocess_pgauc_approx}
&~\min_{\vw\in\cW} f_0(\vw)\\\nonumber
\text{s.t.}
&~\frac{\sum_{i=n_k^\alpha+1}^{n_k^\beta}\sum_{j=n_{k'}^\alpha+1}^{n_{k'}^\beta} \sigma(h_{\vw}(\vxi_{[i]}^k) - h_{\vw}(\vxi_{[j]}^{k'}))}{(n_k^\beta-n_k^\alpha)(n_{k'}^\beta-n_{k'}^\alpha)}\leq 0.5+\kappa,~\forall k,k'\in\mathcal{G}.
\end{align}
\normalsize
Note that, depending on $\sigma$, the constraint in \eqref{eq:inprocess_pgauc_approx} may not necessarily be symmetric in $k$ and $k'$, so there can be $G^2$ constraints in general.

Following a process similar to the one we used  to derive the approximation \eqref{eq:inprocess_wpdp_approx} of \eqref{eq:inprocess_wpdp}, we can construct the approximation of \eqref{eq:inprocess_peop}. In particular, we first introduce the partition $\mathcal{D}=(\cup_{k\in\cG}\mathcal{D}_{k,+})\cup(\cup_{k\in\cG}\mathcal{D}_{k,-})$, where
\begin{align*}
\mathcal{D}_{k,+}=&\{(\vxi,\zeta,\gamma)\in\mathcal{D}|\gamma=k,\zeta=1\}=\{(\vxi_i^{k,+},1,k)\}_{i=1}^{n_{k,+}},\\
\mathcal{D}_{k,-}=&\{(\vxi,\zeta,\gamma)\in\mathcal{D}|\gamma=k,\zeta=-1\}=\{(\vxi_i^{k,-},-1,k)\}_{i=1}^{n_{k,-}},
\end{align*}
and $n_{k,+}$ and $n_{k,-}$ are the corresponding group sizes ($n_{k,+}+n_{k,-}=n_{k}$). Then given the percentile interval $\mathcal{I}=[\alpha, \beta)\subset[0,1]$ and threshold $\widehat\theta$ in \eqref{eq:peop} and \eqref{eq:peod}, we obtain the following approximation of \eqref{eq:inprocess_peop}:
\begin{align}
\label{eq:inprocess_peop _approx}
&\min_{\vw\in\cW} f_0(\vw)\\\nonumber
\text{s.t.}&
\begin{array}{l}
\min\left\{\frac{1}{n_{k,+}} \sum_{i=1}^{n_{k,+}} \sigma(h_{\vw}(\vxi_i^{k,+})-\widehat\theta),\beta\right\}\\
-\min\left\{\frac{1}{n_{k,+}} \sum_{i=1}^{n_{k,+}} \sigma(h_{\vw}(\vxi_i^{k,+})-\widehat\theta),\alpha\right\}\\
-\min\left\{\frac{1}{n_{k',+}} \sum_{i=1}^{n_{k',+}} \sigma(h_{\vw}(\vxi_i^{k',+})-\widehat\theta),\beta\right\}\\
+\min\left\{\frac{1}{n_{k',+}} \sum_{i=1}^{n_{k',+}} \sigma(h_{\vw}(\vxi_i^{k',+})-\widehat\theta),\alpha\right\}\leq \kappa(\beta-\alpha),
\end{array}\\\nonumber
&~~\forall k, k'\in\mathcal{G}.
\end{align}

Similarly, we can obtain the following approximation of \eqref{eq:inprocess_peod}:
\begin{align}
\label{eq:inprocess_peod_approx}
&\min_{\vw\in\cW} f_0(\vw)\\\nonumber
\text{s.t.}&
\begin{array}{l}
\max_{c \in \{ +, - \}}\Big\{\min\left\{\frac{1}{n_{k,c}} \sum_{i=1}^{n_{k,c}} \sigma(h_{\vw}(\vxi_i^{k,c})-\widehat\theta),\beta\right\}\\
-\min\left\{\frac{1}{n_{k,c}} \sum_{i=1}^{n_{k,c}} \sigma(h_{\vw}(\vxi_i^{k,c})-\widehat\theta),\alpha\right\}\\
-\min\left\{\frac{1}{n_{k',c}} \sum_{i=1}^{n_{k',c}} \sigma(h_{\vw}(\vxi_i^{k',c})-\widehat\theta),\beta\right\}\\
+\min\left\{\frac{1}{n_{k',c}} \sum_{i=1}^{n_{k',c}} \sigma(h_{\vw}(\vxi_i^{k',c})-\widehat\theta),\alpha\right\} \Big\}\leq \kappa(\beta-\alpha),
\end{array}\\\nonumber
&~~\forall k, k'\in\mathcal{G}.
\end{align}

\section{How to choose $\mathcal{I}$ for classification problem:}\label{sec:chooseI}
The algorithm and theoretical results we develop are applicable to any interval $\mathcal{I}$ that users choose based on their specific fairness concerns. When users do not have a targeted interval, we recommend selecting $\mathcal{I}$ to cover an estimated positive decision rate $p$. For instance, in applications such as school admissions, $p$ can be the historical admission rate. Then one may choose $\mathcal{I}$ such that $p \in \mathcal{I}$, as this score region is likely the most contested for fairness issues. To refine $\mathcal{I}$, one can first train a model without fairness constraints, i.e., $\vw^* = \arg\min_{\vw \in \cW} f_0(\vw)$, and generate a candidate set of intervals that all cover $p$. Then, among these candidates, one may select an interval $\mathcal{I}$ over which the model $h_{\vw^*}(\vxi)$ exhibits the most significant unfairness. If the historical positive decision rate $p$ is not available, one may specify a decision threshold $\widehat{\theta}$ for the predicted scores (e.g., $\widehat{\theta} = 0$ for binary classification) according to the application context, and define $p = \text{Pr}(h_{\vw^*}(\vxi) > \widehat{\theta})$.

\section{Equivalent formulations of statistical parity and demographic parity partial fairness constraints}
\label{sec:equivreform}
In this section, we present the proofs of Lemma~\ref{eq:pdp_equiv} and  Lemma~\ref{eq:wpdp_equiv}. Before that, we first present a few observations that will be used in the proofs. 

Let 
$$
\bar{F}_{\vw,k}^{-1}(p):= \inf \{ \theta \in \mathbb{R}| \bar{F}_{\vw,k}(\theta) \leq p \},
$$
which is the inverse CCDF of $h_{\vw}(\vxi)$ for group $k$. Basically, $\bar{F}_{\vw,k}^{-1}(p)$ represents the cutoff value above which the top $100p\%$  scores fall, or equivalently, $\bar{F}_{\vw,k}^{-1}(p)$ is the $(1-p)$-quantile of the conditional distribution of $h_{\vw}(\vxi)$ conditioning on $\gamma=k$. 
Therefore, we can then rewrite
\begin{align}
\nonumber
&~\text{Pr}\left(h_{\vw}(\vxi) >\theta | \gamma=k, \bar{F}_{\vw,k}(h_{\vw}(\vxi))\in\mathcal{I}\right)\\\nonumber
=&~\frac{\text{Pr}\left(h_{\vw}(\vxi) >\theta, \bar{F}_{\vw,k}(h_{\vw}(\vxi))\in\mathcal{I}|\gamma=k\right)}{\text{Pr}\left(\bar{F}_{\vw,k}(h_{\vw}(\vxi))\in\mathcal{I}|\gamma=k\right)}\\\nonumber
=&~\frac{\text{Pr}\big(h_{\vw}(\vxi) >\theta, h_{\vw}(\vxi)\in \big[\bar{F}_{\vw,k}^{-1}(\beta),\bar{F}_{\vw,k}^{-1}(\alpha)\big]\big|\gamma=k\big)}{\beta-\alpha}\\\nonumber
=&~\frac{\text{Pr}\big(h_{\vw}(\vxi) >\theta, h_{\vw}(\vxi)>\bar{F}_{\vw,k}^{-1}(\beta)\big|\gamma=k\big)}{\beta-\alpha}
-
\frac{\text{Pr}\big(h_{\vw}(\vxi) >\theta, h_{\vw}(\vxi)>\bar{F}_{\vw,k}^{-1}(\alpha)\big|\gamma=k\big)}{\beta-\alpha}\\\nonumber
=&~\frac{\text{Pr}\big(h_{\vw}(\vxi) >\max\big\{\theta,\bar{F}_{\vw,k}^{-1}(\beta)\big\} \big|\gamma=k\big)}{\beta-\alpha}
-
\frac{\text{Pr}\big(h_{\vw}(\vxi) >\max\big\{\theta,\bar{F}_{\vw,k}^{-1}(\alpha)\big\} \big|\gamma=k\big)}{\beta-\alpha}\\\label{eq:CCDF_onI_all}
=&~\frac{\min\left\{\textup{Pr}\big(h_{\vw}(\vxi) >\theta \big| \gamma=k\big),\beta\right\}}{\beta-\alpha}
-
\frac{\min\left\{\textup{Pr}\big(h_{\vw}(\vxi) >\theta \big| \gamma=k\big),\alpha\right\}}{\beta-\alpha}.
\end{align}
Since we have assumed that $h_{\vw}(\vxi)$ has an absolutely continuous distribution in each group,  \eqref{eq:CCDF_onI_all} increases monotonically and continuously from $0$ to $1$ as $\theta$ changes from $+\infty$ to $-\infty$.
Note that, when $\bar{F}_{\vw,k}(\theta)=\textup{Pr}\big(h_{\vw}(\vxi) >\theta \big| \gamma=k\big)\in [\alpha,\beta]$,  \eqref{eq:CCDF_onI_all} can be further simplified to 
\begin{align}
\label{eq:CCDF_onI}
\text{Pr}\left(h_{\vw}(\vxi) >\theta | \gamma=k, \bar{F}_{\vw,k}(h_{\vw}(\vxi))\in\mathcal{I}\right)
=\frac{\text{Pr}\left(h_{\vw}(\vxi) >\theta |\gamma=k\right)-\alpha}{\beta-\alpha}.
\end{align}

\begin{proof}[Proof of Lemma~\ref{eq:pdp_equiv}]
Suppose $\vw\in\cW$ is feasible to \eqref{eq:inprocess_pdp}, or equivalently,  
\begin{align}
\label{eq:pdpkappa}
&\left|\begin{array}{l}
\text{Pr}\left(h_{\vw}(\vxi) >\theta \big| \gamma=k, \bar{F}_{\vw,k}(h_{\vw}(\vxi))\in\mathcal{I}\right)\\[0.5mm]
-\,\text{Pr}\left(h_{\vw}(\vxi) > \theta \big| \gamma=k', \bar{F}_{\vw,k'}(h_{\vw}(\vxi))\in\mathcal{I}\right)
\end{array}
\right|\leq \kappa,
~\forall \theta\in\mathbb{R},~\forall k,k'\in\mathcal{G}.
\end{align}
Consider any $p\in[\alpha,\beta-\kappa(\beta-\alpha))$. Let $k\in\mathcal{G}$ be the group such that $\bar{F}_{\vw,k}^{-1}(p)\leq\bar{F}_{\vw,k'}^{-1}(p)$ for any $k'\in\mathcal{G}$.
Let $\theta_p=\bar{F}_{\vw,k}^{-1}(p)$ so 
\begin{align}
\label{eq:pkinterval}
\textup{Pr}\big(h_{\vw}(\vxi) >\theta_p \big| \gamma=k\big)=p\in[\alpha, \beta].
\end{align}
By~\eqref{eq:CCDF_onI_all}, it holds that
\begin{align}
\label{eq:ineqkappa1}
\text{Pr}\left(h_{\vw}(\vxi) >\theta_p | \gamma=k, \bar{F}_{\vw,k}(h_{\vw}(\vxi))\in\mathcal{I}\right)=&~\frac{p-\alpha}{\beta-\alpha},\\\label{eq:ineqkappa2}
\text{Pr}\left(h_{\vw}(\vxi) >\theta_p | \gamma=k', \bar{F}_{\vw,k'}(h_{\vw}(\vxi))\in\mathcal{I}\right)\geq&~\frac{p-\alpha}{\beta-\alpha},\quad\forall k'\neq k.
\end{align}
Combining \eqref{eq:pdpkappa} with $\theta=\theta_p$ and the two inequalities above gives 
\begin{align}
\label{eq:ineqkappa3}
\text{Pr}\left(h_{\vw}(\vxi) >\theta_p | \gamma=k, \bar{F}_{\vw,k}(h_{\vw}(\vxi))\in\mathcal{I}\right)\leq&~\frac{p-\alpha}{\beta-\alpha}+\kappa,\\\label{eq:ineqkappa4}
\text{Pr}\left(h_{\vw}(\vxi) >\theta_p | \gamma=k', \bar{F}_{\vw,k'}(h_{\vw}(\vxi))\in\mathcal{I}\right)\leq&~\frac{p-\alpha}{\beta-\alpha}+\kappa,\quad\forall k'\neq k.
\end{align}
Consider any $k'\neq k$. By \eqref{eq:CCDF_onI_all}, \eqref{eq:ineqkappa2} and the fact that $\frac{p-\alpha}{\beta-\alpha}+\kappa<1$, we must have 
$\textup{Pr}\big(h_{\vw}(\vxi) >\theta_p \big| \gamma=k'\big)<\beta$. If $p=\alpha$, we have $\theta_p=\bar{F}_{\vw,k}^{-1}(\alpha)\leq \bar{F}_{\vw,k'}^{-1}(\alpha)$ so $\textup{Pr}\big(h_{\vw}(\vxi) >\theta_p \big| \gamma=k'\big)\geq\alpha$. If $p>\alpha$, by \eqref{eq:CCDF_onI_all} and \eqref{eq:ineqkappa2}, we must have $\textup{Pr}\big(h_{\vw}(\vxi) >\theta_p \big| \gamma=k'\big)>\alpha$.  In any case, we have
\begin{align}
\label{eq:pkprimeinterval}
\textup{Pr}\big(h_{\vw}(\vxi) >\theta_p \big| \gamma=k'\big)\in[\alpha, \beta],\quad\forall k'\neq k.
\end{align}
Applying \eqref{eq:CCDF_onI_all}, \eqref{eq:pkinterval} and \eqref{eq:pkprimeinterval} to \eqref{eq:ineqkappa1}, \eqref{eq:ineqkappa2}, \eqref{eq:ineqkappa3} and \eqref{eq:ineqkappa4} gives \eqref{eq:pdpnew}. 


Suppose \eqref{eq:pdpnew} holds. We will prove \eqref{eq:pdpkappa} by contradiction. Suppose there exist $\bar\theta$ and $k$ and $k'$ in $\mathcal{G}$ such that 
\begin{align}
\label{eq:pdpkappa_violate}
\text{Pr}\left(h_{\vw}(\vxi) >\bar\theta | \gamma=k, \bar{F}_{\vw,k}(h_{\vw}(\vxi))\in\mathcal{I}\right)
-\text{Pr}\left(h_{\vw}(\vxi) > \bar\theta | \gamma=k', \bar{F}_{\vw,k'}(h_{\vw}(\vxi))\in\mathcal{I}\right)> \kappa.
\end{align}
Let 
\begin{align*}
p=0.5(\beta-\alpha)\cdot\left(
\begin{array}{l}\text{Pr}\left(h_{\vw}(\vxi) >\bar\theta \big| \gamma=k, \bar{F}_{\vw,k}(h_{\vw}(\vxi))\in\mathcal{I}\right)\\[0.5mm]
+\,\text{Pr}\left(h_{\vw}(\vxi) > \bar\theta \big| \gamma=k', \bar{F}_{\vw,k'}(h_{\vw}(\vxi))\in\mathcal{I}\right)
\end{array}
\right)
-0.5\kappa(\beta-\alpha)+\alpha.
\end{align*}
By \eqref{eq:pdpkappa_violate}, we have 
\begin{align*}
\alpha=&~ 0.5(\beta-\alpha)\kappa -0.5\kappa(\beta-\alpha)+\alpha\\
<&~p<0.5(\beta-\alpha)(2-\kappa) -0.5\kappa(\beta-\alpha)+\alpha=\beta-\kappa(\beta-\alpha).
\end{align*}
By \eqref{eq:pdpnew}, there exists $\theta_p$ such that 
\begin{align}
\label{eq:pdpnew1}
&~\textup{Pr}(h_{\vw}(\vxi) >\theta_p | \gamma=k )\geq p>\alpha,\\
\label{eq:pdpnew2}
&~\textup{Pr}(h_{\vw}(\vxi) >\theta_p | \gamma=k')\geq p>\alpha,\\\label{eq:pdpnew3}
&~\textup{Pr}(h_{\vw}(\vxi) >\theta_p | \gamma=k )\leq p+\kappa(\beta-\alpha)<\beta,\\\label{eq:pdpnew4}
&~\textup{Pr}(h_{\vw}(\vxi) >\theta_p | \gamma=k')\leq p+\kappa(\beta-\alpha)<\beta,
\end{align}
so \eqref{eq:CCDF_onI} holds for $k$ and $k'$. Applying \eqref{eq:CCDF_onI} to \eqref{eq:pdpnew2} and \eqref{eq:pdpnew3} leads to \eqref{eq:ineqkappa2} and \eqref{eq:ineqkappa3}.

By \eqref{eq:ineqkappa2}, \eqref{eq:pdpkappa_violate}, and the definition of $p$, we can show that 
\begin{align*}
&~\text{Pr}\left(h_{\vw}(\vxi) > \bar\theta | \gamma=k', \bar{F}_{\vw,k'}(h_{\vw}(\vxi))\in\mathcal{I}\right)\\
<&~0.5\left(
\begin{array}{l}
\text{Pr}\left(h_{\vw}(\vxi) >\bar\theta \big| \gamma=k, \bar{F}_{\vw,k}(h_{\vw}(\vxi))\in\mathcal{I}\right)\\[0.5mm]
+\,\text{Pr}\left(h_{\vw}(\vxi) > \bar\theta \big| \gamma=k', \bar{F}_{\vw,k'}(h_{\vw}(\vxi))\in\mathcal{I}\right)
\end{array}
\right)
-0.5\kappa\\
=&~\frac{p-\alpha}{\beta-\alpha}\\
\leq&~\text{Pr}\left(h_{\vw}(\vxi) >\theta_p | \gamma=k', \bar{F}_{\vw,k'}(h_{\vw}(\vxi))\in\mathcal{I}\right),
\end{align*}
which implies $ \bar\theta>\theta_p$ and thus 
$$
\text{Pr}\left(h_{\vw}(\vxi) >\theta_p | \gamma=k, \bar{F}_{\vw,k}(h_{\vw}(\vxi))\in\mathcal{I}\right)\geq
\text{Pr}\left(h_{\vw}(\vxi) > \bar\theta | \gamma=k, \bar{F}_{\vw,k}(h_{\vw}(\vxi))\in\mathcal{I}\right).
$$
This inequality, \eqref{eq:ineqkappa3},  and the definition of $p$ together imply  
\begin{align*}
&~0.5\left(
\begin{array}{l}\text{Pr}\left(h_{\vw}(\vxi) >\bar\theta \big| \gamma=k, \bar{F}_{\vw,k}(h_{\vw}(\vxi))\in\mathcal{I}\right)\\[0.5mm]
+\,\text{Pr}\left(h_{\vw}(\vxi) > \bar\theta \big| \gamma=k', \bar{F}_{\vw,k'}(h_{\vw}(\vxi))\in\mathcal{I}\right)
\end{array}
\right)
+0.5\kappa\\
=&~\frac{p-\alpha}{\beta-\alpha}+\kappa\\
\geq&~\text{Pr}\left(h_{\vw}(\vxi) > \theta_p | \gamma=k, \bar{F}_{\vw,k}(h_{\vw}(\vxi))\in\mathcal{I}\right)\\
\geq&~\text{Pr}\left(h_{\vw}(\vxi) > \bar\theta | \gamma=k, \bar{F}_{\vw,k}(h_{\vw}(\vxi))\in\mathcal{I}\right),
\end{align*}
which further contradicts with \eqref{eq:pdpkappa_violate}. Suppose there exists $\bar\theta$ such that 
\begin{align*}
&\text{Pr}\left(h_{\vw}(\vxi) >\bar\theta | \gamma=k, \bar{F}_{\vw,k}(h_{\vw}(\vxi))\in\mathcal{I}\right)
-\text{Pr}\left(h_{\vw}(\vxi) > \bar\theta | \gamma=k', \bar{F}_{\vw,k'}(h_{\vw}(\vxi))\in\mathcal{I}\right)<- \kappa.
\end{align*}
A similar contradiction can be derived by following a similar argument. This indicates that \eqref{eq:pdpkappa} must hold, which completes the proof.


\end{proof}

\begin{proof}[Proof of Lemma~\ref{eq:wpdp_equiv}]
The conclusion is a direct consequence of \eqref{eq:CCDF_onI_all} with $\theta=\widehat\theta$.
\end{proof}

\section{Reformulate Training Problems with Partial Fairness Constraints as DC Optimization}
\label{sec:reform}
This section is organized as follows. In Section~\ref{sec:DCobj}, we first present multiple examples of the DC objective function in \eqref{eq:inprocess_pdp_approx}, \eqref{eq:inprocess_wpdp_approx}, \eqref{eq:inprocess_pgauc}, \eqref{eq:inprocess_peop} and \eqref{eq:inprocess_peod}. Next, in Section~\ref{sec:DCcst1}, we present some examples of \eqref{eq:inprocess_pdp_approx} whose constraints are DC. Then, in Section~\ref{sec:DCcst2}, we present some examples of \eqref{eq:inprocess_wpdp_approx} whose constraints are DC. Finally, in Section~\ref{sec:DCcst3}, we present some examples of \eqref{eq:inprocess_pgauc_approx}, \eqref{eq:inprocess_peop _approx} and \eqref{eq:inprocess_peod_approx}.

Before that, we first present some useful facts. When each $f_i(\vw)$ in \eqref{DC} is $\rho$-weakly convex, meaning that $f_i(\vw)+\frac{\rho}{2}\|\vw\|^2$ is convex, $f_i(\vw)$ is a DC function because 
\begin{align}
\label{eq:wcisdc}
f_i(\vw)=\underbrace{f_i(\vw)+\frac{\rho}{2}\|\vw\|^2}_{=f_i^+(\vw)} - \underbrace{\frac{\rho}{2}\|\vw\|^2}_{=f_i^-(\vw)}.
\end{align}
Therefore, the IDCA in this work can be applied to weakly convex constrained optimization with oracle complexity of $\tilde O(\epsilon^{-4})$  as shown in Theorem~\ref{thm:idca}. This complexity matches the best complexity of first-order methods in the literature for weakly convex constrained optimization up to a logarithmic factor. Moreover, if $f_i(\vw)$ is differentiable and $\nabla f_i(\vw)$ is $L$-Lipschitz continuous, $f_i(\vw)$ is $L$-weakly convex and thus also DC. However, for differentiable problems, there exist more efficient first-order algorithms in the literature with oracle complexity lower than $O(\epsilon^{-4})$, so we do not recommend using IDCA for differentiable problems in general. In the following subsections, we present the scenarios where \eqref{eq:inprocess_pdp_approx} and \eqref{eq:inprocess_wpdp_approx}  are instances of \eqref{DC}.

\subsection{DC objective function in \eqref{eq:inprocess_pdp_approx} and \eqref{eq:inprocess_wpdp_approx} }
\label{sec:DCobj}
Given a training  set $\mathcal{D}=\{(\vxi_i,\zeta_i,\gamma_i)\}_{i=1}^n$, the parameter $\vw$ in model $h_{\vw}(\cdot)$ is typically obtained by minimizing the following empirical loss
\begin{align}
\label{eq:erm}
f_0(\vw)=\frac{1}{n}\sum_{i=1}^{n}\ell(h_{\vw}(\vxi_i),\zeta_i),
\end{align}
where $\ell(h_{\vw}(\vxi),\zeta)$ is a loss function measuring the discrepancy between the predicted score  $h_{\vw}(\vxi)$ and the corresponding target variable $\zeta$. 

\begin{example}
\label{ex:objerm}
Suppose Assumption~\ref{assume:hsmooth} holds. In the following three scenarios: 
\begin{itemize}
\item Quadratic loss: $\ell(h_{\vw}(\vxi_i),\zeta_i)=\frac{1}{2}(h_{\vw}(\vxi_i)-\zeta_i)^2$ and there exists $B\geq0$ such that $|h_{\vw}(\vxi_i)|\leq B$ for $i=1,\dots,n$  for any $\vw\in\cW$;
\item Hinge loss: $\ell(h_{\vw}(\vxi_i),\zeta_i)=(1-\zeta_i\cdot h_{\vw}(\vxi_i))_+$ and $\zeta_i\in\{1,-1\}$;
\item Logistic loss: $\ell(h_{\vw}(\vxi_i),\zeta_i)=\ln(1+\exp(-\,\zeta_i\cdot h_{\vw}(\vxi_i)))$ and $\zeta_i\in\{1,-1\}$;
\end{itemize}
there exists $\rho\geq0$ such that $f_0(\vw)$ in \eqref{eq:erm} is $\rho$-weakly convex and thus is DC in the form of \eqref{eq:wcisdc}.
\end{example}

Consider a binary classification problem where $\zeta\in\{1,-1\}$. Suppose the goal is to optimize the AUC of $h_{\vw}(\vxi)$, that is, 
\begin{align}
\label{eq:auc}
\text{Pr}\big(h_{\vw}(\vxi)\geq h_{\vw}(\vxi') \big| \zeta=1 , \zeta'=-1\big).
\end{align}
We partition the  $\mathcal{D}$ into positive and negative sets, namely, $\mathcal{D}=\mathcal{D}_+\cup \mathcal{D}_-$, where
\begin{align*}
\mathcal{D}_+=&~\{(\vxi,\zeta,\gamma)\in\mathcal{D}|\zeta=1\}=\{(\vxi_i^+,1,\gamma_i^+)\}_{i=1}^{n_+}\\
\mathcal{D}_-=&~\{(\vxi,\zeta,\gamma)\in\mathcal{D}|\zeta=-1\}=\{(\vxi_i^-,-1,\gamma_i^-)\}_{i=1}^{n_-}.
\end{align*}
To maximize AUC in \eqref{eq:auc}, one can minimize an objective function $f_0(\vw)$ defined with pairwise losses, i.e., 
\begin{align}
\label{eq:aucloss}
f_0(\vw)=\frac{1}{n_+n_-}\sum_{i=1}^{n_+}\sum_{j=1}^{n_-}\ell(h_{\vw}(\vxi_i^+)-h_{\vw}(\vxi_j^-)),
\end{align}
where $\ell(h_{\vw}(\vxi_i^+)-h_{\vw}(\vxi_i^-))$ is a surrogate loss approximating the indicator $\mathbf{1}_{h_{\vw}(\vxi_i^+)\leq h_{\vw}(\vxi_i^-)}$. In some applications, the goal is to maximize the partial AUC of $h_{\vw}(\vxi)$ with the false positive rate restricted in an interval $\mathcal{I}=[\alpha,\beta]$, that is, 
\begin{align}
\label{eq:partialauc}
\text{Pr}\big(h_{\vw}(\vxi)\geq h_{\vw}(\vxi') \big| \zeta=1 , \zeta'=-1, \bar{F}_{\vw,k'}(h_{\vw}(\vxi'))\in\mathcal{I} \big),
\end{align}
where  $(\vxi,\zeta,\gamma)$ and $ (\vxi',\zeta',\gamma')$ are two i.i.d. random data points.
Similar to \eqref{eq:aucloss}, let $n_-^\alpha=\lceil \alpha n_-\rceil$ and  $n_-^\beta=\lceil \beta n_-\rceil$ so $n_-^\beta-n_-^\alpha$ represents the number of data points in $\mathcal{D}_-$ whose ranks are in $\mathcal{I}$. Let $h_{\vw}(\vxi_{[i]}^-)$ be the $i$-th largest score in $\{h_{\vw}(\vxi_i^-)| i=1,\dots,n_-\}$ with ties broken in an arbitrary way. Then, to maximize the partial AUC in \eqref{eq:partialauc}, one can minimize the following objective function 
\begin{align}
\nonumber
f_0(\vw)=&~\frac{1}{n_+(n_-^\beta-n_-^\alpha)}\sum_{i=1}^{n_+}\sum_{j=n_-^\alpha+1}^{n_-^\beta}\ell(h_{\vw}(\vxi_i^+)-h_{\vw}(\vxi_{[j]}^-))\\
\label{eq:paucloss}
=&~\frac{1}{n_+(n_-^\beta-n_-^\alpha)}\sum_{i=1}^{n_+}\sum_{j=1}^{n_-^\beta}\ell(h_{\vw}(\vxi_i^+)-h_{\vw}(\vxi_{[j]}^-))
-\frac{1}{n_+(n_-^\beta-n_-^\alpha)}\sum_{i=1}^{n_+}\sum_{j=1}^{n_-^\alpha}\ell(h_{\vw}(\vxi_i^+)-h_{\vw}(\vxi_{[j]}^-)).
\end{align}

\begin{example}
\label{ex:objauc}
Suppose Assumption~\ref{assume:hsmooth} holds. In the following three scenarios: 
\begin{itemize}
\item Quadratic loss: $\ell(h_{\vw}(\vxi_i^+)-h_{\vw}(\vxi_j^-))=\frac{1}{2}(1-h_{\vw}(\vxi_i^+)+h_{\vw}(\vxi_j^-))^2$ and there exists $B\geq0$ such that $|h_{\vw}(\vxi_i)|\leq B$ for $i=1,\dots,n$  for any $\vw\in\cW$;
\item Hinge loss: $\ell(h_{\vw}(\vxi_i^+)-h_{\vw}(\vxi_j^-))=(1-h_{\vw}(\vxi_i^+)+h_{\vw}(\vxi_j^-))_+$;
\item Logistic loss: $\ell(h_{\vw}(\vxi_i^+)-h_{\vw}(\vxi_j^-))=\ln(1+\exp(-\,h_{\vw}(\vxi_i^+)+h_{\vw}(\vxi_j^-)))$;
\end{itemize}
there exists $\rho\geq0$ such that the two terms on the right-hand side of \eqref{eq:paucloss} are both $\rho$-weakly convex and thus $f_0(\vw)$ in \eqref{eq:paucloss}  is DC in the form of 
\begin{align*}
f_0(\vw)=&~f_0^+(\vw)-f_0^-(\vw),\\
f_0^+(\vw)=&~\frac{1}{n_+(n_-^\beta-n_-^\alpha)}\sum_{i=1}^{n_+}\sum_{j=1}^{n_-^\beta}\ell(h_{\vw}(\vxi_i^+)-h_{\vw}(\vxi_{[j]}^-))+\frac{\rho}{2}\|\vw\|^2,\\
f_0^-(\vw)=&~\frac{1}{n_+(n_-^\beta-n_-^\alpha)}\sum_{i=1}^{n_+}\sum_{j=1}^{n_-^\alpha}\ell(h_{\vw}(\vxi_i^+)-h_{\vw}(\vxi_{[j]}^-))+\frac{\rho}{2}\|\vw\|^2.
\end{align*}
\end{example}
Note that $f_0$ in \eqref{eq:aucloss} is DC in the three scenarios above also because it is a special case of \eqref{eq:paucloss} with $\mathcal{I}=[0,1]$, i.e., $n_-^\alpha=0$ and $n_-^\beta=n_-$.



\subsection{DC constraints in \eqref{eq:inprocess_pdp_approx}}
\label{sec:DCcst1}
\begin{example}
\label{ex:hingesurrogate_pdp}
Suppose Assumption~\ref{assume:hsmooth} holds and $\sigma(x)=\max\{\min\{x+0.5,1\},0\}=\sigma^+(x)-\sigma^-(x)$ in \eqref{eq:inprocess_pdp_approx}, where
$$
\sigma^+(x)=\max\{x+0.5,0\}\text{  and  }\sigma^-(x)=\max\{x-0.5,0\}.
$$
Since $\sigma^+(x)$ and $\sigma^-(x)$ are convex and $1$-Lipschitz continuous, there exists $\rho\geq0$ such that
$$
\frac{1}{n_k} \sum_{i=1}^{n_k} \sigma^+(h_{\vw}(\vxi_i^k)-\theta_{p})
\text{  and  }
\frac{1}{n_k} \sum_{i=1}^{n_k} \sigma^-(h_{\vw}(\vxi_i^k)-\theta_{p})
$$
are both $\rho$-weakly convex jointly in $\vw$ and $\theta_p$ for any $p\in\widehat{\mathcal{I}}$ and any $k\in\mathcal{G}$. Therefore, the constraints in \eqref{eq:inprocess_pdp_approx} are instances of the constraints in \eqref{DC} in the form of  
\begin{align*}
f_i(\vw)
=\underbrace{\frac{1}{n_k} \sum_{i=1}^{n_k} \sigma^-(h_{\vw}(\vxi_i^k)-\theta_{p})+\frac{\rho}{2}\|\vw\|^2}_{f_i^+(\vw)}
-\underbrace{\left[\frac{1}{n_k} \sum_{i=1}^{n_k} \sigma^+(h_{\vw}(\vxi_i^k)-\theta_{p})-p+\frac{\rho}{2}\|\vw\|^2\right]}_{f_i^-(\vw)}
\leq 0
\end{align*}
and
\begin{align*}
f_i(\vw)
=
\underbrace{\frac{1}{n_k} \sum_{i=1}^{n_k} \sigma^+(h_{\vw}(\vxi_i^k)-\theta_{p})+\frac{\rho}{2}\|\vw\|^2}_{f_i^+(\vw)}
-\underbrace{\left[\frac{1}{n_k} \sum_{i=1}^{n_k} \sigma^-(h_{\vw}(\vxi_i^k)-\theta_{p})+p+\kappa(\beta-\alpha)+\frac{\rho}{2}\|\vw\|^2\right]}_{f_i^-(\vw)}
\leq 0
\end{align*}
for any $p\in\widehat{\mathcal{I}}$ and any $k\in\mathcal{G}$.
\end{example}

\begin{example}
\label{ex:sigmoidsurrogate_pdp}
Suppose Assumption~\ref{assume:hsmooth} holds and $\sigma(x)=\exp(x)/(1+\exp(x))$ in \eqref{eq:inprocess_pdp_approx}. Also, suppose  there exists $B\geq0$ such that $\|\nabla h_{\vw}(\vxi_i)\|\leq B$ for $i=1,\dots,n$  for any $\vw\in\cW$. 
In this case, $\frac{1}{n_k} \sum_{i=1}^{n_k} \sigma(h_{\vw}(\vxi_i^k)-\theta_{p})$, as a function of $\vw$ and $\theta_p$, 
is differentiable with a Lipschitz continuous gradient on $\cW$. Hence, there exists $\rho\geq0$ such that $\frac{1}{n_k} \sum_{i=1}^{n_k} \sigma(h_{\vw}(\vxi_i^k)-\theta_{p})$ is $\rho$-weakly convex jointly in $\vw$ and $\theta_p$ for any $p\in\widehat{\mathcal{I}}$ and any $k\in\mathcal{G}$. Therefore, the constraints in \eqref{eq:inprocess_pdp_approx} are DC constraints in the form of  
\begin{align*}
f_i(\vw)=\underbrace{\frac{\rho}{2}\|\vw\|^2}_{f_i^+(\vw)}-\underbrace{\left[\frac{1}{n_k} \sum_{i=1}^{n_k} \sigma(h_{\vw}(\vxi_i^k)-\theta_{p})-p+\frac{\rho}{2}\|\vw\|^2\right]}_{f_i^-(\vw)}\leq 0
\end{align*}
and
\begin{align*}
f_i(\vw)=\underbrace{\frac{1}{n_k} \sum_{i=1}^{n_k} \sigma(h_{\vw}(\vxi_i^k)-\theta_{p})+\frac{\rho}{2}\|\vw\|^2}_{f_i^+(\vw)}-\underbrace{\left[p+\kappa(\beta-\alpha)+\frac{\rho}{2}\|\vw\|^2\right]}_{f_i^-(\vw)}\leq 0
\end{align*}
for any $p\in\widehat{\mathcal{I}}$ and any $k\in\mathcal{G}$.
\end{example}

\subsection{DC constraints in \eqref{eq:inprocess_wpdp_approx}}
\label{sec:DCcst2}
Suppose  $\sigma(x)$, $\sigma^+(x)$ and $\sigma^-(x)$ are the same as in Example~\ref{ex:hingesurrogate_pdp}. We first observe that
\small
\begin{align}
\label{eq:DCexample1_wpdp}
\begin{split}
&~\min\left\{\frac{1}{n_k} \sum_{i=1}^{n_k} \sigma(h_{\vw}(\vxi_i^k)-\widehat\theta),\alpha\right\}\\
=&~\frac{1}{n_k} \sum_{i=1}^{n_k} \sigma^+(h_{\vw}(\vxi_i^k)-\widehat\theta)+\alpha-\max\left\{\frac{1}{n_k} \sum_{i=1}^{n_k} \sigma^-(h_{\vw}(\vxi_i^k)-\widehat\theta)+\alpha,\sum_{i=1}^{n_k} \sigma^+(h_{\vw}(\vxi_i^k)-\widehat\theta)\right\},\\
&~\min\left\{\frac{1}{n_k} \sum_{i=1}^{n_k} \sigma(h_{\vw}(\vxi_i^k)-\widehat\theta),\beta\right\}\\
=&~\frac{1}{n_k} \sum_{i=1}^{n_k} \sigma^+(h_{\vw}(\vxi_i^k)-\widehat\theta)+\beta-\max\left\{\frac{1}{n_k} \sum_{i=1}^{n_k} \sigma^-(h_{\vw}(\vxi_i^k)-\widehat\theta)+\beta,\sum_{i=1}^{n_k} \sigma^+(h_{\vw}(\vxi_i^k)-\widehat\theta)\right\},
\end{split}
\end{align}
\normalsize
where the right-hand sides of both equalities are differences of two weakly convex functions and thus are DC.

\begin{example}
\label{ex:hingesurrogate_wpdp}
Suppose Assumption~\ref{assume:hsmooth} holds and $\sigma(x)$, $\sigma^+(x)$ and $\sigma^-(x)$ in \eqref{eq:inprocess_wpdp_approx} are the same as in Example~\ref{ex:hingesurrogate_pdp}. For the same reason as in Example~\ref{ex:hingesurrogate_pdp}, 
$$
\frac{1}{n_k} \sum_{i=1}^{n_k} \sigma^+(h_{\vw}(\vxi_i^k)-\widehat\theta)
\text{  and  }
\frac{1}{n_k} \sum_{i=1}^{n_k} \sigma^-(h_{\vw}(\vxi_i^k)-\widehat\theta)
$$
are both weakly convex for any $k\in\mathcal{G}$. So are 
$$
\max\left\{\frac{1}{n_k} \sum_{i=1}^{n_k} \sigma^-(h_{\vw}(\vxi_i^k)-\widehat\theta)+\alpha,\sum_{i=1}^{n_k} \sigma^+(h_{\vw}(\vxi_i^k)-\widehat\theta)\right\}
$$
and
$$
\max\left\{\frac{1}{n_k} \sum_{i=1}^{n_k} \sigma^-(h_{\vw}(\vxi_i^k)-\widehat\theta)+\beta,\sum_{i=1}^{n_k} \sigma^+(h_{\vw}(\vxi_i^k)-\widehat\theta)\right\}
$$
for any $k\in\mathcal{G}$. Therefore, by \eqref{eq:DCexample1_wpdp}, there exists $\rho\geq0$ such that
the constraints in \eqref{eq:inprocess_wpdp_approx} are instances of the constraints in \eqref{DC} in the form of  
\small
\begin{align*}
f_i(\vw)=&~f_i^+(\vw)-f_i^-(\vw),\\
f_i^+(\vw)=&~\frac{1}{n_k} \sum_{i=1}^{n_k} \sigma^+(h_{\vw}(\vxi_i^k)-\widehat\theta)+\max\left\{\frac{1}{n_k} \sum_{i=1}^{n_k} \sigma^-(h_{\vw}(\vxi_i^k)-\widehat\theta)+\alpha,\sum_{i=1}^{n_k} \sigma^+(h_{\vw}(\vxi_i^k)-\widehat\theta)\right\}\\
&+\frac{1}{n_{k'}} \sum_{i=1}^{n_{k'}} \sigma^+(h_{\vw}(\vxi_i^{k'})-\widehat\theta)+\max\left\{\frac{1}{n_{k'}} \sum_{i=1}^{n_{k'}} \sigma^-(h_{\vw}(\vxi_i^{k'})-\widehat\theta)+\beta,\sum_{i=1}^{n_{k'}} \sigma^+(h_{\vw}(\vxi_i^{k'})-\widehat\theta)\right\}\\
&+\frac{\rho}{2}\|\vw\|^2,\\
f_i^-(\vw)=&~\frac{1}{n_{k'}} \sum_{i=1}^{n_{k'}} \sigma^+(h_{\vw}(\vxi_i^{k'})-\widehat\theta)+\max\left\{\frac{1}{n_{k'}} \sum_{i=1}^{n_{k'}} \sigma^-(h_{\vw}(\vxi_i^{k'})-\widehat\theta)+\alpha,\sum_{i=1}^{n_{k'}} \sigma^+(h_{\vw}(\vxi_i^{k'})-\widehat\theta)\right\}\\
&+\frac{1}{n_k} \sum_{i=1}^{n_k} \sigma^+(h_{\vw}(\vxi_i^k)-\widehat\theta)+\max\left\{\frac{1}{n_k} \sum_{i=1}^{n_k} \sigma^-(h_{\vw}(\vxi_i^k)-\widehat\theta)+\beta,\sum_{i=1}^{n_k} \sigma^+(h_{\vw}(\vxi_i^k)-\widehat\theta)\right\}\\
&+\frac{\rho}{2}\|\vw\|^2+\kappa(\beta-\alpha),
\end{align*}
\normalsize
for any $k,k'\in\mathcal{G}$.
\end{example}

\begin{example}
\label{ex:sigmoidsurrogate_wpdp}
Suppose Assumption~\ref{assume:hsmooth} holds and $\sigma(x)=\exp(x)/(1+\exp(x))$ in \eqref{eq:inprocess_wpdp_approx}. Also, suppose  there exists $B\geq0$ such that $\|\nabla h_{\vw}(\vxi_i)\|\leq B$ for $i=1,\dots,n$  for any $\vw\in\cW$. 
In this case, $-\frac{1}{n_k} \sum_{i=1}^{n_k} \sigma(h_{\vw}(\vxi_i^k)-\widehat\theta)$ is differentiable with a Lipschitz continuous gradient on $\cW$. Hence,  $-\frac{1}{n_k} \sum_{i=1}^{n_k} \sigma(h_{\vw}(\vxi_i^k)-\widehat\theta)$ is weakly convex for any $k\in\mathcal{G}$. So are
\small
$$
-\min\left\{\frac{1}{n_k} \sum_{i=1}^{n_k} \sigma(h_{\vw}(\vxi_i^k)-\widehat\theta),\beta\right\}=\max\left\{-\frac{1}{n_k} \sum_{i=1}^{n_k} \sigma(h_{\vw}(\vxi_i^k)-\widehat\theta),-\beta\right\}
$$
\normalsize
and
\small
$$
-\min\left\{\frac{1}{n_k} \sum_{i=1}^{n_k} \sigma(h_{\vw}(\vxi_i^k)-\widehat\theta),\alpha\right\}=\max\left\{-\frac{1}{n_k} \sum_{i=1}^{n_k} \sigma(h_{\vw}(\vxi_i^k)-\widehat\theta),-\alpha\right\}
$$
\normalsize
for any $k\in\mathcal{G}$. Therefore,  there exists $\rho\geq0$ such that the constraints in \eqref{eq:inprocess_wpdp_approx} are DC constraints in the form of  
\small
\begin{align*}
f_i(\vw)=&~f_i^+(\vw)-f_i^-(\vw),\\
f_i^+(\vw)=&\max\left\{-\frac{1}{n_k} \sum_{i=1}^{n_k} \sigma(h_{\vw}(\vxi_i^k)-\widehat\theta),-\alpha\right\}+\max\left\{-\frac{1}{n_{k'}} \sum_{i=1}^{n_{k'}} \sigma(h_{\vw}(\vxi_i^{k'})-\widehat\theta),-\beta\right\}+\frac{\rho}{2}\|\vw\|^2,\\
f_i^-(\vw)=&\max\left\{-\frac{1}{n_k} \sum_{i=1}^{n_k} \sigma(h_{\vw}(\vxi_i^k)-\widehat\theta),-\beta\right\}+\max\left\{-\frac{1}{n_{k'}} \sum_{i=1}^{n_{k'}} \sigma(h_{\vw}(\vxi_i^{k'})-\widehat\theta),-\alpha\right\}+\frac{\rho}{2}\|\vw\|^2\\
&+\kappa(\beta-\alpha),
\end{align*}
\normalsize
for any $k,k'\in\mathcal{G}$. 
\end{example}

\subsection{DC constraints in \eqref{eq:inprocess_pgauc_approx}, \eqref{eq:inprocess_peop _approx} and \eqref{eq:inprocess_peod_approx}} \label{sec:DCcst3}
\begin{example}
\label{ex:hingesurrogate_gauc}
Suppose Assumption~\ref{assume:hsmooth} holds and $\sigma(x)=\exp(x)/(1+\exp(x))$ in \eqref{eq:inprocess_wpdp_approx}. Also, suppose  there exists $B\geq0$ such that $\|\nabla h_{\vw}(\vxi_i)\|\leq B$ for $i=1,\dots,n$  for any $\vw\in\cW$. We first show that the constraints in \eqref{eq:inprocess_pgauc_approx} can be formulated as DC constraints.

Let $n_k^\alpha=\lceil \alpha n_k\rceil$ and  $n_k^\beta=\lceil \beta n_k\rceil$ so $n_k^\beta-n_k^\alpha$ represents the number of data points in $\mathcal{D}_k$ whose ranks are in $\mathcal{I}$. We define $[i]$ is the index of the $i$-th largest coordinate in vector $(h_{\boldsymbol{w}}(\boldsymbol{\xi}_i^{k}))_{i=1}^{n_{k}}$ with ties broken arbitrarily. Given any $k$ and $k'$ in $\mathcal{G}$, we first define
\begin{align*}
\nonumber
G_{j} :=&~ \sum_{i=n_{k}^{\alpha}+1}^{n_k^{\beta}} \sigma\left(h_{\boldsymbol{w}}(\boldsymbol{\xi}_{[i]}^{k}) -h_{\boldsymbol{w}}(\boldsymbol{\xi}_{j}^{k'})\right)\\\nonumber
 =&~ \underbrace{\sum_{i=1}^{n_k^{\beta}} \sigma\left(h_{\boldsymbol{w}}(\boldsymbol{\xi}_{[i]}^{k}) -h_{\boldsymbol{w}}(\boldsymbol{\xi}_{j}^{k'})\right)}_{=:G_j^+} - \underbrace{\sum_{i=1}^{n_k^{\alpha}} \sigma\left(h_{\boldsymbol{w}}(\boldsymbol{\xi}_{[i]}^{k}) -h_{\boldsymbol{w}}(\boldsymbol{\xi}_{j}^{k'})\right)}_{=:G_j^-}, \quad \forall j = 1, \dots, n_{k'}.\nonumber
\end{align*}
Next, we define $\left \langle j \right \rangle$ as the index of the $j$-th smallest coordinate in vector $(G—_j)_{j=1}^{n_{k'}}$ with ties broken arbitrarily, so we can write 
\begin{align}
\label{eq:numerator}
    \sum_{i=n_{k}^{\alpha}+1}^{n_k^{\beta}} \sum_{j=n_{k'}^{\alpha}+1}^{n_{k'}^{\beta}} \sigma\left(h_{\boldsymbol{w}}(\boldsymbol{\xi}_{[i]}^{k}) -h_{\boldsymbol{w}}(\boldsymbol{\xi}_{[j]}^{k'})\right)
    = \sum_{j=n_{k'}^{\alpha}+1}^{n_{k'}^{\beta}} G_{\left \langle j \right \rangle}
    = \sum_{j=1}^{n_{k'}^{\beta}}G_{\left \langle j \right \rangle} - \sum_{j=1}^{n_{k'}^{\alpha}} G_{\left \langle j \right \rangle}.
\end{align}
For any positive integers $n$ and $N$ with $n\leq N$, we define 
\begin{align*}
    \nonumber
    \Delta_{n}^{N} :=&~ \Bigg\{ \vp \in \mathbb{R}^{N} \Bigg| \sum_{j=1}^{N} p_j = n, 0 \leq p_j \leq 1, \forall j=1, \dots, N \Bigg\}.
\end{align*}
Also, let $G^+:= \sum_{j=1}^{n_{k'}} G_j^+$. Then we can write
\begin{align*}
    \nonumber 
    \sum_{j=1}^{n_{k'}^{\beta}}G_{\left \langle j \right \rangle} = &~\min_{\vp \in \Delta_{n_{k'}^{\beta}}^{n_{k'}}} \sum_{j=1}^{n_{k'}} p_j (G_j^+ - G_j^-)\\\nonumber
    = &~ \min_{\vp \in \Delta_{n_{k'}^{\beta}}^{n_{k'}}} \sum_{j=1}^{n_{k'}} p_j (G_j^+ - G_j^- + G^+ - G^+)\\\nonumber
    = &~ n_{k'}^{\beta} G^+ + \min_{\vp \in \Delta_{n_{k'}^{\beta}}^{n_{k'}}} \sum_{j=1}^{n_{k'}} p_j (-(G^+-G_j^+)-G_j^-)\\\nonumber
    = &~ n_{k'}^{\beta} G^+ - \max_{\vp \in \Delta_{n_{k'}^{\beta}}^{n_{k'}}}\sum_{j=1}^{n_{k'}} p_j ((G^+-G_j^+)+G_j^-),
\end{align*}
which is a difference of two weakly convex functions. Similarly, we can write $\sum_{j=1}^{n_{k'}^{\alpha}} G_{\left \langle j \right \rangle} = n_{k'}^{\alpha} G^+ - \max_{\vp \in \Delta_{n_{k'}^{\alpha}}^{n_{k'}}}\sum_{j=1}^{n_{k'}} p_j ((G^+-G_j^+)+G_j^-)$, which is also a difference of two weakly convex functions. Therefore, \eqref{eq:numerator} is also a difference of two weakly convex functions, so
there exists $\rho\geq0$ such that the constraints in \eqref{eq:inprocess_pgauc_approx} are DC constraints in the form of
\begin{align*}
    f_i(\vw) =&~ \underbrace{\frac{(n_{k'}^\beta-n_{k'}^\alpha)G^+ + \max_{\vp \in \Delta_{n_{k'}^{\alpha}}^{n_{k'}}}\sum_{j=1}^{n_{k'}} p_j ((G^+-G_j^+)+G_j^-)}{(n_k^\beta-n_k^\alpha)(n_{k'}^\beta-n_{k'}^\alpha)}+\frac{\rho}{2}\|\vw\|^2}_{f_i^+(\vw)} \\
    & - \underbrace{\frac{\max_{\vp \in \Delta_{n_{k'}^{\beta}}^{n_{k'}}}\sum_{j=1}^{n_{k'}} p_j ((G^+-G_j^+)+G_j^-)}{(n_k^\beta-n_k^\alpha)(n_{k'}^\beta-n_{k'}^\alpha)}+\frac{\rho}{2}\|\vw\|^2+0.5+\kappa}_{f_i^-(\vw)}
    \leq 0
\end{align*}
for any $k, k' \in \mathcal{G}$.
\end{example}

For DC constraints in \eqref{eq:inprocess_peop _approx} and \eqref{eq:inprocess_peod_approx}, their structures are almost identical to that of \eqref{eq:inprocess_wpdp_approx}. Hence, they can be formulated by the same procedure as in Example~\ref{ex:hingesurrogate_wpdp} and \ref{ex:sigmoidsurrogate_wpdp}.

\section{Proofs for Complexity Analysis}\label{sec:proofs_complexity}
In this section, we present the proofs for the lemmas, propositions, and theorems related to the complexity analysis of IDCA. All proofs are under Assumption~\ref{assume:DC}.

\subsection{Proof of Proposition~\ref{thm:ssg}}
\label{sec:ssg}
\begin{proof}
Since $f_i(\vw^{(k)})\leq 0$ for $i\in[m]$, \eqref{DCA} is feasible. Recall that $\widehat\vw^{(k+1)}$ is the optimal solution of \eqref{DCA}. We prove this proposition by contradiction. Suppose $T_k\geq\frac{M^2}{\epsilon_k\mu}\ln\left(1+\frac{\mu \|\vw^{(k)}-\widehat\vw^{(k+1)}\|^2}{\epsilon_k}\right)$ but $g_0(\vw^{(k+1)})-g_0(\widehat\vw^{(k+1)})>\epsilon_k$, or equivalently, $g_0(\vv^{(t)})-g_0(\widehat\vw^{(k+1)})>\epsilon_k$ for all $t\in\mathcal{T}$, after Algorithm~\ref{alg:swg} terminates.

If $t\in\mathcal{T}$, we have 
\begin{align}
\nonumber
\|\vv^{(t+1)}-\widehat\vw^{(k+1)}\|^2
\leq& ~\|\vv^{(t)}-\widehat\vw^{(k+1)}\|^2-\frac{2\epsilon_k}{\|\vg_0^{(t)}\|^2}\left\langle\vg_0^{(t)},\vv^{(t)}-\widehat\vw^{(k+1)}\right\rangle+\frac{\epsilon_k^2}{\|\vg_0^{(t)}\|^2}\\\nonumber
\leq&\left(1-\frac{\epsilon_k\mu}{\|\vg_0^{(t)}\|^2}\right)\|\vv^{(t)}-\widehat\vw^{(k+1)}\|^2-\frac{2\epsilon_k}{\|\vg_0^{(t)}\|^2}\left(g_0(\vv^{(t)})-g_0(\widehat\vw^{(k+1)})\right)+\frac{\epsilon_k^2}{\|\vg_0^{(t)}\|^2}\\\nonumber
<&\left(1-\frac{\epsilon_k\mu}{\|\vg_0^{(t)}\|^2}\right)\|\vv^{(t)}-\widehat\vw^{(k+1)}\|^2-\frac{\epsilon_k^2}{\|\vg_0^{(t)}\|^2}\\\nonumber
\leq&\left(1-\frac{\epsilon_k\mu}{M^2}\right)\|\vv^{(t)}-\widehat\vw^{(k+1)}\|^2-\frac{\epsilon_k^2}{M^2},
\end{align}
where the second inequality is by the $\mu$-strong convexity of $g_0$, the third is because $g_0(\vv^{(t)})-g_0(\widehat\vw^{(k+1)})>\epsilon_k$ and the last is because of Assumption~\ref{assume:DC}D.

If $t\notin\mathcal{T}$, by similar arguments, we have
\begin{align}
\nonumber
\|\vv^{(t+1)}-\widehat\vw^{(k+1)}\|^2
\leq& ~\|\vv^{(t)}-\widehat\vw^{(k+1)}\|^2-\frac{2\epsilon_k}{\|\vg^{(t)}\|^2}\left\langle\vg^{(t)},\vv^{(t)}-\widehat\vw^{(k+1)}\right\rangle+\frac{\epsilon_k^2}{\|\vg^{(t)}\|^2}\\\nonumber
\leq&\left(1-\frac{\epsilon_k\mu}{\|\vg^{(t)}\|^2}\right)\|\vv^{(t)}-\widehat\vw^{(k+1)}\|^2-\frac{2\epsilon_k}{\|\vg^{(t)}\|^2}\left(g(\vv^{(t)})-g(\widehat\vw^{(k+1)})\right)+\frac{\epsilon_k^2}{\|\vg^{(t)}\|^2}\\\nonumber
<&\left(1-\frac{\epsilon_k\mu}{\|\vg^{(t)}\|^2}\right)\|\vv^{(t)}-\widehat\vw^{(k+1)}\|^2-\frac{\epsilon_k^2}{\|\vg^{(t)}\|^2}\\\nonumber
\leq&\left(1-\frac{\epsilon_k\mu}{M^2}\right)\|\vv^{(t)}-\widehat\vw^{(k+1)}\|^2-\frac{\epsilon_k^2}{M^2}.
\end{align}
\normalsize

The two inequalities above together imply
$$
\|\vv^{(t+1)}-\widehat\vw^{(k+1)}\|^2<\left(1-\frac{\epsilon_k\mu}{M^2}\right)\|\vv^{(t)}-\widehat\vw^{(k+1)}\|^2-\frac{\epsilon_k^2}{M^2},
$$
for $t=0,1,\dots,T_k-1$. For brevity, for $t=0,1,\dots,T_k-1$, set
\begin{align*}
    a_t:=\|\vv^{(t)}-\widehat\vw^{(k+1)}\|^2,~~ 
    q:=1-\frac{\epsilon_k\mu}{M^2},~~
    c:=\frac{\epsilon_k^2}{M^2}.
\end{align*}
So the inequality is $a_{t+1}< qa_t-c$. By fully applying this recursively, we get
\begin{align*}
    a_1<&~qa_0-c,\\
    a_2<&~qa_1-c
    <q(qa_0-c)-c=q^2a_0-c(q+1),\\
    a_3<&~qa_2-c
    <q(q^2a_0-c(q+1))-c
    =q^3a_0-c(q^2+q+1),
\end{align*}
and in general
\begin{align*}
    a_{T_k}<q^{T_k}a_0-c\sum_{j=0}^{T_k-1}q^j
    =q^{T_k}a_0-c\frac{1-q^{T_k}}{1-q}=q^{T_k}\left(a_0+\frac{c}{1-q}\right)-\frac{c}{1-q}.
\end{align*}
by noting that $q\in(0,1)$. Substituting back $a_t,q$ and $c$, we get
\begin{align*}
    \|\vv^{(T_k)}-\widehat\vw^{(k+1)}\|^2<&\left(1-\frac{\epsilon_k\mu}{M^2}\right)^{T_k}\left(\|\vv^{(0)}-\widehat\vw^{(k+1)}\|^2+\frac{\epsilon_k^2}{M^2}\Big/\left(\frac{\epsilon_k\mu}{M^2}\right)\right)-\frac{\epsilon_k^2}{M^2}\Big/\left(\frac{\epsilon_k\mu}{M^2}\right)\\
    =&\left(1-\frac{\epsilon_k\mu}{M^2}\right)^{T_k}\left(\|\vv^{(0)}-\widehat\vw^{(k+1)}\|^2+\frac{\epsilon_k}{\mu}\right)-\frac{\epsilon_k}{\mu}\\
\leq&~\exp\left(-\frac{\epsilon_k\mu}{M^2}T_k\right){\left(\|\vv^{(0)}-\widehat\vw^{(k+1)}\|^2+\frac{\epsilon_k}{\mu}\right)}-\frac{\epsilon_k}{\mu}\\
\leq&~0,
\end{align*}
where the last inequality is by the definition of $T_k$, which ensures
\begin{align*}
    \exp\left(-\frac{\epsilon_k\mu}{M^2}T_k\right)\leq\frac{\epsilon_k/\mu}{\|\vv^{(0)}-\widehat\vw^{(k+1)}\|^2 +\epsilon_k/\mu}
    \iff T_k\geq\frac{M^2}{\epsilon_k\mu}\ln\left(1+\frac{\mu \|\vw^{(k)}-\widehat\vw^{(k+1)}\|^2}{\epsilon_k}\right), 
\end{align*}
and the fact that $\vv^{(0)}=\vw^{(k)}$. This contradiction indicates that $g_0(\vw^{(k+1)})-g_0(\widehat\vw^{(k+1)})\leq\epsilon_k$. Note that $g(\vw^{(k+1)})\leq \epsilon_k$ since $\vw^{(k+1)}=\vv^{(\tau)}$ with $\tau\in\mathcal{T}$.
\end{proof}

\subsection{Distance between $\vw^{(k)}$ and $\widehat\vw^{(k+1)}$}
As follows, we present subproblem \eqref{DCA} defined by a generic solution $\vw\in\cW$ instead of just $\vw^{(k)}$.
\begin{equation} \label{DCAw}
    \begin{split}
       \min_{\vv \in \cW} &~ g_0(\vv):=f_0^+(\vv) - \widehat{f}_0^-(\vv;\vw)\\
        \text{s.t.}~ &~ g_i(\vv):=f_i^+(\vv) - \widehat{f}_i^-(\vv;\vw) \leq 0,~ i\in[m],
    \end{split}
\end{equation}
where
\begin{equation}
\label{eq:convexifyfw}
    \widehat{f}_i^-(\vv;\vw):= f_i^-(\vw) + (\vf_i^-)^\top(\vv - \vw),
\end{equation}
and $\vf_i^-$ is an arbitrary subgradient in $\partial f_i^-(\vw)$. Like $\widehat\vw^{(k+1)}$, let $\widehat\vw$ be the optimal solution of \eqref{DCAw}. Furthermore, when $f_i(\vw)\leq 0$ for $i\in[m]$,  by Assumption~\ref{assume:DC}B, the constraints in \eqref{DCAw} satisfy the Slater's condition, so there exists a vector of Lagrangian multipliers $\vlam=(\lambda_1,\dots,\lambda_m)^\top\in\mathbb{R}_+^m$ such that the following KKT conditions of \eqref{DCAw} hold.  
\begin{align}
\label{eq:stationarityw}
\vf_0^+-\vf_0^-+\textstyle\sum_{i=1}^m\lambda_i(\vf_i^+-\vf_i^-)+\vn_{\cW}=\mathbf{0},\\\label{eq:feasibilityw}
g_i(\widehat\vw)\leq 0,~ i\in[m],\\\label{eq:complementaryw}
\lambda_ig_i(\widehat\vw)=0,~i\in[m].
\end{align}
where $\vn_{\cW}\in \cN_\cW(\widehat\vw)$, $\vf_i^+\in\partial f_i^+(\widehat\vw)$, $\vf_i^-\in\partial f_i^-(\vw)$ for $i=0,1,\dots,m$. Using the three conditions above, we can show that $\vw$ is a nearly $\epsilon$-KKT point when $\|\widehat\vw-\vw\|\leq O(\epsilon)$. 
\begin{lemma}
\label{thm:nearlyKKT}
Suppose $f_i(\vw)\leq 0$ for $i\in[m]$ and $\|\widehat\vw - \vw\|\leq\min\{1,\frac{1}{2M},\frac{1}{2M\max_{i\in[m]}\lambda_i}\}\cdot\epsilon$,\footnote{Here, we follow the convention that $\frac{1}{\max_{i\in[m]}\lambda_i}=+\infty$ if $\max_{i\in[m]}\lambda_i=0$.} where $\widehat\vw$ is the optimal solution of \eqref{DCAw} and $\lambda_i$ for $i\in[m]$ are the Lagrangian multipliers corresponding to $\widehat\vw$. Then $\vw$ is a nearly $\epsilon$-KKT solution of \eqref{DC}.
\end{lemma}
\begin{proof}
Since $f_i(\vw)\leq 0$ for $i\in[m]$, we have \eqref{eq:stationarityw}, which means \eqref{eq:stationarity} holds with $\|\widehat\vw - \vw\|\leq \epsilon$. Moreover, since $g_i$ is $2M$-Lipschitz continuous and $f_i(\vw)=g_i(\vw)$, \eqref{eq:feasibilityw} and \eqref{eq:complementaryw} imply that 
$$
f_i(\vw) = g_i(\vw) \leq g_i(\widehat\vw)+2M\|\widehat\vw - \vw\|\leq \epsilon
$$
and
$$
|\lambda_if_i(\vw)|= |\lambda_ig_i(\vw)| \leq |\lambda_ig_i(\widehat\vw)|+2M\lambda_i\|\widehat\vw - \vw\|\leq \epsilon.
$$
This means $\vw$ is a nearly $\epsilon$-KKT solution of \eqref{DC}.
\end{proof}

\subsection{Bounded Lagrangian multipliers}
Suppose  $f_i(\vw^{(k)})\leq 0$ for $i\in[m]$. Let $\vlam^{(k)}=(\lambda_1^{(k)},\dots,\lambda_m^{(k)})^\top\in\mathbb{R}_+^m$ be the Lagrangian multipliers corresponding to  $\widehat\vw^{(k+1)}$. In other words, \eqref{eq:stationarityw}, \eqref{eq:feasibilityw} and \eqref{eq:complementaryw} hold with $\vw=\vw^{(k)}$, $\vlam=\vlam^{(k)}$ and $\widehat\vw=\widehat\vw^{(k+1)}$. According to Lemma~\ref{thm:nearlyKKT}, to analyze the oracle complexity of IDCA, it suffices to analyze the complexity for finding $\vw^{(k)}$ satisfying $\|\vw^{(k)}-\widehat\vw^{(k+1)}\|\leq \min\{1,\frac{1}{2M},\frac{1}{2M\max_{i\in[m]}\lambda_i^{(k)}}\}\cdot\epsilon$.  However, this inequality will be difficult to guarantee if $\max_{i\in[m]}\lambda_i^{(k)}$ increases to infinity as $k$ increases. Fortunately, we can provide a uniform upper bound for $\max_{i\in[m]}\lambda_i^{(k)}$ for each $k$ as long as  $f_i(\vw^{(k)})\leq 0$ for $i\in[m]$ for each $k$.

\begin{lemma}
\label{thm:dualbound}
Suppose $f_i(\vw)\leq 0$ for $i\in[m]$. Let $\widehat\vw$ be the optimal solution of \eqref{DCAw} and $\lambda_i$ for $i\in[m]$ be the Lagrangian multipliers corresponding to $\widehat\vw$.  We have 
\begin{align}
    \label{eq:ch3_Lambdabound_wc}
    \|\widehat\vw-\vw\|\leq 2M/\mu\quad\text{ and }\quad
    \textstyle{\sum_{i=1}^m}\lambda_i\leq  \Lambda:=2M/\sqrt{2 \mu\nu}.
\end{align}
\end{lemma}
\begin{proof}
Since $f_i(\vw)\leq 0$ for $i\in[m]$. According to Assumption~\ref{assume:DC}B, there exists $\vv\in\cW$ such that
$$
g_i (\vv)\leq -\nu,~\forall i\in[m],
$$
so the conditions \eqref{eq:stationarityw}, \eqref{eq:feasibilityw} and \eqref{eq:complementaryw} hold.

Choose any $\widehat\vf_i^-\in\partial f_i^-(\widehat\vw)$ for $i=0,1,\dots,m$. 
Since $\vn_{\cW}\in \cN_\cW(\widehat\vw)$, we have $0\geq  -\left\langle\vn_{\cW},\widehat\vw - \vw\right\rangle$. This inequality and \eqref{eq:stationarityw} imply
\begin{align*}
    0\geq& \left\langle\vf_0^+-\vf_0^-+\textstyle\sum_{i=1}^m\lambda_i(\vf_i^+-\vf_i^-),\widehat\vw - \vw\right\rangle\\
   =& \left\langle\vf_0^+-\widehat\vf_0^-+\textstyle\sum_{i=1}^m\lambda_i(\vf_i^+-\widehat\vf_i^-),\widehat\vw - \vw\right\rangle+\left\langle\widehat\vf_0^--\vf_0^-+\textstyle\sum_{i=1}^m\lambda_i(\widehat\vf_i^--\vf_i^-),\widehat\vw - \vw\right\rangle\\
   \geq& -\|\vf_0^+-\widehat\vf_0^-+\textstyle\sum_{i=1}^m\lambda_i(\vf_i^+-\widehat\vf_i^-)\|\cdot\|\widehat\vw - \vw\|+\mu(1+\textstyle\sum_{i=1}^m\lambda_i)\|\widehat\vw - \vw\|^2\\
\geq& -2(1+\textstyle\sum_{i=1}^m\lambda_i)M\|\widehat\vw - \vw\|+\mu(1+\textstyle\sum_{i=1}^m\lambda_i)\|\widehat\vw - \vw\|^2,
\end{align*}
where the second inequality is because of the $\mu$-strongly convexity of $f_i^-$ for $i\in[m]$ and the Cauchy-Schwarz inequality, and the last inequality is by Assumption~\ref{assume:DC}D. Reorganizing the inequality above gives the first inequality in \eqref{eq:ch3_Lambdabound_wc}.
 
If $\max_{i\in[m]}\lambda_i=0$, the conclusion holds trivially. Suppose $\max_{i\in[m]}\lambda_i > 0$. Since $\vf_i^+-\vf_i^-\in\partial g_i(\widehat\vw)$, $g_i(\cdot)$ is $\mu$-strongly convex by Assumption~\ref{assume:DC}A and $\vn_{\cW}\in \cN_\cW(\widehat\vw)$,  we have
\begin{align*}
    -\nu\geq g_i(\vv) 
    \geq&~ g_i(\widehat\vw ) + \left\langle\vf_i^+-\vf_i^- +\vn_{\cW}/\left(\textstyle\sum_{i=1}^m\lambda_i\right), \vv - \widehat\vw  \right\rangle + \frac{\mu}{2} \| \vv - \widehat\vw  \|^2.
\end{align*}
Multiplying both sides of this inequality by $\lambda_i$ and summing up for $i\in[m]$, we obtain
\begin{align*}
    -\nu\left({\textstyle\sum_{i=1}^m\lambda_i} \right) \geq&~{\textstyle\sum_{i=1}^m\lambda_ig_i(\widehat\vw ) } +\left\langle{\textstyle\sum_{i=1}^m\lambda_i(\vf_i^+-\vf_i^-)+\vn_{\cW}}, \vv - \widehat\vw  \right\rangle + \frac{\mu}{2}\left({\textstyle\sum_{i=1}^m\lambda_i}\right) \| \vv - \widehat\vw  \|^2\\
     \geq&-\frac{\|\textstyle\sum_{i=1}^m\lambda_i(\vf_i^+-\vf_i^-)+\vn_{\cW}\|^2}{2\mu(\sum_{i=1}^m\lambda_i)}\\
     =&-\frac{\|\vf_0^+-\vf_0^-\|^2}{2\mu(\sum_{i=1}^m\lambda_i)}\\
     \geq&-\frac{2M^2}{\mu(\sum_{i=1}^m\lambda_i)},
\end{align*}
where the second inequality is because of \eqref{eq:complementaryw} and the Young's inequality, the equality is by \eqref{eq:stationarityw}, and the last inequality is by Assumption~\ref{assume:DC}D. Reorganizing the inequality above gives the second inequality in \eqref{eq:ch3_Lambdabound_wc}.
\end{proof}

\subsection{Near feasibility of solutions}
In order to apply Lemma~\ref{thm:dualbound} and Lemma~\ref{thm:nearlyKKT} to each iteration of Algorithm~\ref{alg:dca}, we need to prove that the following result, which is motivated by Lemma 3.5 in \citet{jia2025first} for weakly convex problems.
\begin{lemma}
\label{thm:nearfeasibility}
Suppose $\vw^{(k)}$ is not a nearly $\epsilon$-KKT solution of \eqref{DC}, $f_i(\vw^{(k)})\leq 0$ for $i\in[m]$ and 
\begin{align}
\label{eq:epsilonkcon1}
\epsilon_k\leq \frac{\mu}{8}\min\left\{1,\frac{1}{4M^2},\frac{\mu\nu}{8M^4}\right\}\cdot\epsilon^2
\end{align}
in the $k$-th iteration of Algorithm~\ref{alg:dca}. Then $f_i(\vw^{(k+1)})\leq 0$ for $i\in[m]$.
\end{lemma}
\begin{proof}
Let $g_i(\cdot)$ be defined as in \eqref{DCA}. Let $\vlam^{(k)}=(\lambda_1^{(k)},\dots,\lambda_m^{(k)})^\top\in\mathbb{R}_+^m$ be the vector of Lagrangian multipliers corresponding to $\widehat\vw^{(k+1)}$. Since the Lagrangian function $g_0(\vv)+\sum_{i=1}^m\lambda_i^{(k)}g_i(\vv)$ is $\mu(1+\sum_{i=1}^m\lambda_i^{(k)})$-strongly convex and is minimized at $\widehat\vw^{(k+1)}$ over $\cW$, we have
\begin{align}
\nonumber
&~g_0(\vw^{(k+1)})+ {\textstyle\sum_{i=1}^m\lambda_i^{(k)}g_i(\vw^{(k+1)})}\\\label{eq:gapwk1andhatwk}
\geq&~ g_0(\widehat\vw^{(k+1)})+{\textstyle\sum_{i=1}^m\lambda_i^{(k)}g_i(\widehat\vw^{(k+1)})}+\frac{\mu}{2}\left(1+{\textstyle\sum_{i=1}^m\lambda_i^{(k)}}\right)\|\widehat\vw^{(k+1)}-\vw^{(k+1)}\|^2.
\end{align}
Since $\lambda_i^{(k)}g_i(\widehat\vw^{(k+1)})=0$ for $i\in[m]$ and $\vw^{(k+1)}$ satisfies \eqref{eq:epsilonwk}, the inequality above implies
\begin{align}
\label{eq:inducefeas1}
\|\widehat\vw^{(k+1)}-\vw^{(k+1)}\|\leq \sqrt{\frac{2\epsilon_k}{\mu}}.
\end{align}

If $\|\vw^{(k)}-\widehat\vw^{(k+1)}\|\leq \min\{1,\frac{1}{2M},\frac{\sqrt{2\mu\nu}}{4M^2}\}\cdot\epsilon$ holds, then we must have 
$\|\vw^{(k)}-\widehat\vw^{(k+1)}\|\leq \min\{1,\frac{1}{2M},\frac{1}{2M\max_{i\in[m]}\lambda_i^{(k)}}\}\cdot\epsilon$ by Lemma~\ref{thm:dualbound} and thus $\vw^{(k)}$ is a nearly $\epsilon$-KKT solution of \eqref{DC} by Lemma~\ref{thm:nearlyKKT}. Since we assume  $\vw^{(k)}$ is not a nearly $\epsilon$-KKT solution, we must have 
\begin{align}
\label{eq:inducefeas2}
\|\vw^{(k)}-\widehat\vw^{(k+1)}\|> \min\left\{1,\frac{1}{2M},\frac{\sqrt{2\mu\nu}}{4M^2}\right\}\cdot\epsilon.
\end{align}

By \eqref{eq:inducefeas1} and \eqref{eq:inducefeas2},
\begin{align}
\nonumber
\|\vw^{(k)}-\vw^{(k+1)}\|^2\geq&~\frac{1}{2}\|\vw^{(k)}-\widehat\vw^{(k+1)}\|^2-\|\vw^{(k+1)}-\widehat\vw^{(k+1)}\|^2\\\label{eq:inducefeas3}
>&~\frac{1}{2}\min\left\{1,\frac{1}{4M^2},\frac{\mu\nu}{8M^4}\right\}\cdot\epsilon^2-\frac{2\epsilon_k}{\mu}.
\end{align}
\normalsize
The $\mu$-strong convexity of $f_i^-$, \eqref{eq:epsilonwk} and condition \eqref{eq:epsilonkcon1} thus implies
\begin{align}
\nonumber
f_i(\vw^{(k+1)})\leq&~g_i(\vw^{(k+1)})-\frac{\mu}{2}\|\vw^{(k+1)}-\vw^{(k)}\|^2\\\label{eq:inducefeas4}
\leq&~\epsilon_k-\frac{\mu}{4}\min\left\{1,\frac{1}{4M^2},\frac{\mu\nu}{8M^4}\right\}\cdot\epsilon^2+\epsilon_k
\leq0.
\end{align}
\end{proof}
\subsection{Proof of Theorem~\ref{thm:idca}}
\label{sec:idca}
\begin{proof}[Proof of Theorem~\ref{thm:idca}.]
Recall that $\epsilon_k$ satisfies \eqref{eq:epsilonkcon1} for  $k=0,1,\dots,K-1$ and $f_i(\vw^{(0)})\leq 0$ for $i\in[m]$. If $\vw^{(0)}$ is not a nearly $\epsilon$-KKT solution of \eqref{DC}, we must have $f_i(\vw^{(1)})\leq 0$ for $i\in[m]$ according to Lemma~\ref{thm:nearfeasibility}. Repeating this argument for $k=1,\dots,K-1$, we can show that, if $\vw^{(k)}$ is not a nearly $\epsilon$-KKT solution for $k=0,1,\dots,K-1$, it must hold that $f_i(\vw^{(k+1)})\leq 0$ for $i\in[m]$ and for $k=0,1,\dots,K-1$. This allows us to apply Lemma~\ref{thm:dualbound} and Lemma~\ref{thm:nearlyKKT} to each iteration of Algorithm~\ref{alg:dca}.

Similar to \eqref{eq:gapwk1andhatwk}, we have
\begin{align}
\nonumber
&~f_0(\vw^{(k)})+ {\textstyle\sum_{i=1}^m\lambda_i^{(k)}f_i(\vw^{(k)})}\\\nonumber
=&~g_0(\vw^{(k)})+ {\textstyle\sum_{i=1}^m\lambda_i^{(k)}g_i(\vw^{(k)})}\\\nonumber
\geq&~ g_0(\widehat\vw^{(k+1)})+{\textstyle\sum_{i=1}^m\lambda_i^{(k)}g_i(\widehat\vw^{(k+1)})}+\frac{\mu}{2}\left(1+{\textstyle\sum_{i=1}^m\lambda_i^{(k)}}\right)\|\widehat\vw^{(k+1)}-\vw^{(k)}\|^2\\\nonumber
\geq&~g_0(\vw^{(k+1)})-\epsilon_k+\frac{\mu}{2}\left(1+{\textstyle\sum_{i=1}^m\lambda_i^{(k)}}\right)\|\widehat\vw^{(k+1)}-\vw^{(k)}\|^2\\\nonumber
\geq&~f_0(\vw^{(k+1)})+\frac{\mu}{2}\|\vw^{(k)}-\vw^{(k+1)}\|^2-\epsilon_k+\frac{\mu}{2}\left(1+{\textstyle\sum_{i=1}^m\lambda_i^{(k)}}\right)\|\widehat\vw^{(k+1)}-\vw^{(k)}\|^2\\\label{eq:gapwkandhatwk}
>&~f_0(\vw^{(k+1)})+\frac{\mu}{2}\min\left\{1,\frac{1}{4M^2},\frac{\mu\nu}{8M^4}\right\}\cdot\epsilon^2,
\end{align}
where the third inequality is by the $\mu$-strong convexity of $f_0^-$. Recall that $f_i(\vw^{(k)})\leq 0$. Summing up both sides of \eqref{eq:gapwkandhatwk} for $k=0,1,\dots,K-1$ and dividing them by $K$ gives 
\begin{align}
\nonumber
\frac{\mu}{2}\min\left\{1,\frac{1}{4M^2},\frac{\mu\nu}{8M^4}\right\}\cdot\epsilon^2
<\frac{1}{K}\left(f_0(\vw^{(0)})-f_0(\vw^{(K)})\right)
\leq\frac{1}{K}\left( f_0(\vw^{(0)})-f_{\text{lb}}\right),
\end{align}
contradicting with \eqref{eq:K}. In fact, with 
$$
K=\left\lceil2\max\left\{1,4M^2,\frac{8M^4}{\mu\nu}\right\}\frac{\left( f_0(\vw^{(0)})-f_{\text{lb}}\right)}{\mu\epsilon^2}\right\rceil,
$$
we have
\begin{align*}
    \frac{\mu}{2}\min\left\{1,\frac{1}{4M^2},\frac{\mu\nu}{8M^4}\right\}\cdot\epsilon^2
    =&~\frac{\mu\epsilon^2}{2\max\left\{1,4M^2,\frac{8M^4}{\mu\nu}\right\}}\\
    <&~\frac{\mu\epsilon^2 (f_0(\vw^{(0)})-f_0(\vw^{(K)}))}{2\max\left\{1,4M^2,\frac{8M^4}{\mu\nu}\right\}( f_0(\vw^{(0)})-f_{\text{lb}})}
    \leq\frac{1}{K}\left(f_0(\vw^{(0)})-f_0(\vw^{(K)})\right).
\end{align*}
This contradiction means one of $\vw^{(k)}$ is a nearly $\epsilon$-KKT solution of \eqref{DC}.
\end{proof}

\section{Generalization bounds for models solved from~\eqref{eq:inprocess_pdp_approx}}
\label{sec:generalization}
In this section, we provide statistical guarantees for the fairness constraints and the objective in problem~\eqref{eq:inprocess_pdp_approx}. Since the objective corresponds to a standard expected loss (e.g., logistic or squared loss), its generalization properties follow from classical results. Therefore, we focus on the generalization behavior of the fairness constraints.

We analyze the statistical guarantees of the fairness constraints for solutions of~\eqref{eq:inprocess_pdp_approx} in three steps. For simplicity, we assume $n_k$ is the same for each $k\in\mathcal{G}$.

\paragraph{Step 1: Discretization error.}
By Lemma~\ref{eq:pdp_equiv}, the constraints in~\eqref{eq:inprocess_pdp} are equivalent to~\eqref{eq:pdpnew}. We approximate~\eqref{eq:pdpnew} using a finite index set $\widehat{\mathcal{I}} \subset \mathcal{I}$. Assume that the points in $\widehat{\mathcal{I}}$ are equally spaced with spacing $(\beta-\alpha)/|\widehat{\mathcal{I}}|$. Then, by Lipchitz continuity, if $\mathbf{w}$ satisfies~\eqref{eq:pdpnew} for all $p \in \widehat{\mathcal{I}}$, it also satisfies the constraints for all $p \in \mathcal{I}$ up to an error of order $O\left(\frac{\beta-\alpha}{|\widehat{\mathcal{I}}|}\right)$. Consequently, by Lemma~\ref{eq:pdp_equiv}, the original constraints  in~\eqref{eq:inprocess_pdp} are satisfied up to an error of order $O\left(\frac{1}{|\widehat{\mathcal{I}}|}\right)$.

\paragraph{Step 2: Statistical estimation error.}
For each $p \in \widehat{\mathcal{I}}$, we approximate the conditional probabilities in~\eqref{eq:pdpnew} using empirical averages of indicator functions. More specifically, for each $p \in \widehat{\mathcal{I}}$, we approximate \eqref{eq:pdpnew} by 
\begin{align}
\label{eq:pdpnew_approx}
    &\textstyle{\frac{1}{n_k} \sum_{i=1}^{n_k} \mathbf{1}_{h_{\vw}(\vxi_i^k)>\theta_{p}}}\geq p~~\text{  and  }\\\nonumber
    &\textstyle{\frac{1}{n_k} \sum_{i=1}^{n_k} \mathbf{1}_{h_{\vw}(\vxi_i^k)>\theta_{p}}}\leq p+\kappa(\beta-\alpha),~~\forall k\in\mathcal{G}.
\end{align}
Standard uniform convergence results (see, e.g.,~\cite{donini2018empirical, liang2016cs}) imply that, with probability at least $1-\delta$,
\begin{align*}
    \sup_{\mathbf{w}\in\mathcal{W}, ~\theta_p\in\mathbb{R}}
    \left|
    \text{Pr}\left(h_{\mathbf{w}}(\xi) > \theta_p \mid \gamma = k\right)
    - \frac{1}{n_k}\sum_{i=1}^{n_k} \mathbf{1}\{h_{\mathbf{w}}(\xi_i^k) > \theta_p\}
    \right|
    \leq E(n_k),
\end{align*}
for all $k \in \mathcal{G}$ and $p\in\widehat{\mathcal{I}}$, where
$$
E(n_k)
:=\sqrt{\frac{8\,VC\,(\log n_k + 1)}{n_k}}
+ \sqrt{\frac{2\log( 4|\widehat{\mathcal{I}}||\mathcal{G}|/\delta )}{n_k}},
$$
and $VC$ denotes the VC-dimension of the function class
$$
\left\{
\xi \mapsto \mathbf{1}\{h_{\mathbf{w}}(\xi) > \theta_p\}~|~ 
\mathbf{w} \in \mathcal{W},~\theta_p \in \mathbb{R}
\right\}.
$$
This implies that any $\mathbf{w}$ satisfying the constraints in~\eqref{eq:pdpnew_approx} for each $p \in \widehat{\mathcal{I}}$ also satisfies the population constraints in~\eqref{eq:pdpnew} $p \in \widehat{\mathcal{I}}$ up to an error of $O(E(n_k))$.

\paragraph{Step 3: Surrogate approximation error.}
In~\eqref{eq:inprocess_pdp_approx}, the indicator function $\mathbf{1}\{h_{\mathbf{w}}(\xi)>\theta_p\}$ in \eqref{eq:pdpnew_approx} has been replaced by a smooth surrogate $\sigma(h_{\mathbf{w}}(\xi)-\theta_p)$. The resulting approximation error, known as the \emph{surrogate fairness gap}, has been studied in~\cite{yao2024understanding}. Following arguments similar to Theorem~2 in~\cite{yao2024understanding}, we obtain that if $\mathbf{w}$ satisfies a surrogate constraint in~\eqref{eq:inprocess_pdp_approx} for some $k \in \mathcal{G}$ and some $p\in\widehat{\mathcal{I}}$, then for any $\gamma \in (0,1)$, it also satisfies the constraint in \eqref{eq:pdpnew_approx} for the same $k \in \mathcal{G}$ and $p\in\widehat{\mathcal{I}}$ up to an error of
$$
O\left(\gamma + \frac{n_k^\gamma}{n_k}\right),
$$
where $n_k^\gamma$ denotes the number of training samples in group $k$ such that
$$
\sigma\big(|h_{\mathbf{w}}(\xi) - \theta_p|\big) \leq 1 - \gamma.
$$

Combining the three steps with the triangle inequality easily yields the following generalization guarantees.

\begin{theorem}[Generalization bound for fairness constraints]
\label{thm:generalization_constraint}
Let $\mathbf{w}$ be a feasible solution of~\eqref{eq:inprocess_pdp_approx} with parameter $\kappa$. Then, with probability at least $1-\delta$, $\mathbf{w}$ satisfies the constraints of~\eqref{eq:inprocess_pdp} with $\kappa$ there replaced by  $\kappa+O\left(\frac{1}{|\widehat{\mathcal{I}}|}+\frac{1}{\beta-\alpha}
\max_{k\in\mathcal{G}} \left(\gamma + \frac{n_k^\gamma}{n_k} + E(n_k)
\right)\right)$. Equivalently, $\mathbf{w}$ satisfies partial statistical parity up to an error of $\kappa+O\left(\frac{1}{|\widehat{\mathcal{I}}|}+\frac{1}{\beta-\alpha}
\max_{k\in\mathcal{G}} \left(\gamma + \frac{n_k^\gamma}{n_k} + E(n_k)
\right)\right)$.
\end{theorem}


Let $f_0(\mathbf{w}) = \mathbb{E}[\ell(h_{\mathbf{w}}(\xi), \zeta)]$ denote the population risk, where $\ell(\cdot,\cdot)$ is a loss function, and let
$$
\hat f_0(\mathbf{w})
= \frac{1}{n_0} \sum_{i=1}^{n_0}
\ell(h_{\mathbf{w}}(\xi_i^0), \zeta_i^0)
$$
be the empirical risk. Let $\widehat{\mathbf{w}}$ be the optimal solution of the following optimization problem
\begin{align}
\label{eq:inprocess_pdp_approx_obj}
    \min_{\vw\in\cW,~(\theta_p)_{p\in\widehat{\mathcal{I}}}}
    &~\hat{f}_0(\vw)\\\nonumber
    \text{s.t. }~\,\quad
    &~\frac{1}{n_k} \sum_{i=1}^{n_k} \sigma(h_{\vw}(\vxi_i^k)-\theta_{p})\geq p,
    ~\forall p\in\widehat{\mathcal{I}},k\in\mathcal{G},\\\nonumber
    &~\frac{1}{n_k} \sum_{i=1}^{n_k} \sigma(h_{\vw}(\vxi_i^k)-\theta_{p})\leq p+\kappa(\beta-\alpha),
    ~\forall p\in\widehat{\mathcal{I}},k\in\mathcal{G},
\end{align}
and let $\mathbf{w}^*$ be the optimal solution of the following problem 
\begin{align}
\label{eq:inprocess_pdp_smallkappa}
    \min_{\vw\in\cW} f_0(\vw) 
    &~\text{  s.t. }
    ~\text{SP}_{k,k'}^{\mathcal{I}}(\vw)\leq \kappa',~\forall k, k'\in\mathcal{G},
\end{align}
which is the same as~\eqref{eq:inprocess_pdp} except that $\kappa$ in~\eqref{eq:inprocess_pdp} is replaced by 
$$
\kappa'=\kappa
- O\left(\frac{1}{|\widehat{\mathcal{I}}|}+\frac{1}{\beta-\alpha}
\max_{k\in\mathcal{G}} \left(\gamma + \frac{n_k^\gamma}{n_k} + E(n_k)
\right)\right).
$$
According to the proof of Theorem~\ref{thm:generalization_constraint}, we can show that, with probability at least $1-\delta$, $\mathbf{w}^*$ must be a feasible solution of~\eqref{eq:inprocess_pdp_approx_obj} and thus $\hat f_0(\widehat{\mathbf{w}})\leq\hat f_0(\mathbf{w}^*)$. 

Therefore, we have 
$$
f_0(\widehat{\mathbf{w}}) - f_0(\mathbf{w}^*)
=\left[f_0(\widehat{\mathbf{w}})-\hat f_0(\widehat{\mathbf{w}})\right]+\left[\hat f_0(\widehat{\mathbf{w}})-\hat f_0(\mathbf{w}^*)\right]+\left[\hat f_0(\mathbf{w}^*)- f_0(\mathbf{w}^*)\right],
$$
and the second term here is less than or equal to zero. The first and third terms can be bounded by the standard uniform convergence results (see, e.g.,~\cite{donini2018empirical, liang2016cs}). Therefore, we have the following statistical guarantee on the objective value of $\widehat{\mathbf{w}}$ relative to the optimal value of \eqref{eq:inprocess_pdp_smallkappa}. 

\begin{theorem}[Generalization bound for the objective]
Let $\widehat{\mathbf{w}}$ be the optimal solution of \eqref{eq:inprocess_pdp_approx_obj} and $\mathbf{w}^*$  be the optimal solution of \eqref{eq:inprocess_pdp_smallkappa}. Then, with probability at least $1-\delta$,
$$
f_0(\widehat{\mathbf{w}}) - f_0(\mathbf{w}^*)
\leq
\mathcal{R}_{n_0}
+
\sqrt{\frac{2\log(4|\widehat{\mathcal{I}}||\mathcal{G}|/\delta)}{n_0}},
$$
where $\mathcal{R}_{n_0}$ denotes the Rademacher complexity of the function class
$$
\left\{
(\xi,\zeta) \mapsto \ell(h_{\mathbf{w}}(\xi), \zeta)~|~\mathbf{w} \in \mathcal{W}
\right\}.
$$
\end{theorem}

A similar argument yields generalization guarantees for the partial demographic parity formulation in~\eqref{eq:inprocess_wpdp_approx}.

\section{Additional details of numerical experiments}
\label{sec:additionalexp}
In this section, additional details and results are provided for the numerical experiments in Section~\ref{sec:exp}.

\subsection{Details of Datasets}
\label{sec:data}
The details of the three public datasets we used in our numerical experiments are as follows.
\begin{itemize}
    \item With \textit{a9a} dataset, the task is to predict if the annual income of an individual exceeds \$50K. Gender is  the sensitive attribute that splits the data into female ($\gamma=1$) and male ($\gamma=2$) groups.
    \vspace{-0.05in}
    \item With \textit{bank} dataset, the task is to predict if a client will subscribe a term deposit with a bank. Age is the sensitive attribute that splits the data into the clients with ages between 25 and 60 ($\gamma=1$) and clients with other ages ($\gamma=2$).
    \vspace{-0.05in}
    \item With \textit{law school} dataset, the task is to predict if a student will pass the bar exam of law schools on the first try. Race is the sensitive attribute that splits the data into white ($\gamma=1$) or non-white ($\gamma=2$) groups.
    \vspace{-0.05in}
    \item With \textit{ACS} dataset, the task is to predict whether a US working adults’ yearly income is above $50,000$. Gender is the sensitive attribute that splits the data into female ($\gamma=1$) and male ($\gamma=2$) groups.
\end{itemize}
Some statistics of these datasets are given in Table \ref{tbl:data}. Data \textit{a9a} was provided with separated training and testing sets, so we further randomly split the original training data into a new training set (90\%) and a validation set (10\%). For \textit{bank} dataset, we randomly split it into training (60\%), validation (20\%) and testing (20\%) sets. For \textit{law school} dataset, we randomly split it into training (56.25\%), validation (18.75\%) and testing (25\%) sets. For \textit{ACS} dataset, we randomly split it into training (60\%), validation (20\%) and testing (20\%) sets.

The training sets are used to build the models. The validation sets are only used for tuning hyper-parameters. The testing sets are used to evaluate the prediction performance and the fairness of the obtained models. The variation of the performance of each method caused by the aforementioned random data splitting is displayed by the error bars (95\% confidence intervals) in Figure~\ref{fig:pDP_and_wpDP_efficiency_frontier}.

\begin{table}[!ht]
\caption{Statistics of the datasets.}
\label{tbl:data}
\vskip 0.1in
\centering
\begin{tabular}{c|ccccc}
\toprule
 Datasets & \#Instances & \#Attributes & Class Label & Sensitive Attribute \\
\midrule
a9a & 48,842 & 123 & Income & Gender\\
bank & 41,188 & 54 & Subscription & Age\\
law school & 20,798 & 12 & Passing exam & Race\\
ACS & 1,664,500 & 813 & Income & Gender\\
\bottomrule
\end{tabular}
\end{table}

\subsection{Implementation details of the proposed methods}
\label{sec:details_IDCA}
In this section, we present the implementation details of the proposed methods that produce the numerical results in Section~\ref{sec:exp}, including the choices of $\kappa$ in \eqref{eq:inprocess_pdp_approx} and \eqref{eq:inprocess_wpdp_approx} and the choices of tuning parameters in IDCA (Algorithm~\ref{alg:dca}).

\subsubsection{Choices of $\kappa$ in \eqref{eq:inprocess_pdp_approx} and \eqref{eq:inprocess_wpdp_approx} }
\label{sec:kappa}
Through the experiments, the prediction performance and the fairness of a model solved from \eqref{eq:inprocess_pdp_approx} or \eqref{eq:inprocess_wpdp_approx} do not change with $\kappa$ at a consistent pace. Therefore, the points in the Pareto frontier will not be nicely spaced if we choose $\kappa$ from an equally spaced sequence within $[0,1]$. For a better visualization, we choose $\kappa$ manually for different datasets and different fairness metrics such that the points along the Pareto frontiers in Figure~\ref{fig:pDP_and_wpDP_efficiency_frontier} are reasonably spaced. Also, we always make sure a small enough $\kappa$ is included such that the frontiers can approach a high level of fairness (towards the right end of the $x$-axis).

The values of $\kappa$ we used  in different cases are presented below.
\begin{itemize}
    \item For problem \eqref{eq:inprocess_pdp_approx} (pSP fairness):

    \textit{a9a}: $\kappa \in \{ 0.01, 0.05, 0.1, 0.2, 0.28 \}$.

    \textit{bank}: $\kappa \in \{ 0.01, 0.08, 0.22, 0.3 \}$.

    \textit{law school}: $\kappa \in \{ 0.005, 0.08, 0.1, 0.15, 0.2 \}$.

    \item For problem \eqref{eq:inprocess_wpdp_approx} (pDP fairness):

    \textit{a9a}: $\kappa \in \{ 0.005, 0.03, 0.05, 0.1, 0.15 \} $.

    \textit{bank}: $\kappa \in \{ 0.01, 0.03, 0.05, 0.07, 0.1 \}$.

    \textit{law school}: $\kappa \in \{ 0.005, 0.04, 0.06, 0.1, 0.15 \}$.
\end{itemize}

\subsubsection{Choices of tuning parameters in IDCA}
\label{sec:parameter_IDCA}
For each $\kappa$ listed in Section~\ref{sec:kappa}, we apply IDCA to \eqref{eq:inprocess_pdp_approx} and \eqref{eq:inprocess_wpdp_approx} with the initial solution $\vw^{(0)}=\mathbf{0}$. For each $\kappa$ and each random splitting of training, validation, and testing sets, we select $T$ from $\{150, 200\}$ and set $\epsilon_k=\epsilon$ with $\epsilon$ selected from $\{5 \times 10^{-4}, 10^{-3}, 2 \times 10^{-3}, 5 \times 10^{-3}\}$. We first select the combination of $T$ and $\epsilon$ that produces the highest classification accuracy on the validation set after 50 outer iterations. Then we use the best combination of $T$ and $\epsilon$ to run IDCA with the total number of outer iterations $K$ selected from $\{100, 150, 200, 250, 300, 350, 400\}$ that produces the highest accuracy on the validation set. Then we plot the classification accuracies and fairness values of the final solution on the testing set to create the Pareto frontiers in the figures.

\subsection{Implementation details of the in-processing baselines in comparison}
\label{sec:benchmark}
In this section, we present the optimization models of the in-processing methods we compare with in Section~\ref{sec:exp} and describe the numerical methods used to solve these models. 

\subsubsection{Constrained optimization models for the in-processing baselines}
\label{sec:baselinemodel}
We compare our methods with other in-processing methods that solve constrained optimization problems with the same objective function as our approach but different fairness constraints. More specifically, we obtain a model by minimizing the same $f_0(\vw)$ as in \eqref{eq:inprocess_pdp_approx} and \eqref{eq:inprocess_wpdp_approx} subject to the constraints that enforce Group AUC Fairness~\citep{yao2023stochastic,vogel2021learning}, Inter-Group  Pairwise Fairness~\citep{kallus2019fairness,beutel2019fairness}, and Intra-Group Pairwise Fairness~\citep{beutel2019fairness}. The three models are defined below.

\begin{baseline}
\label{ex:gauc}
\textbf{Group AUC Fairness}. Suppose the data points are ranked in descending order based on their scores $h_{\vw}(\vxi)$. Group AUC fairness requires that, with probability $50\%$, a random data point sampled from any group is not ranked lower than a random data point sampled from a different group,  that is, \eqref{eq:gauc} holds. Recall that we have partitioned the training data into subsets based on $\gamma$, namely, $\mathcal{D}=\cup_{k\in\cG}\mathcal{D}_k$, where
$$
\mathcal{D}_k=\{(\vxi,\zeta,\gamma)\in\mathcal{D}|\gamma=k\}=\{(\vxi_i^k,\zeta_i^k,k)\}_{i=1}^{n_k},
$$
and $n_k$ is the size of data from group $k$. Similar to \eqref{eq:aucloss}, we will approximate the probability above using the sample average of indicator $\mathbf{1}_{h_{\vw}(\vxi)-h_{\vw}(\vxi')\geq0}$ over all data pairs between groups $k$ and $k'$. Then we approximate the indicator function by a surrogate $\sigma(h_{\vw}(\vxi)-h_{\vw}(\vxi'))$. This leads to the following optimization problem similar to \eqref{eq:inprocess_pdp_approx} and \eqref{eq:inprocess_wpdp_approx},
\begin{align}
\label{eq:inprocess_gauc_approx}
\min_{\vw\in\cW} &~f_0(\vw)\\\nonumber
\text{s.t.}~&~
\frac{1}{n_kn_{k'}}\sum_{i=1}^{n_k}\sum_{j=1}^{n_{k'}}\sigma(h_{\vw}(\vxi_i^k)-h_{\vw}(\vxi_j^{k'}))-0.5\leq \kappa,~\forall k, k'\in\mathcal{G}.
\end{align}
Following~\citet{yao2023stochastic}, we choose $\sigma(\cdot)$ to be the quadratic surrogate in the numerical experiments, namely, 
\begin{align}
\label{eq:ellquadratic}
\sigma(x)=\frac{1}{2}(1+x)^2.
\end{align}
\end{baseline}

\begin{baseline}
\label{ex:igpf}
\textbf{Inter-Group Pairwise Fairness}. Suppose the data points are ranked in descending order based on their scores $h_{\vw}(\vxi)$. Inter-Group Pairwise Fairness requires that the probability of ranking a random positive data point from group $k$ higher than a random negative data point from group $k'$ must be the same as the probability of ranking a random positive data point from group $k'$ higher than a random negative data point from group $k$. In other words, it requires 
\begin{align*}
&~\text{Pr}(h_{\vw}(\vxi)\geq h_{\vw}(\vxi') | \gamma=k , \gamma'=k', \zeta=1, \zeta'=-1)\\
=&~\text{Pr}(h_{\vw}(\vxi')\geq h_{\vw}(\vxi) | \gamma=k , \gamma'=k', \zeta=-1, \zeta'=1 ),~\forall k, k'\in\mathcal{G},
\end{align*}
where $ (\vxi,\zeta,\gamma)$ and $ (\vxi',\zeta',\gamma')$ are two i.i.d. random data points. It can be reformulated equivalently as 
\begin{align}
\nonumber
&~\text{Pr}(h_{\vw}(\vxi)\geq h_{\vw}(\vxi') | \gamma=k , \gamma'=k', \zeta=1, \zeta'=-1)\\\label{eq:igpf}
+&~\text{Pr}(h_{\vw}(\vxi)\geq h_{\vw}(\vxi') | \gamma=k , \gamma'=k', \zeta=-1, \zeta'=1 )=1,~\forall k, k'\in\mathcal{G}.
\end{align}
To derive a tractable approximation of \eqref{eq:igpf}, we first partition the training data into subsets based on $\gamma$ and $\zeta$, namely, $\mathcal{D}=(\cup_{k\in\cG}\mathcal{D}_{k,+})\cup(\cup_{k\in\cG}\mathcal{D}_{k,-})$, where
$$
\mathcal{D}_{k,+}=\{(\vxi,\zeta,\gamma)\in\mathcal{D}|\gamma=k,\zeta=1\}=\{(\vxi_i^{k,+},1,k)\}_{i=1}^{n_{k,+}}
$$
and
$$
\mathcal{D}_{k,-}=\{(\vxi,\zeta,\gamma)\in\mathcal{D}|\gamma=k,\zeta=-1\}=\{(\vxi_i^{k,-},-1,k)\}_{i=1}^{n_{k,-}},
$$
where $n_{k,+}$ and $n_{k,-}$ are the group sizes ($n_{k,+}+n_{k,-}=n_{k}$). Similar to \eqref{eq:inprocess_gauc_approx}, using sample approximation and surrogates, we obtain the following optimization problem based on fairness metric~\eqref{eq:igpf},
\begin{align}
\label{eq:inprocess_igpf_approx}
\min_{\vw\in\cW} &~f_0(\vw)\\\nonumber
\text{s.t.}~&~
\frac{1}{n_{k,+}n_{k',-}}\sum_{i=1}^{n_{k,+}}\sum_{j=1}^{n_{k',-}}\sigma(h_{\vw}(\vxi_i^{k,+})-h_{\vw}(\vxi_j^{k',-}))\\\nonumber
&
+\frac{1}{n_{k,-}n_{k',+}}\sum_{i=1}^{n_{k,-}}\sum_{j=1}^{n_{k',+}}\sigma(h_{\vw}(\vxi_j^{k,-})-h_{\vw}(\vxi_i^{k',+}))-1
\leq \kappa,~\forall k, k'\in\mathcal{G}.
\end{align}
We choose $\sigma(\cdot)$ to be \eqref{eq:ellquadratic} also.
\end{baseline}

\begin{baseline}
\label{ex:itgpf}
\textbf{Intra-Group Pairwise Fairness}. Suppose the data points are ranked in descending order based on their scores $h_{\vw}(\vxi)$. Intra-Group Pairwise Fairness requires that the probability of ranking a random positive data point from group $k$ higher than a random negative data point from group $k$ must be the same as the probability of ranking a random positive data point from group $k'$ higher than a random negative data point from group $k'$. In other words, it requires 
\begin{align*}
&~\text{Pr}(h_{\vw}(\vxi)\geq h_{\vw}(\vxi') | \gamma=\gamma'=k, \zeta=1, \zeta'=-1)\\
=&~\text{Pr}(h_{\vw}(\vxi)\geq h_{\vw}(\vxi') | \gamma=\gamma'=k', \zeta=1, \zeta'=-1 ),~\forall k, k'\in\mathcal{G},
\end{align*}
where $ (\vxi,\zeta,\gamma)$ and $ (\vxi',\zeta',\gamma')$ are two i.i.d. random data points. It can be reformulated equivalently as 
\begin{align}
\nonumber
&~\text{Pr}(h_{\vw}(\vxi)\geq h_{\vw}(\vxi') | \gamma=\gamma'=k, \zeta=1, \zeta'=-1)\\\label{eq:itgpf}
+&~\text{Pr}(h_{\vw}(\vxi')\geq h_{\vw}(\vxi)  | \gamma=\gamma'=k', \zeta=1, \zeta'=-1 )=1,~\forall k, k'\in\mathcal{G}.
\end{align}
Similar to \eqref{eq:inprocess_igpf_approx}, we obtain the following optimization problem based on fairness metric~\eqref{eq:itgpf},
\begin{align}
\label{eq:inprocess_itgpf_approx}
\min_{\vw\in\cW} &~f_0(\vw)\\\nonumber
\text{s.t.}~&~
\frac{1}{n_{k,+}n_{k,-}}\sum_{i=1}^{n_{k,+}}\sum_{j=1}^{n_{k,-}}\sigma(h_{\vw}(\vxi_i^{k,+})-h_{\vw}(\vxi_j^{k,-}))\\\nonumber
&
+\frac{1}{n_{k',-}n_{k',+}}\sum_{i=1}^{n_{k',-}}\sum_{j=1}^{n_{k',+}}\sigma(h_{\vw}(\vxi_j^{k',-})-h_{\vw}(\vxi_i^{k',+}))-1
\leq \kappa,~\forall k, k'\in\mathcal{G}.
\end{align}
We choose $\sigma(\cdot)$ to be \eqref{eq:ellquadratic} once again.
\end{baseline}

\subsubsection{Training algorithm for the in-processing baselines}
\label{sec:baselinesetup}
When implementing the three in-processing baselines in the  numerical experiments, we set the objective function $f_0(\vw)$ in  \eqref{eq:inprocess_gauc_approx}, \eqref{eq:inprocess_igpf_approx} and \eqref{eq:inprocess_itgpf_approx} to be the same as the one used in our methods, which is defined in Section~\ref{sec:pdpexp}. We then solve \eqref{eq:inprocess_gauc_approx}, \eqref{eq:inprocess_igpf_approx} and \eqref{eq:inprocess_itgpf_approx} using the projected gradient descent method. In particular, starting with $\vw^{(0)}=\mathbf{0}$, we generate  $\vw^{(k)}$ for $k=0,1,\dots,K-1$ as 
\begin{align}
\label{eq:gd_inprocess_gauc_approx}
\vw^{(k+1)}=\argmin_{\vw\in\cW} \frac{1}{2}\left\|\vw-\left(\vw^{(k)}-\eta_k\nabla f_0(\vw^{(k)})\right)\right\|^2\text{ s.t. }\vw\text{ satisfies the constraints in }\eqref{eq:inprocess_gauc_approx},
\end{align}
\begin{align}
\label{eq:gd_inprocess_igpf_approx}
\vw^{(k+1)}=\argmin_{\vw\in\cW} \frac{1}{2}\left\|\vw-\left(\vw^{(k)}-\eta_k\nabla f_0(\vw^{(k)})\right)\right\|^2\text{ s.t. }\vw\text{ satisfies the constraints in }\eqref{eq:inprocess_igpf_approx},
\end{align}
\begin{align}
\label{eq:gd_inprocess_itgpf_approx}
\vw^{(k+1)}=\argmin_{\vw\in\cW} \frac{1}{2}\left\|\vw-\left(\vw^{(k)}-\eta_k\nabla f_0(\vw^{(k)})\right)\right\|^2\text{ s.t. }\vw\text{ satisfies the constraints in }\eqref{eq:inprocess_itgpf_approx},
\end{align}
for solving  \eqref{eq:inprocess_gauc_approx}, \eqref{eq:inprocess_igpf_approx} and \eqref{eq:inprocess_itgpf_approx}, respectively, where $\eta_k$ is the step size.

Because of \eqref{eq:ellquadratic} and the fact that the model $h_{\vw}(\vxi)$ we choose is linear in $\vw$, the constraints in  \eqref{eq:inprocess_gauc_approx}, \eqref{eq:inprocess_igpf_approx} and \eqref{eq:inprocess_itgpf_approx} are all convex quadratic constraints on $\vw$ and the Hessian matrices of the constraint functions only need to be computed once and used in each iteration. Also, the projection problems \eqref{eq:gd_inprocess_gauc_approx}, \eqref{eq:gd_inprocess_igpf_approx} and \eqref{eq:gd_inprocess_itgpf_approx} are convex quadratically constrained quadratic programs (QCQP), which we solve by using \texttt{fmincon} function from MATLAB. \texttt{fmincon} solves the convex QCQPs using the interior-point method with parameters `MaxIterations'=1000, `OptimalityTolerance'=1e-6, and `ConstraintTolerance'=1e-6. We terminate the training algorithm after $K=1000$ iterations for the \textit{a9a} and \textit{bank} datasets and $K=500$ iterations for \textit{the law school} dataset. The numbers of iterations we choose are large enough to ensure the change of the objective value is less than $10^{-5}$ at termination in all cases.

In order to create the Pareto frontiers by the models solved from \eqref{eq:inprocess_gauc_approx}, \eqref{eq:inprocess_igpf_approx} and \eqref{eq:inprocess_itgpf_approx}, we also need to solve them with different values of $\kappa$. Moreover, in order to compare the prediction performances of different models at similar levels of fairness, we choose a sequence of values for $\kappa$ in \eqref{eq:inprocess_gauc_approx}, \eqref{eq:inprocess_igpf_approx} and \eqref{eq:inprocess_itgpf_approx} so the models solved from them produce $\max_{k,k}\text{SP}_{k,k'}^{\mathcal{I}}(\vw)$ and $\max_{k,k}\text{DP}_{k,k'}^{\mathcal{I}}(\vw)$ similar to our methods on the validation set. The values of $\kappa$ we used in the baseline optimization models to produce Figure~\ref{fig:pDP_and_wpDP_efficiency_frontier} are presented below.
\begin{itemize}
    \item For problem \eqref{eq:inprocess_gauc_approx} (pSP fairness):

    \textit{a9a}: $\kappa \in \{ 0.7, 1, 2, 3.5 \}$.

    \textit{bank}: $\kappa \in \{ 0.1, 0.3, 0.56, 0.7, 1 \}$.

    \textit{law school}: $\kappa \in \{ 0.5, 0.75, 1.15, 1.8, 2.7 \}$.

    \item For problem \eqref{eq:inprocess_gauc_approx} (pDP fairness):

    \textit{a9a}: $\kappa \in \{ 0.3, 0.4, 0.5, 0.7, 1 \} $.

    \textit{bank}: $\kappa \in \{ 0.06, 0.5, 2.35, 2.62, 3.28, 4 \}$.

    \textit{law school}: $\kappa \in \{ 1, 3, 4, 5, 6, 10 \}$.

    \item For problem \eqref{eq:inprocess_igpf_approx} (pSP fairness):

    \textit{a9a}: $\kappa \in \{ 1.5, 3, 3.5, 6 \}$.

    \textit{bank}: $\kappa \in \{ 1.5, 1.8, 2, 3 \}$.

    \textit{law school}: $\kappa \in \{ 0.4, 1.6, 2.4, 4 \}$.

    \item For problem \eqref{eq:inprocess_igpf_approx} (pDP fairness):

    \textit{a9a}: $\kappa \in \{ 0.9, 1.2, 1.5, 2, 2.7 \} $.

    \textit{bank}: $\kappa \in \{ 0.3, 1, 5, 5.4, 6 \}$.

    \textit{law school}: $\kappa \in \{ 2, 6.5, 8, 9, 10, 15 \}$.

    \item For problem \eqref{eq:inprocess_itgpf_approx} (pSP fairness):

    \textit{a9a}: $\kappa \in \{ 3, 5, 8 \}$.

    \textit{bank}: $\kappa \in \{ 2, 3, 4 \}$.

    \textit{law school}: $\kappa \in \{ 3, 5, 10 \}$.

    \item For problem \eqref{eq:inprocess_itgpf_approx} (pDP fairness):

    \textit{a9a}: $\kappa \in \{ 0.7, 1, 1.3, 1.7, 2.2 \} $.

    \textit{bank}: $\kappa \in \{ 0.1, 0.3, 0.42, 0.5, 1, 2.5 \}$.

    \textit{law school}: $\kappa \in \{ 0.1, 0.8, 2, 3.5, 10 \}$.
\end{itemize}

\subsection{Implementation details for the regularization method in comparison}
\label{sec:regularization_method}
In this section, we present the optimization models of the in-processing methods we compare with in Section~\ref{sec:exp} and describe the numerical methods used to solve these models. 

\subsubsection{Optimization model for the regularization method}
\label{sec:regularization_method_model}
We can also solve \eqref{eq:inprocess_pdp} using a regularization method (also called a penalty method) using either $\max_{k, k'\in\mathcal{G}}\text{SP}_{k,k'}^{\mathcal{I}}(\vw)$ in \eqref{eq:pdp} or $\max_{k, k'\in\mathcal{G}}\text{DP}_{k,k'}^{\mathcal{I}}(\vw)$ in \eqref{eq:wpdp} as a regularization term (also called a penalty term). In particular, given a regularization parameter $\lambda\geq0$, we solve 
\begin{align}
\label{eq:inprocess_pdp_regularized}
&~\min_{\vw\in\cW} f_0(\vw) +\lambda \max_{k, k'\in\mathcal{G}}\text{SP}_{k,k'}^{\mathcal{I}}(\vw),\\\label{eq:inprocess_wpdp_regularized}
&~\min_{\vw\in\cW} f_0(\vw)+\lambda \max_{k, k'\in\mathcal{G}}\text{DP}_{k,k'}^{\mathcal{I}}(\vw),
\end{align}
respectively. We will reformulate \eqref{eq:inprocess_pdp_regularized} and \eqref{eq:inprocess_wpdp_regularized} as unconstrained optimization problems by following a procedure similar to the one we used in \eqref{eq:inprocess_pdp_approx} to derive  \eqref{eq:inprocess_pdp_approx} and \eqref{eq:inprocess_wpdp_approx}. We present this procedure in detail below.

First, computing the subgradient of $\max_{k, k'\in\mathcal{G}}\text{SP}_{k,k'}^{\mathcal{I}}(\vw)$ is difficult because of the two challenges mentioned at the beginning of Section~\ref{sec:optimization_models} (i.e., how to approximate the probabilities) as well as the challenge in taking the maximization over $\theta$ in \eqref{eq:pdp}.  According to Lemma~\ref{eq:pdp_equiv} with $\kappa=0$, the constraint $\text{SP}_{k,k'}^{\mathcal{I}}(\vw)=0$ 
holds if and only if, for any $p\in[\alpha,\beta)$, there exists $\theta_p\in\mathbb{R}$ such that
\begin{equation}
\begin{split}
&~\textup{Pr}(h_{\vw}(\vxi) >\theta_p | \gamma=k )=p,~\forall k\in\cG.\label{eq:pdpnew_kappa0}
\end{split}
\end{equation}


Following the same logic and proof as Lemma~\ref{eq:pdp_equiv}, we can show that 
$$
\max_{k, k'\in\mathcal{G}}\text{SP}_{k,k'}^{\mathcal{I}}(\vw)=\frac{2}{\beta-\alpha}\max_{p\in[\alpha,\beta)}\min_{\theta_p}\max_{k\in\mathcal{G}}\left|\text{Pr}\big(h_{\vw}(\vxi)>\theta_p\big|\gamma=k\big) -p\right|.
$$
Similar to Section~\ref{sec:optimization_models}, we approximate each probability by the empirical distribution and a continuous surrogate $\sigma(x)\approx \mathbf{1}_{x>0}$, and approximate the maximization $\max_{p\in[\alpha,\beta)}$ by $\max_{p\in\widehat{\mathcal{I}}}$ where $\widehat{\mathcal{I}}$ is a finite subset satisfying $\widehat{\mathcal{I}}\subset[\alpha,\beta)$. We then obtain 
\begin{align*}
\max_{k, k'\in\mathcal{G}}\text{SP}_{k,k'}^{\mathcal{I}}(\vw)
\approx&~\frac{2}{\beta-\alpha}\max_{p\in\widehat{\mathcal{I}}}\min_{\theta_p}\max_{k\in\mathcal{G}}\left|\text{Pr}\big(h_{\vw}(\vxi)>\theta_p\big|\gamma=k\big) -p\right|\\
\approx&~\frac{2}{\beta-\alpha}\max_{p\in\widehat{\mathcal{I}}}\min_{\theta_p}\max_{k\in\mathcal{G}}\left|\frac{1}{n_k} \sum_{i=1}^{n_k} \sigma(h_{\vw}(\vxi_i^k)-\theta_{p}) -p\right|.
\end{align*}
Applying this approximation to \eqref{eq:inprocess_pdp_regularized}, we obtain the following regularized variant of \eqref{eq:inprocess_pdp_approx}
\begin{align}
\label{eq:inprocess_pdp_regularized_approx}
\min_{\vw\in\cW,~(\theta_p)_{p\in\widehat{\mathcal{I}}}} f_0(\vw) +\frac{2\lambda }{\beta-\alpha}\max_{p\in\widehat{\mathcal{I}}}\max_{k\in\mathcal{G}}\left|\frac{1}{n_k} \sum_{i=1}^{n_k} \sigma(h_{\vw}(\vxi_i^k)-\theta_{p}) -p\right|.
\end{align}

According to the proof of Lemma~\ref{eq:wpdp_equiv}, we have 
\begin{align*}
\text{DP}_{k,k'}^{\mathcal{I}}(\vw)=\frac{1}{\beta-\alpha}\left|
\begin{array}{l}
\min\{\textup{Pr}(h_{\vw}(\vxi) >\widehat\theta | \gamma=k),\beta\}\\[0.5mm]
-\min\{\textup{Pr}(h_{\vw}(\vxi) >\widehat\theta | \gamma=k),\alpha\}\\[0.5mm]
-\min\{\textup{Pr}(h_{\vw}(\vxi) >\widehat\theta | \gamma=k'),\beta\}\\[0.5mm]
+\min\{\textup{Pr}(h_{\vw}(\vxi) >\widehat\theta | \gamma=k'),\alpha\}
\end{array}
\right|.
\end{align*}
\normalsize
Similar to Section~\ref{sec:optimization_models}, we approximate each probability by the empirical distribution and a continuous surrogate $\sigma(x)\approx \mathbf{1}_{x>0}$ on the right-hand side above, and obtain 
\begin{align*}
\text{DP}_{k,k'}^{\mathcal{I}}(\vw)
\approx\frac{1}{\beta-\alpha}\left|\begin{array}{l}
\min\big\{\frac{1}{n_k} \sum_{i=1}^{n_k} \sigma(h_{\vw}(\vxi_i^k)-\widehat\theta),\beta\big\}\\[0.5mm]
-\min\big\{\frac{1}{n_k} \sum_{i=1}^{n_k} \sigma(h_{\vw}(\vxi_i^k)-\widehat\theta),\alpha\big\}\\[0.5mm]
-\min\big\{\frac{1}{n_{k'}} \sum_{i=1}^{n_{k'}} \sigma(h_{\vw}(\vxi_i^{k'})-\widehat\theta),\beta\big\}\\[0.5mm]
+\min\big\{\frac{1}{n_{k'}} \sum_{i=1}^{n_{k'}} \sigma(h_{\vw}(\vxi_i^{k'})-\widehat\theta),\alpha\big\}
\end{array}\right|.
\end{align*}
Applying this approximation to \eqref{eq:inprocess_wpdp_regularized}, we obtain the following regularized variant of \eqref{eq:inprocess_wpdp_approx}
\begin{align}
\label{eq:inprocess_wpdp_regularized_approx}
\min_{\vw\in\cW} f_0(\vw) +\frac{\lambda}{\beta-\alpha}\left|\begin{array}{l}
\min\big\{\frac{1}{n_k} \sum_{i=1}^{n_k} \sigma(h_{\vw}(\vxi_i^k)-\widehat\theta),\beta\big\}\\[0.5mm]
-\min\big\{\frac{1}{n_k} \sum_{i=1}^{n_k} \sigma(h_{\vw}(\vxi_i^k)-\widehat\theta),\alpha\big\}\\[0.5mm]
-\min\big\{\frac{1}{n_{k'}} \sum_{i=1}^{n_{k'}} \sigma(h_{\vw}(\vxi_i^{k'})-\widehat\theta),\beta\big\}\\[0.5mm]
+\min\big\{\frac{1}{n_{k'}} \sum_{i=1}^{n_{k'}} \sigma(h_{\vw}(\vxi_i^{k'})-\widehat\theta),\alpha\big\}
\end{array}\right|.
\end{align}

\subsubsection{Training algorithm for the regularization method}
\label{sec:regularization_method_training}
We first create a sequence of values of $\lambda$ (given later). For each value of $\lambda$, we apply the standard subgradient descent (SGD) algorithm to \eqref{eq:inprocess_pdp_regularized_approx} and \eqref{eq:inprocess_wpdp_regularized_approx} by fine-tuning the step size. In particular, we select the step size in SGD from $\{5 \times 10^{-4}, 10^{-3}, 2 \times 10^{-3}, 5 \times 10^{-3}\}$ and select the one that produces the highest accuracy on the validation set after $10^{4}$ iterations. Then we run SGD using the selected step size for the same total number of iterations as our method, i.e., $TK$ iterations, and use the last iteration to evaluate the classification accuracies and fairness values on the testing set to create the Pareto frontiers in the figures.


In order to create the Pareto frontiers by the models solved from \eqref{eq:inprocess_pdp_regularized_approx} and \eqref{eq:inprocess_wpdp_regularized_approx}, we also need to solve them with different values of $\lambda$. To compare the prediction performances of different models at similar levels of fairness, we choose a sequence of values for $\lambda$ in \eqref{eq:inprocess_pdp_regularized_approx} and \eqref{eq:inprocess_wpdp_regularized_approx} so the output models produce $\max_{k,k}\text{SP}_{k,k'}^{\mathcal{I}}(\vw)$ and $\max_{k,k}\text{DP}_{k,k'}^{\mathcal{I}}(\vw)$ similar to our methods on the validation sets. The values of $\lambda$ we used in \eqref{eq:inprocess_pdp_regularized_approx} and \eqref{eq:inprocess_wpdp_regularized_approx} to produce Figure~\ref{fig:pDP_and_wpDP_efficiency_frontier} are presented below.
\begin{itemize}
    \item  For problem \eqref{eq:inprocess_pdp_regularized_approx} (pSP fairness):
    
    \textit{a9a}: $\lambda \in \{0.2, 0.3, 0.37, 1\}$.
    
    \textit{bank}: $\lambda \in \{0.4, 0.405, 0.41, 1, 1.5\}$.
    
    \textit{law school}: $\lambda \in \{0.39, 0.4, 0.5, 0.8, 1.5\}$.
    
    \item  For problem \eqref{eq:inprocess_wpdp_regularized_approx} (pDP fairness):
    
    \textit{a9a}: $\lambda \in \{0.13, 0.17, 0.2, 0.23, 1\}$.
    
    \textit{bank}: $\lambda \in \{0.005, 0.015, 0.018, 0.1, 2\}$.
    
    \textit{law school}: $\lambda \in \{0.001, 0.01, 1\}$.
\end{itemize}

\subsection{Changes of score densities with $\kappa$}
\label{sec:density}
\begin{figure*}[!ht]
     \begin{tabular}{@{}c|cccc@{}}
      & unconstrained & $\kappa = 0.1$ & $\kappa = 0.05$ & $\kappa = 0.01$ \\
		\hline \vspace*{-0.1in}\\
		\raisebox{10ex}{\small{\rotatebox[origin=c]{90}{Male}}}
		& \hspace*{-0.06in}\includegraphics[width=0.22\textwidth]{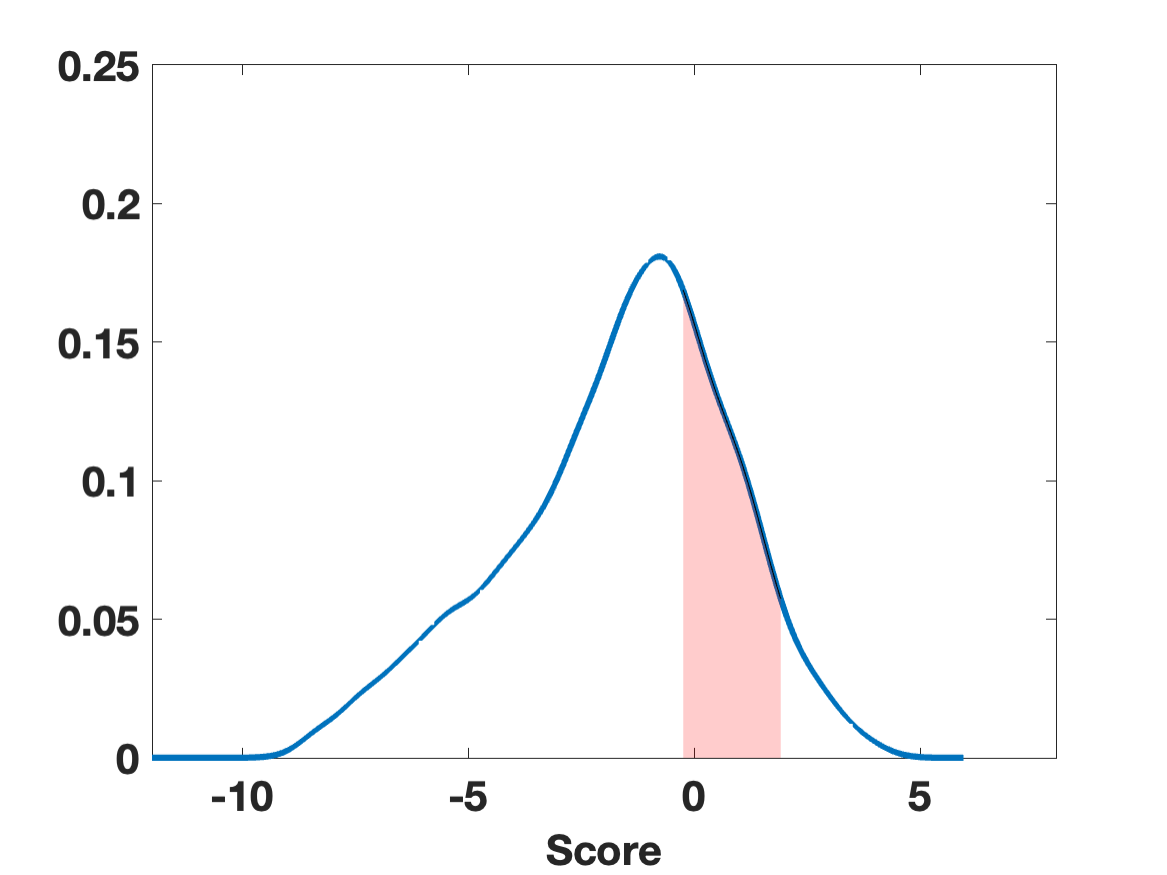}
		& \hspace*{-0.06in}\includegraphics[width=0.22\textwidth]{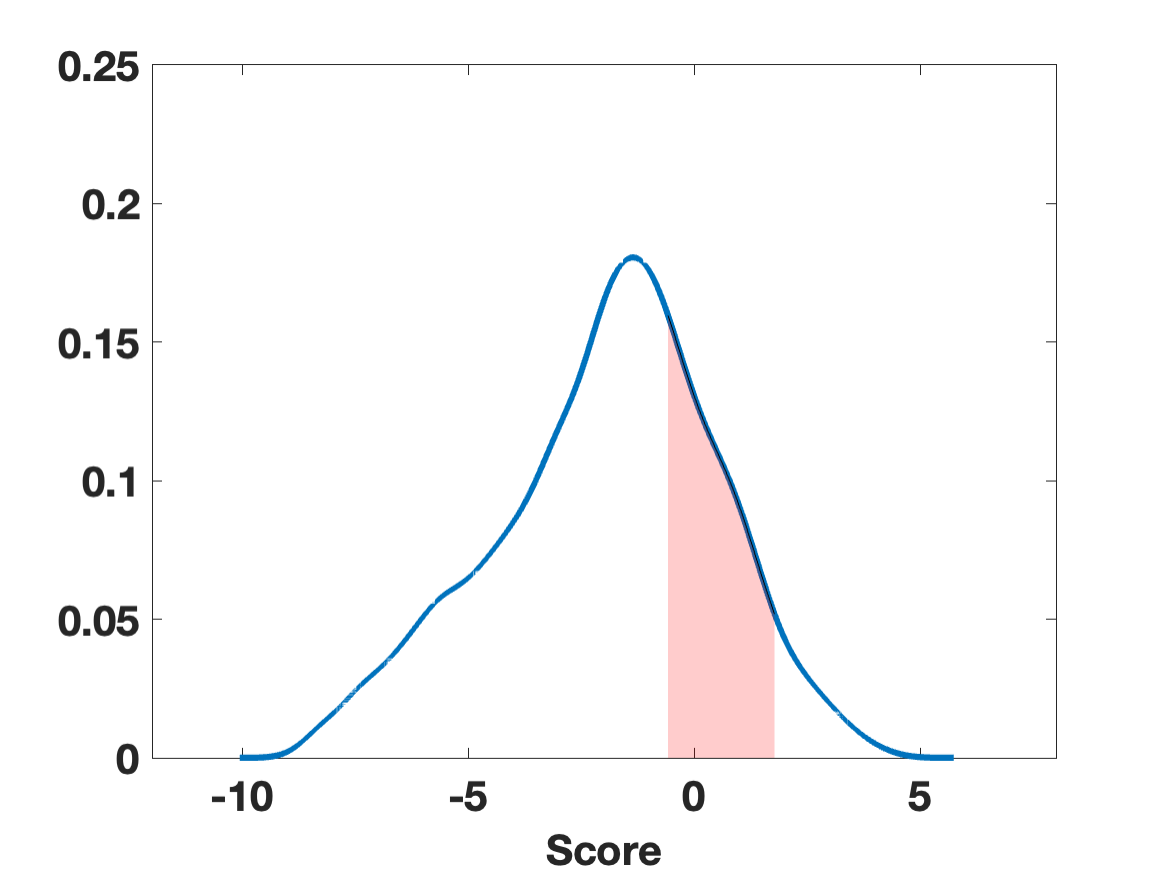}
		& \hspace*{-0.06in}\includegraphics[width=0.22\textwidth]{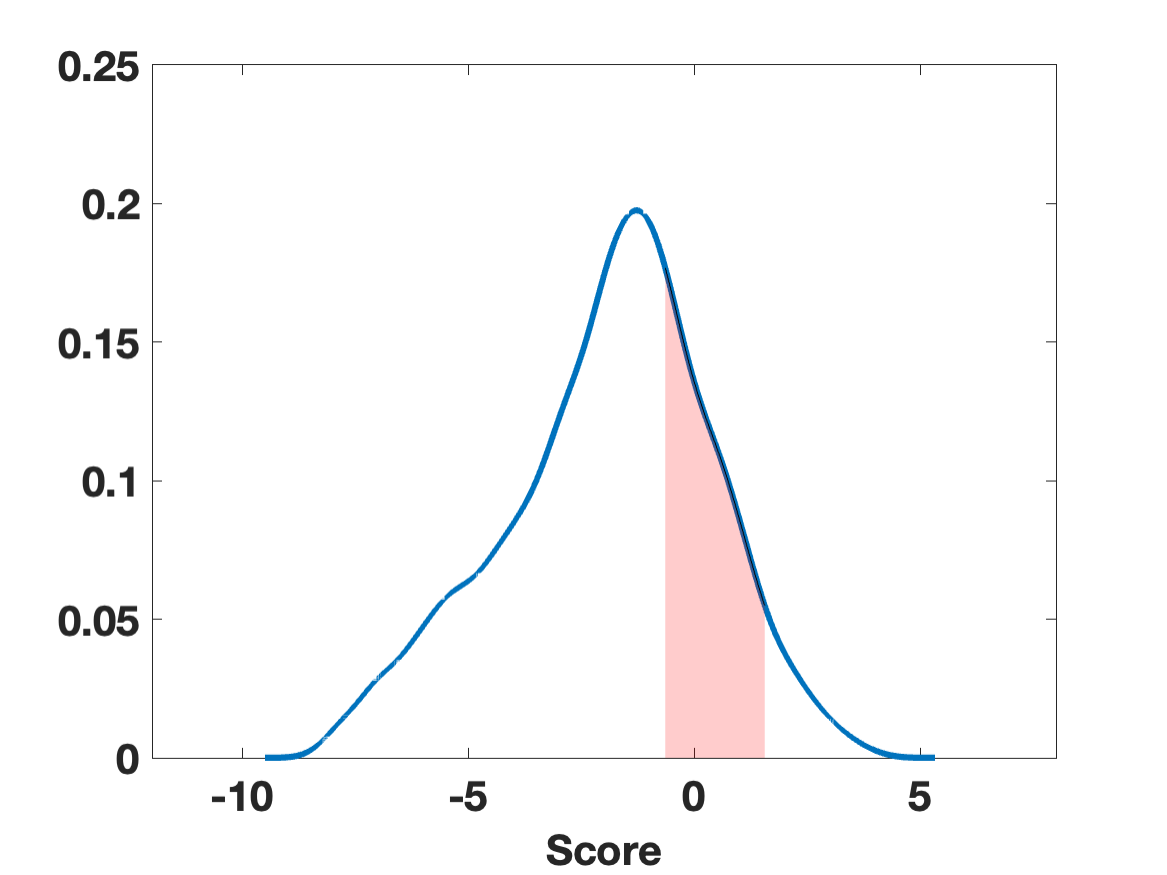}
		& \hspace*{-0.06in}\includegraphics[width=0.22\textwidth]{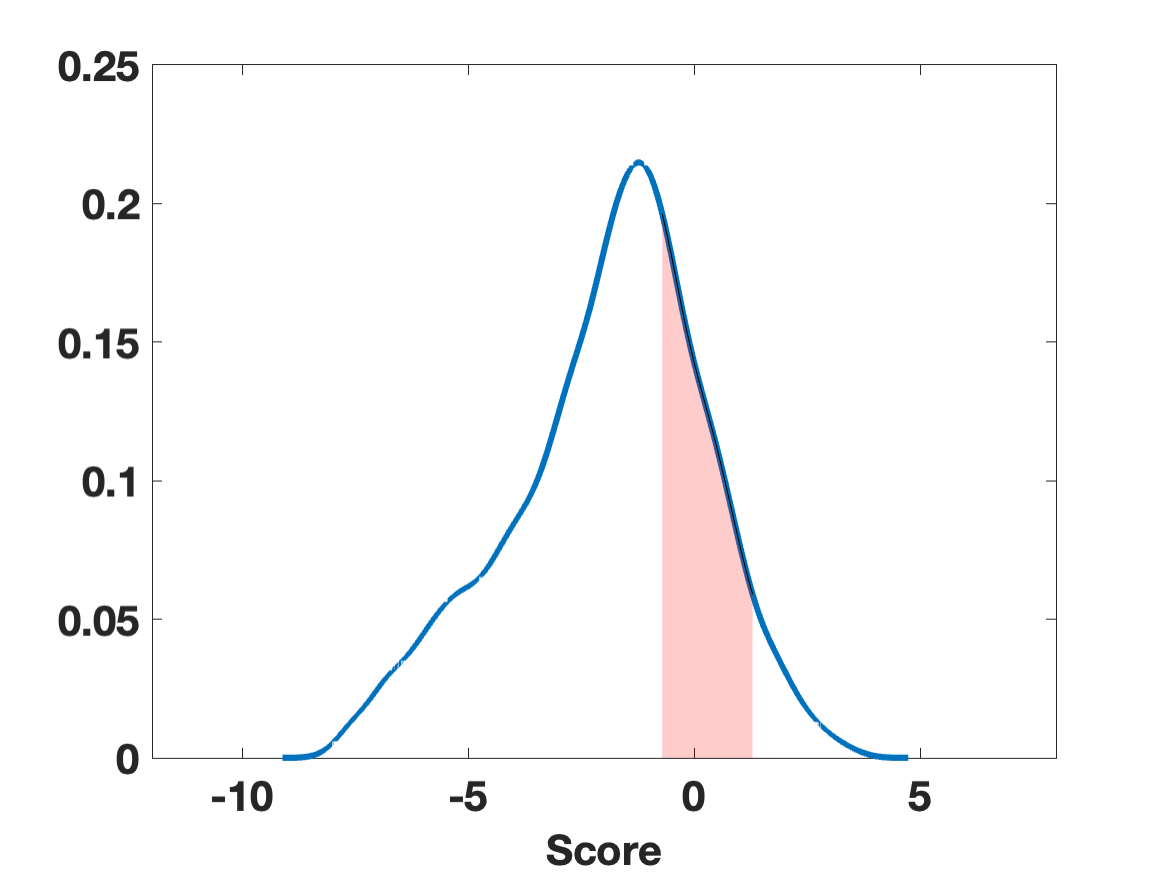}
        
          \\
		\raisebox{10ex}{\small{\rotatebox[origin=c]{90}{Female}}}
		& \hspace*{-0.06in}\includegraphics[width=0.22\textwidth]{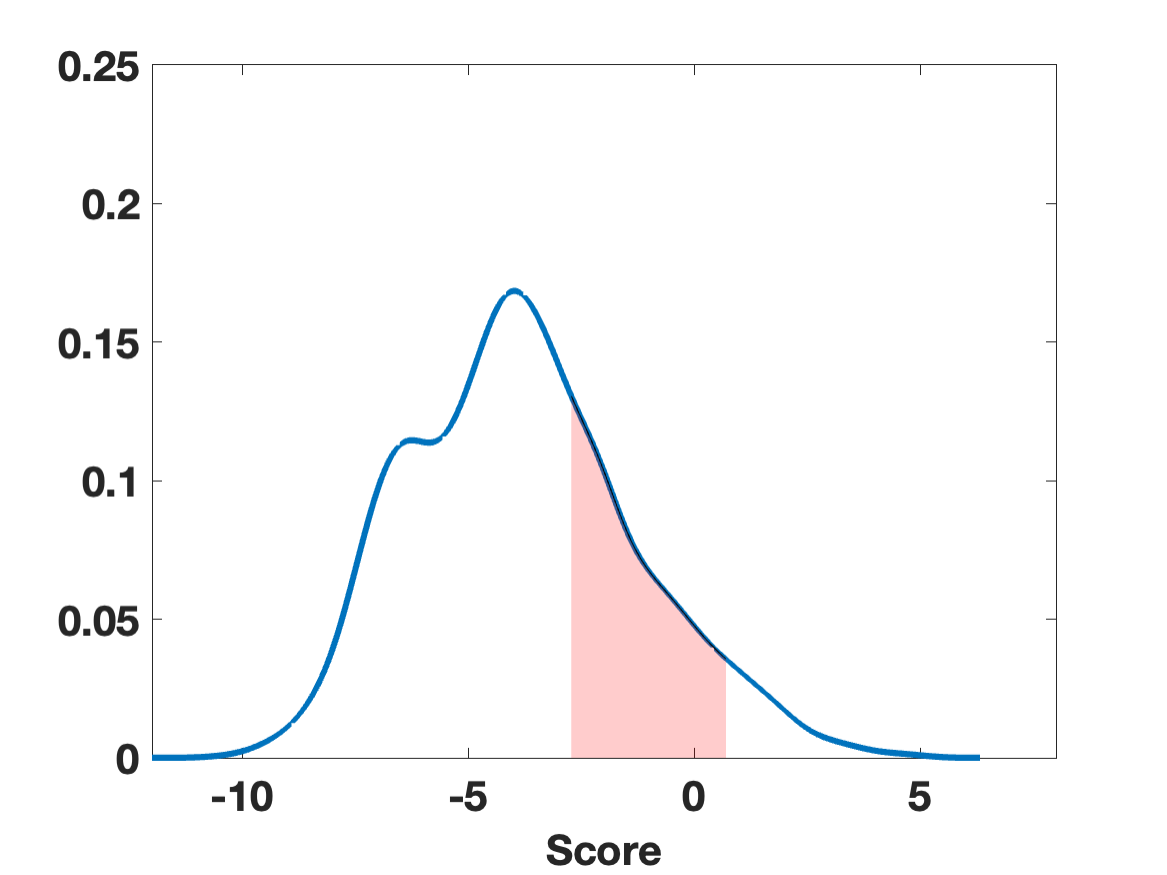}
		& \hspace*{-0.06in}\includegraphics[width=0.22\textwidth]{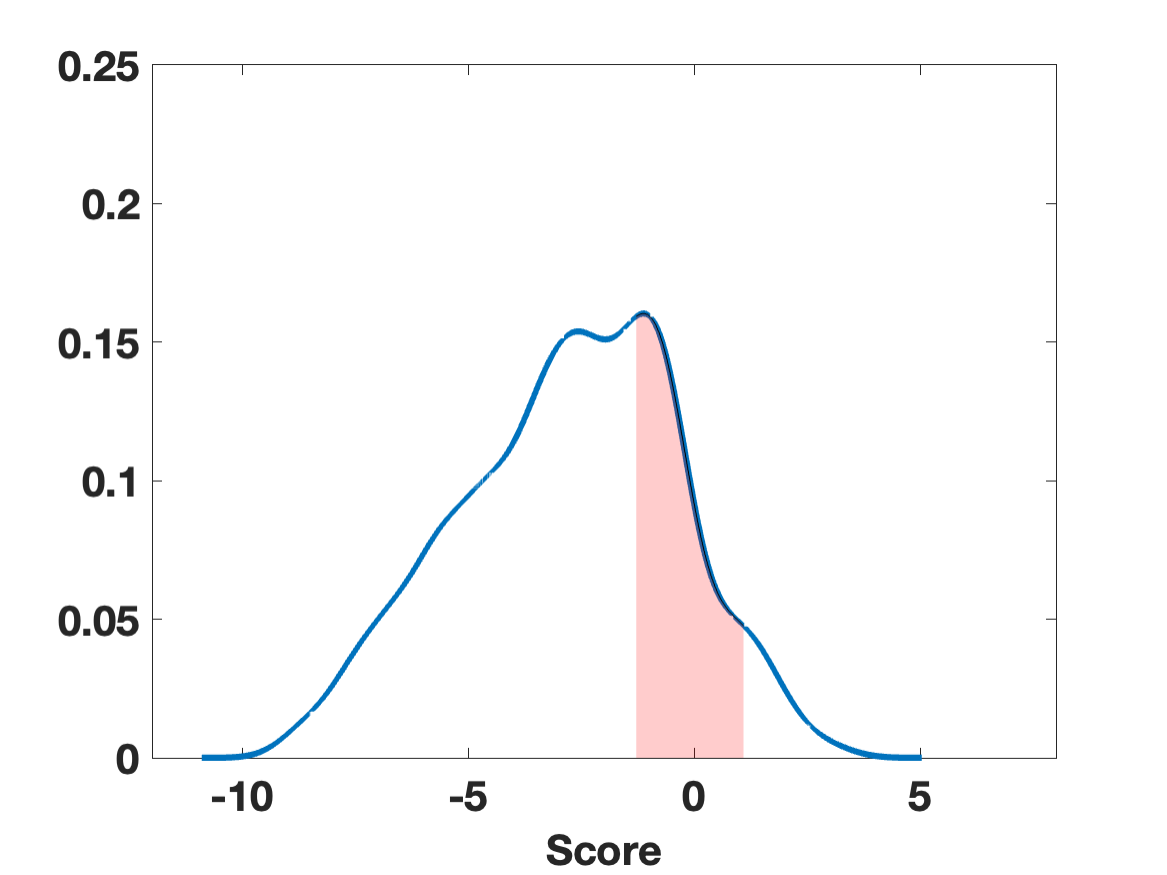}
		& \hspace*{-0.06in}\includegraphics[width=0.22\textwidth]{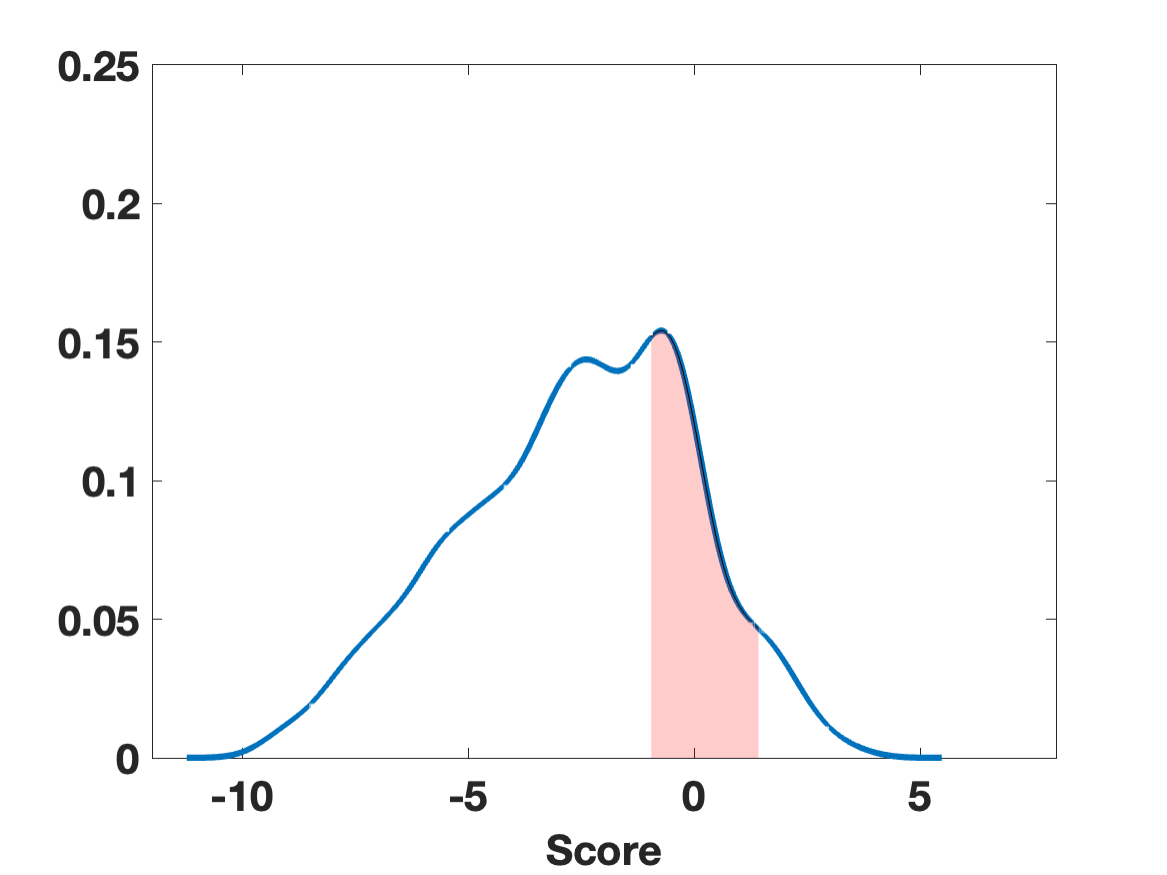}
		& \hspace*{-0.06in}\includegraphics[width=0.22\textwidth]{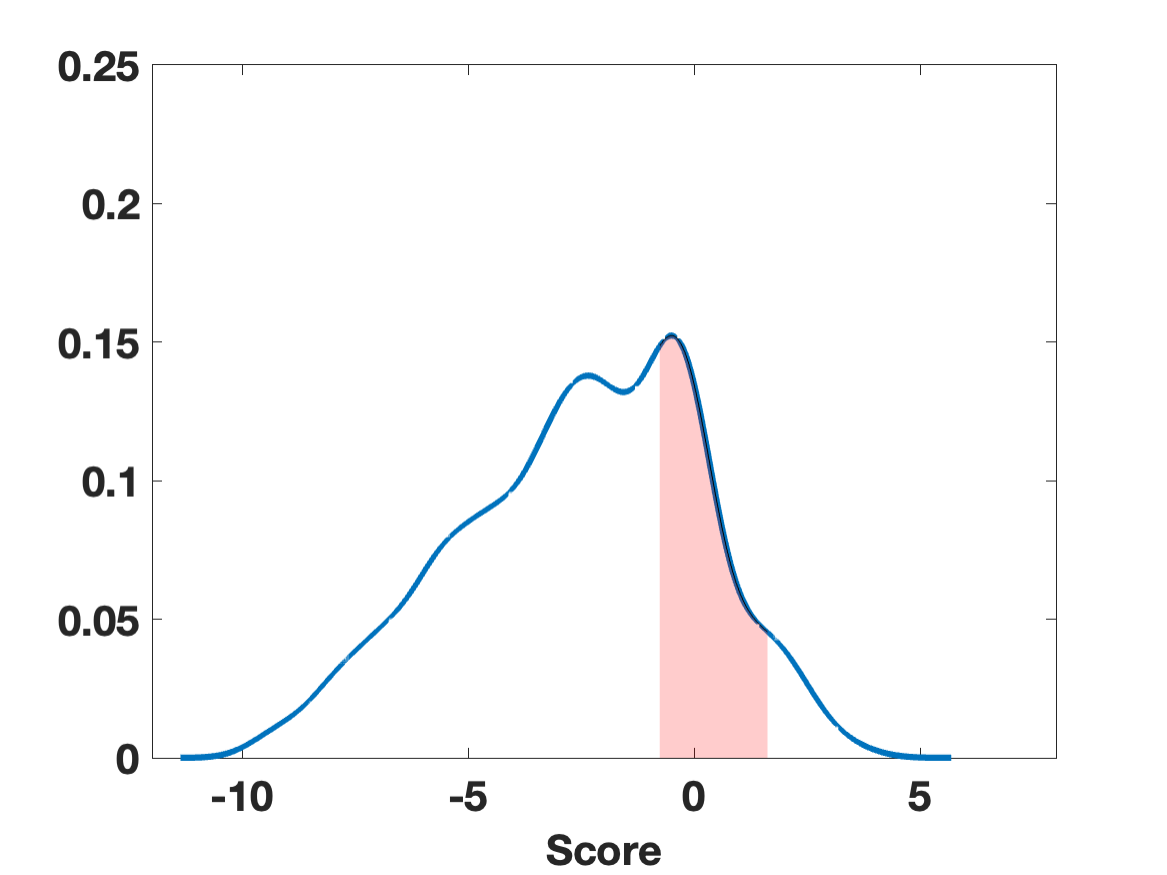}
        \end{tabular}
	\caption{Distributions of predicted scores of different sensitive groups on \textit{a9a} dataset by the unconstrained model and the models solved from \eqref{eq:inprocess_pdp_approx} (pSP constraints) with different $\kappa$'s. The interval $\mathcal{I}$ is $[5\%, 30\%]$ and is highlighted in red.}  
    \label{fig:pDP_accuracy_a9a_distribution}
	\vspace{-0.1in}
\end{figure*}

\begin{figure*}[!ht]
     \begin{tabular}{@{}c|cccc@{}}
      & unconstrained & $\kappa = 0.3$ & $\kappa = 0.08$ & $\kappa = 0.01$ \\
		\hline \vspace*{-0.1in}\\
		\raisebox{7ex}{\small{\rotatebox[origin=c]{90}{Other ages}}}
		& \hspace*{-0.06in}\includegraphics[width=0.22\textwidth]{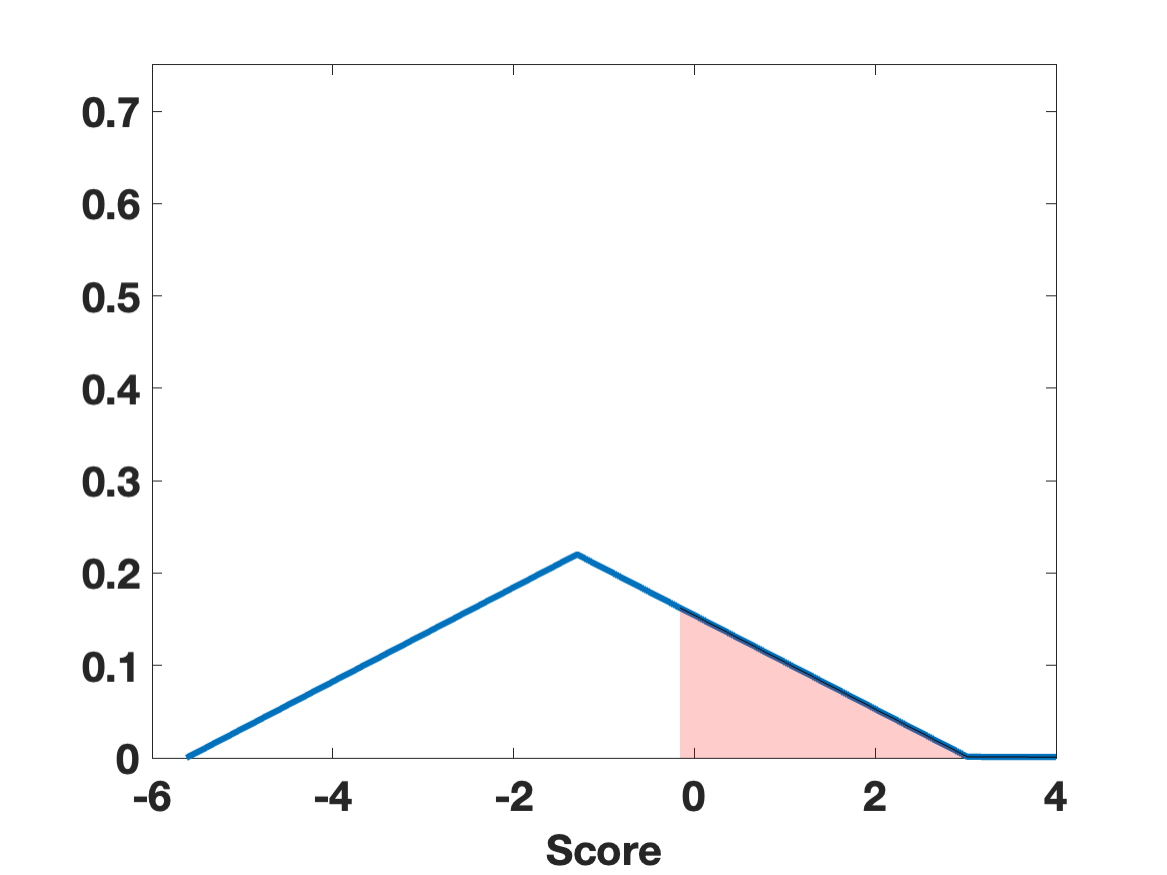}
		& \hspace*{-0.06in}\includegraphics[width=0.22\textwidth]{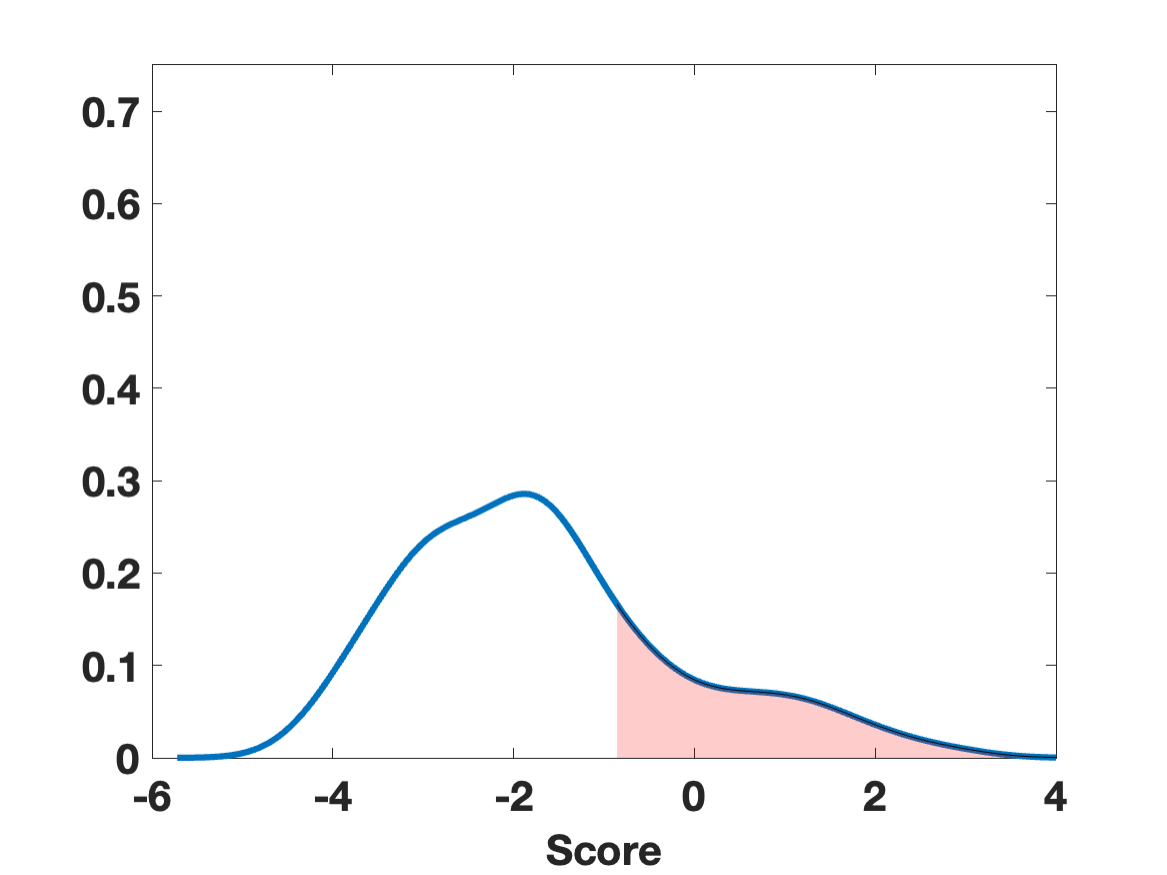}
		& \hspace*{-0.06in}\includegraphics[width=0.22\textwidth]{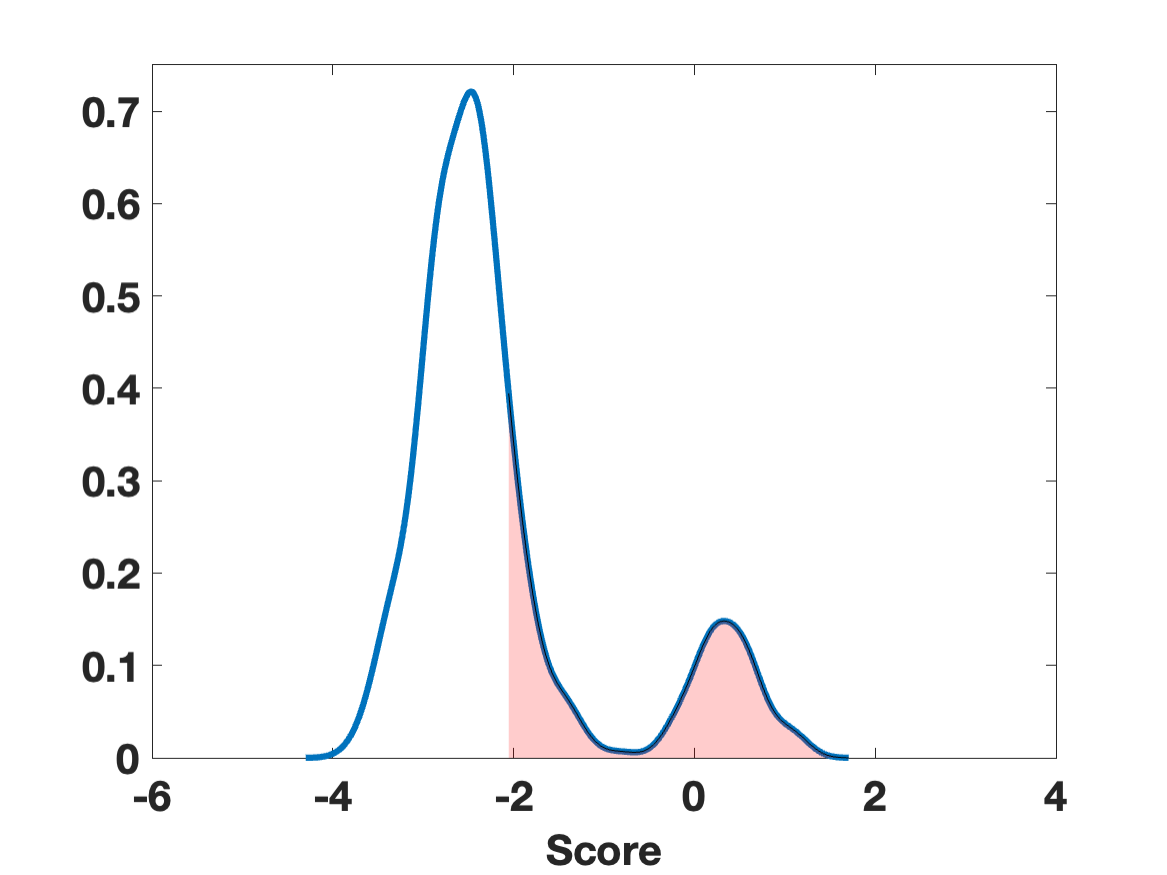}
		& \hspace*{-0.06in}\includegraphics[width=0.22\textwidth]{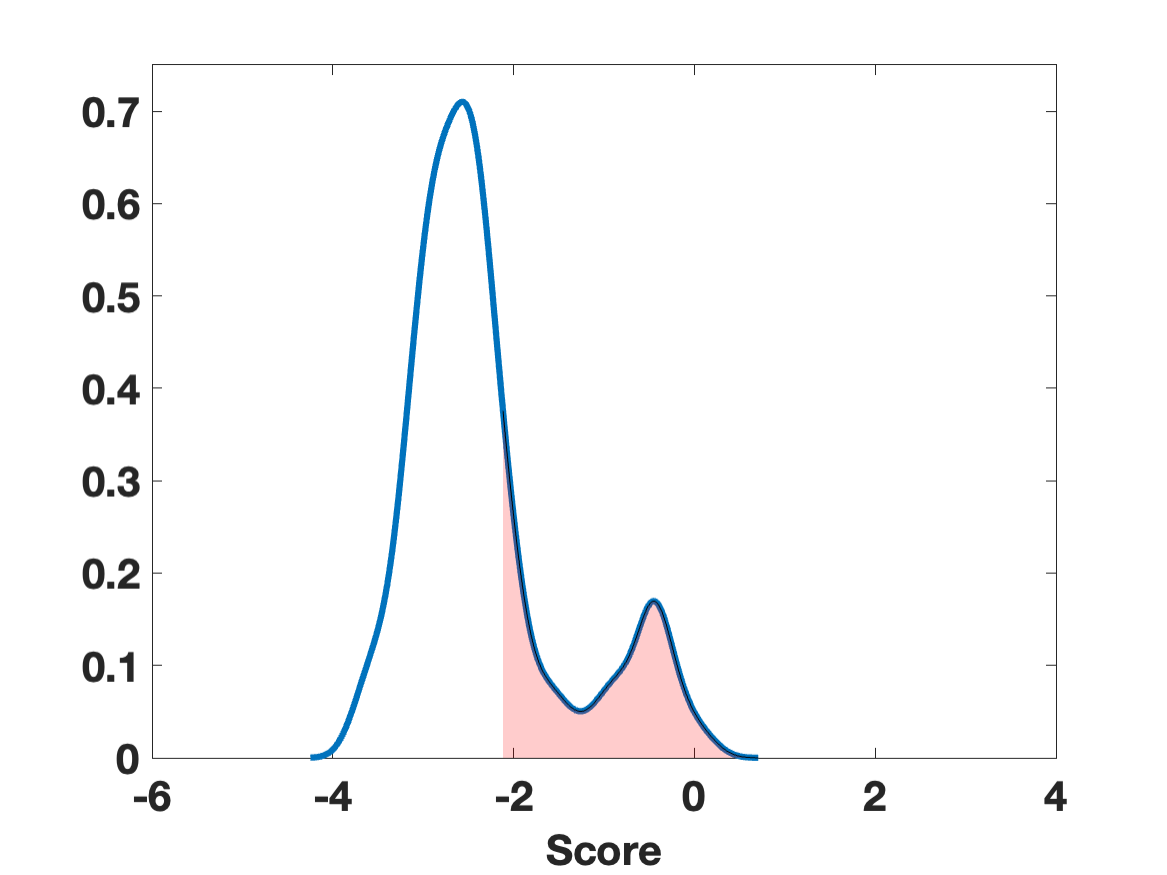}
        
          \\
		\raisebox{7ex}{\small{\rotatebox[origin=c]{90}{Ages between 25 and 60}}}
		& \hspace*{-0.06in}\includegraphics[width=0.22\textwidth]{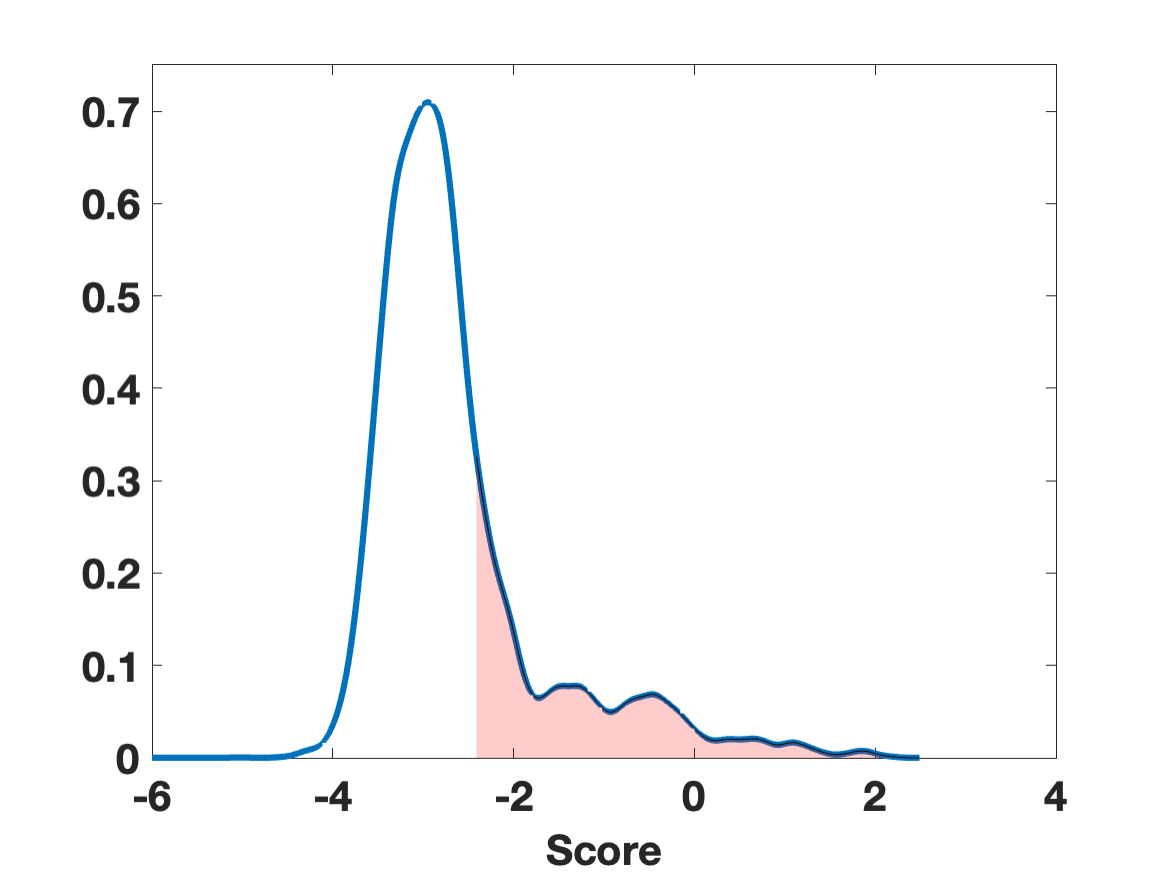}
		& \hspace*{-0.06in}\includegraphics[width=0.22\textwidth]{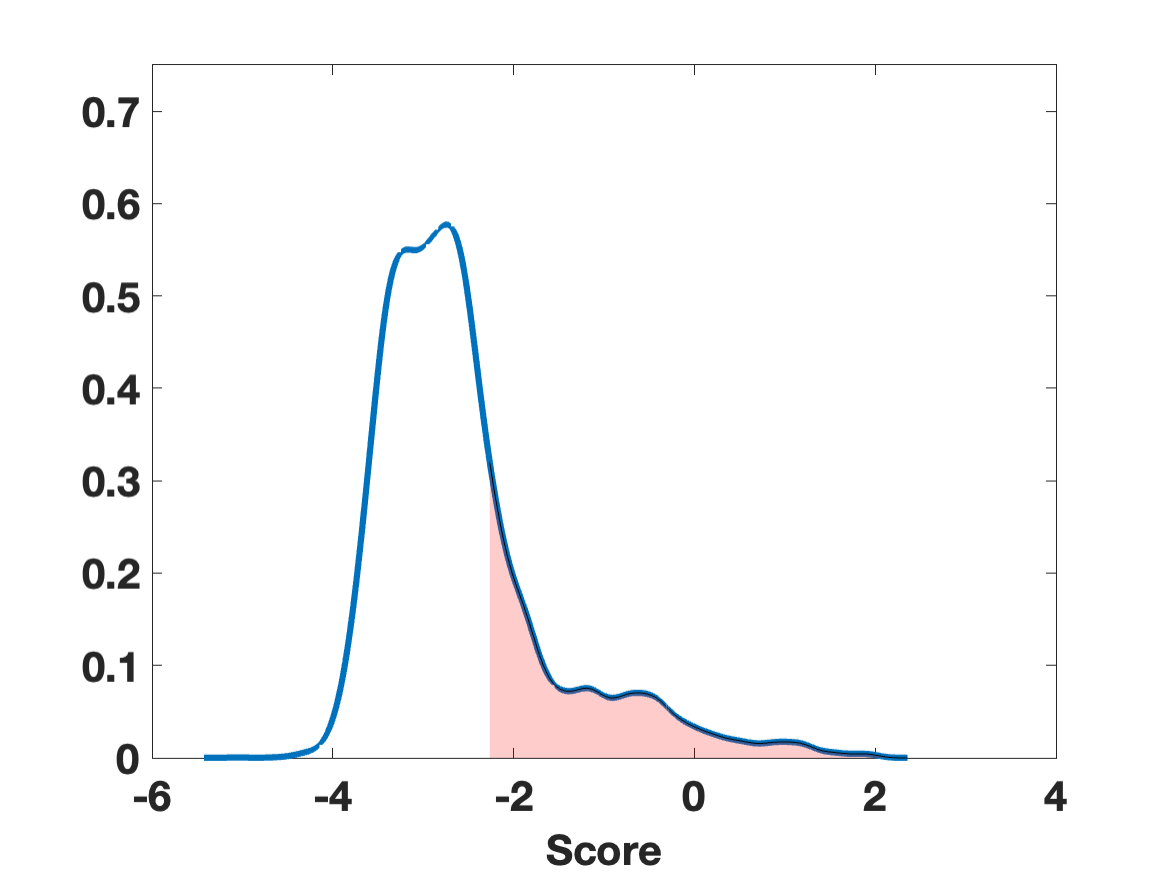}
		& \hspace*{-0.06in}\includegraphics[width=0.22\textwidth]{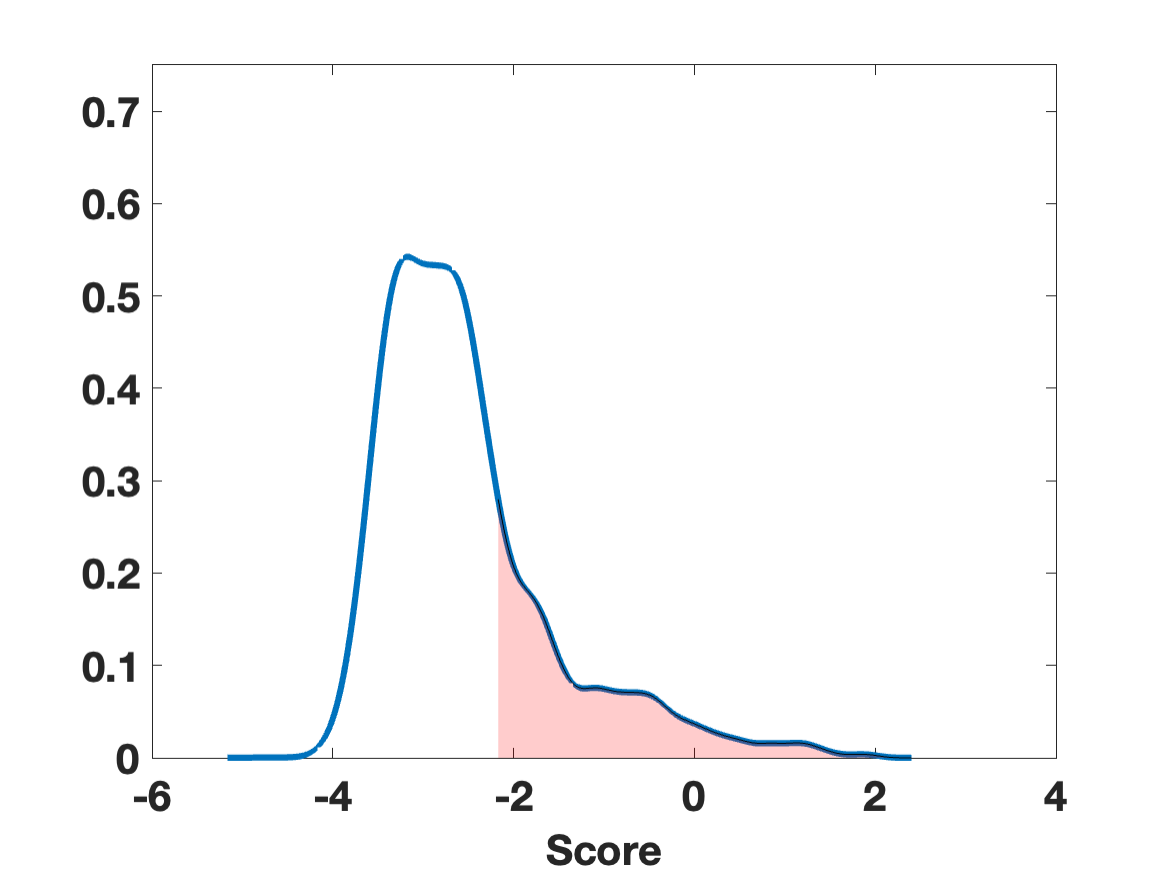}
		& \hspace*{-0.06in}\includegraphics[width=0.22\textwidth]{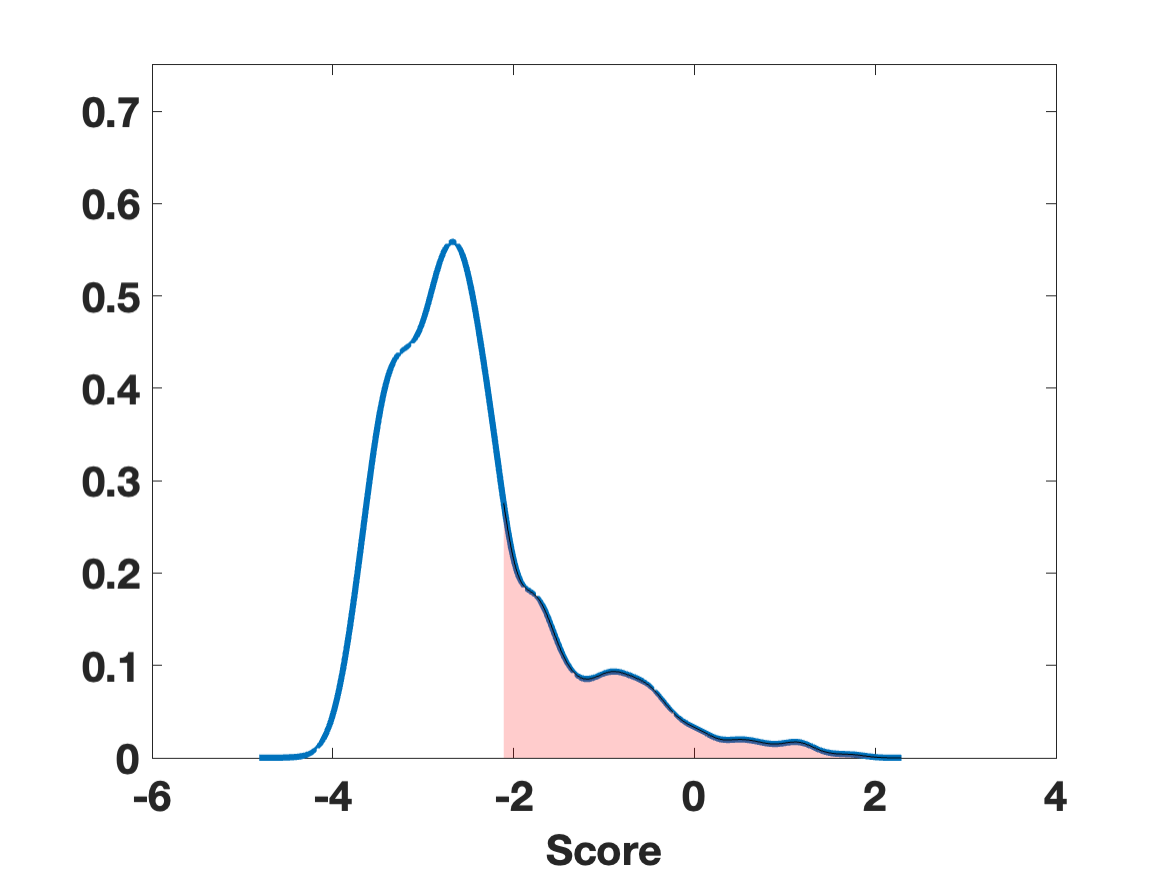}
        \end{tabular}
	\caption{Distributions of predicted scores of different sensitive groups on \textit{bank} dataset by the unconstrained model and the models solved from \eqref{eq:inprocess_pdp_approx} (pSP constraints) with different $\kappa$'s. The interval $\mathcal{I}$ is $[0\%, 25\%]$ and is highlighted in red.}  
    \label{fig:pDP_accuracy_bank_distribution}
	\vspace{-0.1in}
\end{figure*}

\begin{figure*}[!ht]
     \begin{tabular}{@{}c|cccc@{}}
      & unconstrained & $\kappa = 0.15$ & $\kappa = 0.1$ & $\kappa = 0.005$ \\
		\hline \vspace*{-0.1in}\\
		\raisebox{7ex}{\small{\rotatebox[origin=c]{90}{Non-white}}}
		& \hspace*{-0.06in}\includegraphics[width=0.22\textwidth]{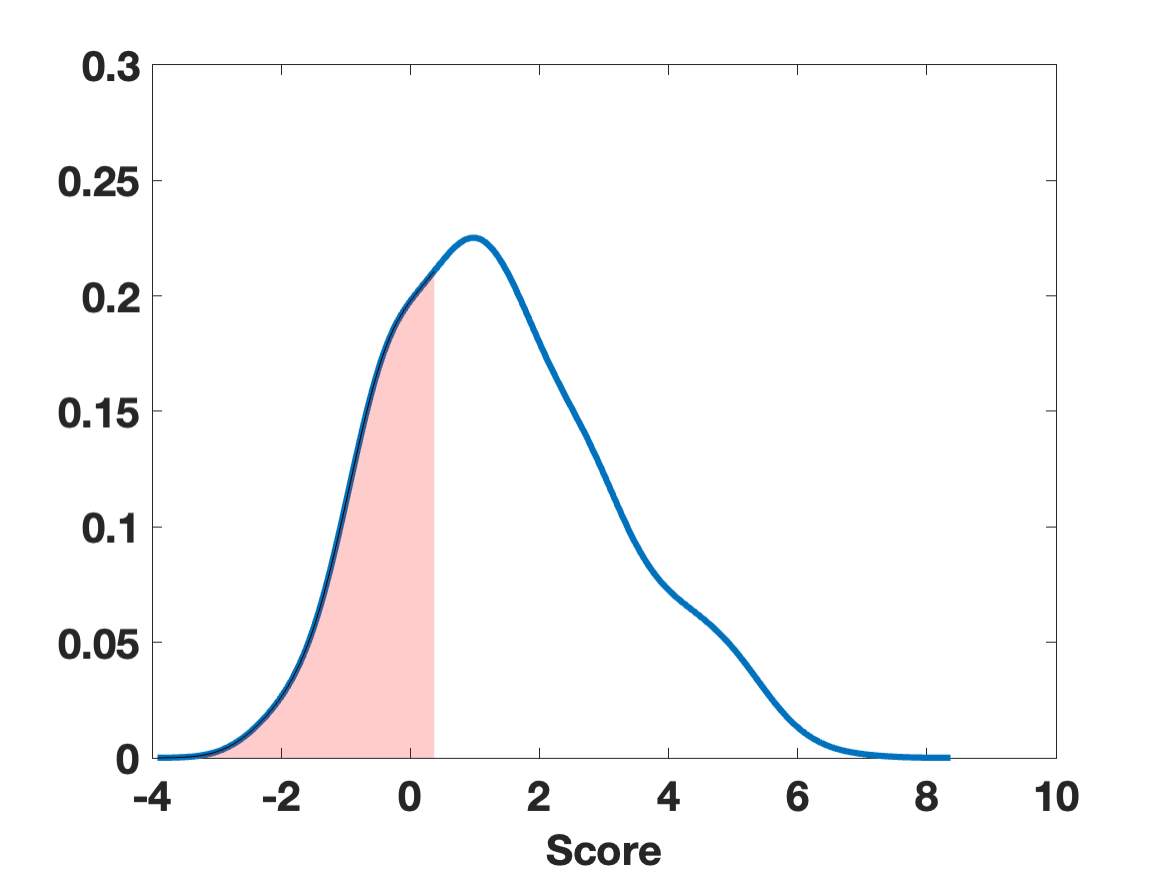}
		& \hspace*{-0.06in}\includegraphics[width=0.22\textwidth]{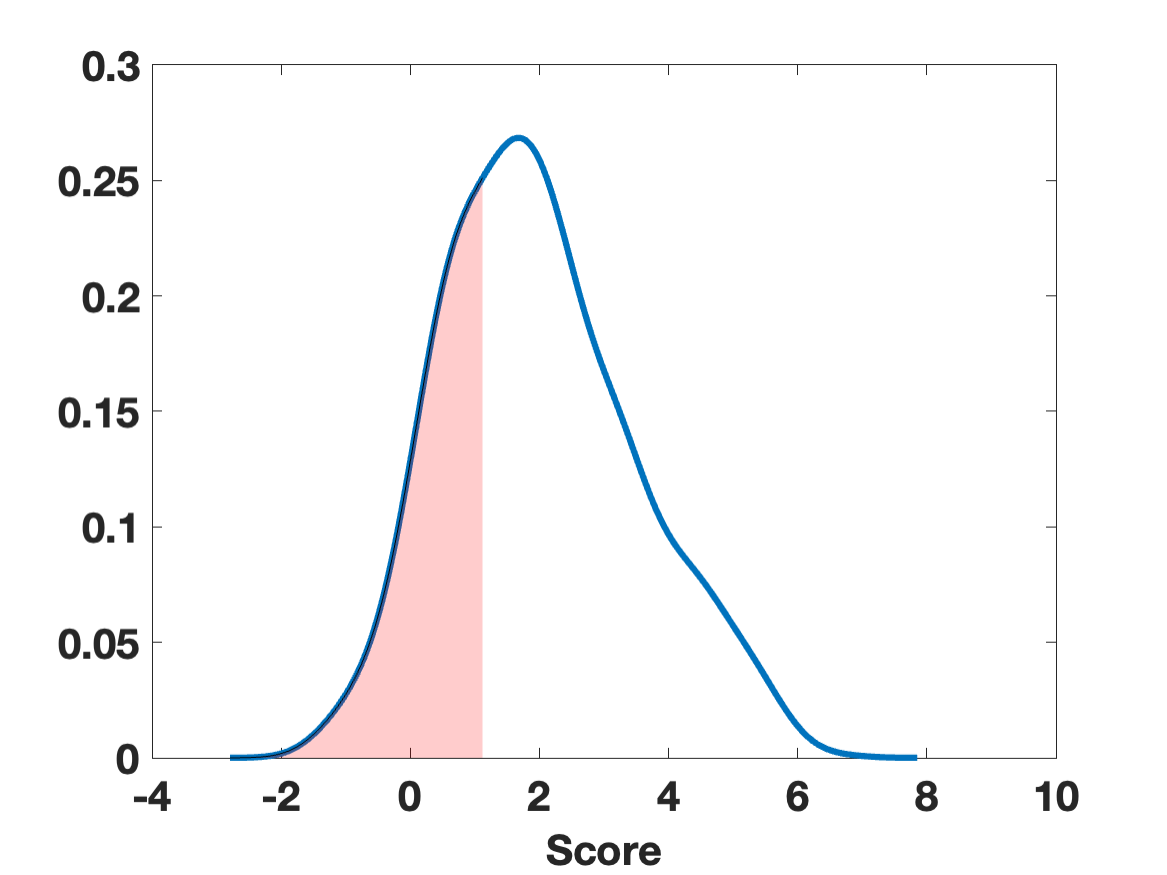}
		& \hspace*{-0.06in}\includegraphics[width=0.22\textwidth]{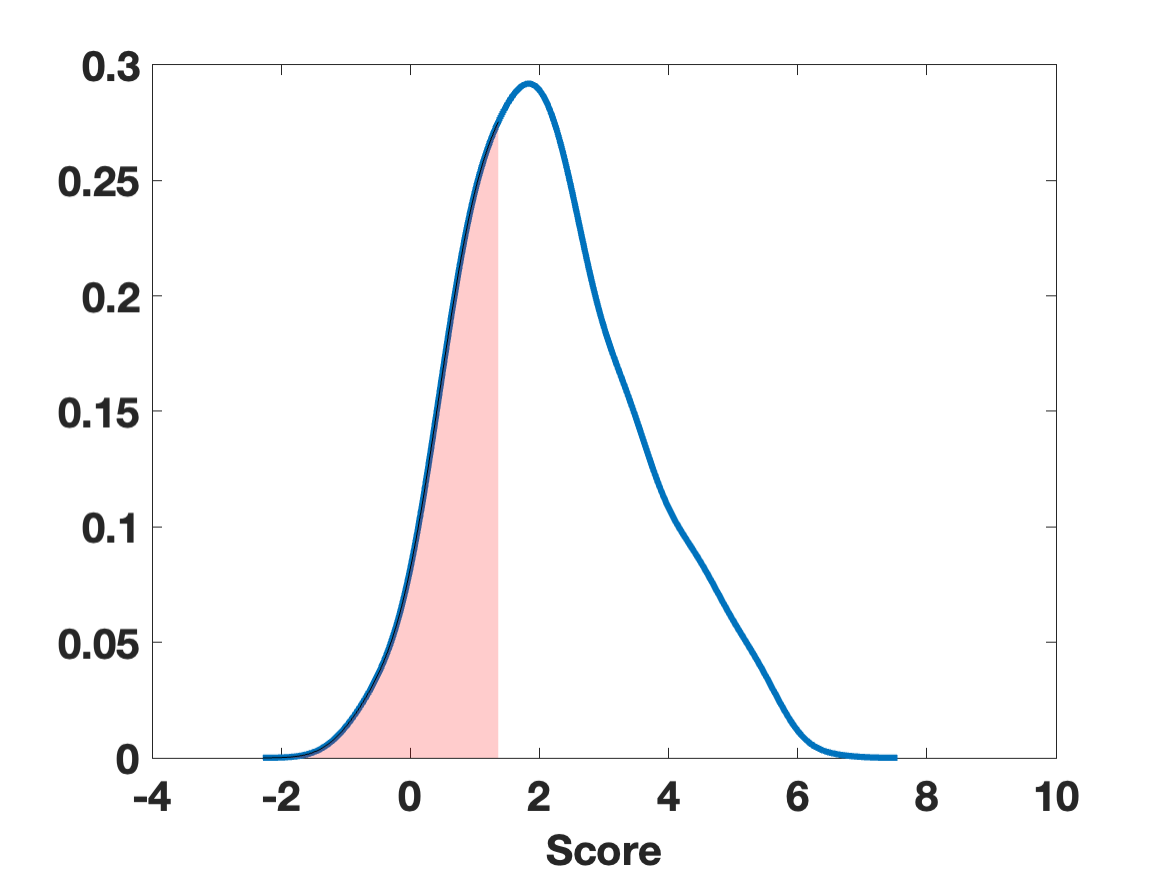}
		& \hspace*{-0.06in}\includegraphics[width=0.22\textwidth]{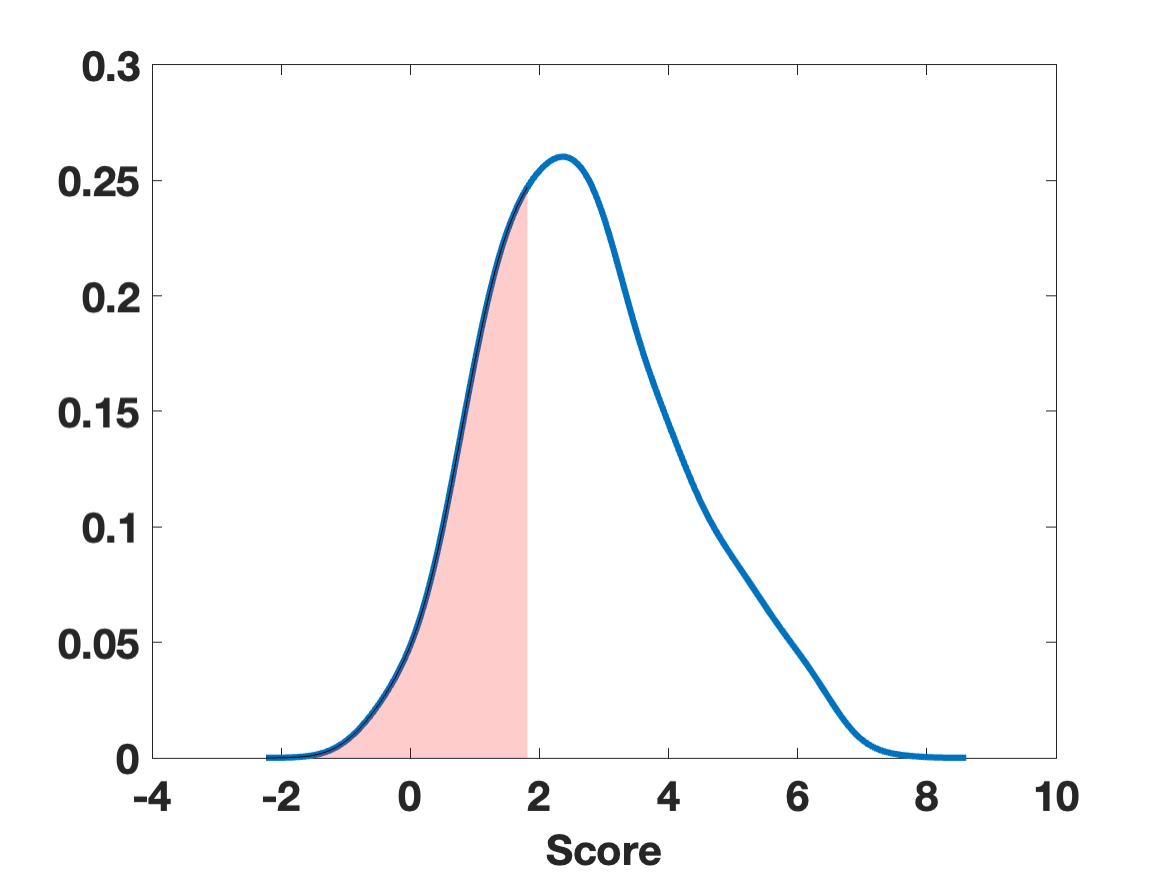}
        
          \\
		\raisebox{7ex}{\small{\rotatebox[origin=c]{90}{White}}}
		& \hspace*{-0.06in}\includegraphics[width=0.22\textwidth]{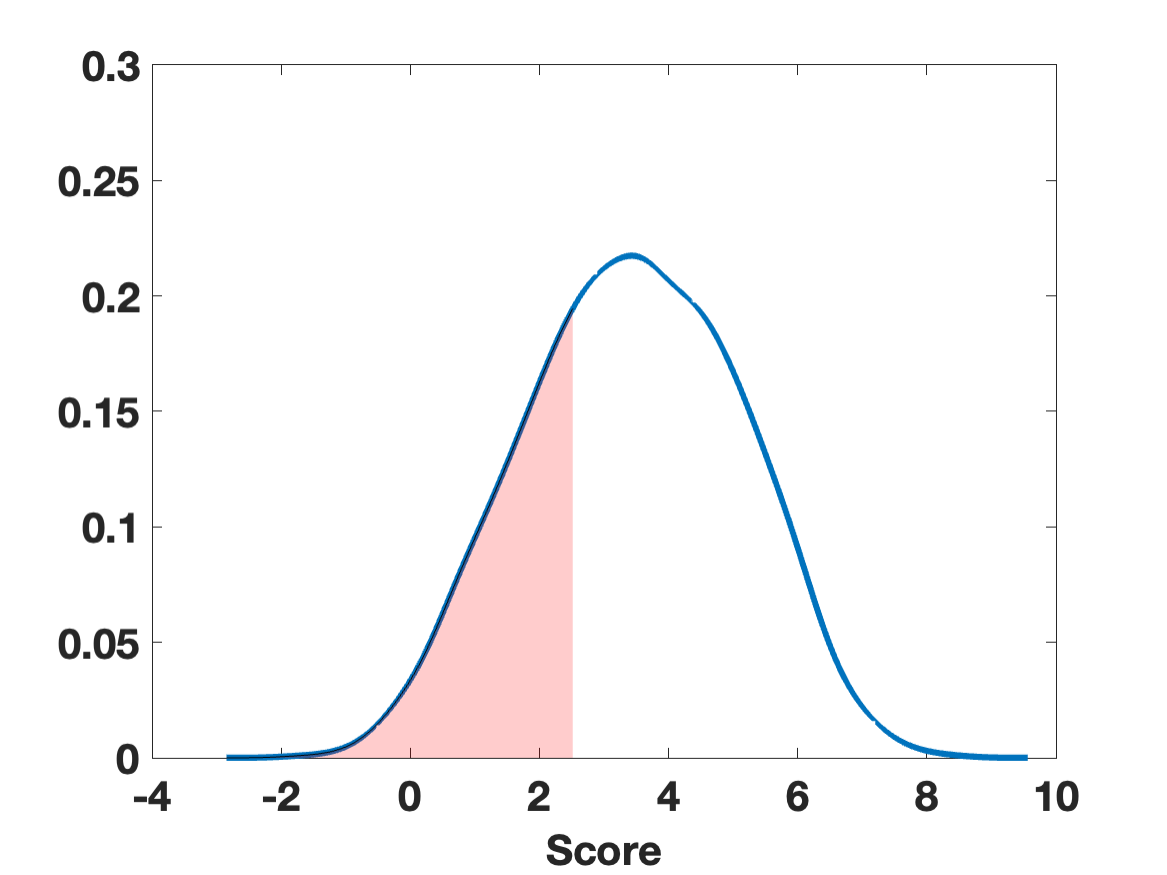}
		& \hspace*{-0.06in}\includegraphics[width=0.22\textwidth]{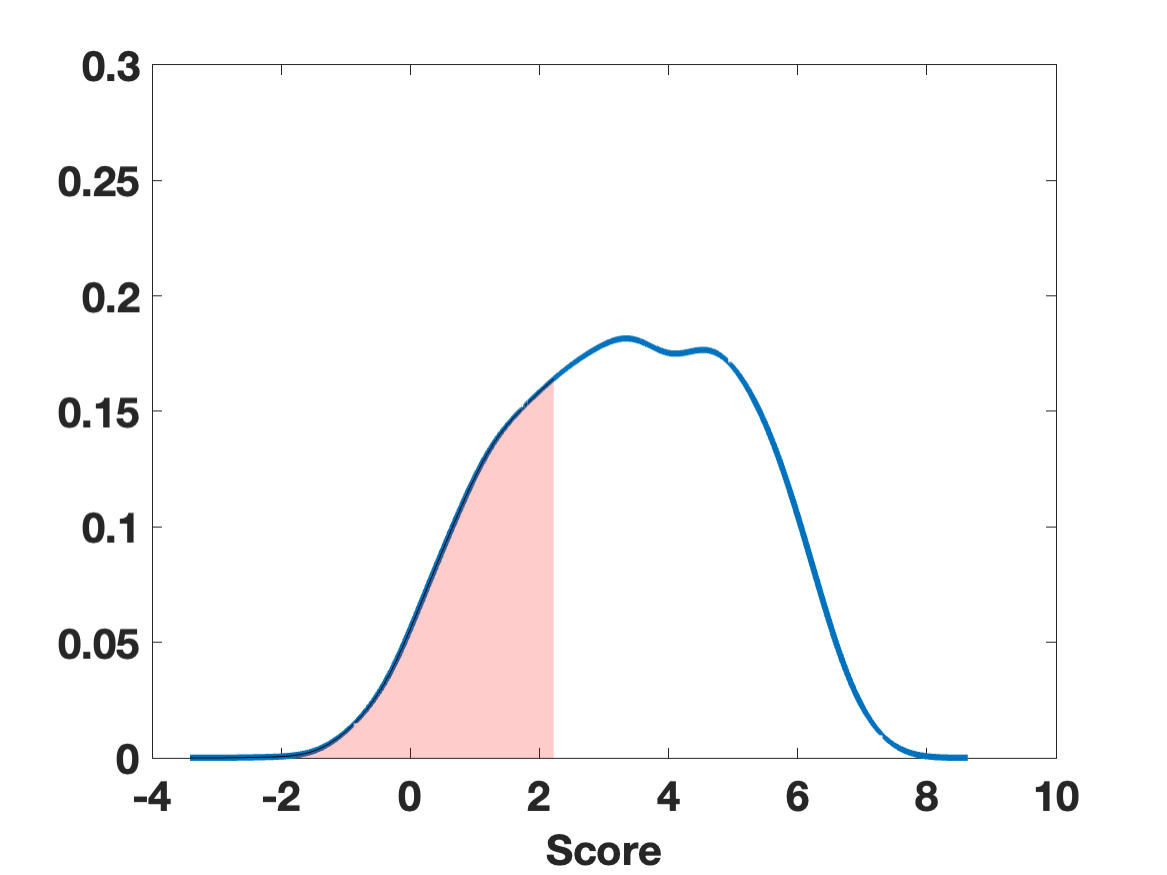}
		& \hspace*{-0.06in}\includegraphics[width=0.22\textwidth]{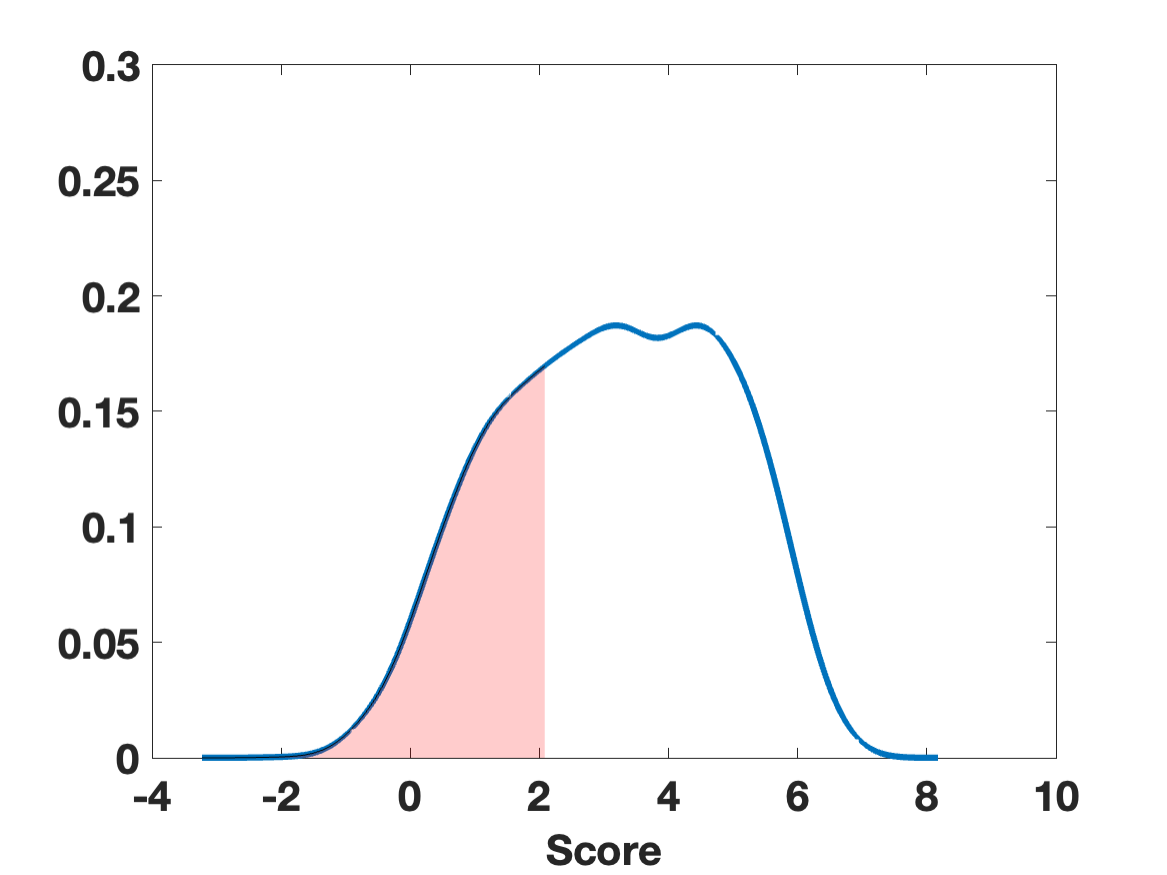}
		& \hspace*{-0.06in}\includegraphics[width=0.22\textwidth]{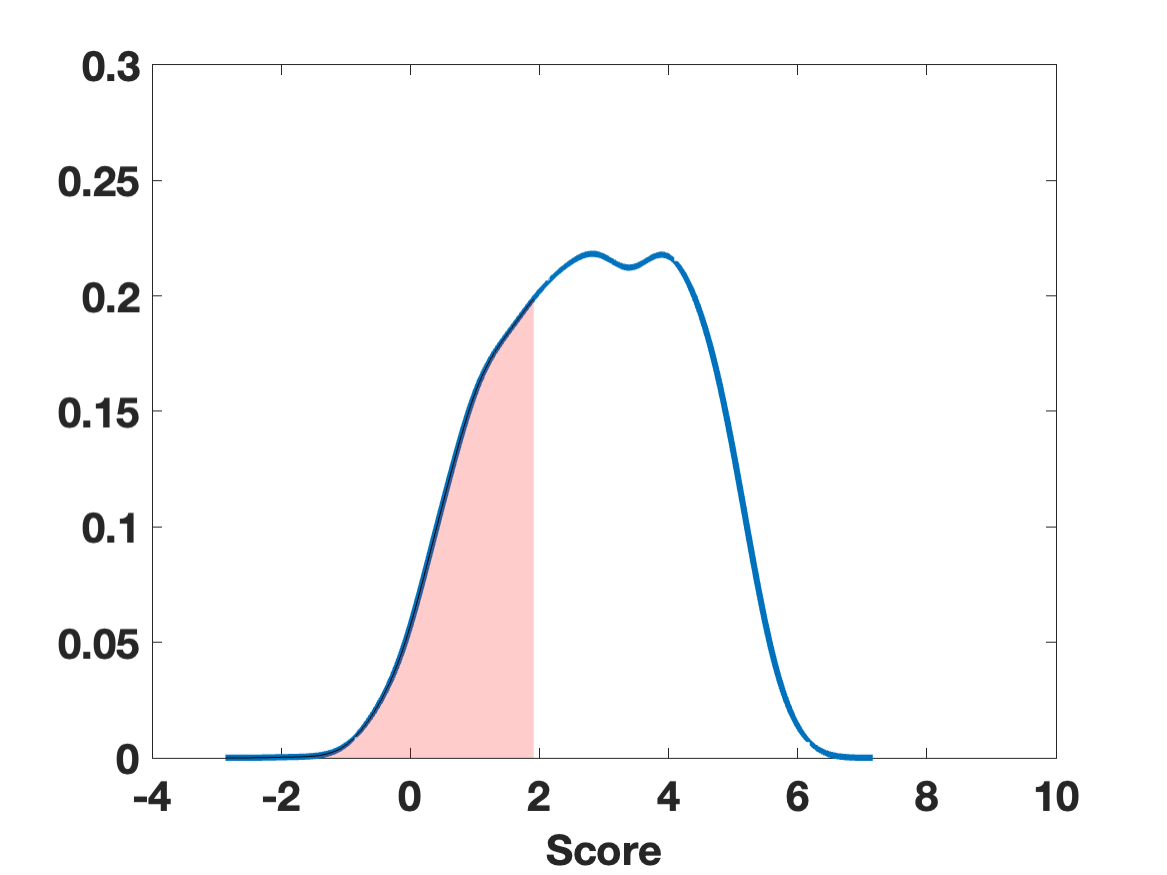}
        \end{tabular}
	\caption{Distributions of predicted scores of different sensitive groups on \textit{law school} dataset by the unconstrained model and the models solved from \eqref{eq:inprocess_pdp_approx} (pSP constraints) with different $\kappa$'s. The interval $\mathcal{I}$ is $[70\%, 100\%]$ and is highlighted in red.}  
    \label{fig:pDP_accuracy_law_distribution}
	\vspace{-0.1in}
\end{figure*}

\begin{figure*}[!ht]
     \begin{tabular}{@{}c|cccc@{}}
      & unconstrained & $\kappa = 0.15$ & $\kappa = 0.05$ & $\kappa = 0.005$ \\
		\hline \vspace*{-0.1in}\\
		\raisebox{7ex}{\small{\rotatebox[origin=c]{90}{Male}}}
		& \hspace*{-0.06in}\includegraphics[width=0.22\textwidth]{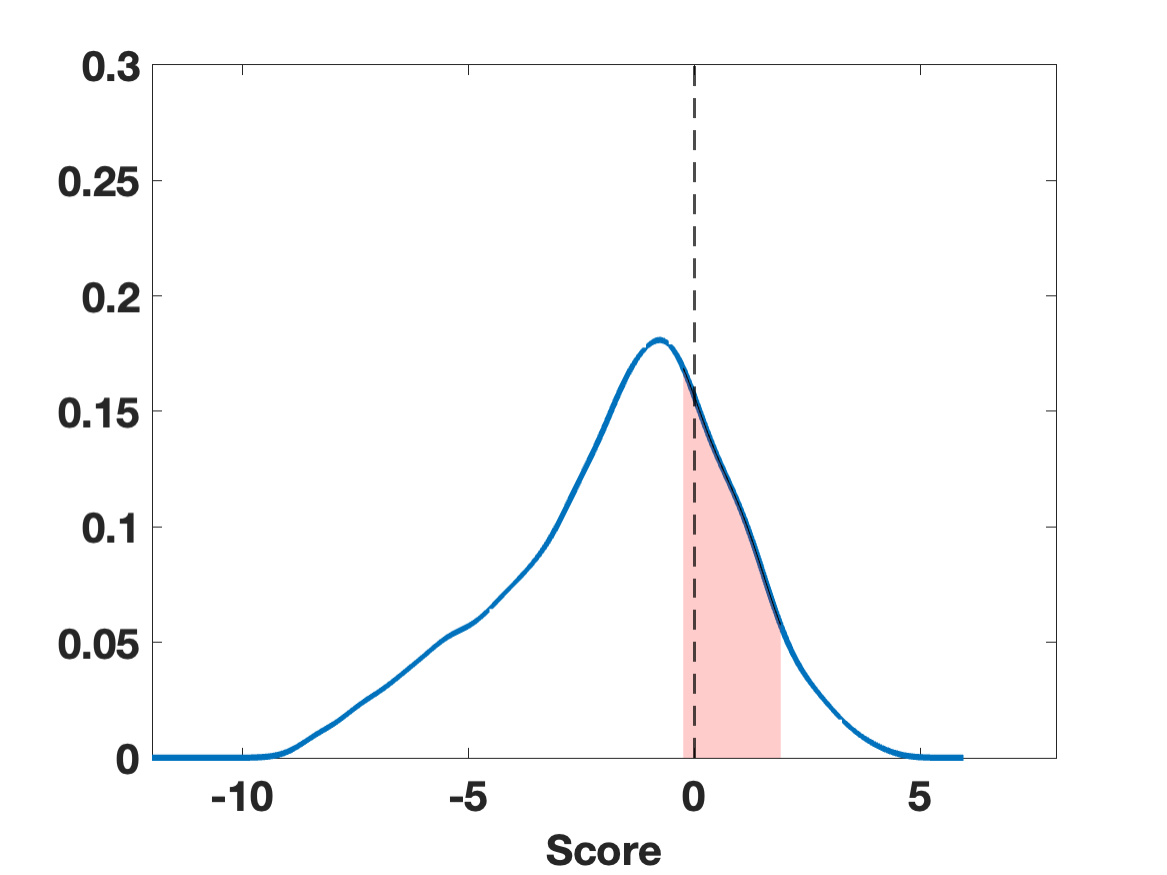}
		& \hspace*{-0.06in}\includegraphics[width=0.22\textwidth]{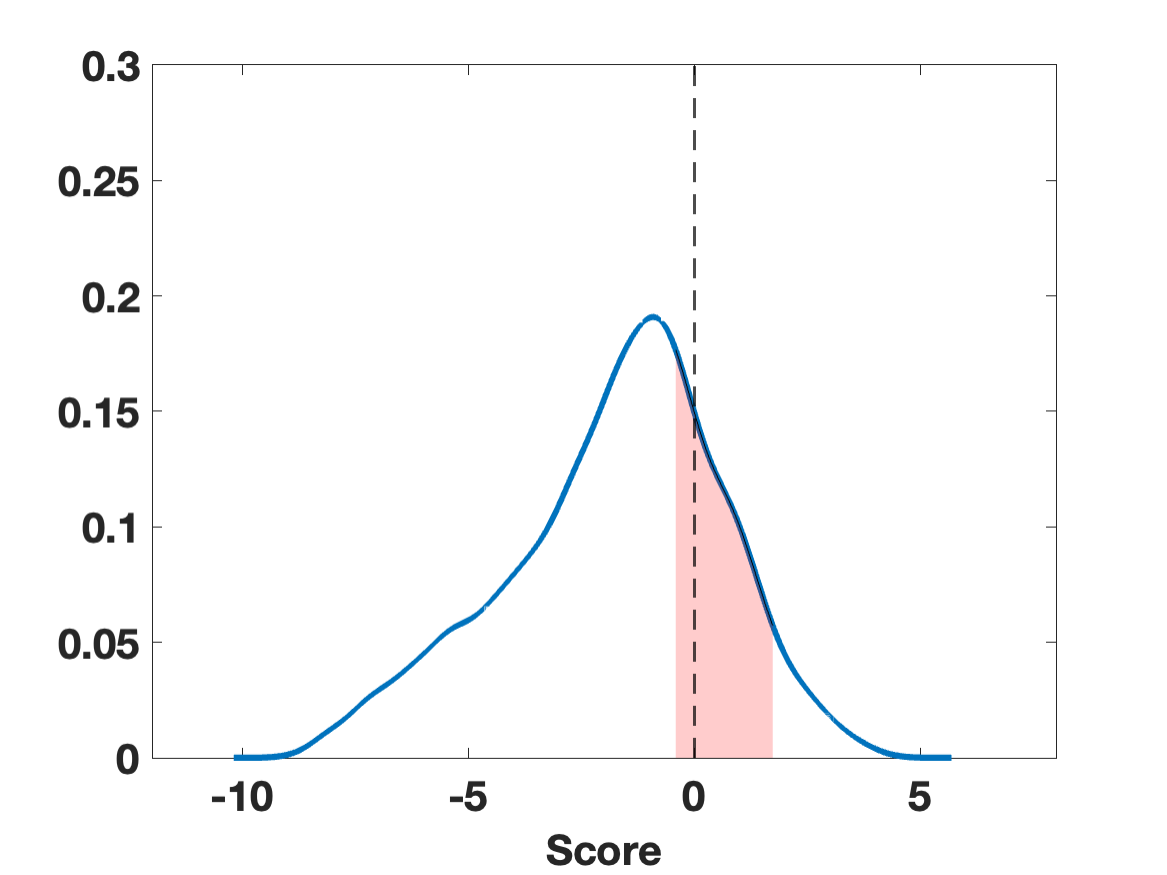}
		& \hspace*{-0.06in}\includegraphics[width=0.22\textwidth]{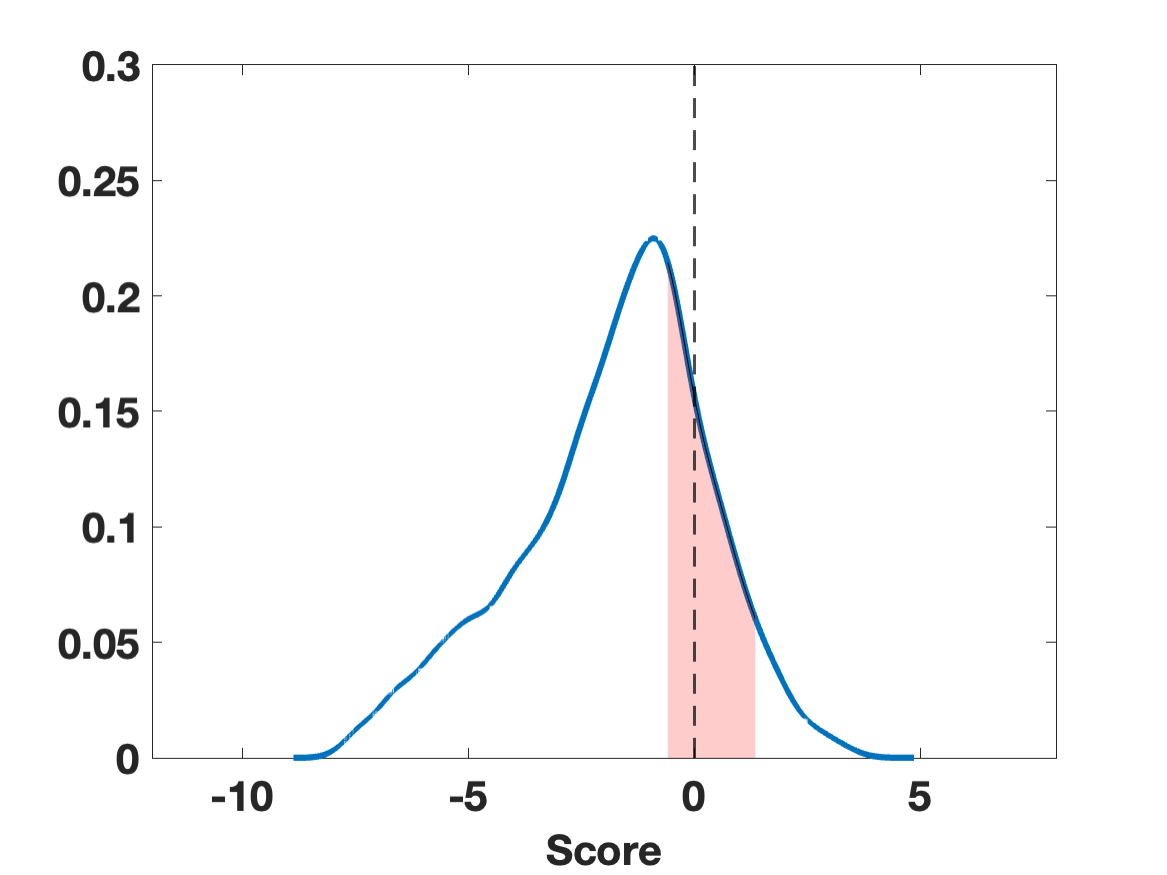}
		& \hspace*{-0.06in}\includegraphics[width=0.22\textwidth]{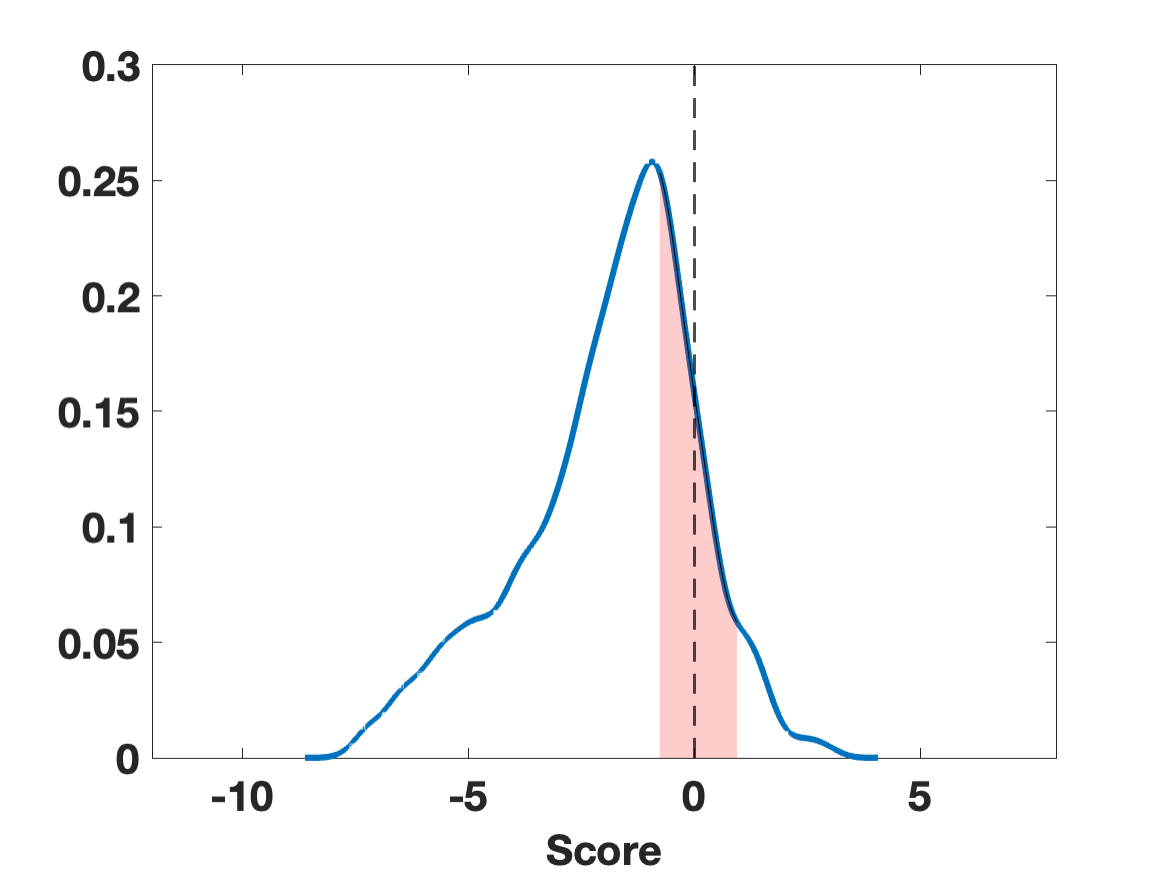}
        
          \\
		\raisebox{7ex}{\small{\rotatebox[origin=c]{90}{Female}}}
		& \hspace*{-0.06in}\includegraphics[width=0.22\textwidth]{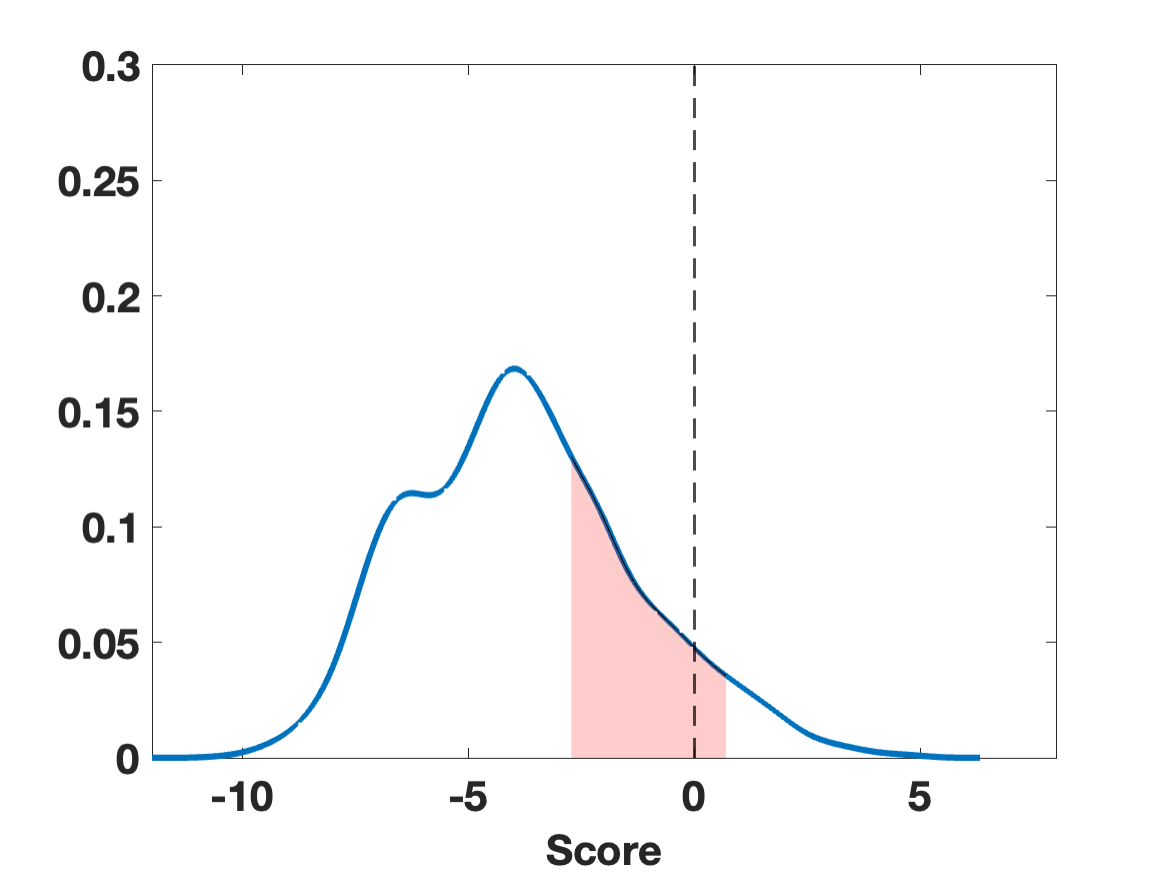}
		& \hspace*{-0.06in}\includegraphics[width=0.22\textwidth]{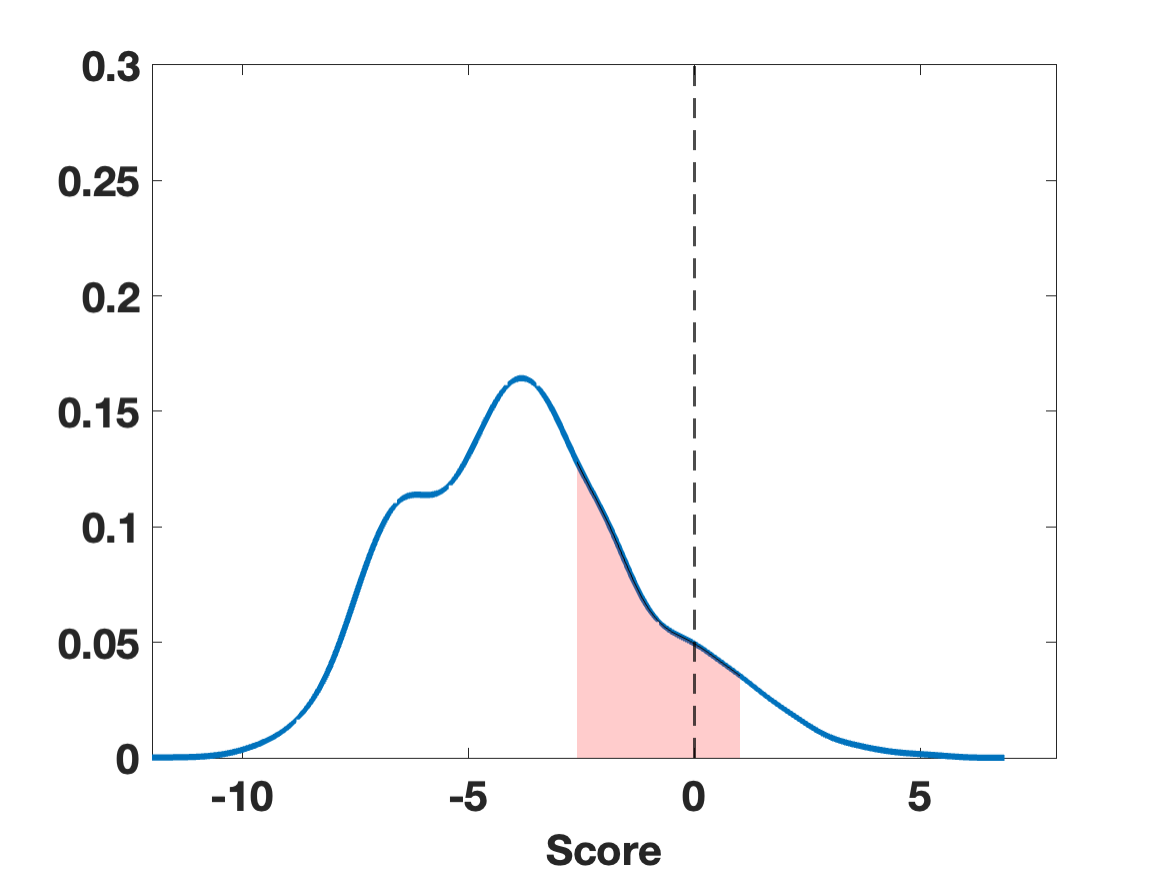}
		& \hspace*{-0.06in}\includegraphics[width=0.22\textwidth]{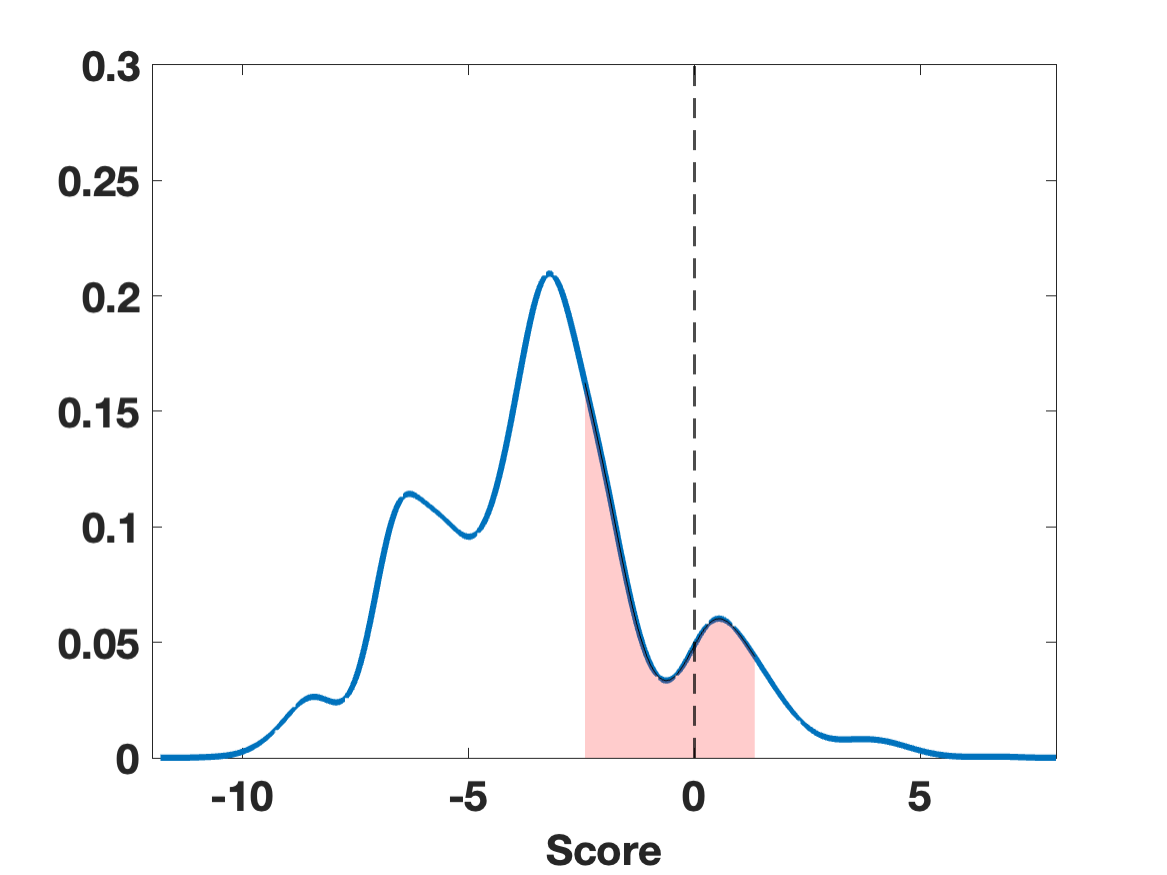}
		& \hspace*{-0.06in}\includegraphics[width=0.22\textwidth]{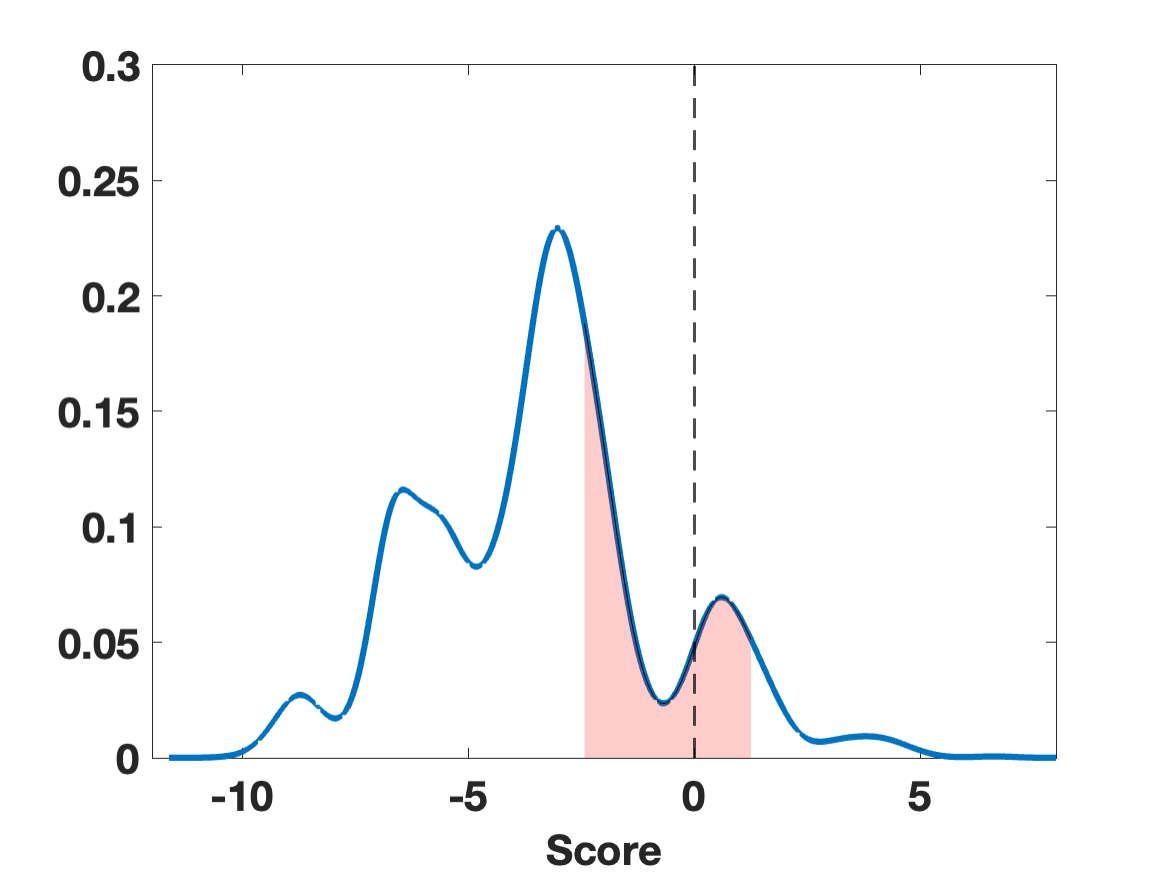}
        \end{tabular}
	\caption{Distributions of predicted scores of different sensitive groups on \textit{a9a} dataset by the unconstrained model and the models solved from \eqref{eq:inprocess_wpdp_approx} (pDP constraints) with different $\kappa$'s. The interval $\mathcal{I}$ is $[5\%, 30\%]$ and is highlighted in red.}  
    \label{fig:wpDP_accuracy_a9a_distribution}
	\vspace{-0.1in}
\end{figure*}

\begin{figure*}[!ht]
     \begin{tabular}{@{}c|cccc@{}}
      & unconstrained & $\kappa = 0.1$ & $\kappa = 0.05$ & $\kappa = 0.01$ \\
		\hline \vspace*{-0.1in}\\
		\raisebox{7ex}{\small{\rotatebox[origin=c]{90}{Other ages}}}
		& \hspace*{-0.06in}\includegraphics[width=0.22\textwidth]{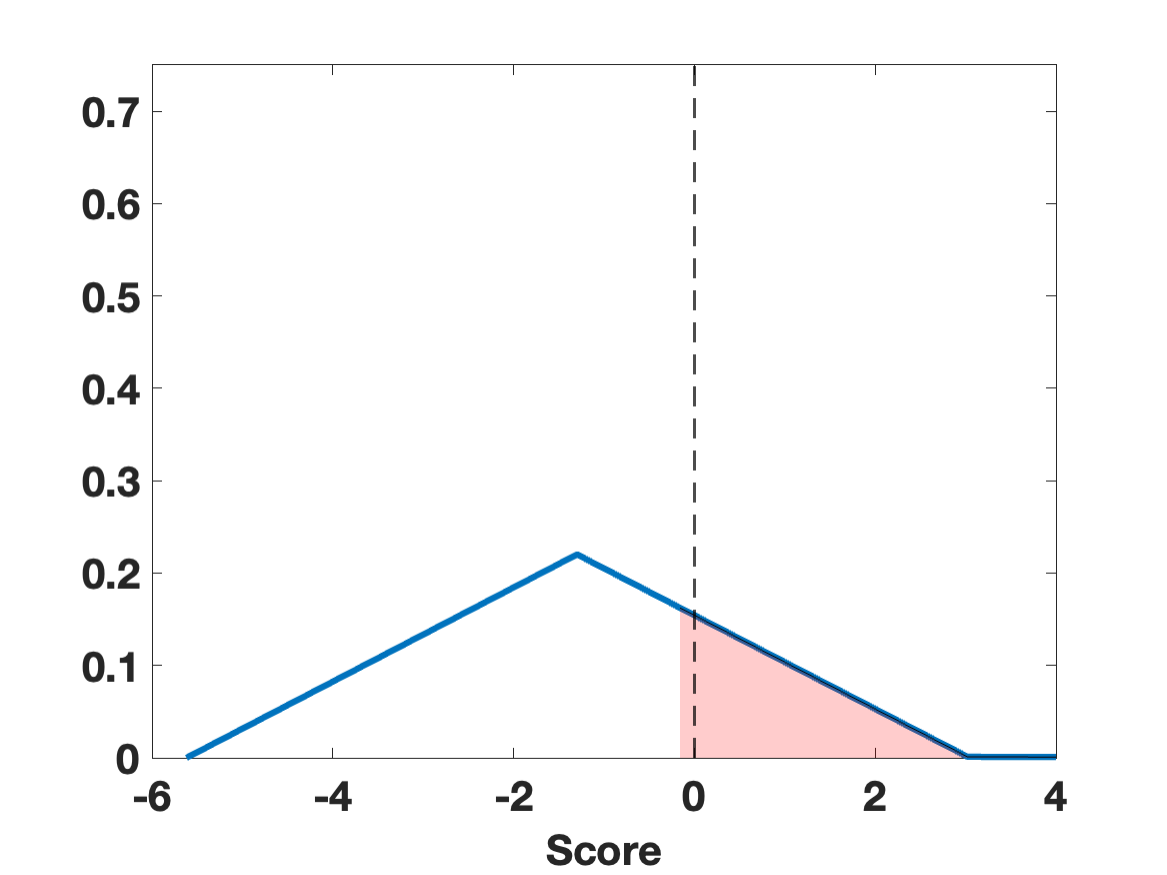}
		& \hspace*{-0.06in}\includegraphics[width=0.22\textwidth]{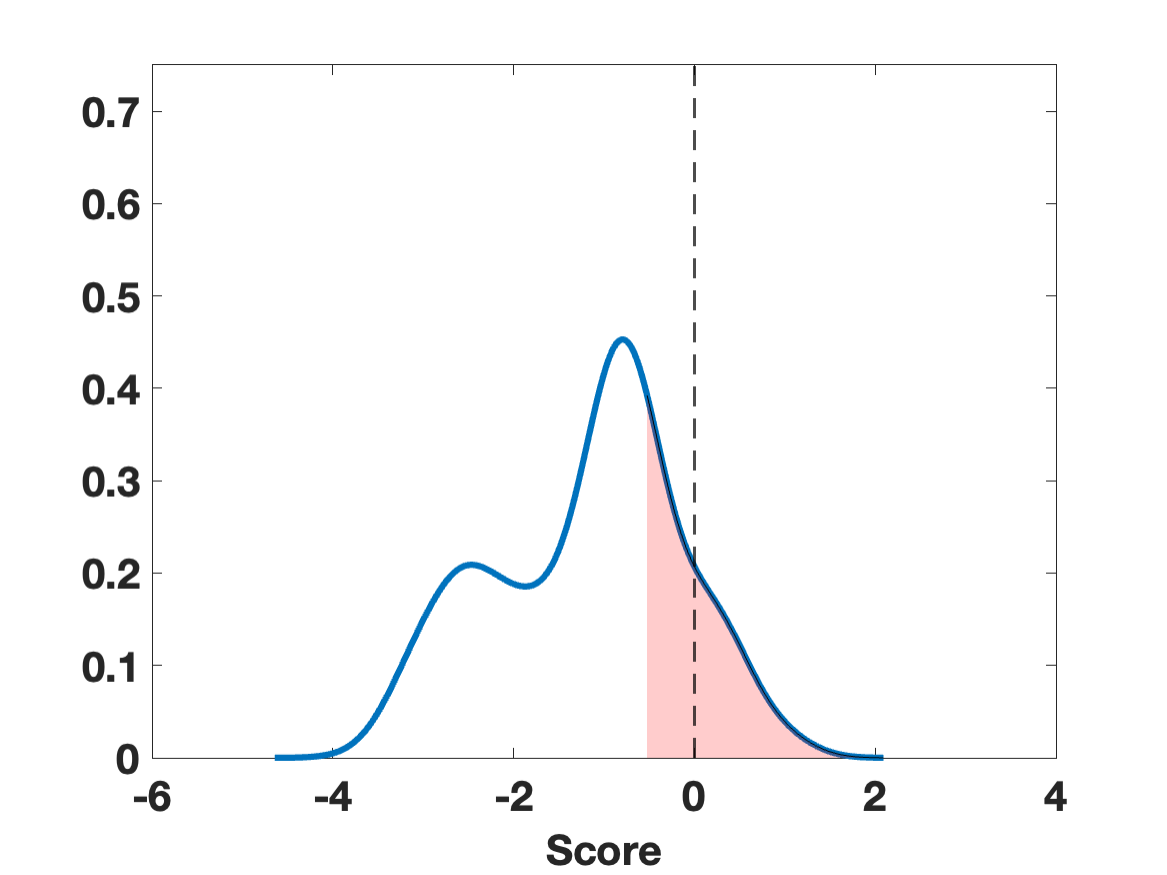}
		& \hspace*{-0.06in}\includegraphics[width=0.22\textwidth]{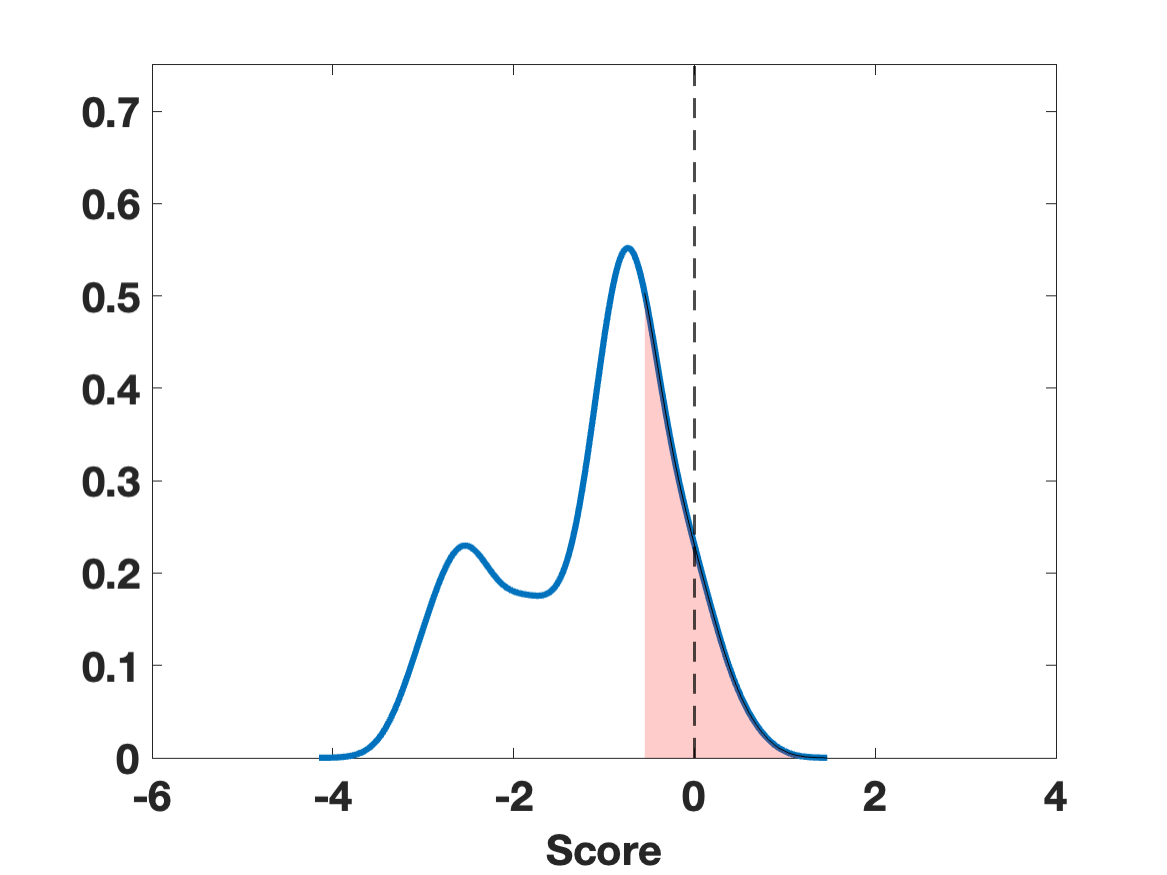}
		& \hspace*{-0.06in}\includegraphics[width=0.22\textwidth]{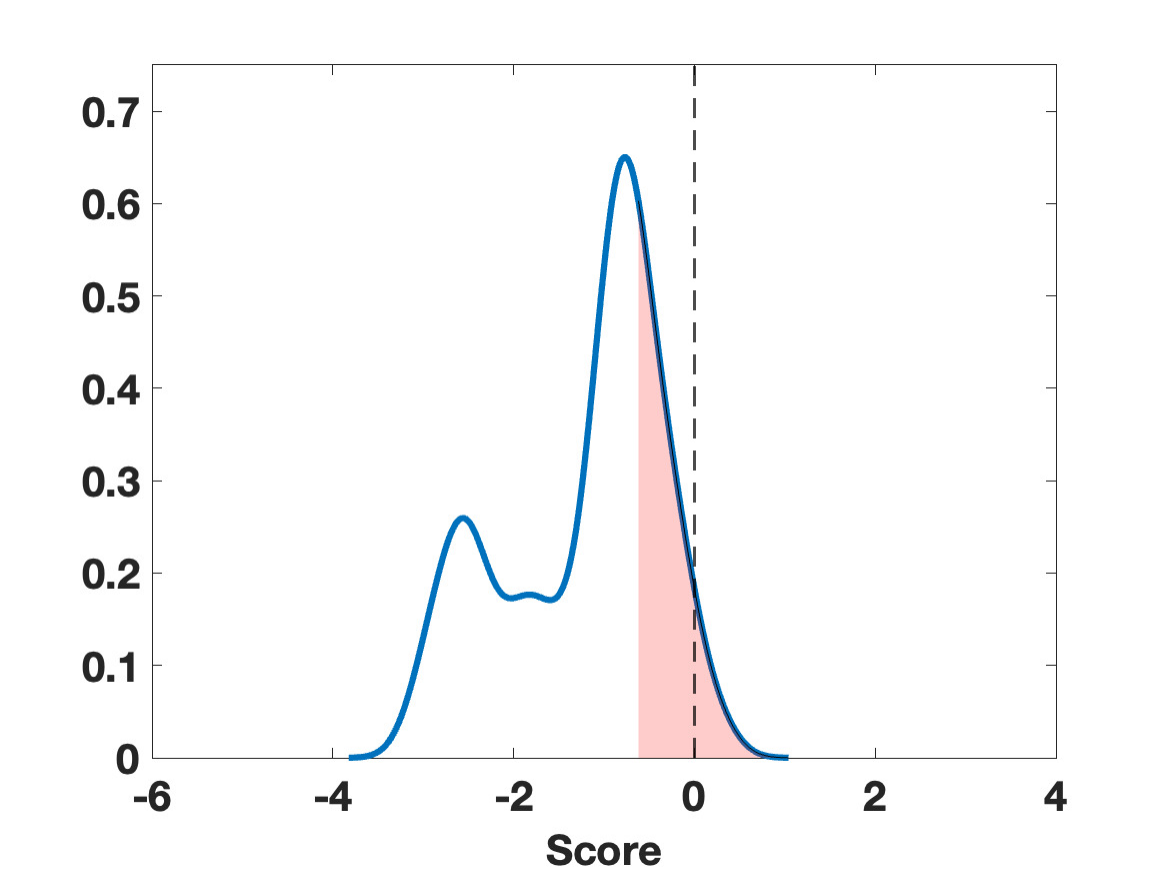}
        
          \\
		\raisebox{7ex}{\small{\rotatebox[origin=c]{90}{Ages between 25 and 60}}}
		& \hspace*{-0.06in}\includegraphics[width=0.22\textwidth]{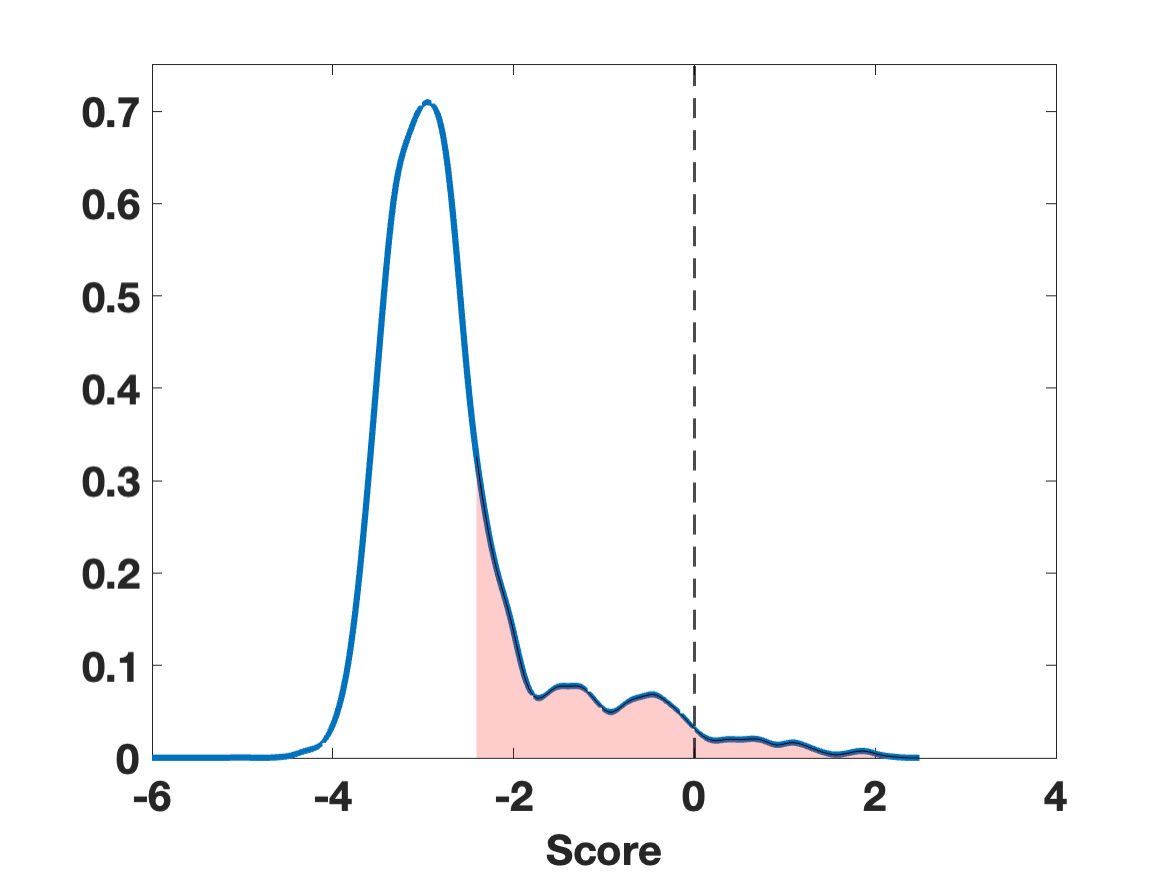}
		& \hspace*{-0.06in}\includegraphics[width=0.22\textwidth]{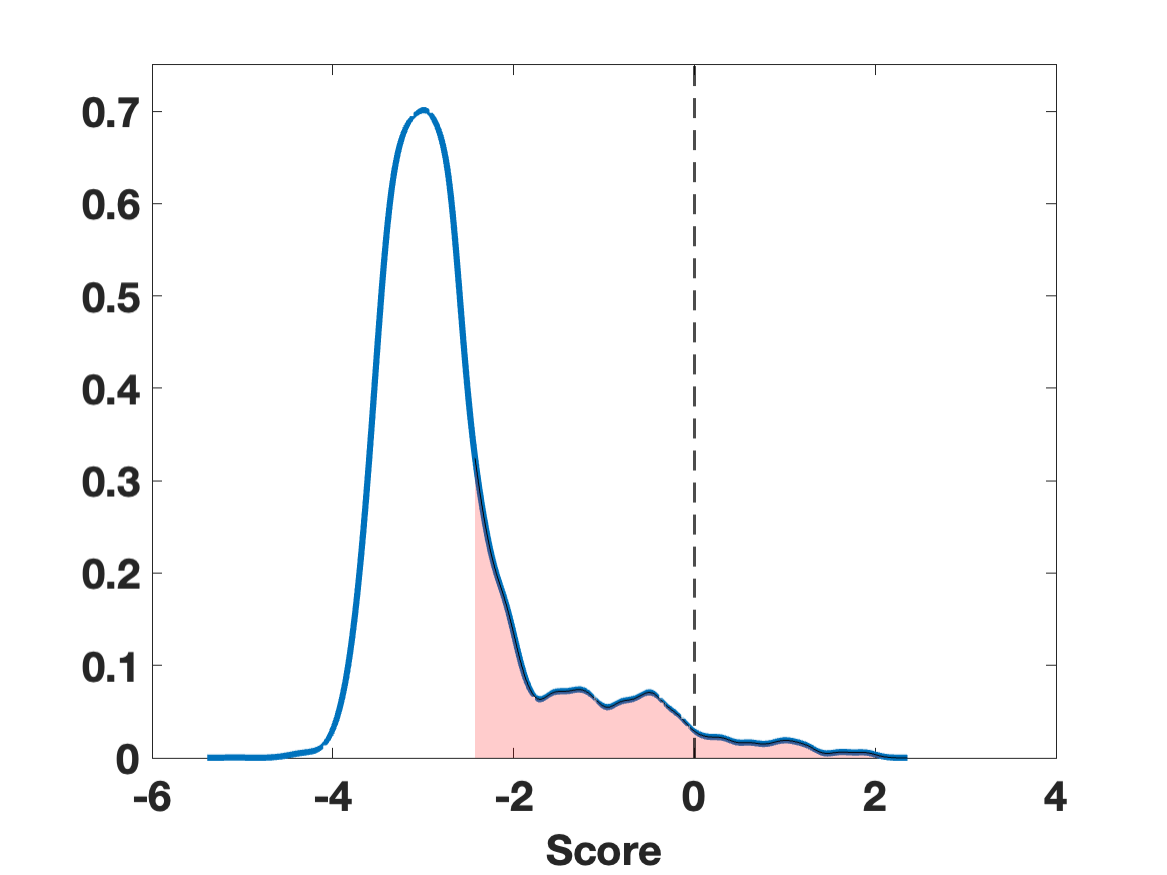}
		& \hspace*{-0.06in}\includegraphics[width=0.22\textwidth]{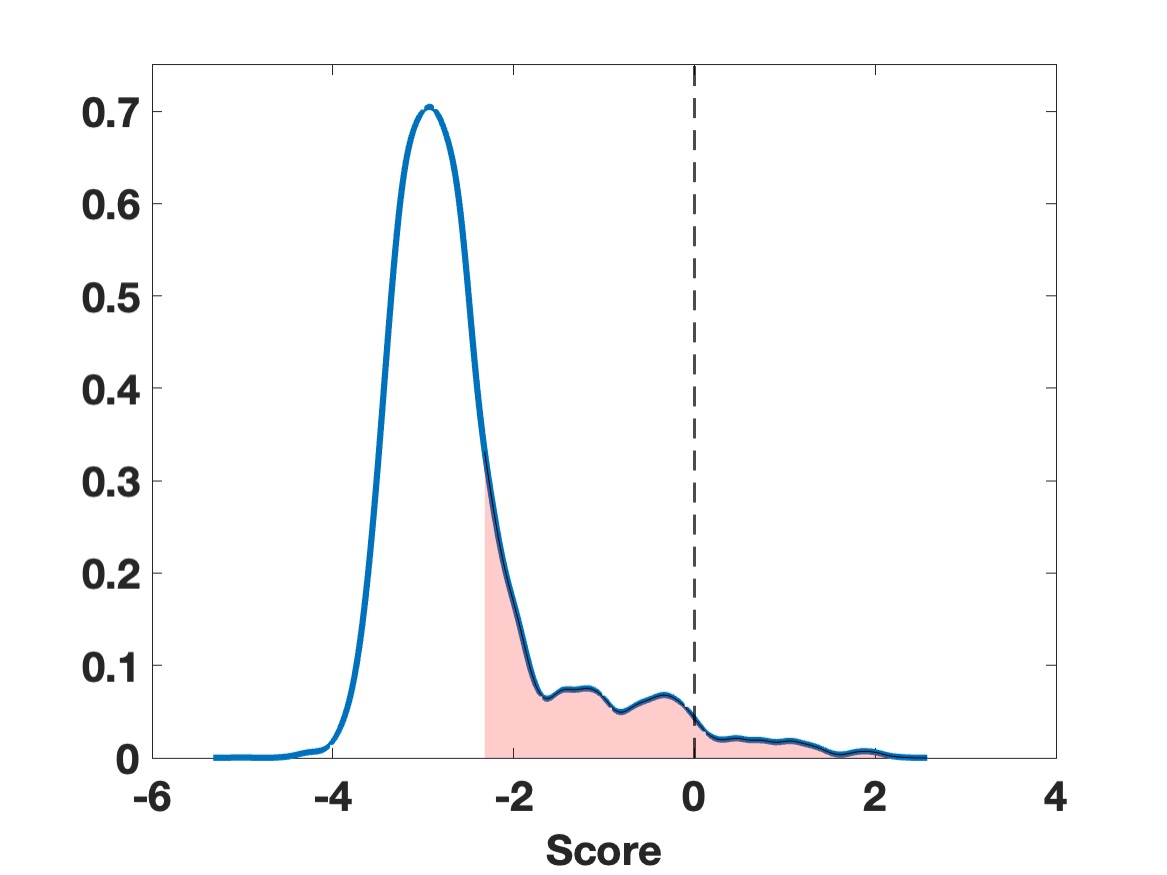}
		& \hspace*{-0.06in}\includegraphics[width=0.22\textwidth]{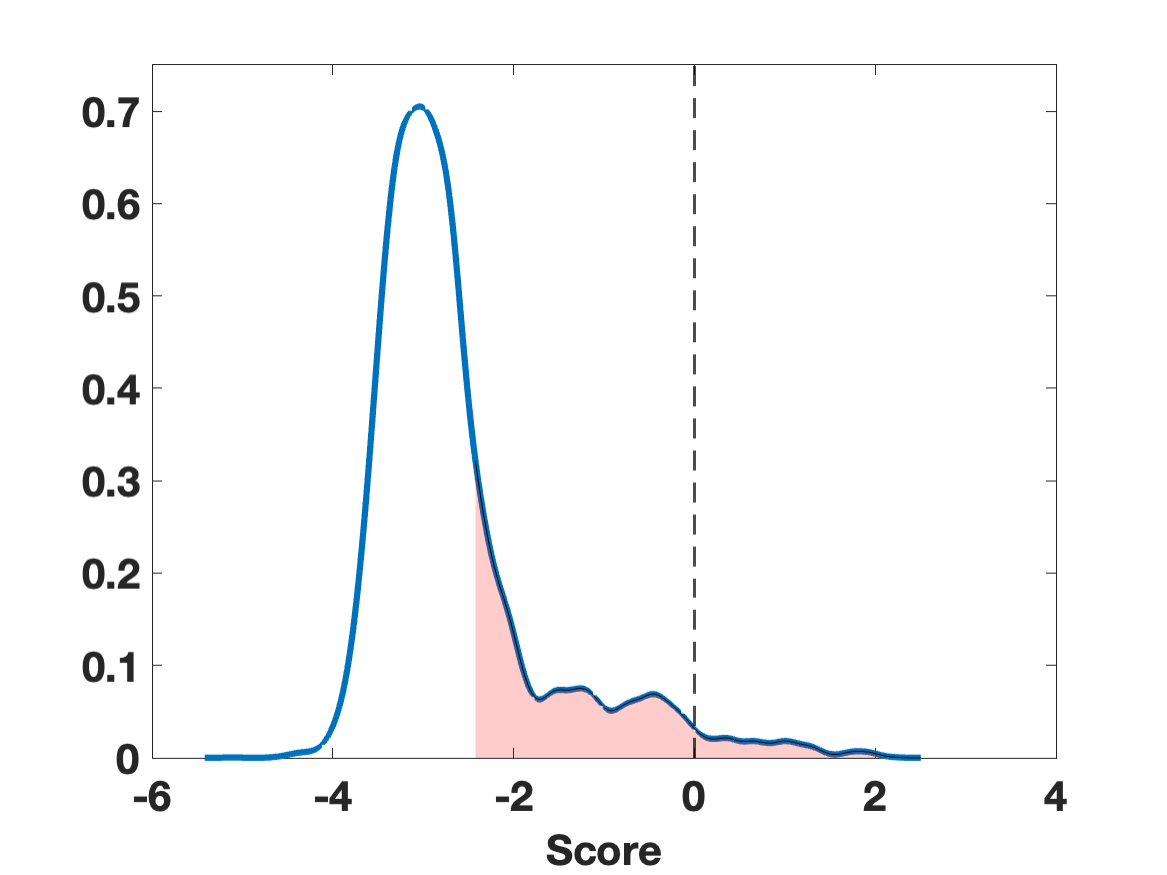}
        \end{tabular}
	\caption{Distributions of predicted scores of different sensitive groups on \textit{bank} dataset by the unconstrained model and the models solved from \eqref{eq:inprocess_wpdp_approx} (pDP constraints) with different $\kappa$'s. The interval $\mathcal{I}$ is top $[0\%, 25\%]$ and is highlighted in red.}  
    \label{fig:wpDP_accuracy_bank_distribution}
	\vspace{-0.1in}
\end{figure*}

\begin{figure*}[!ht]
     \begin{tabular}{@{}c|cccc@{}}
      & unconstrained & $\kappa = 0.1$ & $\kappa = 0.04$ & $\kappa = 0.005$ \\
		\hline \vspace*{-0.1in}\\
		\raisebox{7ex}{\small{\rotatebox[origin=c]{90}{Non-white}}}
		& \hspace*{-0.06in}\includegraphics[width=0.22\textwidth]{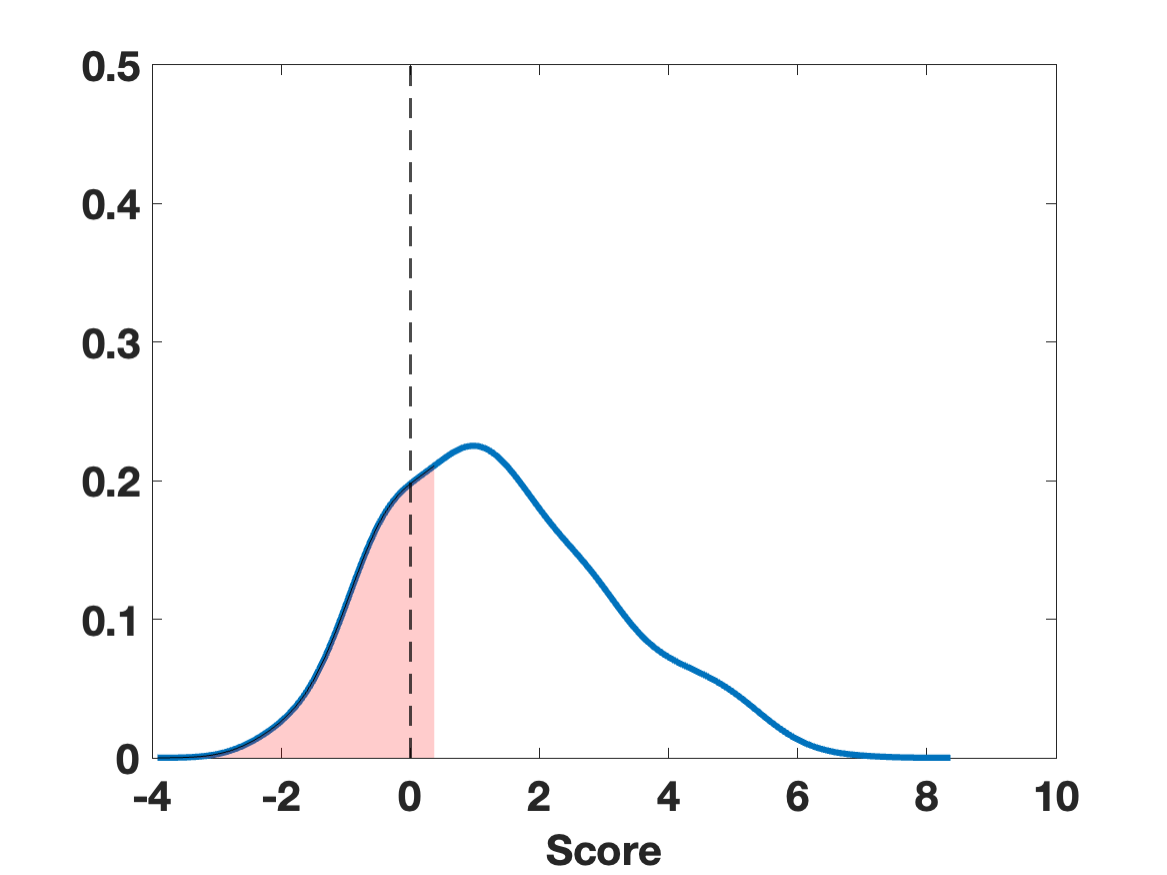}
		& \hspace*{-0.06in}\includegraphics[width=0.22\textwidth]{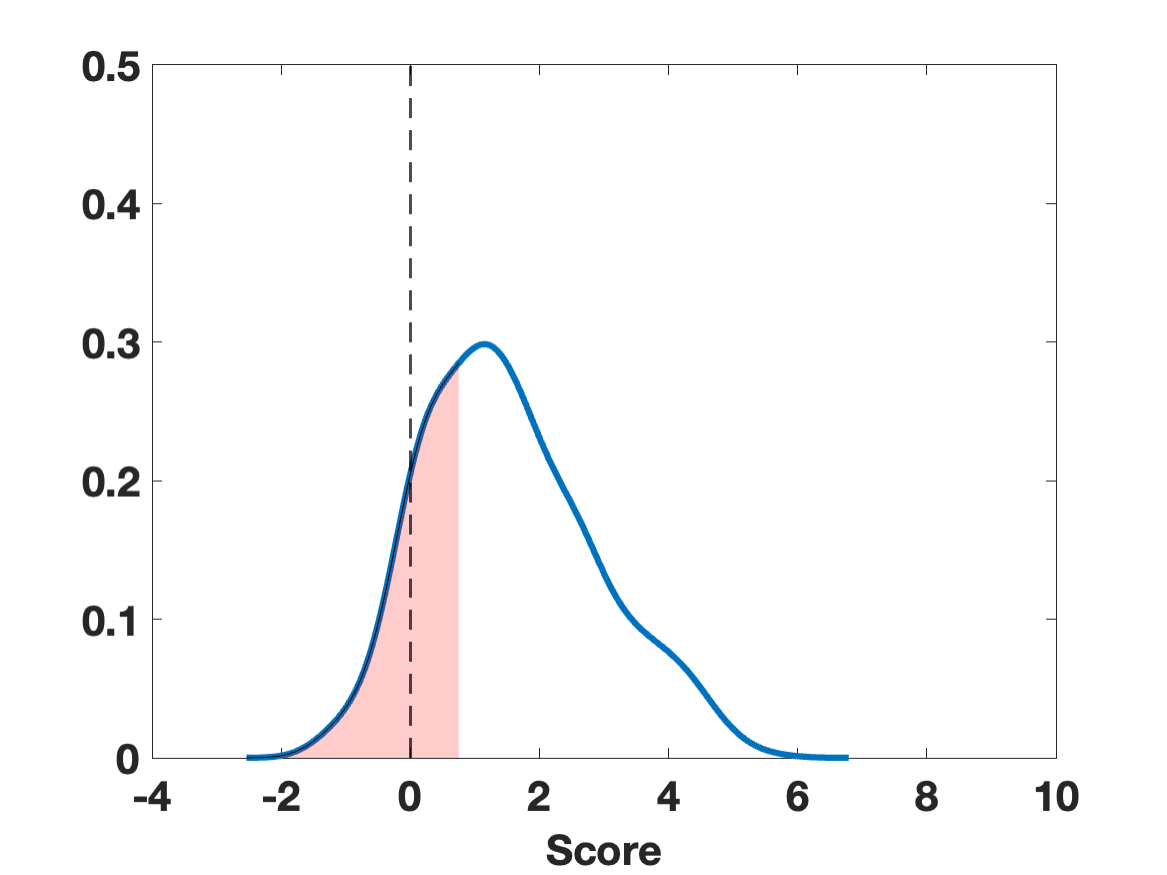}
		& \hspace*{-0.06in}\includegraphics[width=0.22\textwidth]{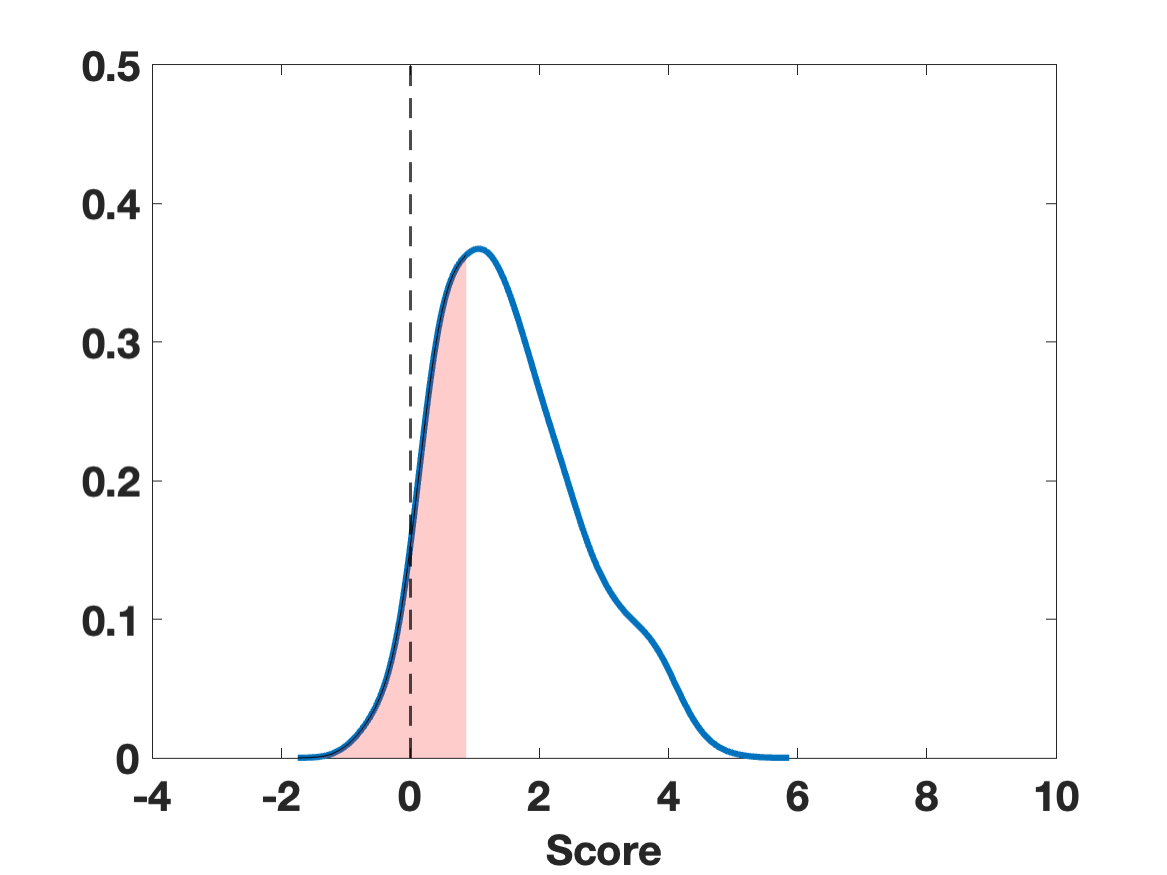}
		& \hspace*{-0.06in}\includegraphics[width=0.22\textwidth]{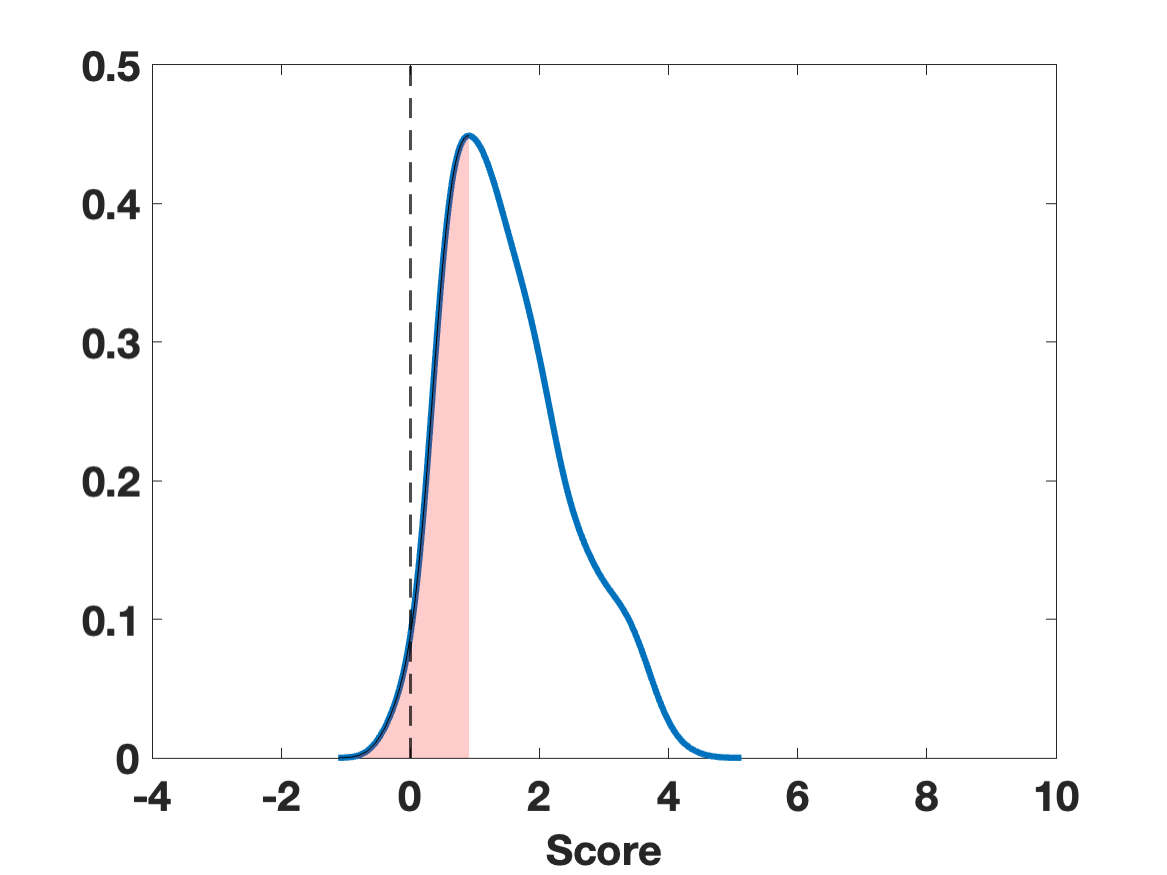}
        
          \\
		\raisebox{7ex}{\small{\rotatebox[origin=c]{90}{White}}}
		& \hspace*{-0.06in}\includegraphics[width=0.22\textwidth]{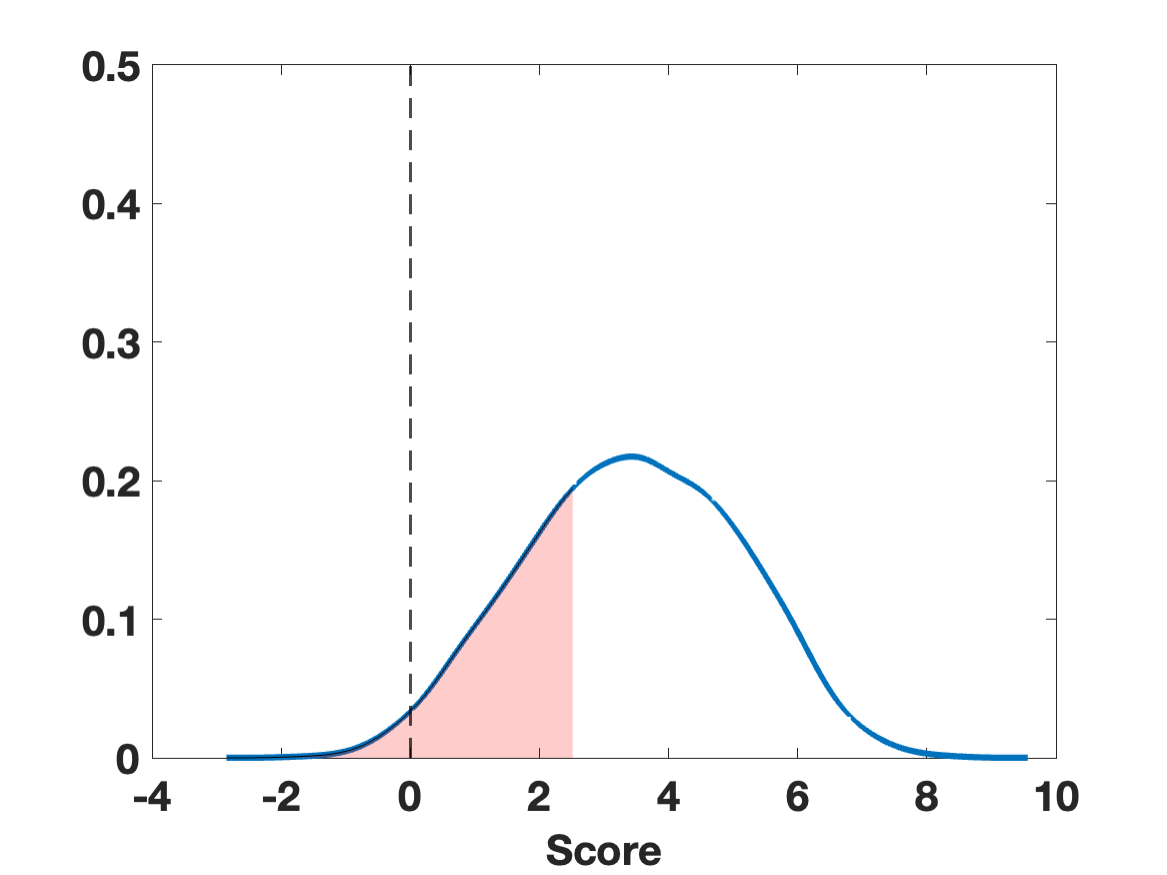}
		& \hspace*{-0.06in}\includegraphics[width=0.22\textwidth]{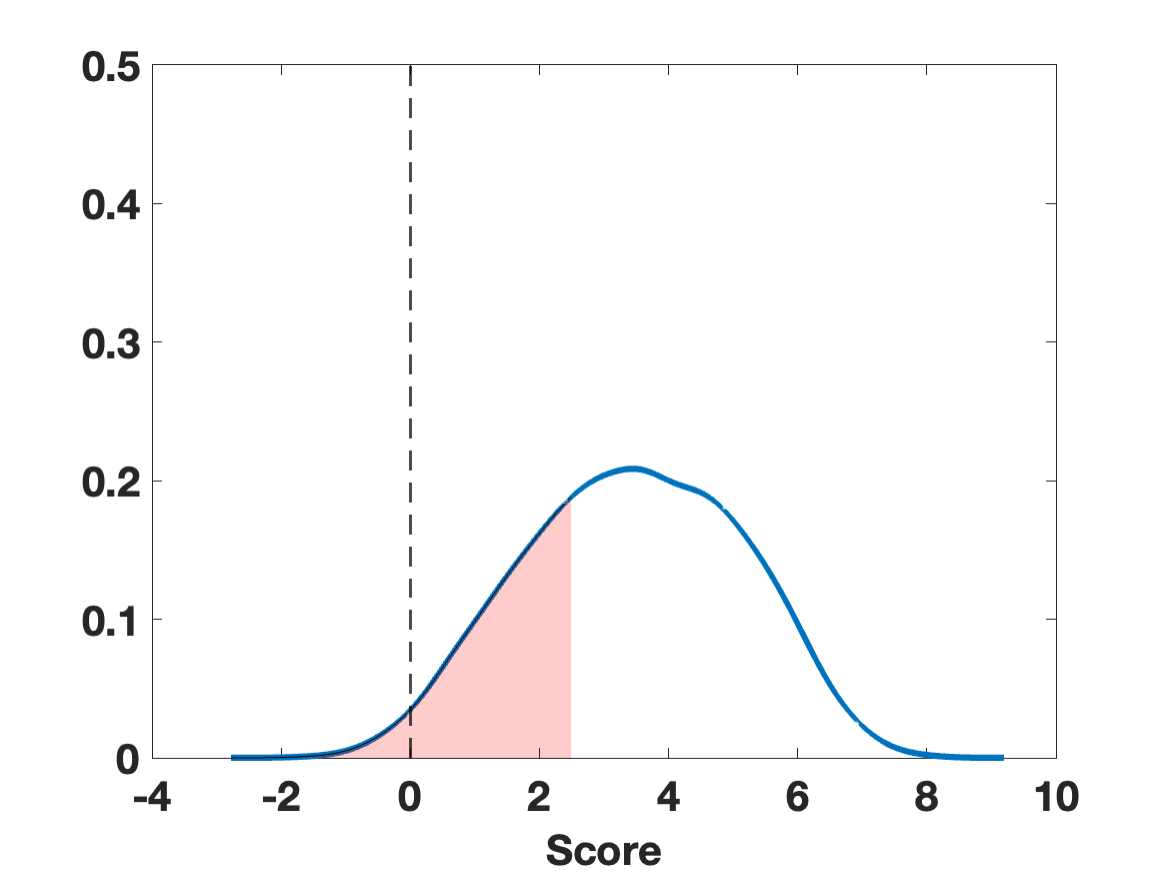}
		& \hspace*{-0.06in}\includegraphics[width=0.22\textwidth]{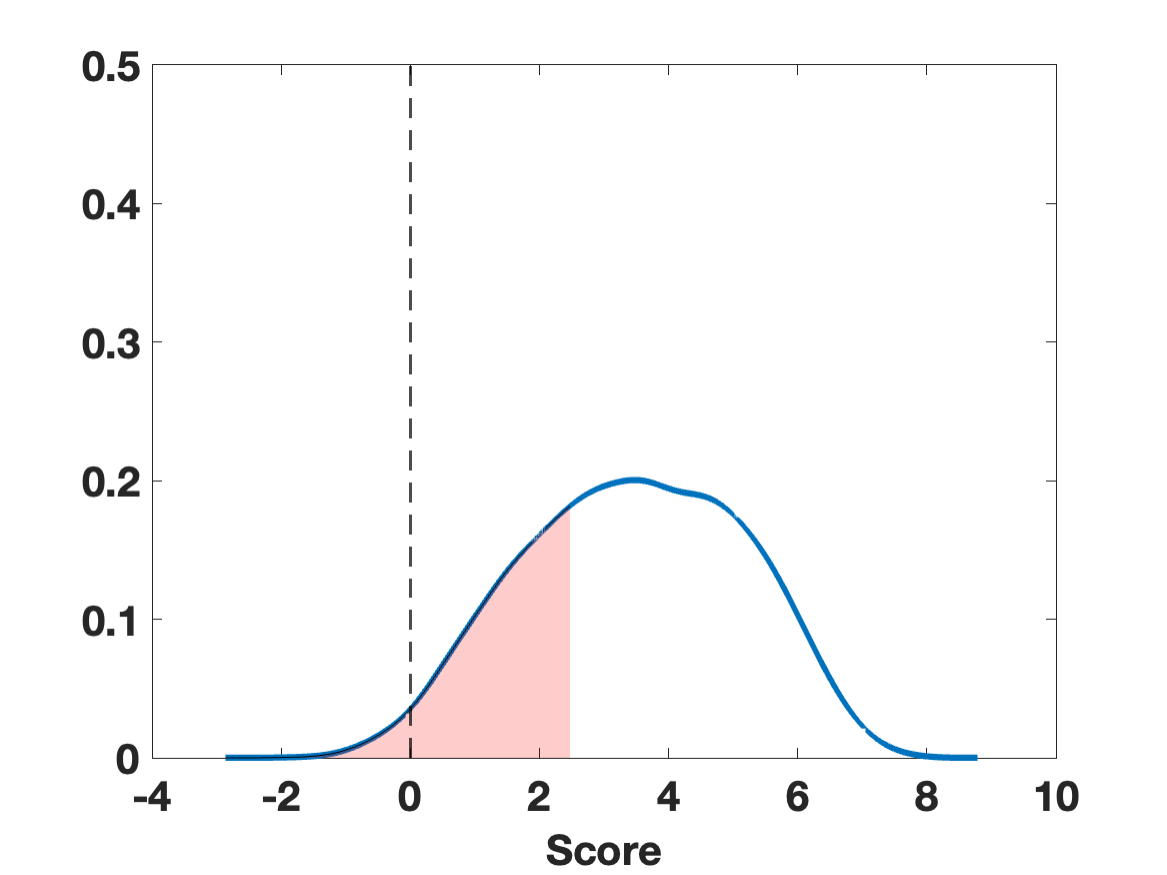}
		& \hspace*{-0.06in}\includegraphics[width=0.22\textwidth]{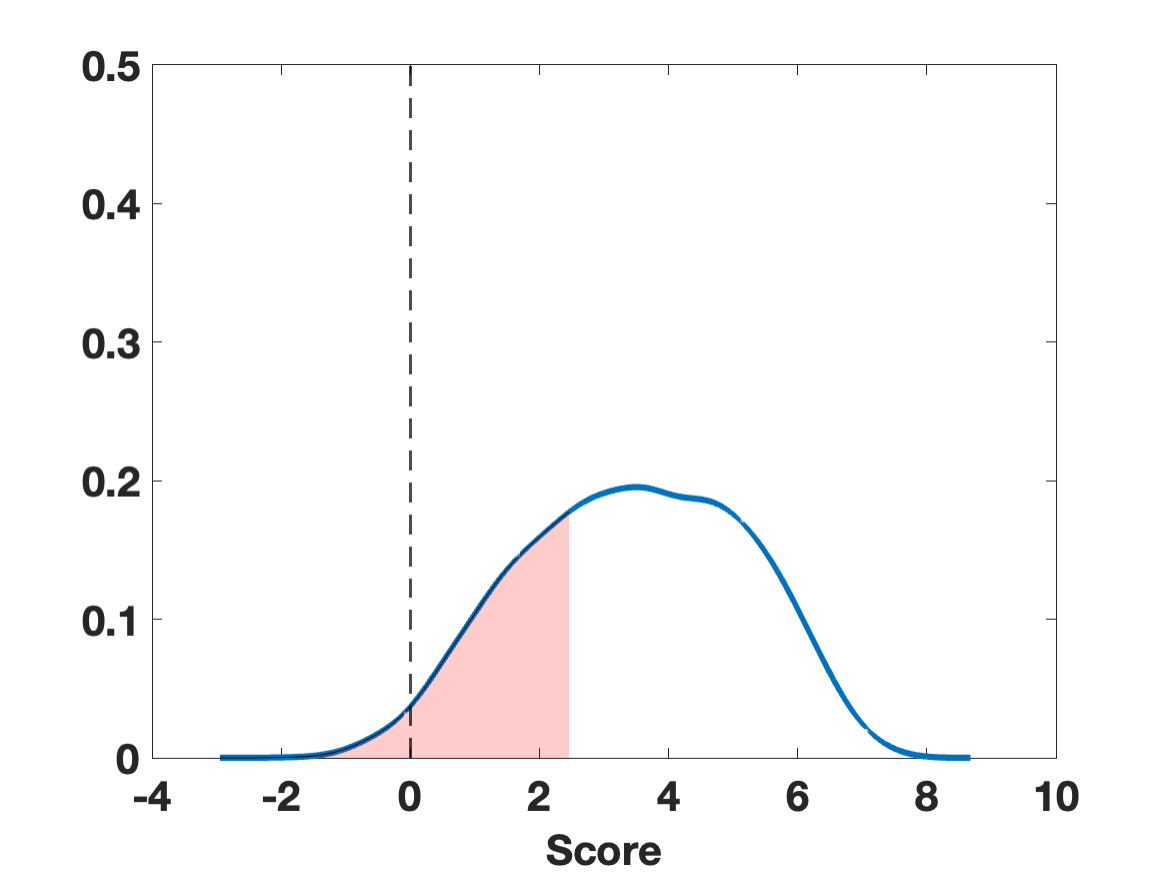}
        \end{tabular}
	\caption{Distributions of predicted scores of different sensitive groups on \textit{law school} dataset by the unconstrained model and the models solved from \eqref{eq:inprocess_wpdp_approx} (pDP constraints) with different $\kappa$'s. The interval $\mathcal{I}$ is $[70\%, 100\%]$ and is highlighted in red.}  
    \label{fig:wpDP_accuracy_law_distribution}
	\vspace{-0.1in}
\end{figure*}

In this section, we present how the densities of the prediction scores produced by our methods change with $\kappa$ on different datasets. In particular, in Figures~\ref{fig:pDP_accuracy_a9a_distribution}-\ref{fig:wpDP_accuracy_law_distribution}, we plot the estimated densities of $h_{\vw}(\vxi)$ by the models solved from \eqref{eq:inprocess_pdp_approx} and \eqref{eq:inprocess_wpdp_approx} with different values of $\kappa$. The density curves are estimated by the \texttt{ksdensity} function from MATLAB using the values of $h_{\vw}(\vxi)$ on the testing sets. 
For comparisons, the densities of the two sensitive groups are plotted separately, and 
the areas within the interval $\mathcal{I}$ are highlighted in red. 

According to Figures~\ref{fig:pDP_accuracy_a9a_distribution}-\ref{fig:wpDP_accuracy_law_distribution}, we have the following observations:
\begin{itemize}
    \item When pSP fairness is enforced (Figure~\ref{fig:pDP_accuracy_a9a_distribution}, Figure~\ref{fig:pDP_accuracy_bank_distribution}, and Figure~\ref{fig:pDP_accuracy_law_distribution}), as $\kappa$ decreases, the distributions of $h_{\vw}(\vxi)$ on $\mathcal{I}$ on the two groups become more and more similar.
    \item When pDP fairness is enforced (Figure~\ref{fig:wpDP_accuracy_a9a_distribution}, Figure~\ref{fig:wpDP_accuracy_bank_distribution}, and Figure~\ref{fig:wpDP_accuracy_law_distribution}), as $\kappa$ decreases,  the proportions of the red areas above 0, i.e., $\text{Pr}\big(h_{\vw}(\vxi) >\widehat\theta \big| \gamma=k, \bar{F}_{\vw,k}(h_{\vw}(\vxi))\in\mathcal{I}\big)$ with $\widehat\theta=0$, on the two groups become more and more similar.
\end{itemize}
Both observations are intuitive and consistent with the goals of the two partial fairness metrics we enforce through the constraints in \eqref{eq:inprocess_pdp_approx} and \eqref{eq:inprocess_wpdp_approx}.

\subsection{Robustness to the choice of $\mathcal{I}$}\label{sec:diffI}
\begin{figure*}[tb]
     \begin{tabular}{@{}ccc@{}}
       a9a & bank & law school \\
		\hline \vspace*{-0.1in}\\
		 \hspace*{0.06in}\includegraphics[width=0.30\textwidth]{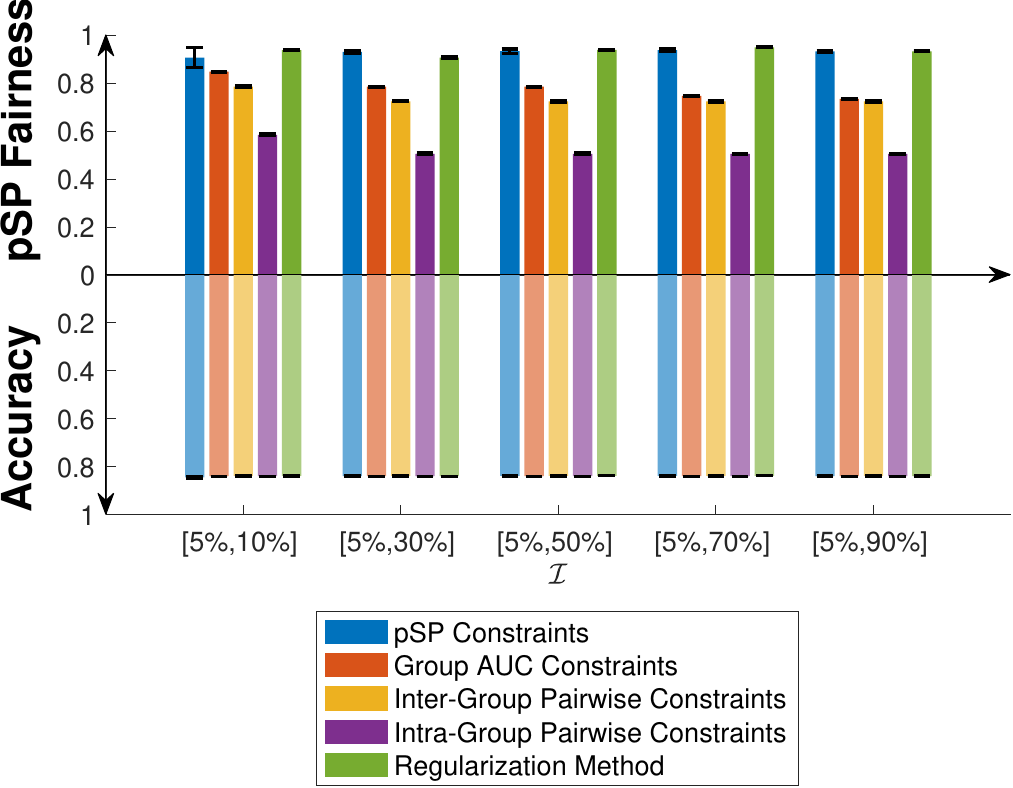}
		& \hspace*{-0.06in}\includegraphics[width=0.30\textwidth]{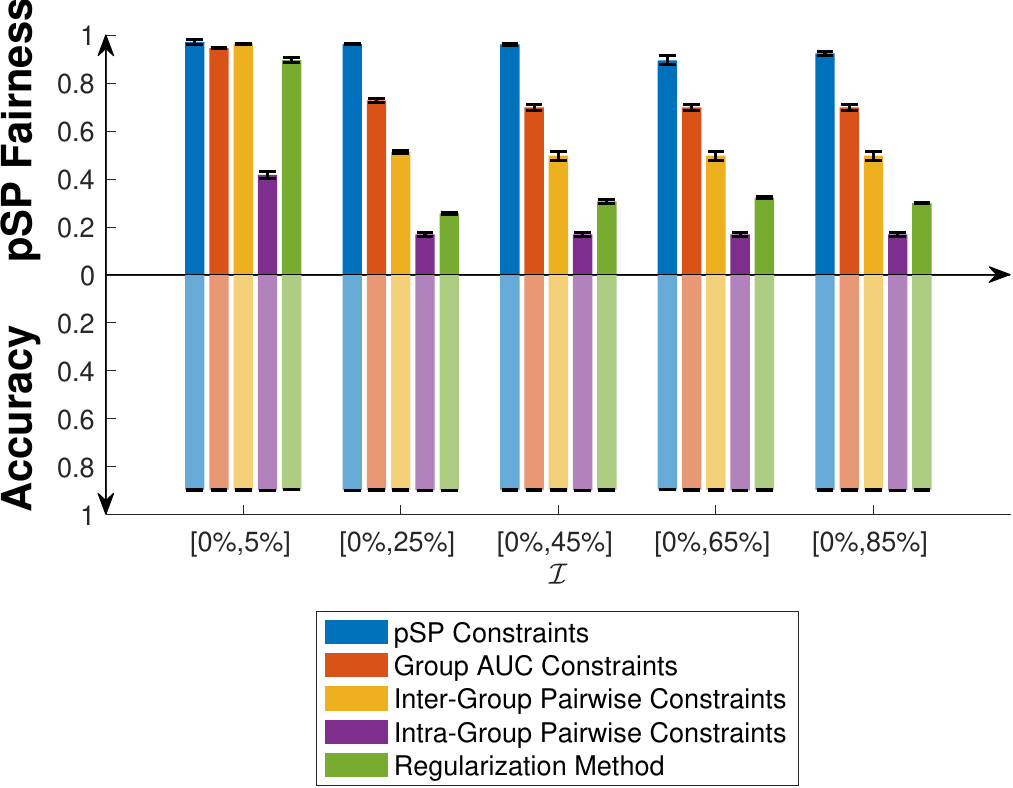}
		& \hspace*{-0.06in}\includegraphics[width=0.30\textwidth]{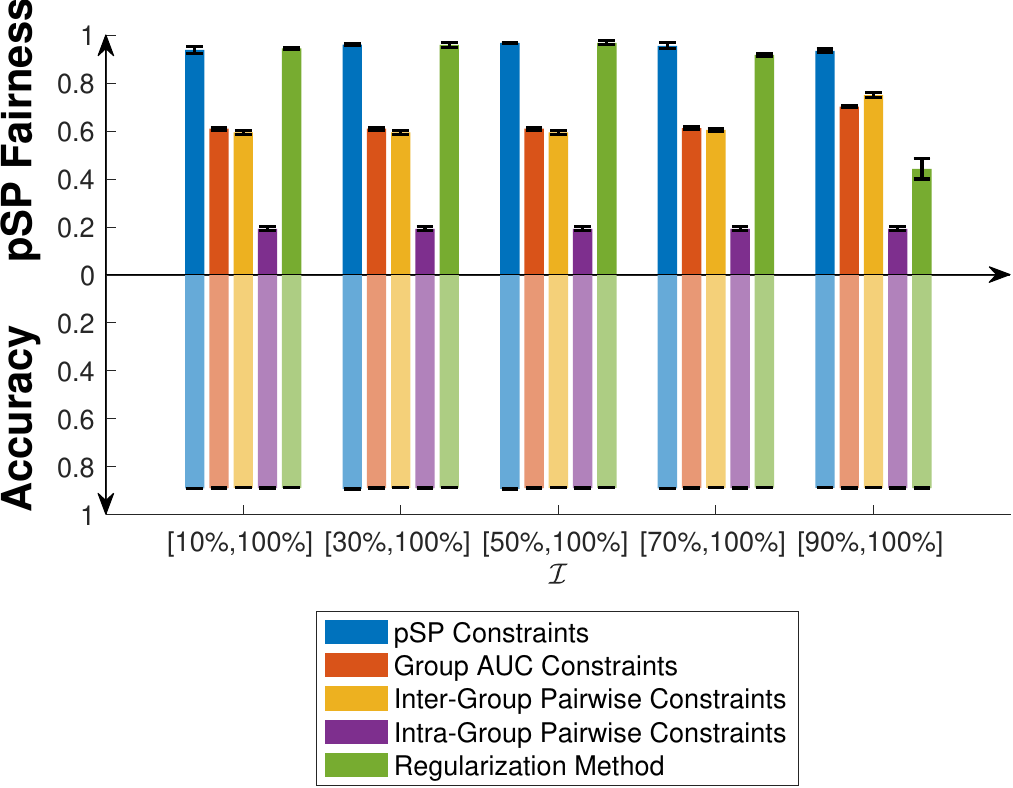}
        \end{tabular}
	\caption{Comparison of
    different in-processing methods in 
    pSP fairness they guarantee when they achieve the same classification accuracy. Results are presented with 
    different intervals  $\mathcal{I}$ on \textit{a9a}, \textit{bank}, and \textit{law school} datasets. }  
    \label{fig:pDP_diff_interval}
	\vspace{-0.1in}
\end{figure*}

A key remaining question is whether the effectiveness of our method is robust to the choice of interval $\mathcal{I}$. In this section, we provide an affirmative answer to this question by comparing our method with other approaches when different intervals $\mathcal{I}$'s are used. For each dataset, we create a collection of intervals $\mathcal{I}$'s (see below). Then, for each interval, we use the same $\kappa$ in \eqref{eq:inprocess_pdp_approx} to train a model, and then evaluate its classification accuracy and pSP fairness. For the other in-processing methods in (\ref{eq:inprocess_gauc_approx}), (\ref{eq:inprocess_igpf_approx}), (\ref{eq:inprocess_itgpf_approx}), and (\ref{eq:inprocess_pdp_regularized_approx}), we choose the value of $\kappa$ or $\lambda$ in their optimization models such that their output solutions achieve an accuracy similar to the one from \eqref{eq:inprocess_pdp_approx}. After this, we evaluate the pSP fairness of their solutions. Finally, we compare the accuracy and pSP fairness of all methods using the mirrored grouped bar charts shown in Figure~\ref{fig:pDP_diff_interval}. As expected, the accuracies of all methods are the same for each interval because $\kappa$ or $\lambda$ is chosen intentionally to ensure that happens. The differences therefore lie in their pSP fairness, and our method demonstrates the most robust performance across different intervals in the sense of achieve the same highest and most stable fairness as the interval changes.  

The interval we choose in this experiment and the values of $\kappa$ or $\lambda$ used in each method are given below.

\begin{itemize}
\item Choices of $\mathcal{I}$'s:

\textit{a9a}: $\mathcal{I} \in \{ [5\%, 10\%], \textbf{[5\%, 30\%]},[5\%, 50\%],[5\%, 70\%],[5\%, 90\%] \}$.

\textit{bank}: $\mathcal{I} \in \{ [0\%, 5\%], \textbf{[0\%, 25\%]},[0\%, 45\%],[0\%, 65\%],[0\%, 85\%] \}$.

\textit{law school}: $\mathcal{I} \in \{ [10\%, 100\%], [30\%, 100\%], [50\%, 100\%], \textbf{[70\%, 100\%]}, [90\%, 100\%] \}$.

where the bold intervals are ones we used for Figure \ref{fig:pDP_and_wpDP_efficiency_frontier}.

    \item For problem (\ref{eq:inprocess_pdp_approx}) (pSP fairness):

    \textit{a9a}: $\kappa = 0.05$; \textit{bank} : $\kappa = 0.01$; \textit{law school} : $\kappa = 0.005$.

    \item For problem (\ref{eq:inprocess_gauc_approx}) (pSP fairness):

    \textit{a9a}: $\kappa = 1.5$; \textit{bank} : $\kappa = 0.7$; \textit{law school} : $\kappa = 2.7$.

    \item For problem (\ref{eq:inprocess_igpf_approx}) (pSP fairness):

    \textit{a9a}: $\kappa = 4.6$; \textit{bank} : $\kappa = 3$; \textit{law school} : $\kappa = 4$.

    \item For problem (\ref{eq:inprocess_itgpf_approx}) (pSP fairness):

    \textit{a9a}: $\kappa = 4.2$; \textit{bank} : $\kappa = 2$; \textit{law school} : $\kappa = 0.9$.

    \item For problem (\ref{eq:inprocess_pdp_regularized_approx}) (pSP fairness):

    \textit{a9a}: $\lambda = 0.37$; \textit{bank} : $\lambda = 0.3$; \textit{law school} : $\lambda = 0.8$.
\end{itemize} 

\subsection{Optimization performance of IDCA}
\label{sec:expopt}
In this section, we compare the IDCA with the switching subgradient (SSG) method directly applied to \eqref{DC}  (instead of being used as a subroutine), which includes \eqref{eq:inprocess_pdp_approx} and \eqref{eq:inprocess_wpdp_approx} as special cases. The problem instances we use here are the same as those in Section~\ref{sec:exp}. Note that the complexity of SSG has only been analyzed for convex problems~\citep[(3.2.24)]{nesterov2018lectures} and weakly convex problems~\citep{huang2023oracle}, while its complexity for solving DC optimization is still unknown. 

We first present the SSG method directly applied to \eqref{DC} in Algorithm~\ref{alg:ssgdirect}, where we denote 
$$
f(\cdot):=\max_{i\in[m]}f_i(\cdot).
$$
The tuning parameter in Algorithm~\ref{alg:ssgdirect} is $\epsilon_t$.  Two different ways of setting $\epsilon_t$ are considered and the better one is applied in the comparison with our IDCA method. One way is to set a static $\epsilon_t$, where $\epsilon_t=\epsilon$ for some $\epsilon>0$ for any $t$. The other way is to set a diminishing $\epsilon_t$, where $\epsilon_t=\frac{c}{t+1}$ for some $c>0$. We select $\epsilon$ in the static setup from $\{10^{-4},2\times10^{-4},5\times10^{-4},10^{-3}\}$ and select $c$ in the dynamic setup from $\{0.2,0.5,1,2\}$. For each of the candidate values of $\epsilon_t$ set in both the dynamic and static ways, we run Algorithm~\ref{alg:ssgdirect} for $T=10000$ iterations and choose $\epsilon_t$ that produces the smallest objective value at termination. Then we run Algorithm~\ref{alg:ssgdirect} again for $T=50000$ iterations using the best  $\epsilon_t$.

\begin{algorithm}[t]
   \caption{Switching Subgradient (SSG) Method for \eqref{DC} (including \eqref{eq:inprocess_pdp_approx} and \eqref{eq:inprocess_wpdp_approx} as special cases)}
   \label{alg:ssgdirect}
\begin{algorithmic}[1]
   \STATE {\bfseries Input:} Infeasibility tolerance $\epsilon >0$, initial solution $\vw^{(0)} \in \cW$ with $f_i(\vw^{(0)})\leq \epsilon$ for $i\in[m]$ and the number of iterations $T$. 
   \STATE $\cT\leftarrow\emptyset$.
    \FOR {$t=0$ {\bfseries to} $T-1$} 
   \IF{$f(\vw^{(t)}) \leq \epsilon_t$}
   \STATE Compute $\vf_0^{(t)}\in \partial f_0(\vw^{(t)})$.
   \STATE $\vw^{(t+1)}\leftarrow \text{Proj}_{\cW}(\vw^{(t)}-\frac{\epsilon_t}{\|\vf_0^{(t)}\|^2} \vf_0^{(t)})$
   \STATE $\cT\leftarrow\cT\cup\{t\}$.
   \ELSE
   \STATE Compute $\vf^{(t)}\in \partial f(\vw^{(t)})$.
   \STATE $\vw^{(t+1)}\leftarrow  \text{Proj}_{\cW}(\vw^{(t)}-\frac{f(\vw^{(t)})}{\|\vf^{(t)}\|^2} \vf^{(t)})$
   \ENDIF
   \ENDFOR
   \STATE {\bfseries Output: }$\vw^{(\tau)}$ with $\tau=\argmin_{t\in\cT} f_0(\vw^{(t)})$
\end{algorithmic}
\end{algorithm}

\begin{figure*}[t]
     \begin{tabular}{@{}c|cccc@{}}
      & a9a & bank & law school 
      \\
		\hline \vspace*{-0.1in}\\
		\raisebox{9ex}{\small{\rotatebox[origin=c]{90}{Objective value}}}
		& \hspace*{-0.06in}\includegraphics[width=0.30\textwidth]{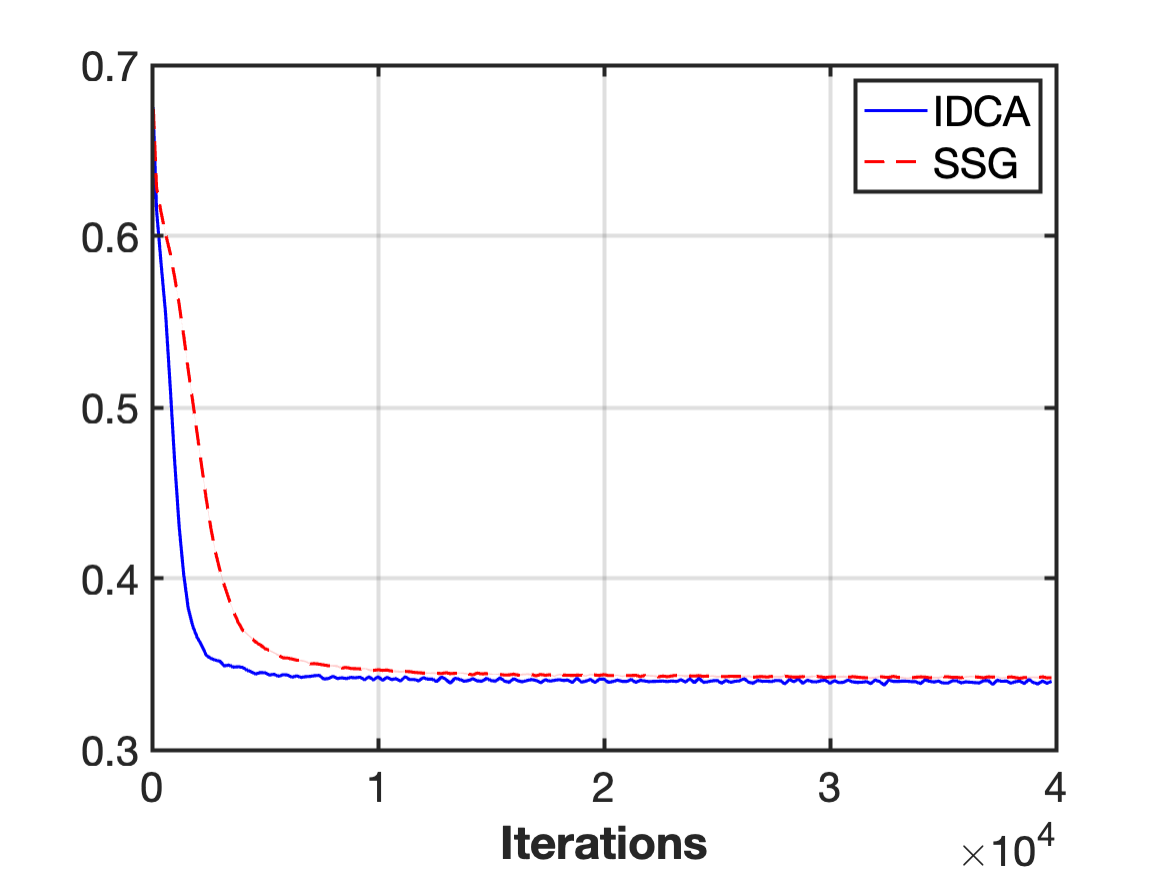}
		& \hspace*{-0.06in}\includegraphics[width=0.30\textwidth]{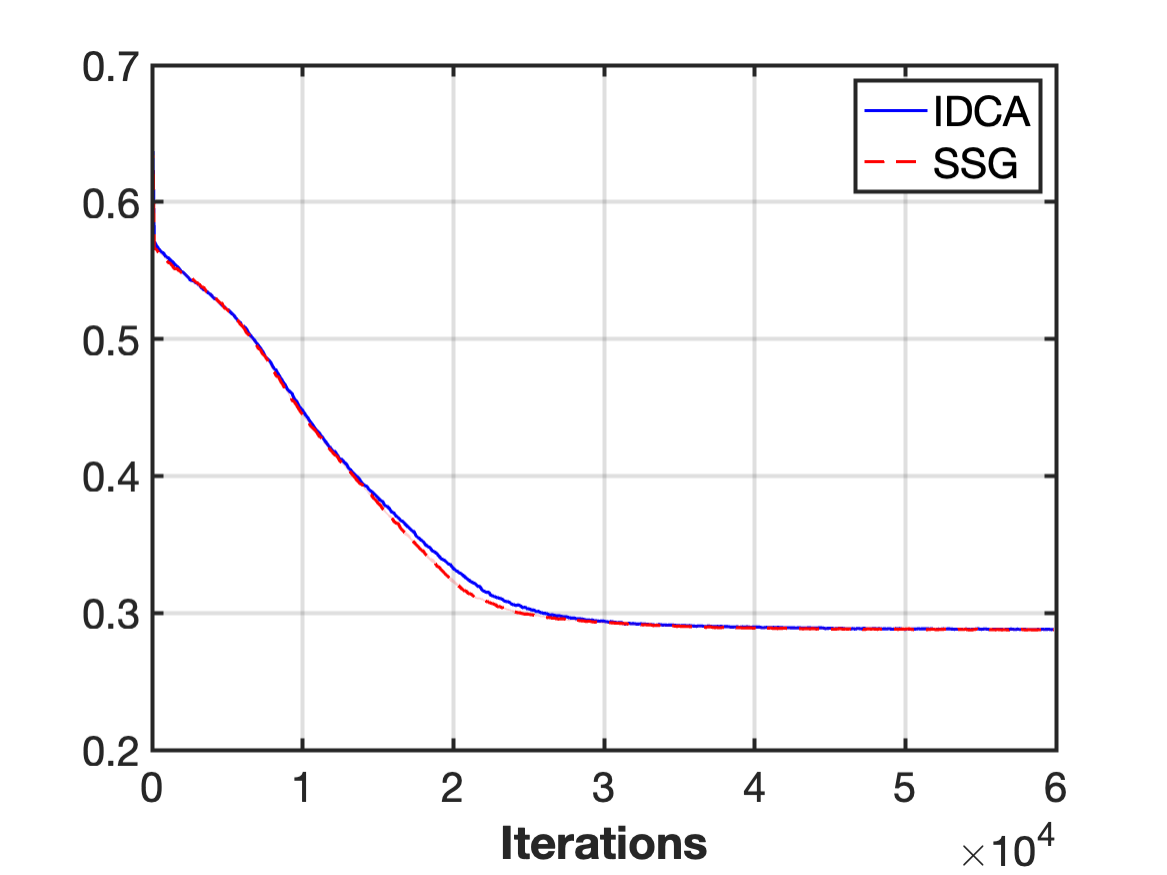}
		& \hspace*{-0.06in}\includegraphics[width=0.30\textwidth]{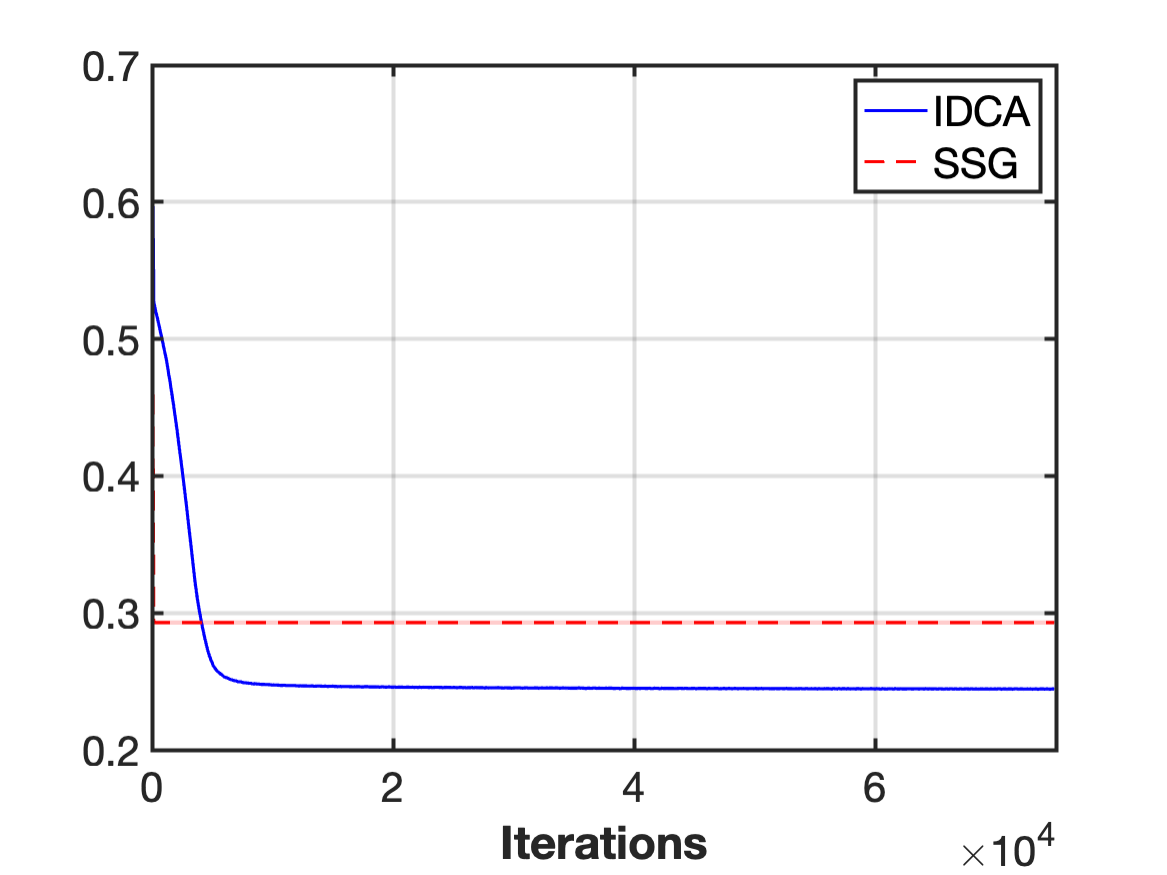}
        
          \\
		\raisebox{9ex}{\small{\rotatebox[origin=c]{90}{Infeasibility}}}
		& \hspace*{-0.06in}\includegraphics[width=0.30\textwidth]{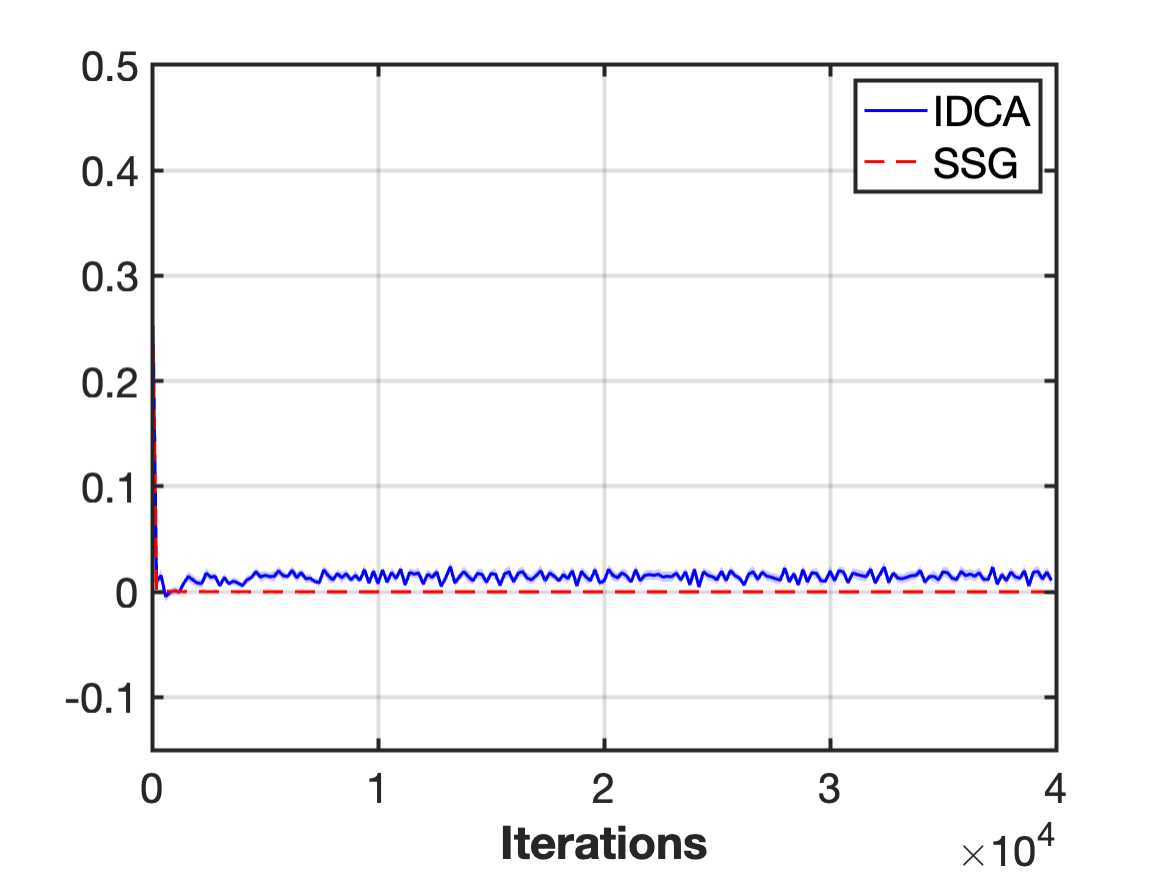}
		& \hspace*{-0.06in}\includegraphics[width=0.30\textwidth]{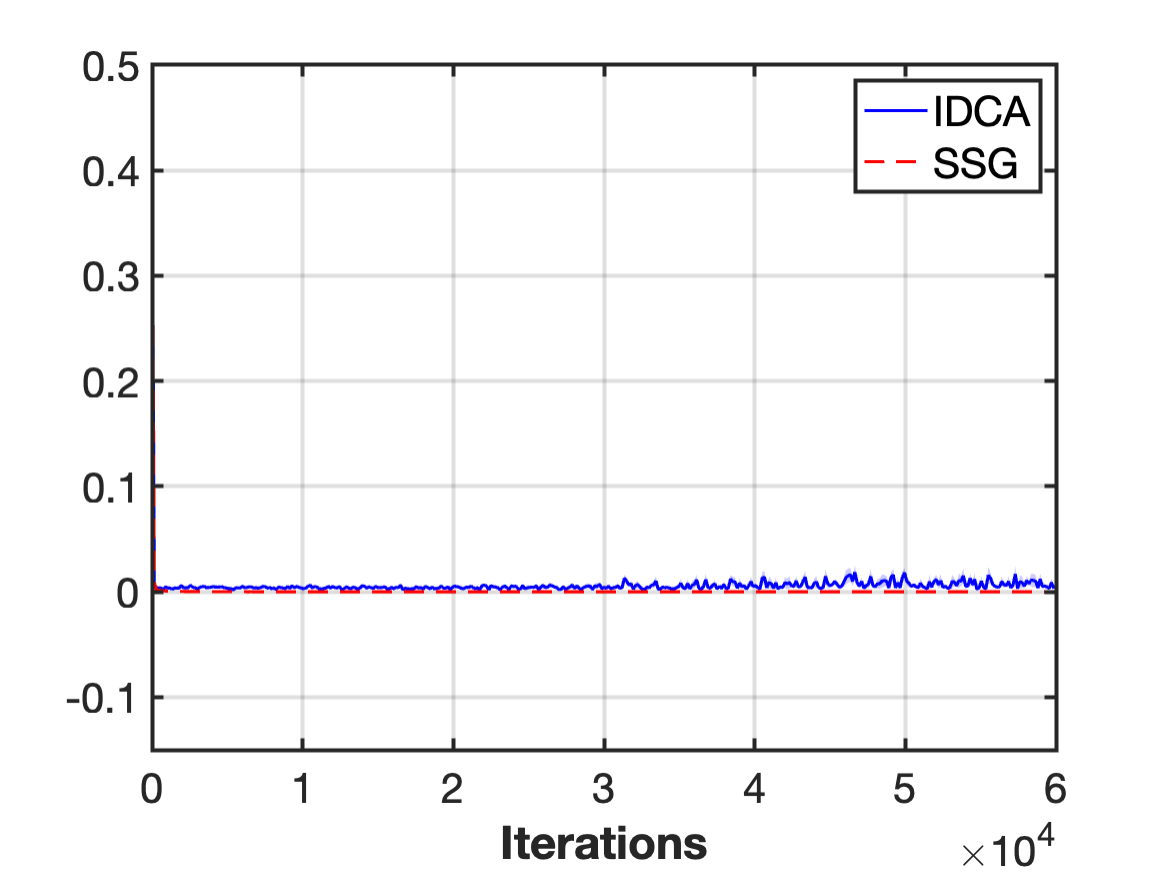}
		& \hspace*{-0.06in}\includegraphics[width=0.30\textwidth]{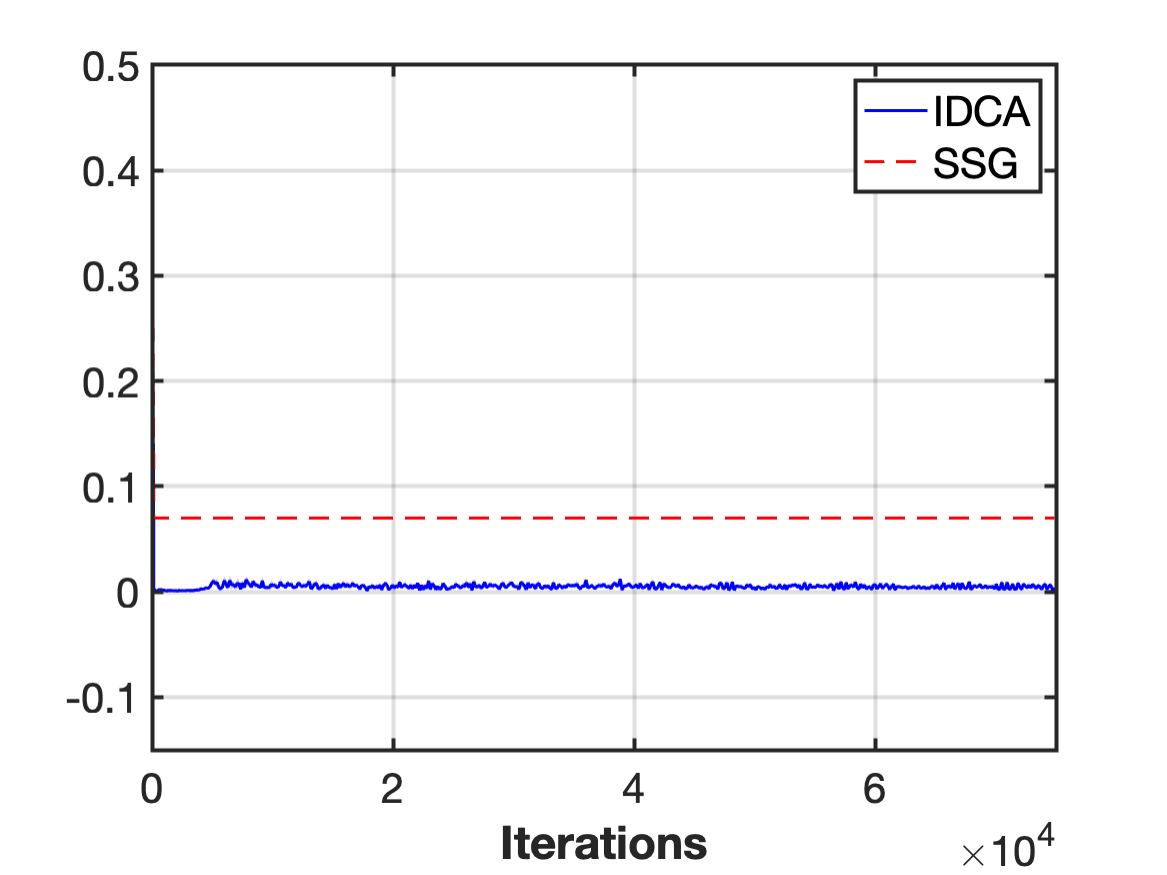}
        \end{tabular}
	\caption{Performances of IDCA and SSG on \eqref{eq:inprocess_pdp_approx}. The interval $\mathcal{I}$ is $[5\%, 30\%]$ for \textit{a9a}, $[0\%, 25\%]$ for \textit{bank}, and $[70\%, 100\%]$ for \textit{law school}, respectively.}
    \label{fig:opt_pdp}
	\vspace{-0.1in}
\end{figure*}

\begin{figure*}[t]
     \begin{tabular}{@{}c|cccc@{}}
      & a9a & bank & law school 
      \\
		\hline \vspace*{-0.1in}\\
		\raisebox{9ex}{\small{\rotatebox[origin=c]{90}{Objective value}}}
		& \hspace*{-0.06in}\includegraphics[width=0.30\textwidth]{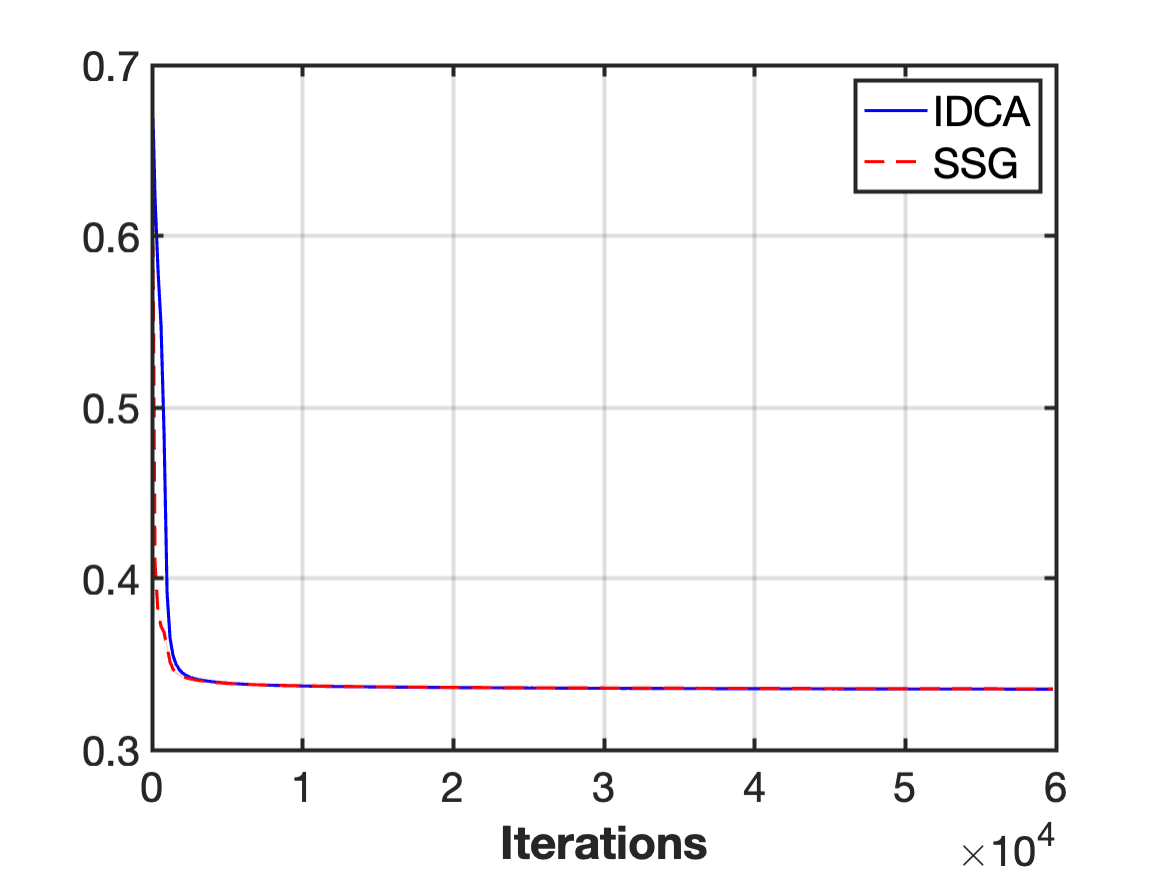}
		& \hspace*{-0.06in}\includegraphics[width=0.30\textwidth]{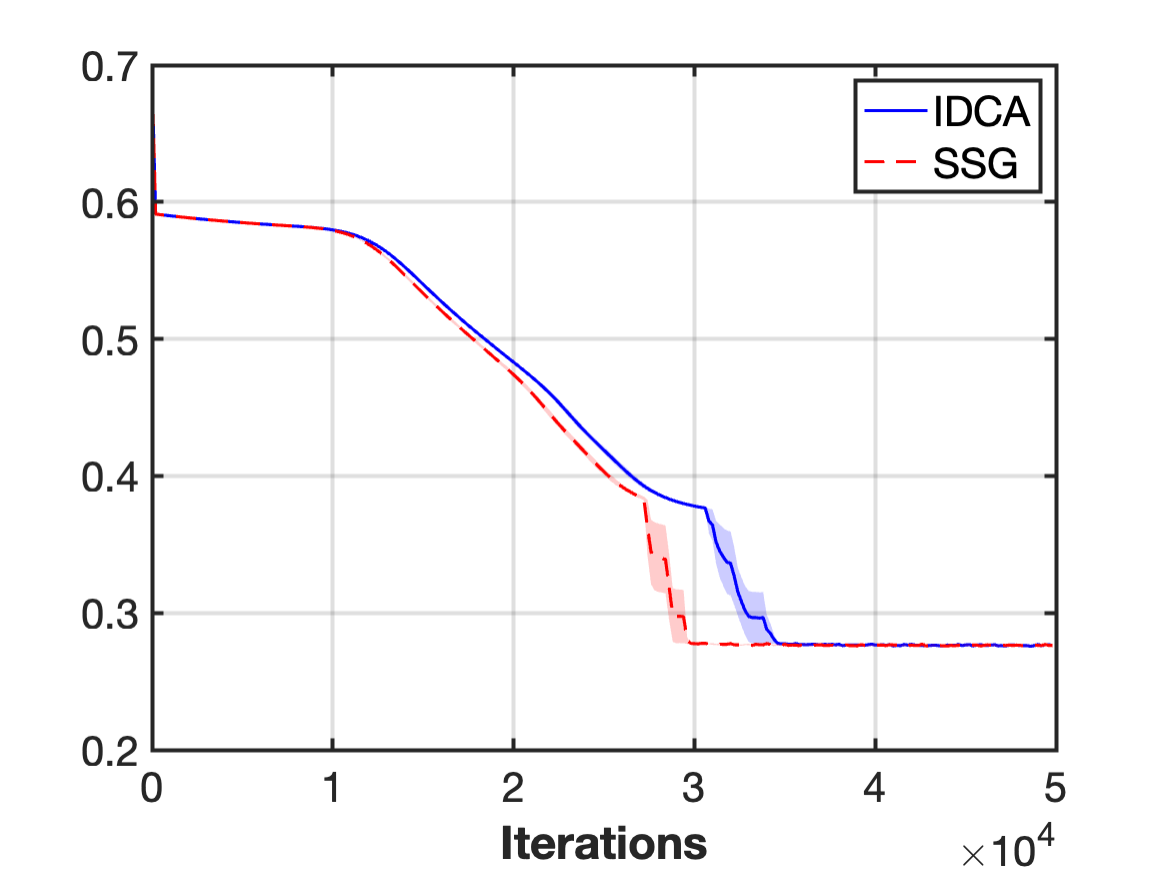}
		& \hspace*{-0.06in}\includegraphics[width=0.30\textwidth]{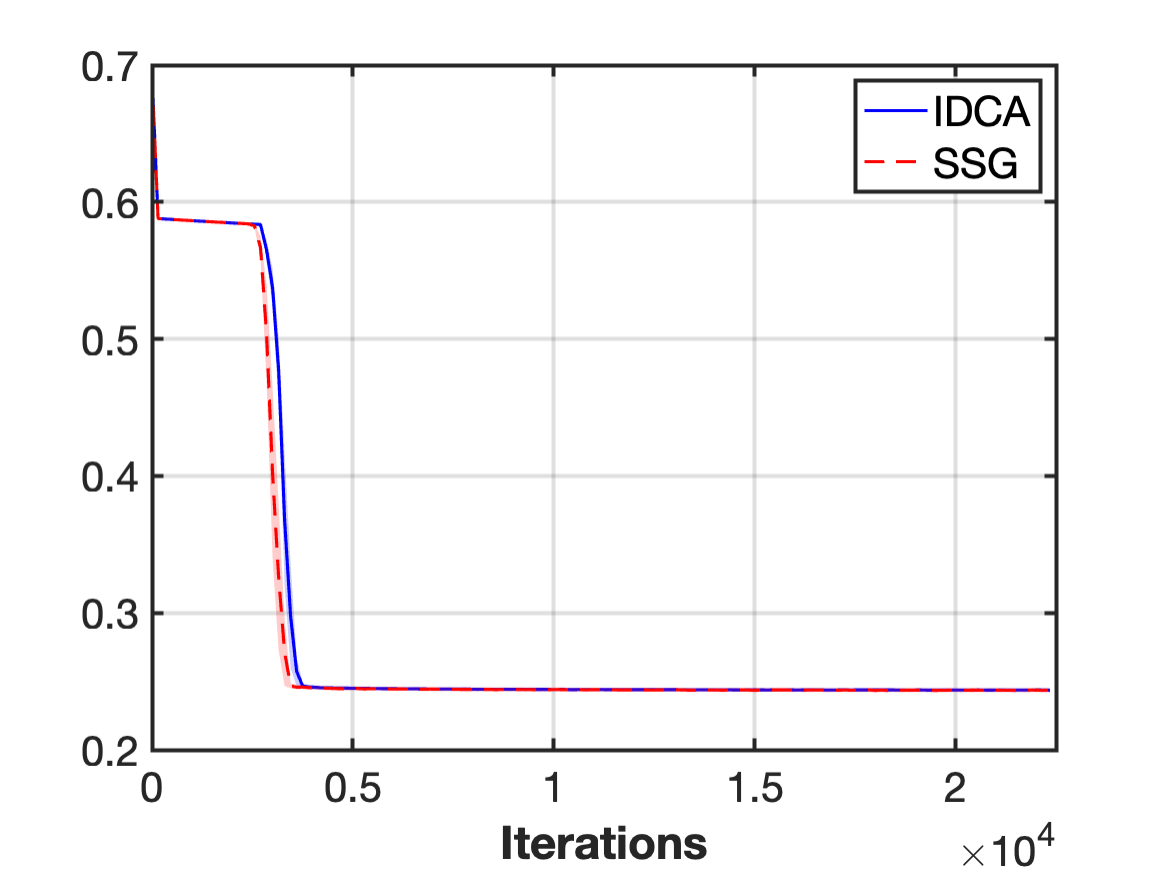}
        
          \\
		\raisebox{9ex}{\small{\rotatebox[origin=c]{90}{Infeasibility}}}
		& \hspace*{-0.06in}\includegraphics[width=0.30\textwidth]{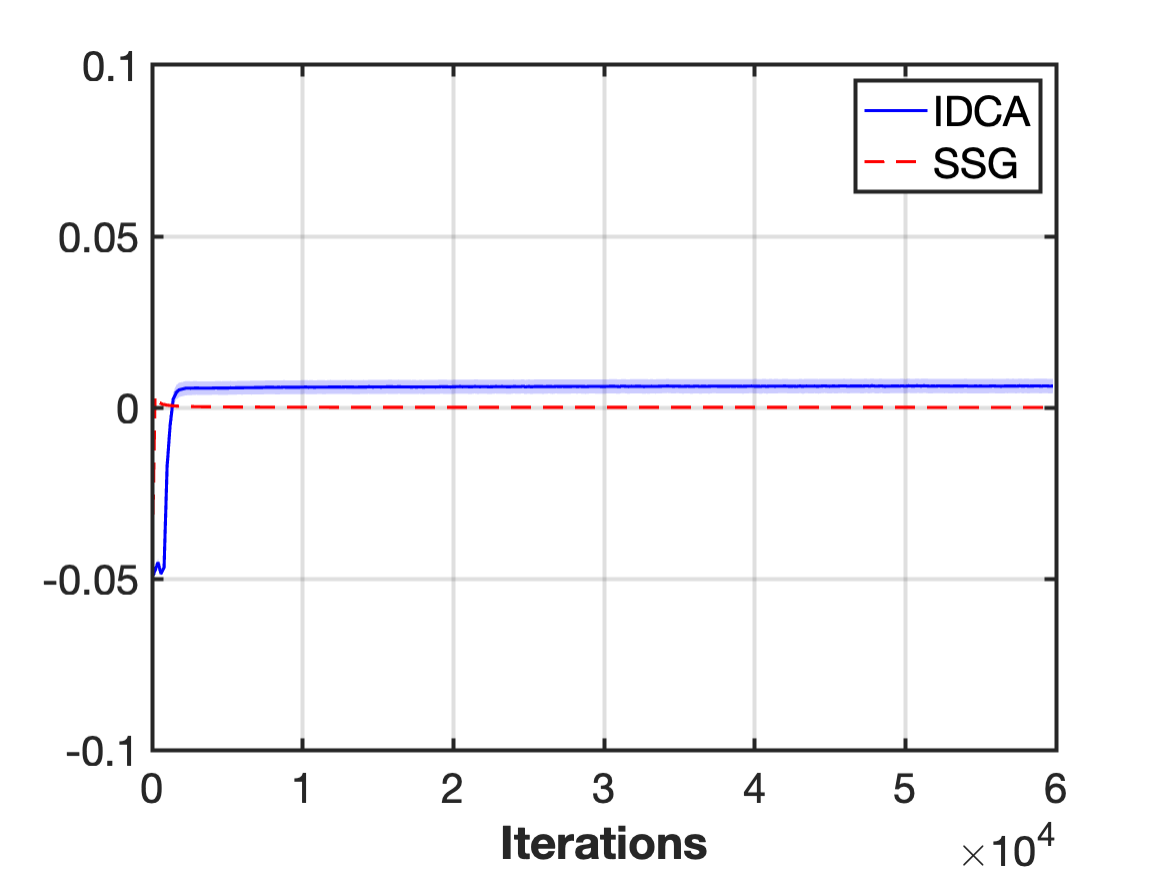}
		& \hspace*{-0.06in}\includegraphics[width=0.30\textwidth]{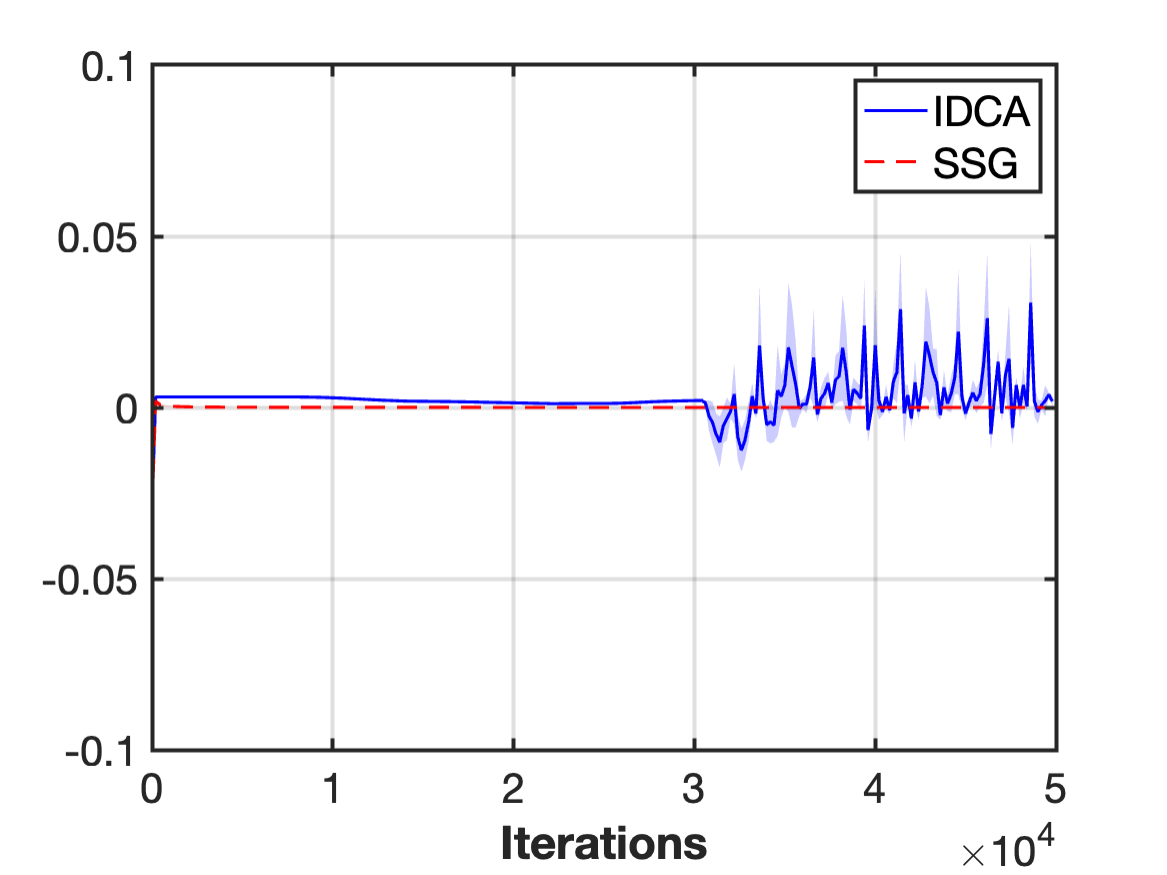}
		& \hspace*{-0.06in}\includegraphics[width=0.30\textwidth]{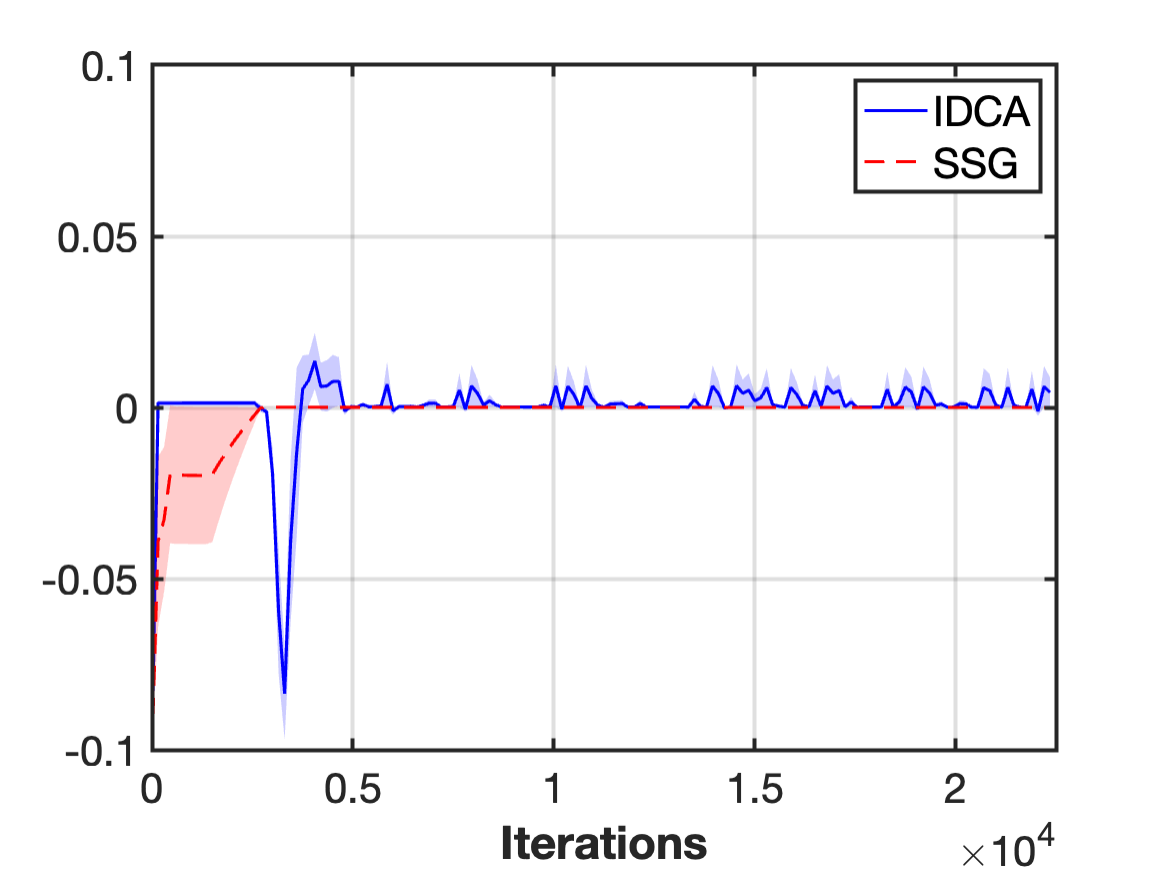}
        \end{tabular}
	\caption{Performances of IDCA and SSG on \eqref{eq:inprocess_wpdp_approx}. The interval $\mathcal{I}$ is $[5\%, 30\%]$ for \textit{a9a}, $[0\%, 25\%]$ for \textit{bank}, and $[70\%, 100\%]$ for \textit{law school}, respectively.}
    \label{fig:opt_wpdp}
	\vspace{-0.1in}
\end{figure*}

We compare the objective value $f_0(\vw)$ and the infeasibility $f(\vw)$ of the solutions generated by IDCA and SSG as the total number of iterations increases. Their performance for solving \eqref{eq:inprocess_pdp_approx} and \eqref{eq:inprocess_wpdp_approx} are presented in  Figures~\ref{fig:opt_pdp} and \ref{fig:opt_wpdp}, respectively. In both figures, the $x$-axis represents the total number of inner iterations for IDCA and represents the total number of iterations for SSG. The $y$-axis represents the objective value and the infeasibility achieved by both methods in the first and second rows, respectively. 

According to Figure~\ref{fig:opt_pdp}, when applied to \eqref{eq:inprocess_pdp_approx}, our IDCA and the SSG method have similar performances on the \emph{a9a} and \emph{bank} datasets. However, IDCA is able to achieve a significantly smaller objective value and smaller infeasibility than SSG on the \emph{law school} dataset. According to Figure~\ref{fig:opt_wpdp}, when applied to \eqref{eq:inprocess_wpdp_approx}, IDCA and the SSG method have similar performances on the \emph{a9a} and \emph{law school} datasets. However, the SSG method has slightly better performance than IDCA on the \emph{bank} dataset. 
We want to emphasize that the SSG method is only a heuristic method for \eqref{eq:inprocess_pdp_approx} and \eqref{eq:inprocess_wpdp_approx} (and for \eqref{DC} in general) as it fails to have any theoretical guarantees. The consequence of the lack of theoretical guarantees is that it may not even return a feasible solution, as shown on the \emph{law school} dataset in Figure~\ref{fig:opt_pdp}. In this case, the iterations of the SSG method likely get trapped in an infeasible stationary solution of the constraint functions. On the contrary, we have provided the complexity analysis for IDCA in Theorem~\ref{thm:idca}, and IDCA always produces a nearly KKT, and thus, a feasible solution.

\subsection{Numerical results on the ACS dataset}
\label{sec:ACSresult}
We also conduct the experiment on the ACS dataset under the same settings and procedures with the same benckmarks as in Section~\ref{sec:pdpexp}. The results are shown in Table~\ref{tab:ACSpSP_compare}. It can be seen that, on the ACS dataset, our method also attains higher accuracy than every baseline models across all levels of pSP fairness, demonstrating its effectiveness in trading accuracy for partial fairness.


\begin{table}[!ht]
\caption{Performances of the models obtained using different fairness constraints on the ACS dataset. $\mathcal{I}$ is set to $[20\%, 40\%]$}
\vskip 0.1in
\centering
\small
\begin{tabular}{cc|cc|cc|cc}
\toprule
\multicolumn{2}{c}{ pSP } 
& \multicolumn{2}{c}{Group AUC } 
& \multicolumn{2}{c}{Inter-group Pairwise } 
& \multicolumn{2}{c}{Intra-group Pairwise} \\
\cmidrule(lr){1-2} \cmidrule(lr){3-4} \cmidrule(lr){5-6} \cmidrule(lr){7-8}
pSP fairness & accuracy 
& pSP fairness & accuracy 
& pSP fairness & accuracy 
& pSP fairness & accuracy \\
\midrule
0.7748 & 0.7986 & 0.7748 & 0.7986 & 0.7748 & 0.7986 & 0.7748 & 0.7986 \\
0.8660 & 0.8100 & 0.8078 & 0.7931 & 0.8163 & 0.7931 & 0.7398 & 0.7920 \\
0.9090 & 0.8068 & 0.8414 & 0.7927 & 0.8410 & 0.7922 & 0.7026 & 0.7902 \\
0.9357 & 0.8043 & 0.9188 & 0.7866 & 0.8821 & 0.7466 & 0.6624 & 0.7848 \\
0.9669 & 0.8001 & 0.9362 & 0.6667 & 0.8685 & 0.6322 & 0.5772 & 0.7648 \\
\bottomrule
\end{tabular}
\label{tab:ACSpSP_compare}
\end{table}

\subsection{Numerical results on deep learning}
\label{sec:CelebA}
The CelebA dataset (Large-scale CelebFaces Attributes Dataset)~\citep{liu2015deep} contains 202,599 face images (split into 60\% training, 20\% validation, and 20\% testing sets) with each image annotated by 40 binary attributes~\citep{liu2015deep}. We consider an image classification problem with a partial statistical parity constraint, formulated as \eqref{eq:inprocess_pdp_approx}, on the CelebA dataset. We instantiate $h_{\vw}(\vxi)$ as a ResNet-18 model (a CNN pretrained on ImageNet) and fine-tune it to predict the “Attractive” attribute, with gender serving as the sensitive attribute in the fairness constraint.

Since SSG (Algorithm~\ref{alg:swg}) is a deterministic method for solving \eqref{DCA}, whereas deep learning training relies on stochastic gradients, we instead implement IDCA (Algorithm~\ref{alg:dca}) using the primal–dual stochastic gradient method of \citet{yu2017online} with a mini-batch size of 64 to solve \eqref{DCA}, ensuring \eqref{eq:epsilonwk} with high probability.

\begin{figure}[!ht]
    \centering
    \includegraphics[width=0.6\textwidth]{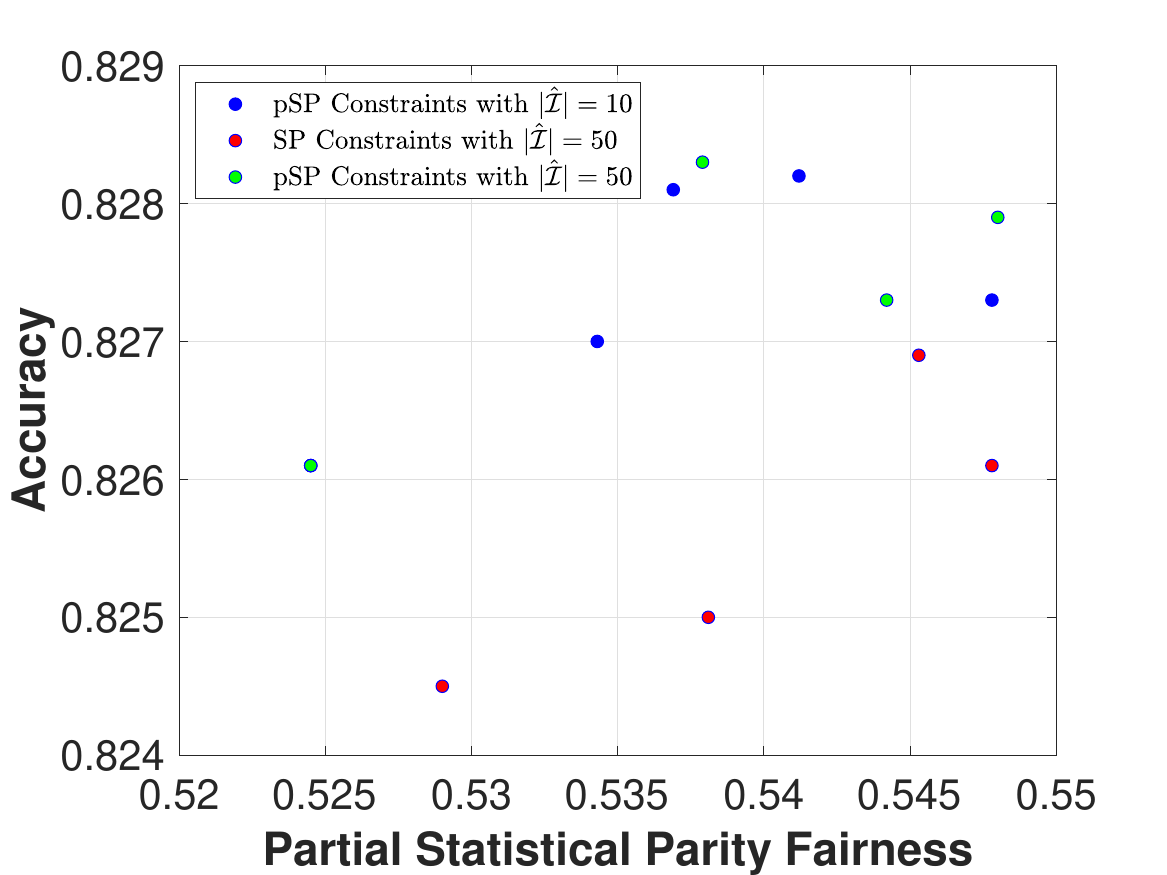}
    \caption{Testing performances of the deep learning models solved from \eqref{eq:inprocess_pdp_approx} under different settings on CelebA dataset.}
    \label{fig:CelebA}
\end{figure}

We evaluate the testing performance of the resulting models under three settings: (i) $\mathcal{I}=[20\%, 40\%]$ with $|\hat{\mathcal{I}}|=10$, (ii) $\mathcal{I}=[0\%, 100\%]$ (i.e., full statistical parity constraint) with $|\hat{\mathcal{I}}|=50$, and (iii) $\mathcal{I}=[20\%, 40\%]$ with $|\hat{\mathcal{I}}|=50$. By varying $\kappa$ within each setting, we obtain different trade-offs between classification accuracy and partial statistical parity, as shown in Figure~\ref{fig:CelebA}.

The results indicate that increasing $|\hat{\mathcal{I}}|$ from 10 to 50 has little impact on performance under partial statistical parity constraints. Moreover, for a fixed level of partial statistical parity, enforcing partial constraints yields higher classification accuracy than imposing full statistical parity.

\subsection{Comparison between partial statistical parity and full statistical parity}
\label{sec:partialfull}
In this section, we compare the performance of models obtained from \eqref{eq:inprocess_pdp_approx} under full statistical parity ($\mathcal{I}=[0\%, 100\%]$) and partial statistical parity ($\mathcal{I}\neq [0\%, 100\%]$). For the partial statistical parity setting, $\mathcal{I}$ and $\hat{\mathcal{I}}$ are chosen as in Section~\ref{sec:pdpexp} for each dataset (see Figure~\ref{fig:pDP_and_wpDP_efficiency_frontier}). For the full statistical parity setting, we set $\mathcal{I}=[0\%, 100\%]$ and construct $\hat{\mathcal{I}}$ as a discrete subset with the same spacing as in the corresponding partial-fairness case. Note that when $\mathcal{I}=[0\%, 100\%]$, problem \eqref{eq:inprocess_pdp_approx} remains a DC program and can therefore be solved using the same IDCA method. 

By varying $\kappa$ in \eqref{eq:inprocess_pdp_approx}, the results for the three datasets are reported in Table~\ref{tab:pSP_SP}. Overall, our partial-fairness approach achieves a superior accuracy–fairness trade-off on the \textit{bank} and \textit{law school} datasets, and a comparable trade-off on the \textit{a9a} dataset.

\begin{table}[!ht]
\caption{Comparison of the performances of the models solved from \eqref{eq:inprocess_pdp_approx} under full statistical parity constraint ($\mathcal{I}=[0\%, 100\%]$) and partial statistical parity constraint ($\mathcal{I}\neq [0\%, 100\%]$).}
\vskip 0.1in
\centering
\small
\begin{tabular}{cc|cc}
\toprule
\multicolumn{4}{c}{a9a} \\
\midrule
\multicolumn{2}{c}{$\mathcal{I}=[0\%, 100\%]$} & \multicolumn{2}{c}{$\mathcal{I}=[5\%, 30\%]$} \\
\cmidrule(lr){1-2} \cmidrule(lr){3-4}
pSP fairness & accuracy & pSP fairness & accuracy \\
\midrule
0.5970 & 0.8499 & 0.5970 & 0.8499 \\
0.6937 & 0.8501 & 0.7017 & 0.8492 \\
0.7799 & 0.8485 & 0.7841 & 0.8479 \\
0.8873 & 0.8440 & 0.8732 & 0.8445 \\
0.9326 & 0.8384 & 0.9310 & 0.8393 \\
0.9742 & 0.8322 & 0.9752 & 0.8311 \\
\midrule

\multicolumn{4}{c}{bank} \\
\midrule
\multicolumn{2}{c}{$\mathcal{I}=[0\%, 100\%]$} & \multicolumn{2}{c}{$\mathcal{I}=[0\%, 25\%]$} \\
\cmidrule(lr){1-2} \cmidrule(lr){3-4}
pSP fairness & accuracy & pSP fairness & accuracy \\
\midrule
0.4366 & 0.9015 & 0.4366 & 0.9015 \\
0.6451 & 0.8957 & 0.6582 & 0.9027 \\
0.7999 & 0.8954 & 0.7376 & 0.9028 \\
0.8456 & 0.8959 & 0.8954 & 0.9025 \\
0.9356 & 0.8941 & 0.9634 & 0.8983 \\
\midrule

\multicolumn{4}{c}{law school} \\
\midrule
\multicolumn{2}{c}{$\mathcal{I}=[0\%, 100\%]$} & \multicolumn{2}{c}{$\mathcal{I}=[70\%, 100\%]$} \\
\cmidrule(lr){1-2} \cmidrule(lr){3-4}
pSP fairness & accuracy & pSP fairness & accuracy \\
\midrule
0.5397 & 0.9004 & 0.5397 & 0.9004 \\
0.5833 & 0.8975 & 0.6038 & 0.8996 \\
0.6976 & 0.8903 & 0.6956 & 0.8937 \\
0.7855 & 0.8892 & 0.7825 & 0.8913 \\
0.8736 & 0.8896 & 0.8479 & 0.8927 \\
0.9615 & 0.8881 & 0.9563 & 0.8909 \\
\bottomrule
\end{tabular}
\label{tab:pSP_SP}
\end{table}

\subsection{Performances of models from \eqref{eq:inprocess_pdp_approx} with different $|\hat{\mathcal{I}}|$}
\label{sec:I_hat}
Theoretically, increasing $|\hat{\mathcal{I}}|$ raises computational cost but can improve the out-of-sample performance of models obtained from \eqref{eq:inprocess_pdp_approx}. To examine this effect, we compare models solved from \eqref{eq:inprocess_pdp_approx} with $|\hat{\mathcal{I}}|=10$ and $|\hat{\mathcal{I}}|=20$ across the three datasets. In particular, we choose $\mathcal{I}$ as in Section~\ref{sec:pdpexp} for each dataset (see Figure~\ref{fig:pDP_and_wpDP_efficiency_frontier}) and then set $\widehat{\mathcal{I}}$ to be a set of equally spaced values in $\mathcal{I}$. The results, reported in Table~\ref{tab:I_hat}, show no significant improvement from increasing $|\hat{\mathcal{I}}|$, except on the \textit{law school} dataset.

\begin{table}[!ht]
\caption{Comparison of the performances of the models solved from \eqref{eq:inprocess_pdp_approx} with $|\hat{\mathcal{I}}|=10$ and $|\hat{\mathcal{I}}|=20$ }
\vskip 0.1in
\centering
\small
\begin{tabular}{c|cc|cc}
\toprule
 & \multicolumn{2}{c}{$|\hat{\mathcal{I}}|=20$} & \multicolumn{2}{c}{$|\hat{\mathcal{I}}|=10$} \\
\cmidrule(lr){2-3} \cmidrule(lr){4-5}
$\kappa$ & pSP fairness & accuracy & pSP fairness & accuracy \\
\midrule

\multicolumn{5}{c}{a9a ($\mathcal{I}=[5\%, 30\%]$)} \\
\midrule
Unconstrained & 0.5970 & 0.8499 & 0.5970 & 0.8499 \\
0.28 & 0.6964 & 0.8491 & 0.7017 & 0.8492 \\
0.2 & 0.7819 & 0.8481 & 0.7841 & 0.8479 \\
0.1 & 0.8751 & 0.8446 & 0.8732 & 0.8445 \\
0.05 & 0.9328 & 0.8387 & 0.9310 & 0.8393 \\
0.01 & 0.9703 & 0.8325 & 0.9752 & 0.8311 \\

\midrule
\multicolumn{5}{c}{bank ($\mathcal{I}=[0\%, 25\%]$)} \\
\midrule
Unconstrained & 0.4366 & 0.9015 & 0.4366 & 0.9015 \\
0.3 & 0.6620 & 0.9026 & 0.6582 & 0.9027 \\
0.22 & 0.7211 & 0.9030 & 0.7376 & 0.9028 \\
0.08 & 0.8683 & 0.9025 & 0.8954 & 0.9025 \\
0.01 & 0.9483 & 0.8985 & 0.9634 & 0.8983 \\

\midrule
\multicolumn{5}{c}{law school ($\mathcal{I}=[70\%, 100\%]$)} \\
\midrule
Unconstrained & 0.5397 & 0.9004 & 0.5397 & 0.9004 \\
0.2 & 0.6360 & 0.8980 & 0.6038 & 0.8996 \\
0.15 & 0.7165 & 0.8929 & 0.6956 & 0.8937 \\
0.1 & 0.7841 & 0.8922 & 0.7825 & 0.8913 \\
0.08 & 0.8683 & 0.8932 & 0.8479 & 0.8927 \\
0.005 & 0.9625 & 0.8910 & 0.9563 & 0.8909 \\

\bottomrule
\end{tabular}
\label{tab:I_hat}
\end{table}

\end{document}